\documentclass[CIT,biber]{nowfnt} %
\vfuzz \hfuzz
\usepackage[utf8]{inputenc}

\usepackage{amsmath}
\usepackage{amsthm}
\usepackage{vmr-symbols-vecbold}
\usepackage{standard-macros}
\newcommand{\ExPow}[3]{\ensuremath{\Exop_{#1}^{#2}\lefto[#3\right]}} 	%
\newcommand{\norm}[1]{\left\Vert #1 \right\Vert}
\newcommand{\subsetm}{\mathcal{M}}

\newcommand{\dataspace}{\mathcal Z}
\newcommand{\featurespace}{\mathcal X}
\newcommand{\labelspace}{\mathcal Y}
\newcommand{\traindata}{\boldsymbol{Z}}
\newcommand{\traindatax}{\boldsymbol{X}}

\newcommand{\traindatasmall}{\boldsymbol{z}}
\newcommand{\traindataxsmall}{\boldsymbol{x}}
\newcommand{\traindataysmall}{\boldsymbol{y}}
\newcommand{\supersample}{\tilde{\boldsymbol{Z}}}
\newcommand{\supersampleloo}{\dot{\boldsymbol{Z}}}
\newcommand{\supersampleloox}{\dot{\boldsymbol{X}}}
\newcommand{\supersamplelooy}{\dot{\boldsymbol{Y}}}
\newcommand{\supersamplelooarg}[1]{\dot{{Z}}_{#1}}
\newcommand{\supersamplelooargb}[1]{\dot{{\boldsymbol{Z}}}_{#1}}
\newcommand{\supersamplez}{\tilde{{Z}}}
\newcommand{\supersamplesmall}{\tilde{\boldsymbol{z}}}
\newcommand{\supersamplesmallx}{\tilde{\boldsymbol{x}}}
\newcommand{\supersamplesmally}{\tilde{\boldsymbol{y}}}
\newcommand{\supersamplex}{\tilde{\boldsymbol{X}}}

\newcommand{\subsetchoice}{\boldsymbol{S}}
\newcommand{\subsetchoicesmall}{\boldsymbol{s}}
\newcommand{\traindatacmi}{\traindata_{\subsetchoice}}
\newcommand{\testdatacmi}{\traindata_{\bar\subsetchoice}}
\newcommand{\traindatacmiloo}{\traindata_{\bar U}}
\newcommand{\testdatacmiloo}{Z_{U}}
\newcommand{\supersampleloss}{\boldsymbol{\Lambda}}
\newcommand{\supersamplelossdiff}{\boldsymbol{\Delta}}

\newcommand{\trainloss}{L_{\traindata}(W)}
\newcommand{\trainlossB}{L_{\traindata_B}(W)}
\newcommand{\trainlossw}{L_{\traindata}(w)}
\newcommand{\trainlosswz}{L_{\traindatasmall}(w)}
\newcommand{\trainlosscmi}{L_{\traindata_{\subsetchoice}}(W)}

\newcommand{\trainlosssub}[1]{L_{\traindata_{#1}}(W)}
\newcommand{\testlosscmi}{L_{\testdatacmi}(W)}

\newcommand{\supersamplelosscmi}{L_{\supersample}(W)}
\newcommand{\poploss}{L_{P_Z}(W)}
\newcommand{\poplossb}{L_{P_{\traindata}}(W)}
\newcommand{\poplossw}{L_{P_Z}(w)}
\newcommand{\loocv}{\text{loo-cv}}

\newcommand{\traindistro}{P_{\traindata}}
\newcommand{\datadistro}{P_Z}
\newcommand{\conddistro}{P_{W\vert \traindata} }
\newcommand{\conddistrorev}{P_{\traindata\vert W} }
\newcommand{\conddistrorevw}{P_{\traindata\vert W=w} }
\newcommand{\conddistroz}{P_{W\vert \traindata=\traindatasmall} }
\newcommand{\conddistrozp}{P_{W\vert \traindata=\traindatasmall'} }
\newcommand{\conddistroi}{P_{W\vert Z_i} }
\newcommand{\jointdistro}{P_{W\! \traindata }}
\newcommand{\auxjointdistro}{Q_{W\! \traindata }}
\newcommand{\auxproductdistro}{Q_{W}\! P_{\traindata }}
\newcommand{\auxconddistro}{Q_{W\vert \traindata }}
\newcommand{\auxconddistroP}{Q_{W\vert \traindata_P }}
\newcommand{\productdistro}{P_{W}\! P_{\traindata }}
\newcommand{\jointdistroi}{P_{W\! Z_i }}
\newcommand{\productdistroi}{P_{W}\! P_{Z_i }}
\newcommand{\traindistrocmi}{P_{\supersample\!\subsetchoice}}
\newcommand{\jointdistrocmi}{P_{W\! \supersample\! \subsetchoice }}

\newcommand{\productdistrocmi}{P_{W\! \supersample}P_{\subsetchoice} }

\newcommand{\conddistrocmi}{P_{W\vert \supersample\! \subsetchoice }}
\newcommand{\conddistrocmii}{P_{W\vert \supersample\! S_i }}
\newcommand{\margdistrocmi}{P_{W\vert \supersample }}
\newcommand{\auxconddistrocmi}{Q_{W\vert \supersample }}

\newcommand{\jointdistroecmi}{P_{\supersampleloss \supersample\! \subsetchoice }}

\newcommand{\jointdistroemi}{P_{\supersampleloss  \subsetchoice }}

\newcommand{\jointdistroedmi}{P_{\supersamplelossdiff  \subsetchoice }}
\newcommand{\productdistroedmi}{P_{\supersamplelossdiff}P_{\subsetchoice} }
\newcommand{\gibbs}{P_{W\vert \traindata}^G}
\newcommand{\gibbsmarginal}{P_W^G}

\newcommand{\genS}{\genop(W,\supersample,\subsetchoice)}
\newcommand{\genSbar}{\genop(W,\supersample,\bar\subsetchoice)}
\newcommand{\genop}{\mathrm{gen}}
\newcommand{\gen}{\genop(W,\traindata)}
\newcommand{\geni}{\genop(W,Z_i)}
\newcommand{\genw}{\genop(w,\traindata)}
\newcommand{\genwz}{\genop(w,\traindatasmall)}

\newcommand{\avgpoploss}{L}
\newcommand{\avgtrainloss}{\hat L}
\newcommand{\avggengap}{\overline{\mathrm{gen}}}
\newcommand{\pacbpoploss}{L(\traindata)}
\newcommand{\pacbtrainloss}{\hat L(\traindata)}
\newcommand{\pacbgengap}{\overline{\mathrm{gen}}(\traindata)}

\newcommand{\lRNderiv}{ \log\frac{\dv\jointdistro}{\dv\auxjointdistro}}
\newcommand{\lRNderivprod}{ \log\frac{\dv\jointdistro}{\dv Q_W\!\traindistro}}
\newcommand{\rnPQ}{\frac{\dv P}{\dv Q}}
\newcommand{\rnPQxy}{\frac{\dv P_{XY}}{\dv P_XP_Y}}
\newcommand{\rnPQxyz}{\frac{\dv P_{X\!Y\!Z} }{\dv P_{X\vert Z}P_{Y\vert Z}P_Z }}

\newcommand{\infdensop}{\imath}
\newcommand{\infdens}{\infdensop(W,\traindata)}
\newcommand{\infdensxy}{\infdensop(X,Y)}
\newcommand{\condinfdens}{\infdensop(W,\subsetchoice\vert \supersample)}
\newcommand{\condinfdensxyz}{\infdensop(X,Y\vert Z)}

\newcommand{\cmi}{I(W;\subsetchoice\vert\supersample)}
\newcommand{\cmii}{I(W;S_i\vert Z_i, Z_{i+n})}
\newcommand{\cmiidis}{I(W;S_i\vert Z_i=z_i, Z_{i+n}=z_{i+n})}
\newcommand{\cmiiz}{I(W;S_i\vert \supersample)}
\newcommand{\ecmi}{I(\supersampleloss;\subsetchoice\vert\supersample)}
\newcommand{\emi}{I(\supersampleloss;\subsetchoice)}
\newcommand{\edmi}{I(\supersamplelossdiff;\subsetchoice)}
\newcommand{\edmii}{I(\Delta_i;S_i)}
\newcommand{\fcmi}{I(\mathbf F;\subsetchoice\vert\supersample)}

\newcommand{\fcmiiz}{I(F_i,F_{i+n};S_i\vert\supersample ) }
\newcommand{\loocmi}{I(W;U\vert\supersampleloo)}
\newcommand{\looemi}{I(\dot{\supersampleloss};U)}
\newcommand{\looecmi}{I(\dot{\supersampleloss};U\vert\supersampleloo)}
\newcommand{\TV}{\mathrm{TV}}
\newcommand{\binrelentinv}[2]{d^{-1}(#1,#2)}
\newcommand{\maxleakage}[2]{\mathcal{L}(#1\rightarrow #2)}
\newcommand{\conmaxleakage}[3]{\mathcal{L}(#1\rightarrow #2\vert #3)}
\newcommand{\wasserstein}[3]{\mathbb{W}_{#1}(#2,#3)}
\newcommand{\chisqdiv}[2]{\chi^2(#1\,||\,#2)}
\newcommand{\Iskl}[2]{ I_{\textnormal{SKL}}(#1;#2)}

\newcommand{\dVC}{d_{\text{VC}}}
\newcommand{\Rloo}{\hat R_{\text{loo}}}

\newcommand{\task}{\tau}
\newcommand{\tasknr}{m}
\newcommand{\Tau}{\mathcal{T}}
\newcommand{\taskspace}{\Tau}
\newcommand{\taskdistro}{P_{\task}}
\newcommand{\traindistroarg}[1]{P_Z^{#1}}
\newcommand{\metatrainsetb}{\hat {\boldsymbol{Z}}}
\newcommand{\metatrainset}{\hat Z}
\newcommand{\metaalg}{P_{U\vert \metatrainsetb}}
\newcommand{\basealg}{P_{W\vert \traindata U}}
\newcommand{\basealgt}{P_{W\vert \traindata^\task \! U}}
\newcommand{\basealgi}{P_{W_i\vert \metatrainsetb_{i,:} U}}

\newcommand{\sourcedistro}{P_Z}
\newcommand{\sourcedistrox}{P_X}
\newcommand{\sourcedistroh}{P_{h_W}}
\newcommand{\sourcedistroyvh}{P_{Y\vert h_W}}
\newcommand{\targetdistro}{P_Z^T}
\newcommand{\targetdistrox}{P_X^T}
\newcommand{\targetdistroh}{P_{h_W}^T}
\newcommand{\targetdistroyvx}{P_{Y\vert X}^T}
\newcommand{\targetdistroyvh}{P_{Y\vert h_W}^T}
\newcommand{\unsuptrain}{\boldsymbol{X}^T}
\newcommand{\conddistrozx}{P_{W\vert \traindata\unsuptrain} }
\newcommand{\jointdistrowzx}{P_{W\! \traindata \! \unsuptrain }}
\newcommand{\representationspace}{\mathcal R}
\newcommand{\productdistroit}{P_{W}\! P_{Z^T_i }}

\newcommand{\conddistrok}{P_{W_k\vert \traindata_k} }
\newcommand{\conddistroaggrw}{P_{W\vert W_1,\dots, W_k} }
\newcommand{\conddistroaggr}{P_{W\vert \traindata} }
\newcommand{\bregmandiv}[3]{\mathcal B_{#1}(#2,#3)}

\newcommand{\statespace}{\mathcal S}
\newcommand{\actionspace}{\mathcal A}

\newcommand{\MBR}{\text{MBR}}

\newcommand{\donskervtext}{Donsker-Varadhan variational representation of the relative entropy}
\newcommand{\aive}{a\"ive}
\newcommand{\boundedlosstext}{Assume that the range of the loss function~$\ell(\cdot,\cdot)$ is~$[0,1]$}
 \newcommand{\newtext}[1]{#1}

\usepackage[capitalise]{cleveref}

\newtheorem{thm}{Theorem}[chapter]
\newtheorem{lem}[thm]{Lemma}

\newtheorem{cor}[thm]{Corollary}
\newtheorem{dfn}[thm]{Definition}
\newtheorem{propo}[thm]{Proposition}

\crefname{thm}{Theorem}{Theorems}
\crefname{lem}{Lemma}{Lemmas}
\crefname{rem}{Remark}{Remarks}
\crefname{cor}{Corollary}{Corollaries}
\crefname{dfn}{Definition}{Definitions}
\crefname{propo}{Proposition}{Propositions}

\usepackage{pgfplots}
\usepackage{tikz}
\usepackage{grffile}
\usepackage{pifont}
\pgfplotsset{compat=newest}
\usetikzlibrary{plotmarks}
\usetikzlibrary{patterns,shapes,fit}
\usetikzlibrary{automata, positioning, arrows}
\usepgfplotslibrary{fillbetween}

\newcommand{\ie}{\emph{i.e.}}
\newcommand{\eg}{\emph{e.g.}}

\usepackage{pageslts}

\title{Generalization Bounds: Perspectives from Information Theory and PAC-Bayes}
\maintitleauthorlist{
Fredrik Hellstr\"om \\
University College London \\
\href{mailto:f.hellstrom@ucl.ac.uk}{f.hellstrom@ucl.ac.uk}
\and
Giuseppe Durisi \\
Chalmers University of Technology \\
\href{mailto:durisi@chalmers.se}{durisi@chalmers.se}
\and
Benjamin Guedj \\
Inria and University College London \\
\href{mailto:benjamin.guedj@inria.fr}{benjamin.guedj@inria.fr}
\and
Maxim Raginsky \\
University of Illinois \\
\href{mailto:maxim@illinois.edu}{maxim@illinois.edu}
}

\author[1]{Hellström, Fredrik}
\author[2]{Durisi, Giuseppe}
\author[3]{Guedj, Benjamin}
\author[4]{Raginsky, Maxim}

\affil[1]{University College London, UK; \href{mailto:frehells@chalmers.se}{frehells@chalmers.se}}
\affil[2]{Chalmers University of Technology, Sweden; \href{mailto:durisi@chalmers.se}{durisi@chalmers.se}}
\affil[3]{Inria and University College London, France and UK; \href{mailto:benjamin.guedj@inria.fr}{benjamin.guedj@inria.fr}}
\affil[3]{University of Illinois, USA; \href{mailto:maxim@illinois.edu}{maxim@illinois.edu}}

\begin{document}

\makeabstracttitle

\begin{abstract}

A fundamental question in theoretical machine learning is generalization.
Over the past decades, the PAC-Bayesian approach has been established as a flexible framework to address the generalization capabilities of machine learning algorithms, and design new ones.
Recently, it has garnered increased interest due to its potential applicability for a variety of learning algorithms, including deep neural networks.
In parallel, an information-theoretic view of generalization has developed, wherein the relation between generalization and various information measures has been established.
This framework is intimately connected to the PAC-Bayesian approach, and a number of results have been independently discovered in both strands.

\newtext{In this monograph, we highlight this strong connection and present a unified treatment of PAC-Bayesian and information-theoretic generalization bounds.}
We present techniques and results that the two perspectives have in common, and discuss the approaches and interpretations that differ.
In particular, we demonstrate how many proofs in the area share a modular structure, through which the underlying ideas can be intuited.
We pay special attention to the conditional mutual information (CMI) framework; analytical studies of the information complexity of learning algorithms; and the application of the proposed methods to deep learning.
This monograph is intended to provide a comprehensive introduction to information-theoretic generalization bounds and their connection to PAC-Bayes, serving as a foundation from which the most recent developments are accessible.
It is aimed broadly towards researchers with an interest in generalization and theoretical machine learning.

\end{abstract}

\chapter{Introduction: On Generalization and Learning}
\label{chap:introduction}

Artificial intelligence and machine learning have emerged as driving forces behind transformative advancements in various fields, becoming increasingly pervasive throughout many industries and in our daily lives.
As these technologies continue to gain momentum, the need to develop a deeper understanding of their underlying principles, capabilities, and limitations grows larger.
In this monograph, we delve into the theory of machine learning, and more specifically statistical learning theory, where a key topic is the generalization capabilities of learning algorithms.

A learning algorithm is a (potentially stochastic) rule for selecting a hypothesis, given a training data set.
\newtext{Generalization bounds for learning algorithms provide guarantees that the performance, as measured by a loss function, is ‘‘good enough,'' given that the training loss is small, when the hypothesis is subjected to new samples that were not necessarily in the training data.}
Such bounds are useful for several reasons.
When applied in a specific use case, a generalization bound provides a certificate that the hypothesis performs well on new data, provided that the assumptions under which the bound was derived are valid.
Furthermore, such bounds can serve as inspiration for the design of new learning algorithms, potentially leading to practical improvements.
Finally, on a deeper level, generalization bounds can enable a more complete understanding of learning algorithms.

While the literature on generalization bounds is vast, making an in-depth review of the full field beyond our scope, we will discuss several key references.
\citet{valiant-84a} formalized a model of learnability, called Probably Approximately Correct (PAC) learning.
Roughly speaking, a problem is PAC learnable if there exists a learning algorithm such that, for any data distribution, the selected hypothesis has satisfactory performance with high probability.
In the preceding decade, \citet{vapnik-71a} studied the uniform convergence of certain events.
They characterized this convergence in terms of a property of the underlying set that would later be termed the Vapnik-Chervonenkis (VC) dimension, which can be thought of as a measure of complexity.
\citet{blumer-89a} connected these two topics, and demonstrated that the VC dimension of a hypothesis class characterizes its PAC learnability.
We discuss these topics and additional results in more detail in \cref{sec:uniform-convergence}.

The two particular strands in the literature on generalization bounds that will be our main focus throughout this monograph are the PAC-Bayesian and information-theoretic lines of research.
Despite the great commonality in techniques and concepts, these two fields have evolved in almost parallel tracks until recently.
One objective of the present monograph is to give a unified treatment of the two approaches and highlight their similarities, despite the differing origins.
The PAC-Bayesian approach---initiated by \citet{shawetaylor-97a,mcallester-98a,mcallester-99a}, with significant later contributions from, \eg,~\citet{catoni-07a}---started as a quest to obtain Bayesian-flavored versions of PAC generalization bounds, as the name implies.
PAC bounds are independent of the specific learning algorithm used, as they hold uniformly over the class of possible hypotheses.
In contrast, PAC-Bayesian bounds take into account the learning algorithm by explicitly incorporating a distribution over hypotheses---hence the Bayesian suffix.

\newtext{The effort of relating generalization and information, with a broad interpretation of these terms, has a long history.
Conventional wisdom, by way of Occam's razor~\citep{blumer-87a}, holds that solutions that are ‘‘simpler'' in some sense tend to generalize better than their more ‘‘complex'' counterparts.
Many different ways of formalizing complexity measures to capture ‘‘information'' of some kind have been studied, with some of the earliest examples being the Fisher information of \citet{edgeworth} and \citet{fisher}, the information theory of \citet{shannon}, and the Kolmogorov complexity of \citet{kolmogorov,solomonoff}.
In seminal works, \citet{barron-99a} and \citet{leung-06a} connected such complexity measures to performance guarantees for density estimation.
Other notable information notions in the context of learning include the Akaike information criterion of \citet{akaike-74a}, the Bayesian information criterion of \citet{schwarz-78a}, and the minimum description length principle, studied by, \eg, \citet{rissanen-78a,rissanen-83a} and \citet{barron-91a}, \citet{barron-98a} (see the book of \cite{grunwald-07a} for an in-depth treatment).
The particular flavor of information-theoretic approach to generalization that we will focus on can be traced back to the work of~\citet{zhang-06a}, and more recently, to the seminal works of~\citet{russo-16a} and~\citet{xu-17a}.
In this line of work, the learning algorithm is viewed as a communication channel from the training data to the hypothesis.}
With this interpretation of the statistical learning process, it is clear that quantities that are common in communication applications, such as the mutual information, have an important role to play.

Despite the historical separation between these lines of work---even within the specific strands, at times---the tools and results that appear in these fields have more similarities than differences, and any discrepancy between them is mainly in the motivation and framing of the work.
This may be due to the interdisciplinary nature of the field: it can naturally be covered as statistics, computer science, electrical engineering, and physics.\footnote{Noting this deep connection, \citet{catoni-07a} referred to the PAC-Bayesian approach as the ‘‘thermodynamics of statistical learning.''}
Thus, the reader will not be surprised that many of these results were re-discovered and re-interpreted in many separate contexts, evolving independently.
\newtext{Still, the connection between PAC-Bayesian and information-theoretic generalization bounds has been noted and explored by, \eg, \citet{russo-16a}, \citet{banerjee-21a}, \citet{grunwald-21a}, and \citet{alquier-21a}.
One of the aims of the present monograph is to solidify the bridge between these strands of the literature, demonstrating the commonalities in the different approaches.}

\section{Notation and Terminology}\label{sec:notation}

To set the stage, we introduce the notation that is used throughout this monograph. Unless otherwise stated, capital letters indicate random variables, with lower-case letters indicating their instances. For random vectors, the same applies, but the letters are in bold.
We consider the training examples to lie in a set~$\mathcal{Z}$, referred to as the \emph{instance space}. In the context of supervised learning, the instance space is a product between a \emph{feature space}~$\mathcal{X}$ and a \emph{label space}~$\mathcal{Y}$, so that~$\mathcal{Z}=\mathcal{X}\times \mathcal{Y}$.
At its disposal, the learning algorithm has a \emph{training set}~$\traindata = (Z_1,\dots,Z_n) \in \dataspace^n$, consisting of~$n$ training examples.\footnote{\newtext{Despite conventionally being called a ‘‘set,'' $\traindata$ is a vector: its elements are ordered, and elements are allowed to be repeated.}}
Usually, we assume that the training examples are independent and identically distributed (\iid),\footnote{This assumption is classical in statistical learning theory. Nevertheless, we will cover recent results that allow one to relax and even remove it (see \cref{chap:probability,chap:alternative-settings}).} with each training example being drawn from a data distribution~$\datadistro$ on~$\dataspace$.
We denote the distribution of~$\traindata$, as well as other product distributions, as~$P_{\traindata}=P_Z^n$.
Throughout, we will use the shorthand~$[n]=\{1,\dots,n\}$ to refer to the indices of the training samples.

Confronted with the training data, the learner selects a hypothesis~$W$ from a set~$\mathcal{W}$, called the \emph{hypothesis space}.
Again, in supervised learning,~$\mathcal{W}$ is typically a subset of all functions from~$\mathcal{X}$ to~$\mathcal{Y}$, or the parameters of such functions, but the general framework can accommodate other notions of hypothesis.
The method by which the learner chooses the hypothesis is described by a (probabilistic) mapping from the training set~$\traindata$ to the hypothesis~$W$, denoted by~$\conddistro$ and referred to as a \emph{learning algorithm}.
Mathematically, it can be seen as a stochastic kernel, which gives rise to a probability distribution on~$\mathcal{W}$ for each instance of~$\traindata$.
Note that~$\conddistro$ is defined for a specific size~$n$ of the training set.
We usually assume that the learning algorithm can be adapted to training sets of different sizes, \ie, we assume that~$\conddistro$ is defined for every~$n$.
While there is often a natural relation between these conditional distributions for various~$n$, we do not require that they are related in general.

The quality of a specific hypothesis $w\in\mathcal{W}$ with respect to a sample~$z\in\mathcal{Z}$ is measured by a \emph{loss function}, $\ell: \mathcal{W}\times \mathcal{Z}\rightarrow \reals^+$.
To give some classical examples of loss functions, consider supervised learning, where the sample is decomposed into features and labels (or inputs and outputs) as $z=(x,y)\in\mathcal{X} \times \mathcal{Y}$ and the hypotheses~$w\in\mathcal{W}$ are functions~$w:\mathcal{X}\rightarrow\mathcal{Y}$.
For classification, where the label space~$\mathcal{Y}$ is discrete, a typical loss function is the classification error~$\ell(w,z) = 1\{w(x)\neq y\}$. Here,~$1\{\cdot\}$ denotes the indicator function. For regression, where the label space is continuous, a common choice is the squared loss~$\ell(w,z) = (w(x)-y)^2$.

The true goal of the learner is to select a hypothesis that performs well on fresh data from the distribution~$P_Z$, as measured by the loss function.
This is formalized by the \emph{population loss}~$$\poplossw=\Ex{P_Z}{\ell(w,Z)},$$ sometimes referred to as the (true) \emph{risk} of a hypothesis.
A key feature of the learning problem is that the true data distribution is assumed to be unknown, which implies that the population loss cannot be computed by the learner.
However, by averaging the loss function over training data, the learner obtains the \emph{training loss}~$$\trainlossw=\frac{1}{n}\sum_{i=1}^n\ell(w,Z_i),$$ which serves as an estimate of the population loss. The training loss is also known as the \emph{empirical risk}.
A natural procedure for selecting a hypothesis is to minimize the training loss.
This is referred to as \emph{empirical risk minimization} (ERM), and is successful in finding a hypothesis with low population loss if the difference between population loss and training loss is small.
This is measured by the \emph{generalization error}~$$\genw=\poplossw-\trainlossw,$$ which is also called the \emph{generalization gap}.

\section{Flavors of Generalization}
\label{sec:flavors}
Since the randomized learning algorithm is described by a conditional probability distribution~$\conddistro$, bounds on the generalization error~$\gen$ come in a variety of forms.
We now introduce three canonical forms that have been studied in the information-theoretic and PAC-Bayesian literature.

Firstly, one possibility that has been widely considered in the information-theoretic strand of the literature is to bound the average generalization error~$\Exop_{\jointdistro}[\gen]$.
Performing an average analysis can often simplify mathematical derivations, and lead to some insights about the studied algorithms. The works of~\citet{russo-16a} and~\citet{xu-17a} both focus on this setting, and the mutual information between training data and hypothesis naturally arises as a fundamental quantity in upper bounds for the average generalization error.
In \cref{sec:basicitbound}, we introduce a first such average generalization bound, as a warm-up to the more general theory presented later in this monograph.
The particular features that are relevant specifically for this scenario are discussed in more detail in \cref{chap:average}.

Secondly, in practical situations, we may be given only one instance of a training set, so an arguably more pertinent question is if we can bound the generalization error with high probability over the draw of the data.
In the PAC-Bayesian literature, initiated in the works of~\citet{shawetaylor-97a} and~\citet{mcallester-98a}, most bounds are on the generalization error when averaged over the learning algorithm,~$\Ex{\conddistro}{\gen}$, and hold with probability at least~$1-\delta$ under~$P_{\traindata}$ for some confidence parameter~$\delta\in (0,1)$.
The change in perspective in the PAC-Bayesian approach, as compared to the classical statistical learning literature, is significant. We no longer ask whether there are specific hypotheses~$w$ that perform well: instead, we ask if there are distributions~$\conddistro$ over hypotheses that do.
To highlight the conceptual connection to Bayesian statistics, the distribution~$\conddistro$ is usually termed \emph{posterior}.
This distribution is compared, via information-theoretic metrics, to a reference measure~$Q_W$ called the \emph{prior}.
Another significant feature that is shared among many PAC-Bayesian bounds is that they hold uniformly for all choices of posterior.
This, and other important properties of PAC-Bayesian bounds, are detailed in \cref{sec:pac-bayesian-bounds}.

Finally, we may be interested in the generalization error when we have a single training set and we use our learning algorithm to select a single hypothesis.
Thus, we seek bounds on~$\gen$ that hold with probability at least~$1-\delta$ under~$\jointdistro$.
In this monograph, we will call this the \emph{single-draw} setting, following~\citet{catoni-07a}, since we are concerned with a single draw of both data and hypothesis.
This type of bound has appeared sporadically in both the information-theoretic and PAC-Bayesian literature.
While this type of bound can arguably be the most relevant in practice---for instance, in deep learning (discussed in \cref{chap:iterative-methods}), one typically uses a deterministic neural network obtained via one instantiation of a randomized learning algorithm---it comes with some drawbacks.
For instance, since the probability is computed with respect to the joint distribution~$\jointdistro$, any single-draw bound is by definition a statement pertaining to a particular posterior~$\conddistro$. Thus, we lose uniformity over posteriors.
Furthermore, for the information-theoretic bounds that we discuss here, we need a stronger technical requirement on the absolute continuity of the distributions involved---at least for data-dependent bounds.
We will discuss this type of bounds in \cref{sec:single-draw-bounds}.

It should be stressed that the terminology used here is not universally accepted, and different names are used by different authors.
Furthermore, bounds of all types have been studied in both the PAC-Bayesian and information-theoretic strands of the literature.
For instance, average bounds have been referred to as ‘‘PAC-Bayesian type'' bounds~\citep{salmon-11a,salmon-12a} or mean approximately correct (MAC)-Bayesian bounds~\citep{grunwald-21a}.
Single-draw bounds have been referred to as pointwise or de-randomized PAC-Bayesian bounds~\citep{catoni-07a,alquier-13a,guedj-13a}.
The term de-randomized PAC-Bayesian bound has also been used for bounds that specifically apply to the average hypothesis, that is, bounds on~$\genop(\Ex{\conddistro}{W},\traindata)$ that hold with probability~$1-\delta$ under~$P_{\traindata}$ \citep{banerjee-21a} (such variants will be discussed in \cref{sec:mean-hypothesis}).
However, throughout this monograph, we will use the terms defined above.

The framework of PAC learnability and the associated uniform-convergence bounds that we mentioned earlier do not fit exactly into any of the flavors that we have mentioned so far (although the single-draw bounds are most closely related).
In the following section, we give a formal definition of PAC learnability, and provide an overview of some generalization bounds based on uniform convergence.

\section{Uniform Convergence-Flavored Generalization Bounds}\label{sec:uniform-convergence}

As previously indicated, demonstrating PAC learnability for a hypothesis class boils down to a very strong type of uniform convergence result.
Roughly speaking, PAC learnability requires that for any data distribution~$P_Z$, there is a learning algorithm that, with sufficient training data, is arbitrarily close to the optimal population loss.
As it turns out, PAC learnability is equivalent to uniform convergence, defined below~\citep[][Chapter 4]{shalev-shwartz-14a}.
\begin{dfn}[Uniform convergence]\label{def:uniform-convergence}
The hypothesis class~$\mathcal W$ has the \emph{uniform convergence property} if there exists a function~$m:(0,1)^2\to \naturals$ such that, for every~$\epsilon,\delta\in (0,1)$ and every data distribution~$\datadistro$, the following holds:
if~$\traindata$ contains~$n\geq m(\epsilon,\delta)$ \iid samples from~$\datadistro$, we have with probability at least~$1-\delta$ that
\begin{equation}
\abs{\trainlossw - \poplossw} \leq \epsilon \quad \text{for all } w\in\mathcal W.
\end{equation}
The function~$m$ is called the \emph{sample complexity}.
\end{dfn}
Thus, if a hypothesis class satisfies the uniform convergence property, we can obtain generalization bounds that are uniform over both data distributions and hypotheses.
The attractiveness of these bounds is clear: no matter what data you are dealing with, independent of the learning algorithm you use, you can trust that the training loss gives a good indication of your population loss.
At the moment, it unfortunately seems as if such requirements are too strict for many modern machine learning settings, such as deep neural networks.\footnote{This is not meant to imply that the bounds discussed in this section have no hope of describing modern models, such as deep neural networks. Indeed, promising steps toward this have been taken in the literature~(\eg, \citealp{neyshabur-19a,negrea-20a}).}
For this model class, some data distributions or some hypotheses lead to poor generalization, while naturally occurring data and commonly used learning algorithms perform well.
This motivates the information-theoretic approach of making statements that are specific to the data distribution and learning algorithm in question.
Still, the framework of uniform generalization has proven immensely powerful for many domains, and has led to a definitive characterization of when learning is possible in this strict sense for binary classification: the VC dimension.
Intuitively, the VC dimension is related to the complexity of a hypothesis class, and measures the size of the biggest data set for which the hypothesis class can induce arbitrary labellings of the features.
We give an overview of the VC dimension in \cref{sec:vc-dimension}.

A step towards incorporating data-dependence in the bounds was taken by~\newtext{\citet{gine-84a,koltchinskii-00a,koltchinskii-01a,bartlett-01a,bartlett-02a}} with the introduction of the Rademacher complexity of a hypothesis class.
The Rademacher complexity similarly measures the ability of a hypothesis class to instantiate arbitrary labels, but can be computed empirically on the basis of a training set.
Still, it has a uniform flavor in terms of the hypothesis class.
We discuss the Rademacher complexity in \cref{sec:rademacher}.

Note that we only provide an exceedingly brief overview of uniform convergence-flavored generalization bounds and their history, in order to provide context for the upcoming sections.
Since properly covering this vast subject is far beyond the scope of the present monograph, the reader is referred to, for instance, the excellent books by~\citet{shalev-shwartz-14a,Mohri18} for further details.

\subsection{VC Dimension}\label{sec:vc-dimension}

We will now focus on binary classification, where the sample space decomposes as~$\dataspace=\mathcal X \times \mathcal Y$.
Here,~$\mathcal X$ is the \emph{feature space}, while~$\mathcal Y=\{0,1\}$ is the \emph{label space}.
Each hypothesis~$w\in \mathcal W$ is a map~$w:\mathcal X\rightarrow \{0,1\}$ that predicts a label for each feature.
We will focus on the~$0-1$ loss function, given by~$\ell(w,z)=1\{w(x)\neq y \}$.
Thus, the hypothesis incurs a loss if and only if it predicts the wrong label.
For this setting, the VC dimension of~$\mathcal W$, denoted as~$\dVC$, provides a fundamental characterization of uniform convergence (defined in \cref{def:uniform-convergence}), and hence of PAC learnability:~$\mathcal W$ satisfies the uniform convergence property if and only if~$\dVC$ is finite.
In order to define the VC dimension, we need to introduce the growth function of a hypothesis class~\citep[Def.~6.5]{shalev-shwartz-14a}.

\begin{dfn}[Growth function and VC dimension]
The \emph{growth function}~$g_\mathcal{W}(m)$ is defined as the maximum number of different ways in which a feature set of size~$m$ can be classified using functions from~$\mathcal W$, that is,
\begin{equation}
\max_{(x_1,\dots,x_m) \in \featurespace^m} \abs{  \{   ( w(x_1), \dots, w(x_m) ) : w\in\mathcal W \}  } .
\end{equation}
Note that~$g_\mathcal{F}(m)\leq 2^m$.
The \emph{VC dimension} of~$\mathcal W$, denoted~$\dVC$, is the largest integer such that this upper bound holds with equality.
Specifically,
\begin{equation}
\dVC = \max\{ m\in\naturals: g_\mathcal{F}(m) = 2^m\} .
\end{equation}
If no such integer exists, we say that~$\dVC=\infty$.
If the VC dimension of a hypothesis class is finite, we will refer to it as a VC class.
\end{dfn}
Intuitively, VC dimension characterizes uniform convergence for the following reason:
if the VC dimension is infinite, we can change the labels of a training set~$\traindata$ arbitrarily and still find a hypothesis that outputs these exact predictions, no matter the size~$n$ of the training set.
Hence, we can find a hypothesis with a minimal or maximal training loss, independent of the underlying population loss.
However, if the VC dimension is finite and~$n\gg \dVC$, we cannot adapt arbitrarily to every sample in the training set, but only to~$\dVC$ of them.
Therefore, in some sense, the remaining~$n-\dVC$ samples provide a reasonable estimate of the population loss.

Re-producing the full proof is beyond our present scope, but essentially, one proceeds by bounding the generalization gap in terms of the growth function by formalizing the intuition above (see, \eg,~\citealp[Chapter~28]{shalev-shwartz-14a}).
Then, the growth function is controlled using the Sauer-Shelah lemma~\citep[Lemma~6.10]{shalev-shwartz-14a}, which provides a bound on the growth function in terms of the VC dimension.\footnote{As we will see in \cref{sec:info-comp-vc}, this is also a key tool for analyzing information-theoretic generalization bounds for the special case of VC classes.}
\begin{lem}[Sauer-Shelah lemma]\label{lem:sauer-shelah}
Let~$g_\mathcal{W}(\cdot)$ denote the growth function of the function class~$\mathcal W$.
For any function class~$\mathcal W$ with VC dimension~$\dVC$,
\begin{equation}\label{eq:lemma_sauer_shelah_for_vc_dim}
g_{\mathcal W}(m) \leq \sum_{i=0}^{\dVC} \binom{m}{i} \leq \begin{cases}
                       \displaystyle 2^{\dVC+1}, &m<\dVC + 1, \\
                        \displaystyle \left(\frac{em}{\dVC} \right)^{\dVC}\!\!\!, &m\geq \dVC+1.
                    \end{cases}
\end{equation}
\end{lem}

With this, we can obtain the following~\citep[Thm.~6.8]{shalev-shwartz-14a}.

\begin{thm}[Generalization from VC dimension]\label{thm:vc-dim-classical}
Consider a hypothesis class~$\mathcal{W}$ with VC dimension~$\dVC$.
Then,~$\mathcal W$ has the uniform convergence property (see \cref{def:uniform-convergence}) with sample complexity~$m$, which is upper and lower bounded as
\begin{equation}
C'\frac{\dVC + \log \frac{1}{\delta}}{\epsilon^2}\leq   m(\epsilon,\delta)\leq C\frac{\dVC + \log \frac{1}{\delta}}{\epsilon^2} = m_+(\epsilon,\delta),
\end{equation}
for some constants $C$, $C'$.
In particular, this implies that for all~$w\in\mathcal W$,
\begin{equation}\label{th_eq:PAC_samplecomp_upperbound-slow}
\abs{\trainlossw - \poplossw} \leq \sqrt{ C \frac{\dVC+\log\frac1\delta}{n} } .
\end{equation}
This implies that~$\mathcal W$ is PAC learnable in the following sense:
for every distribution~$P_Z$, there exists a deterministic learning algorithm~$\conddistro$ such that, for every~$\epsilon,\delta\in(0,1)$, we have that with probability at least~$1-\delta$ over~$P_{\traindata}$,
\begin{equation}\label{th_eq:vc-slow}
\poploss \leq \inf_{w\in W} \poplossw + \epsilon
\end{equation}
provided that~$n\geq m_+(\epsilon,\delta)$.
\end{thm}
Remarkably, the upper and lower bounds on the sample complexity~$m(\varepsilon,\delta)$ differ only by a multiplicative constant, and specifically, the dependence on~$\dVC$ is identical.
Thus, the PAC learnability of a hypothesis class~$\mathcal W$ is fully determined by its VC dimension~$\dVC$ in the sense that~$\mathcal W$ admits a finite sample complexity \emph{if and only if}~$\dVC$ is finite.
As remarked before, PAC learnability is a very strong requirement, as it is equivalent to uniform convergence both with respect to the hypothesis class and the data distribution.
Hence, less stringent notions of generalization are of interest, especially distribution- and algorithm-dependent ones.

Under the assumption of realizability, where~$\inf_{w\in W} \poplossw=0$, it is possible to derive a bound similar to~\eqref{th_eq:PAC_samplecomp_upperbound-slow}, but with a decay of~$1/n$.
This is referred to as a \emph{fast} rate, in contrast to the \emph{slow} rate of~$1/\sqrt n$.
\newtext{For more details on fast rates, the reader is referred to the seminal works of \citet{vapnik-74a}, \citet{lee-98a}, \citet{li-99a}, and the more recent works of~\citet{vanerven-15a} and \citet{grunwald-20a}.}

\subsection{Rademacher Complexity}\label{sec:rademacher}

Another important metric in the theoretical study of generalization is the \textit{Rademacher complexity}~\newtext{\citep{gine-84a,koltchinskii-00a,koltchinskii-01a,bartlett-01a,bartlett-02a}.}
Notably, the Rademacher complexity of a hypothesis class~$\mathcal{W}$ is defined with respect to a given data set (although an average version, where an expectation is taken over the data set, is commonly used).
We now give the definition of Rademacher complexity~\citep[Chap.~26]{shalev-shwartz-14a}. %
\begin{dfn}[Rademacher complexity]
Let~$\traindata\in \dataspace^n$ be a vector of data samples and let~$\ell: \mathcal{W}\times \dataspace\rightarrow\reals^+$ be a loss function.
Let~$\sigma_i$ for $i \in [n]$ be independent Rademacher random variables, so that~$P_{\sigma_i}[\sigma_i = -1]=P_{\sigma_i}[\sigma_i = +1]=1/2 $.
Then, the Rademacher complexity of the function class~$\mathcal{W}$ with respect to~$\traindata$ and~$\ell(\cdot,\cdot)$ is given by
\begin{equation}
    \mathrm{Rad}_{\traindata}(\mathcal{W}) = \frac{1}{n}\Ex{P_{\sigma_1\dots\sigma_n}}{\sup_{w\in\mathcal{W}} \sum_{i=1}^n\sigma_i \ell(w,Z_i) }.
\end{equation}
\end{dfn}
To get some intuition for the Rademacher complexity, one can imagine splitting the data set $\traindata$ into a training set and a test set uniformly at random.
What the Rademacher complexity measures, in a worst-case sense over the hypothesis class, is how big the discrepancy between the loss on the training set and the loss on the test set will be on average.
With this interpretation, it is easy to see how the Rademacher complexity is tied to generalization: it is almost a generalization measure by definition.
In the following theorem, the connection is made more specific~\citep[Thm.~26.5]{shalev-shwartz-14a}.

\begin{thm}[Generalization guarantee from Rademacher complexity]

Assume that, for all $z\in\dataspace$ and all $w\in\mathcal{W}$, we have that $\ell(w,z)\in [0,1]$. With probability at least~$1-\delta$ over $P_{\traindata}$, for all $w\in\mathcal{W}$,
\begin{equation}
    \poplossw - L_{\traindata}(w) \leq 2 \mathrm{Rad}_{\traindata}(\mathcal{W}) + \sqrt{\frac{2\log (2/\delta)}{n}}.
\end{equation}
\end{thm}
A similar bound holds when the sample-dependent Rademacher complexity is replaced by its expectation under~$P_{\traindata}$.

As discussed by~\citet[Part IV]{shalev-shwartz-14a}, the Rademacher complexity can be used to derive generalization bounds for relevant hypothesis classes, such as support vector machines, and can also be used to provide tighter bounds for classes with finite VC dimension.
One issue with the Rademacher complexity is that, while being data-dependent, it is still a worst-case measure over the hypothesis class.
This may typically lead to generalization estimates for modern machine learning algorithms that are overly pessimistic.

\section{Generalization Bounds from Algorithmic Stability}\label{sec:alg-stability-bounds}

We conclude our overview of generalization bounds by discussing an example that takes the learning algorithm into account, namely bounds based on algorithmic stability~\citep{rogers-78a,devroye-79a}.
As for the section on uniform convergence, we will only provide a very short presentation to provide context for upcoming chapters, as an exhaustive discussion is beyond our scope.

The intuition behind generalization bounds based on algorithmic stability is roughly as follows:
if the selected output hypothesis does not depend too strongly on the specific training data it is based on, it should generalize well to unseen samples.
Making this intuition precise, and specifically formalizing the notion of ‘‘strong dependence,'' leads to several different notions of stability that can be related to generalization performance.
In this section, we will focus only on uniform stability, as studied by, \eg,~\citet[Def.~6]{bousquet-02a}.
There is, however, a whole host of alternatives that have been studied in the literature (see, \eg, the works of~\citealp{kutin-02a} and~\citealp{rakhlin-05a}).
As shown by~\citet{shalev-shwartz-10a}, there is also a fundamental relation between stability and uniform convergence in settings beyond standard supervised classification and regression.

We now present a generalization bound for deterministic learning algorithms that satisfy uniform stability~\citep[Def.~6]{bousquet-02a}.
\begin{thm}[Uniform stability and generalization]\label{thm:pointwise-hypothesis-stability-gen}
We denote~$\traindata^{\backslash i}=(Z_1,\dots,Z_{i-1},Z_{i+1},\dots,Z_n)$, and let~$W(\traindata)\in\mathcal W$ denote the output of a deterministic learning algorithm given a training set~$\traindata$.
Assume that the learning algorithm has uniform stability~$\beta$ in the sense that, for all~$\traindata \in \dataspace^n$ and all~$i\in [n]$,
\begin{equation}
\max_{z'\in\dataspace} \left\{ \abs{ \ell( W(\traindata), z') - \ell( W(\traindata^{\backslash i}), z')  } \right\} \leq \beta.
\end{equation}
Then, with probability at least~$1-\delta$ under~$P_{\traindata}$,
\begin{equation}
L_{P_Z}(W(\traindata)) - L_{\traindata}(W(\traindata)) \leq 2\beta + (4n\beta+1)\sqrt{\frac{\log\frac1\delta}{2n}}.
\end{equation}
\end{thm}
For many stable algorithms, such as linear regression and classification with support vector machines, the stability parameter~$\beta$ decays with~$n$, implying that the bound in \cref{thm:pointwise-hypothesis-stability-gen} approaches zero as the number of training samples increases.
For further details, including the relation to regularization, see, for instance,~\citet[Chapter~13]{shalev-shwartz-14a}.

While we will not discuss them in detail, other approaches to generalization have been taken in the literature, for instance, based on margins~\citep{shawetaylor-99a} and norms~\citep{neyshabur-15a}.

\section{Outline}

This monograph is structured as follows.
In Part~I, comprising \crefrange{chap:it-approach-to-gen}{chap:cmi}, we cover the foundations of information-theoretic and PAC-Bayesian generalization bounds for standard supervised learning.
Specifically, %
in \cref{chap:it-approach-to-gen}, we give an intuitive motivation for why information-theoretic tools are suited for the study of generalization, before presenting and proving a first information-theoretic generalization bound as a gentle introduction to the subsequent chapters.
In \cref{chap:tools}, we overview the core tools that are used in deriving generalization bounds in the upcoming chapters, in the form of information measures, change of measure techniques, and concentration inequalities.
We use these tools to derive generalization bounds in expectation in \cref{chap:average} and generalization bounds in probability in \cref{chap:probability}, including PAC-Bayesian generalization bounds.
We conclude Part~I by presenting the conditional mutual information (CMI) framework, as well as the generalization bounds that can be derived through it.

In Part~II, comprising \crefrange{chap:info-complexity}{chap:perspectives}, we turn to applications of the generalization bounds from Part~I, as well as extensions to settings beyond standard supervised learning.
In \cref{chap:info-complexity}, we examine the \emph{information complexity} of several learning algorithms, that is, the value of information measures that the learning algorithms induce.
In \cref{chap:iterative-methods}, we focus specifically on iterative methods, wherein the hypothesis is sequentially updated as training progresses.
This includes neural networks trained through standard methods, such as variants of gradient descent.
In \cref{chap:alternative-settings}, we derive bounds for alternative learning models, namely meta learning, out-of-distribution generalization, federated learning, and reinforcement learning.
Finally, in \cref{chap:perspectives}, we provide concluding remarks and a broader discussion of information-theoretic and PAC-Bayesian generalization bounds as a whole.

\part{Foundations}

\chapter{Information-Theoretic Approach to Generalization}\label{chap:it-approach-to-gen}

In the previous chapter, we introduced the generalization problem and hinted at an information-theoretic approach to addressing it.
In this chapter, we expand upon this connection.
We begin by providing a short introduction to information theory, and the flavor of results it provides.
While this is only a brief overview of a vast area of study, our goal is to provide a glimpse of the field, which can serve to motivate and contextualize the coming results.
After this, we clarify why information theory is a suitable starting point for studying generalization, before finishing the chapter by presenting and proving our first information-theoretic generalization bound.
This serves as a warmup for the following chapters, since it allows us to introduce the general tools and concepts with a concrete, simple example.

\section{An Exceedingly Brief Introduction to Information Theory}\label{sec:brief-intro-to-it}

Information theory, as originally developed by Claude E. Shannon~\citep{shannon} in the late 1940s and early 1950s, provides a rigorous mathematical framework for representing, processing, storing, and transferring information.
Many information-theoretic quantities turn out to characterize fundamental limits for this: the entropy characterizes the minimum compressed size at which information can be stored under a perfect-reconstruction requirement; the relative entropy measures the same under a distribution mismatch; and the mutual information characterizes the limit at which information can be reliably transferred over an unreliable medium.
The definitions of these quantities are provided in \cref{sec:info-measures}.

In the last decades, the information-theoretic approach of seeking fundamental limits without imposing complexity constraints has found applications in many fields beyond data transmission and storage, including statistical estimation, sparse recovery, and adaptive data analysis.
In this monograph, we will see how information-theoretic quantities arise naturally when seeking analytic characterizations of the generalization error of randomized algorithms in the supervised learning setting.

\section{Why Information-Theoretic Generalization Bounds?}\label{sec:why-information-theoretic}

But why is generalization in machine learning related to information theory?
Intuitively, generalization should occur when the learning algorithm captures the relevant aspects of the training data, but disregards irrelevant factors.
In a sense, this can be seen as a variant of Occam's razor, which says that among learners that perform well on the training set, the one that provides the simplest explanation is to be preferred.
One way of interpreting what simplicity means is to say that the learner that extracts the least amount of information from the training data is the simplest one.
Information-theoretic generalization bounds make this intuition precise by characterizing the generalization error of (randomized) learning algorithms in terms of information-theoretic metrics.
\newtext{Crucially, unlike the bounds based on uniform convergence in \cref{sec:uniform-convergence}, these information-theoretic bounds do not solely aim to measure the complexity of the hypothesis class under consideration.
Instead, they also incorporate dependence on the specific learning algorithm and data distribution.
In \crefrange{chap:average}{chap:cmi}, we provide several such results, and discuss their features in terms of assumptions, tightness in various situations, derivations, and relations between them.}

\newtext{Beyond this intuitive appeal, the framework of information-theoretic generalization bounds has several other attractive features.
First, it can be used to recover bounds which were originally derived using a wide range of other approaches.
In this sense, information-theoretic bounds offer a certain unifying (albeit not all-encompassing) perspective.
This is covered in more detail in \cref{chap:info-complexity}.
Moreover, information-theoretic and PAC-Bayesian bounds have been used to obtain some of the tightest numerical performance guarantees for neural networks to date, indicating a promising avenue for furthering our understanding of these models.
New learning algorithms can also be devised on the basis of minimizing the generalization bounds, paving the way for \emph{self-certified} learning---\ie, learning algorithms that use the training data to both learn a hypothesis and provide performance guarantees.
We expand on these points in \cref{chap:iterative-methods}.
Finally, as we cover in \cref{chap:alternative-settings}, the information-theoretic framework is flexible enough to accommodate many settings of interest, beyond the standard learning setting introduced in \cref{chap:introduction}.
}%

As an introduction, we begin by proving a simple information-theoretic bound in \cref{sec:basicitbound}.
This enables us to provide concrete instantiations of the tools and concepts that are relevant for deriving and interpreting information-theoretic bounds, before exploring these tools in greater generality in \cref{chap:tools}.
While the results that are available in the literature vary widely in their details, the general recipe for obtaining them typically includes two crucial steps.
The first step is a \emph{change of measure}, which we cover in \cref{sec:change}.
The second step is a \emph{concentration inequality}, which we discuss in \cref{sec:concentration}.
Variations of these two steps yield the generalization bounds we will discuss in \cref{chap:average} and \cref{chap:probability}.

When studying generalization, the main object of interest is the error event, which occurs when the hypothesis incurs a large loss on new data samples---\ie, the hypothesis does not generalize well.
The probability distribution that governs this event is typically not amenable to direct analysis, because the hypothesis and training sample are jointly distributed.
For this reason, it is convenient to \emph{change measure} to an auxiliary distribution that is easier to analyze.
The cost of replacing the original distribution with the auxiliary distribution is quantified by an \emph{information measure}, which can be seen as gauging the discrepancy between the two probability distributions.

The auxiliary probability distribution is chosen so that, under this distribution, we can control the error event. This is done by  utilizing \emph{concentration of measure inequalities}, which, roughly speaking, characterize the degree to which a random variable tends to deviate from its mean.
Thus, by changing measure to a more easy-to-handle auxiliary distribution and applying concentration of measure results, we can obtain generalization bounds expressed through information measures.

\section{A First Information-Theoretic Generalization Bound}
\label{sec:basicitbound}

To start us off gently within the broad topic of information-theoretic generalization bounds, we begin by giving a specific instantiation of an average bound.
Specifically, in this section, we will present a generalization bound based on the sub-Gaussianity of bounded random variables and the \donskervtext~\citep{donsker-75a,csiszar-75a}.
Throughout, we will highlight the role played by the different proof ingredients, focusing on intuition and providing indications of how these ingredients can later be generalized.

\subsection{The Bound}
Recall that the notation used here, and throughout this monograph, is detailed in \cref{sec:notation}.
Before stating our first information-theoretic bound, we need to define the \emph{relative entropy} between two probability distributions, also known as the Kullback-Leibler (KL) divergence.
We also need the definition of the \emph{mutual information}, which is the relative entropy between the joint distribution of two random variables and the product of their marginals.

\begin{dfn}[Relative entropy and mutual information]\label{def:relent-intro-chap}
Consider two probability distributions $P$ and $Q$ defined on a common measurable space such that~$P$ is absolutely continuous with respect to~$Q$, denoted by~$P\ll Q$. The relative entropy between~$P$ and~$Q$ is given by
\begin{equation}
\relent{P}{Q} = \Ex{P}{ \log \rnPQ }.
\end{equation}
Here, $\rnPQ $ denotes the Radon-Nikodym derivative of~$P$ with respect to~$Q$.
If~$P$ is not absolutely continuous with respect to~$Q$, the Radon-Nikodym derivative is undefined and we let~$\relent{P}{Q} =\infty$.
We will give a precise definition of the Radon-Nikodym derivative in Theorem~\ref{thm:radon-nikodym}, but for now, it is sufficient to think of it as a likelihood ratio.
The relative entropy is non-negative, so that~$\relent{P}{Q}\geq 0$, with equality if and only if~$P=Q$.
While it is tempting to interpret the relative entropy as a distance between~$P$ and~$Q$, it is not a metric: it is not symmetric and it does not satisfy the triangle inequality.

For two random variables $X$ and $Y$ with joint distribution~$P_{XY}$ and product of marginals~$P_XP_Y$, the mutual information between~$X$ and~$Y$ is
\begin{equation}
I(X;Y)=\relent{P_{XY}}{P_XP_Y}.
\end{equation}
\end{dfn}
\noindent Note that, if~$X$ and~$Y$ are independent,~$P_{XY}=P_XP_Y$, and~$I(X;Y)=0$.
Also note that, if~$X$ is a continuous random variable and~$Y=f(X)$ is a deterministic function of~$X$, we have~$I(X;Y)=\infty$.

We are now ready to state our first information-theoretic generalization bound.

\begin{thm}\label{thm:first-it-bound}
Consider a learning setting where the loss function is bounded, and satisfies~$\ell(w,z)\in[0,1]$ for all~$(w,z)\in\mathcal W\times\mathcal Z$. Then,
\begin{equation}\label{eq:first-it-bound}
 \abs{\Ex{\jointdistro}{\gen}} = \abs{\Ex{\jointdistro}{\poploss-\trainloss}}\leq \sqrt{ \frac{I(W;\traindata) }{2n}  }.
\end{equation}
\end{thm}
Before proceeding to the proof of this theorem, let us examine the components of~\eqref{eq:first-it-bound}.
After removing the absolute value, the average population loss can be upper bounded by two terms: the training loss and a so-called complexity term.
Thus, to have any hope of obtaining a small bound on the population loss, we need to achieve a small training loss.
Now, consider the complexity term, \ie, the right-hand side of~\eqref{eq:first-it-bound}, whose key component is the mutual information $I(W;\traindata)$ between the hypothesis and the training data.
On one hand, if the learning algorithm is oblivious to the training data, so that~$\conddistro=P_W$, the mutual information will vanish, and the population loss is guaranteed to equal the training loss (on average). This is not surprising, since the training loss in this case is an unbiased estimator of the population loss.
On the other hand, if the hypothesis is a deterministic function of the training data and both~$W$ and~$\traindata$ are continuous random variables, the mutual information is unbounded, and \cref{thm:first-it-bound} provides a vacuous guarantee, meaning that the upper bound is trivial.

Typically, the value of the mutual information depends on the size~$n$ of the training set~$\traindata$.
For \cref{thm:first-it-bound} to give bounds that improve with~$n$, the rate of increase of the mutual information with~$n$ has to be sublinear.
If this is the case, the complexity term in \cref{thm:first-it-bound} goes to 0 as~$n$ approaches infinity, and we guarantee that the population loss of the hypothesis we learn is arbitrarily close to its training loss, given sufficient samples.

It is tempting to compare this result to the channel coding problem, mentioned in \cref{sec:brief-intro-to-it}.
There, a transmitter encodes a message as a codeword~$X$, which after transmission over a noisy channel~$P_{Y\vert X}$ gives rise to the output~$Y$, which is observed by the receiver whose aim is to decode the original message.
From a mathematical standpoint, we can identify the training data~$\traindata$ with the codewords~$X$, the learning algorithm~$\conddistro$ with the channel law~$P_{Y\vert X}$, and the hypothesis~$W$ with the output~$Y$.
For channel coding, the communication capacity of a noisy channel is given by the mutual information between the input and output, maximized over the input distribution~$P_X$.
By maximizing over the analogue of~$P_X$ in~\eqref{eq:first-it-bound}, \ie, the distribution of the training data~$P_{\traindata}$, we obtain a worst-case upper bound for the generalization error.\footnote{In order to match the setting of learning with independent and identically distributed data, we must restrict ourselves to product distributions in this maximization.}

However, despite these superficial similarities between the two settings, there are fundamental differences between them.
For channel communication, the conditional distribution from input to output is considered fixed, and the aim is to find an input that maximizes the mutual information.
For machine learning, the input distribution is considered fixed, and the goal is to select a conditional distribution from input to output that minimizes the population loss.
More importantly, while the mutual information has a very specific operational meaning in channel coding---it characterizes the maximal rate of reliable communication---its role in \cref{thm:first-it-bound} is much more spurious.
Indeed, it appears as an upper bound simply as a consequence of the particular change of measure that we use.
As we will see in the coming chapters, other changes of measure give rise to upper bounds in terms of other information measures.

\subsection{Proof of the Bound}
We now proceed with proving the information-theoretic generalization bound in \cref{thm:first-it-bound}.
For this, we will need two results: the Donsker-Varadhan variational formula for the relative entropy and a concentration result  for bounded random variables.\footnote{Note that we use the term ‘‘concentration result'' quite liberally to include bounds on the moment-generating function, as such bounds imply concentration inequalities in the more strict sense.}
As previously mentioned, these are the two main ingredients needed for deriving most information-theoretic generalization bounds.
We state the results here without proof and not in the fullest generality possible, and defer further details to \cref{sec:change} and \cref{sec:concentration} respectively.

\begin{thm}[Donsker-Varadhan variational formula]\label{thm:firstbound-donskervaradhan}
Let~$P$ and~$Q$ be two probability distributions on a common measurable space~$\mathcal X$ such that~$P\ll Q$. Then, for every~$f:\mathcal{X}\rightarrow \reals$ such that~$\Ex{Q}{e^{ f(X)}}<\infty$,
\begin{equation}
\relent{P}{Q} \geq \Ex{P}{f(X)} -\log \Ex{Q}{e^{ f(X)}}.
\end{equation}
\end{thm}
\begin{thm}[Concentration of bounded random variables]\label{thm:sub-exponential-bounded}
Let~$X_i$, for $i\in[n]$, be independent random variables distributed according to $P_X$ with range~$[0,1]$ and $\Ex{}{X_i}=\mu$.
Let $X=\sum_{i=1}^n X_i/n$ denote the average of the $X_i$. Then, for every~$\lambda\in\reals$,
\begin{equation}
\log\Ex{}{ e^{\lambda (\mu-X)} } \leq \frac{\lambda^2}{8n}.
\end{equation}
\end{thm}

To get an idea of how these results can be used, consider a situation where we want to know how a random variable~$X$ behaves under the distribution~$P$, but where it is hard to perform an analysis dealing with~$P$ directly.
Then, if we have an auxiliary distribution~$Q$ that allows easier analysis---for instance, if \cref{thm:sub-exponential-bounded} holds under~$Q$---we can first use \cref{thm:firstbound-donskervaradhan} with~$f(X)=\lambda(\mu-X)$ to change distribution from~$P$ to~$Q$, at the price of a relative entropy, and then apply \cref{thm:sub-exponential-bounded} to bound the term~$\log \Ex{Q}{e^{\lambda (\mu-X)}}$.
This is exactly what we will do to prove \cref{thm:first-it-bound}.

\begin{proof}[Proof of \cref{thm:first-it-bound}]
We first apply \cref{thm:firstbound-donskervaradhan} with~$X=(W,\traindata)$, $f(W,\traindata)=\lambda\gen$ for~$\lambda\in\reals$, $P=\jointdistro$, and $Q=\productdistro$.
This implies that
\begin{equation}\label{eq:specific-instance-of-gen-exp-ineq}
\Ex{\jointdistro}{\lambda\gen}\\
  \leq \log \Ex{\productdistro}{e^{\lambda \gen}} + \relent{\jointdistro}{\productdistro}.
\end{equation}
Since~$\relent{\jointdistro}{\productdistro}=I(W;Z)$, we now see how the mutual information arises from this change of measure.
Next, note that, for any fixed~$w\in\mathcal W$,
\begin{align}
\genw&=\poplossw- \frac{1}{n} \sum_{i=1}^n \ell(w,Z_i).%
\end{align}
Since the training losses~$\ell(w,Z_i)$ are bounded to~$[0,1]$ and identically distributed with mean~$\poplossw$, we can invoke \cref{thm:sub-exponential-bounded} to conclude that, for every~$w\in \mathcal{W}$ and~$\lambda \in \reals$,
\begin{equation}\label{eq:deriv_first_pacb_bound_concentration}
\Ex{\traindistro}{e^{\lambda \genw}}  \leq  \exp\lefto(\frac{\lambda^2}{8n}\right).
\end{equation}
By averaging~\eqref{eq:deriv_first_pacb_bound_concentration} over~$P_W$, we obtain
\begin{equation}\label{eq:deriv_first_pacb_bound_concentration_specified}
\log \Ex{\productdistro}{e^{\lambda \gen}} \leq  \frac{\lambda^2}{8n}. %
\end{equation}
We now see that, through the concentration inequality in \cref{thm:sub-exponential-bounded}, we are able to control the generalization gap under the distribution~$\productdistro$.
By inserting~\eqref{eq:deriv_first_pacb_bound_concentration_specified} into~\eqref{eq:specific-instance-of-gen-exp-ineq}, we obtain, for~$\lambda>0$,
\begin{equation}\label{eq:first-it-bound-last-step}
\Ex{\jointdistro}{\gen}
\leq \frac{\lambda}{8n}+  \frac{I(W;\traindata)}{\lambda}.
\end{equation}
All that remains is to select the hitherto unspecified parameter~$\lambda$, which we do by minimizing the right-hand side of~\eqref{eq:first-it-bound-last-step}.
To obtain the absolute value, we perform the same procedure for~$\lambda<0$.
After this, the result in~\eqref{eq:first-it-bound} follows.
\end{proof}

The proof of \cref{thm:first-it-bound} illustrates the key tools needed to establish information-theoretic generalization bounds.
In this example, the change of measure was performed via the Donsker-Varadhan variational formula, the resulting information metric is the mutual information $I(W;\traindata)$, and the concentration of measure relied on the boundedness of the involved random variables.
In the remainder of this monograph, we will present a more general framework for obtaining information-theoretic generalization bounds, through which alternative techniques can be used to obtain tighter bounds or bounds that hold under different assumptions than the ones in this section.

\section{Bibliographic Remarks and Additional Perspectives}\label{sec:bib-remarks-intro-to-it}

The specific bound that we present in \cref{thm:first-it-bound}, along with its proof, are based on the work of~\citet{xu-17a}, which itself extended the results of \citet{russo-16a} to a more general setting.
Arguably, the core of the approach dates back to the work of~\citet{shawetaylor-97a}, who derived PAC bounds for Bayesian predictors in terms of a ‘‘luckiness'' function (which is similar to a prior).
This was extended to more general settings by~\citet{mcallester-98a}, leading to a bound that is very similar in form to the bounds discussed in this monograph.
The proof technique, however, is quite different: it relies on the ‘‘quantifier reversal lemma'' which, in some sense, plays the role of a change of measure.
This PAC-Bayesian strand of the literature then flourished, with generalizations and tighter bounds by, to only give some examples,~\citet{langford-01a},~\citet{mcallester-03a},~\citet{audibert-04a}, and~\citet{catoni-07a}, with proofs of a similar form as we discuss here.
A more extensive overview of the PAC-Bayesian literature is given in \cref{chap:probability}, specifically in \cref{sec:pac-bayesian-bounds,sec:bib-remarks-probability}.
The bounds in the PAC-Bayesian literature focused primarily on bounded losses, and in particular, the~$0-1$ loss.
Around the same time,~\citet{zhang-06a} developed generalization bounds for generic loss functions based on a result termed the ‘‘information exponential inequality.''

We now come to the work of~\citet{russo-16a}, who focused on adaptive data analysis, rather than generalization bounds.
Specifically, given the data, an analyst computes~$m$ different measurements~$\phi=\{\phi_i\}_{i\in[m]}$.
Then, based on the values of these measurements, they report~$\phi_T$ for some~$T\in[m]$.
Since the choice of the measurement to report depends on the measurements themselves, this can introduce a significant bias.
The main result of~\citet{russo-16a} is a bound on this bias in terms of the mutual information~$I(T;\phi)$, under the assumption that the measurements are sub-Gaussian.
This setting can be seen to be equivalent to a statistical learning setting, where the measurements correspond to losses and the index~$T$ corresponds to a hypothesis from a finite set.
While these developments appear to be largely independent from the PAC-Bayesian literature,~\citet{russo-16a} noted the resemblance to PAC-Bayesian bounds, stating that it would be interesting to explore the connections between PAC-Bayes and adaptive data analysis.
\citet{xu-17a} made the connection between statistical learning and the results of~\citet{russo-16a} precise, and in particular, extended the argument to uncountable hypothesis classes.
Prior to this, \citet{raginsky-16a} derived generalization bounds in terms of information-theoretic versions of algorithmic stability, where the bounds were given in terms of the mutual information between the hypothesis and a single training datum, given the rest of the samples, where the sub-Gaussianity assumption was slightly different.

Since we have so far only provided an initial introduction to information-theoretic generalization bounds, we will defer a more detailed discussion and comparison of these results to \cref{chap:average,chap:probability}.

Another tool from information theory that has received significant attention in machine learning is the information bottleneck~\citep{tishby-99a}.
While we will not discuss it much in the remainder of this monograph, we will conclude this chapter with a discussion of the application of the information bottleneck in statistical learning.
Specifically, consider two random variables~$X$ and~$Y$, where~$X$ is an input and~$Y$ is an output.
Assume that we want to find a representation~$T$, which is a compressed version of~$X$, but which should be useful in predicting~$Y$.
The idea of the information bottleneck method is that we want to set the conditional distribution~$P^*_{T\vert X}$ of~$T$ given~$X$ so that, for some parameter~$\beta>0$,
\begin{equation}
P^*_{T\vert X} = \sup_{P_{T\vert X}} \left\{   \beta I(T;Y) - I(X;T) \right\}.
\end{equation}
Here,~$I(T;Y)$ captures the \emph{sufficiency} of~$T$, in the sense that it is informative of~$Y$, while~$I(X;T)$ measures the \emph{minimality} of~$T$, in the sense that it only captures aspects of~$X$ that are necessary for predicting~$Y$.
The parameter~$\beta$ controls the trade-off between these two objectives.
While originally motivated by compression in information theory,~\citet{tishby-17a} argued that the information bottleneck can also be used to explain phenomena in statistical learning, and in particular neural networks.
Specifically, let~$T$ denote the activations of an intermediate layer in a neural network.
Through empirical studies,~\citet{tishby-17a} argued that neural network training consisted of a \emph{fitting} phase, where both~$I(T;Y)$ and~$I(X;T)$ increase and the network achieves good predictive performance, followed by a \emph{compression} phase, where~$I(T;Y)$ remains constant but~$I(X;T)$ decreases, so that the network learns a compressed, well-generalizing representation.
\citet{achille-18a} developed this further to derive a regularized training objective that aims to promote learning minimal representations, and connected this with PAC-Bayesian theory.
The existence of the fitting and compression phases was questioned by~\citet{saxe-18a}, who argued that these empirical phenomena do not occur in general, and depend heavily on implementation details.
More discussion on the information bottleneck and its connection to learning can be found in the works of~\citet{ziv-20a} and~\citet{geiger-20a}, \newtext{as well as \citet{kawaguchi-23a}, who establish generalization bounds.}

\chapter{Tools}\label{chap:tools}

The proofs of the large majority of information-theoretic generalization bounds in the literature have two key steps in common: a \textit{change of measure} and a \textit{concentration of measure}.
In the previous chapter, this was illustrated with a concrete example, leading to our first information-theoretic generalization bound.
As we will see in \cref{sec:pac-bayesian-bounds}, these same tools are also at the heart of the PAC-Bayesian approach, and these two strands can be unified through this lens.

In this chapter, we will introduce the tools that will be used to derive PAC-Bayesian and information-theoretic generalization bounds in the remainder of the monograph in more detail and generality.
Specifically, we will define some common information measures in~\cref{sec:info-measures}, discuss change of measure techniques in~\cref{sec:change}, and present concentration of measure in \cref{sec:concentration}.

\section{Information Measures}\label{sec:info-measures}

In \cref{thm:first-it-bound}, we found that the generalization error of a randomized learning algorithm can be controlled by the mutual information between the training data and the hypothesis.
The mutual information is just one example of an \emph{information measure}.
Formally, given a measurable space~$\mathcal X$ and the associated family~$\mathcal{M}(\mathcal X)$ of probability measures on~$\mathcal X$, an (average) information measure is a mapping~$\mathrm{IM}: \mathcal{M}(\mathcal X)\times \mathcal{M}(\mathcal X) \rightarrow \reals$.
Typically, for all~$P\in \mathcal{M}(\mathcal X)$, we have~$\mathrm{IM}(P,P)=0$.
Thus, an information measure is some way to quantify the discrepancy between two probability measures.
Often, these information measures are not metrics in the formal sense, as they may not satisfy symmetry or the triangle inequality.
An example of this is the relative entropy from \cref{def:relent-intro-chap}, which maps the two distributions~$P,Q\in\mathcal M(\mathcal X)$ to~$\relent{P}{Q}=\Ex{P}{\log\frac{\dv P}{\dv Q}}$.
Note that, in general,~$\relent{P}{Q}\neq \relent{Q}{P}$.
In addition to such average information metrics, which only depend on the distributions, we will also consider pointwise versions, which are mappings from~$\mathcal{M}(\mathcal X)^2\times \mathcal X^2$ to~$\reals$.

Throughout information theory and machine learning, such information measures are exceedingly useful and abundant.
In the context of information-theoretic and PAC-Bayesian generalization bounds, they naturally appear in upper bounds on the population loss of learning algorithms, as exemplified by \cref{thm:first-it-bound}.
In this section, we will introduce some information measures along with their properties, which will be useful in later chapters.
For a more detailed review, the reader is referred to, for example,~\citet{cover-06a} and \citet{polyanskiy-22a}, upon which much of the material in this section is based.

A basic building block of many information measures is some kind of likelihood ratio.
For two probability mass functions~$P$ and~$Q$ on a common space~$\mathcal{X}$, their likelihood ratio at a point~$x\in\mathcal{X}$ is defined as~$P(x)/Q(x)$.
Similarly, if~$p$ and~$q$ are probability densities, the likelihood ratio is~$p(x)/q(x)$.
For generic measures~$P$ and~$Q$, this concept is captured by the Radon-Nikodym derivative, denoted by~$\dv P/\dv Q$.
For the cases of discrete or continuous random variables, it reduces to the aforementioned likelihood ratios.
The precise meaning of this object is captured by the Radon-Nikodym theorem, a change of measure that relates probabilities of events under~$P$ with their probabilities under~$Q$.
We will present this result in \cref{thm:radon-nikodym}.
The Radon-Nikodym derivative exists whenever~$P$ is \emph{absolutely continuous} with respect to~$Q$, as described in the following definition.

\begin{dfn}[Absolute continuity]\label{def:absolute-continuity}
A measure~$P$ is absolutely continuous with respect to a measure~$Q$, denoted as~$P\ll Q$, if, for every measurable set~$\setE$ such that~$Q(\setE)=0$, we also have~$P(\setE)=0$.
\end{dfn}

For the special case where~$P=P_{X\!Y}$ and~$Q=P_XP_Y$ are the joint distribution and product of marginal distributions of two random variables~$X$ and~$Y$, we will refer to the logarithm of the Radon-Nikodym derivative as the \emph{information density}.
\begin{dfn}[Information density]
The information density between two random variables~$X$ and~$Y$ with joint distribution~$P_{X\!Y}$ and marginal distributions~$P_X$ and~$P_Y$ is given by
\begin{equation}
\infdensxy = \log \rnPQxy,
\end{equation}
provided that~$P_{XY}\ll P_XP_Y$.
\newtext{The conditional information density between~$X$ and~$Y$ given~$Z$ is
\begin{equation}\label{eq:conditional_info_density}
\condinfdensxyz = \log \rnPQxyz,
\end{equation}
provided that~$P_{XYZ}\ll P_{X\vert Z}P_{Y\vert Z} P_Z$.}
\end{dfn}

A fundamental information-theoretic quantity is the \emph{entropy}.
\begin{dfn}[Entropy]\label{def:entropy}
Let~$X$ be a discrete random variable on~$\mathcal X$ with probability mass function~$P_X$.
The entropy of~$X$ is given by
\begin{equation}
H(X)  = \sum_{x\in \mathcal X} P_X(x) \log \frac{1}{P_X(x)}.
\end{equation}
Furthermore, let~$Y$ be a discrete random variable on~$\mathcal Y$ with probability mass function~$P_Y$, such that the joint distribution of~$X$ and~$Y$ is~$P_{XY}$.
Then, the conditional entropy of~$X$ given~$Y$ is
\begin{equation}
H(X\vert Y) =  \sum_{x,y\in \mathcal X\times\mathcal Y} P_{XY}(x,y) \log \frac{P_Y(y)}{P_{XY}(x,y)}.
\end{equation}
The entropy satisfies the following key properties:
\begin{enumerate}
\item \textit{Non-negativity:} $H(X)\geq 0$, and equality holds if and only if~$P_X(x)=1$ for some~$x\in\mathcal X$.
\item \textit{Maximum:} $H(X)\leq \log(\abs{\mathcal X})$, and equality holds if and only if~$P_X$ is the uniform distribution on~$\mathcal X$.
\item \textit{Chain rule:} $H(X,Y)=H(Y)+H(X\vert Y)=H(X)+H(Y\vert X)$.
\item \textit{Conditioning reduces entropy:} $H(X\vert Y)\leq H(X)$, and equality holds if and only if~$X$ and~$Y$ are independent.
\end{enumerate}
\end{dfn}

There exists an extension to continuous random variables in the form of the differential entropy.
\begin{dfn}[Differential entropy]\label{def:differential-entropy}
Let~$X$ be a continuous random variable on~$\mathcal X$ with probability density function~$p_X$.
The differential entropy of~$X$ is given by
\begin{equation}
h(X)  = \int_{\mathcal X}  p_X(x) \log \frac{1}{p_X(x)} \dv x .
\end{equation}
Furthermore, let~$Y$ be a continuous random variable on~$\mathcal Y$ with probability density function~$p_Y$, such that the joint density of~$X$ and~$Y$ is~$p_{XY}$.
Then, the conditional differential entropy of~$X$ given~$Y$ is
\begin{equation}
h(X\vert Y) =  \int_{\mathcal X\times\mathcal Y}  p_{XY}(x,y) \log \frac{p_Y(y)}{p_{XY}(x,y)} \dv x \dv y.
\end{equation}
\end{dfn}
The differential entropy is shift invariant, so that for any~$a\in\reals$,~$h(X)=h(X+a)$.
However, the differential entropy does not satisfy many key properties of its discrete counterpart, such as non-negativity, and it is not scale-invariant, meaning that~$h(aX)\neq h(X)$ in general.

Several key features of both the discrete and differential entropy can be described using the \emph{relative entropy}, sometimes called the Kullback-Leibler (KL) divergence.
This is a very commonly used information measure, which we already introduced in \cref{sec:basicitbound}.
We repeat its definition below.
\begin{dfn}[The relative entropy]
Consider two probability distributions $P$ and $Q$ defined on a common measurable space such that~$P\ll Q$.
The relative entropy between~$P$ and~$Q$ is given by
\begin{equation}
\relent{P}{Q} = \Ex{P}{ \log \rnPQ }.
\end{equation}
If~$P$ is not absolutely continuous with respect to~$Q$, the Radon-Nikodym derivative is undefined and we set~$\relent{P}{Q} =\infty$.

Given a distribution~$P_X$ on~$\mathcal{X}$ and two conditional distributions~$P_{Y\vert X}$ and~$Q_{Y\vert X}$ on~$Y$ given~$X$, the conditional relative entropy between~$P_{Y\vert X}$ and~$Q_{Y\vert X}$ given~$P_X$ is defined as
\begin{equation}
\conrelent{P_{Y\vert X}}{Q_{Y\vert X}}{P_X} = \Ex{P_X}{\relent{P_{Y\vert X}}{Q_{Y\vert X}}}.
\end{equation}
\end{dfn}
The relative entropy satisfies a useful property called the \emph{chain rule}.
\begin{thm}[The chain rule of relative entropy]\label{thm:relent-chain-rule}
Given the distributions~$P_{X\!Y}=P_XP_{Y\vert X}$ and~$Q_{X\!Y}=Q_XQ_{Y\vert X}$, we have
\begin{equation}
\relent{P_{X\!Y}}{Q_{X\!Y}} = \conrelent{P_{Y\vert X}}{Q_{Y\vert X}}{P_X} + \relent{P_X}{Q_X}.
\end{equation}
\end{thm}
When~$P$ and~$Q$ are a joint distribution and product of marginals of two random variables, the relative entropy between~$P$ and~$Q$ is referred to as the \emph{mutual information} between the random variables.
\begin{dfn}[Mutual information]
The mutual information between two random variables~$X$ and~$Y$ with joint distribution~$P_{X\!Y}$ and marginal distributions~$P_X$ and~$P_Y$ is given by
\begin{equation}
I(X;Y) = \relent{P_{XY}}{P_XP_Y} = \Ex{P_{X\!Y}}{\infdensxy} . %
\end{equation}
The conditional mutual information between two random variables~$X$ and~$Y$ given~$Z$ is given by
\begin{equation}
I(X;Y\vert Z)=\conrelent{P_{X\!Y\vert Z}}{P_{X\vert Z}P_{Y\vert Z}}{P_Z} = \Ex{P_{X\!Y\!Z}}{\condinfdensxyz}.
\end{equation}
\end{dfn}
We now see the motivation behind the name information density---its average is the mutual information (with an analogous correspondence for the conditional information density).
The mutual information is one of the most fundamental quantities in information theory, and famously characterizes the capacity of a noisy communication channel.
Recently, as discussed in \cref{sec:basicitbound}, the mutual information has garnered interest in the statistical learning community as a measure of generalization.

\newtext{Since it is a relative entropy, the mutual information inherits a version of the chain rule for relative entropy, which follows directly from \cref{thm:relent-chain-rule}.
\begin{thm}[The chain rule of mutual information]\label{thm:mi-chain-rule}
Consider three random variables $X$, $Y$, and $Z$. Then,
\begin{equation}
I(X,Y;Z)=I(X;Z) + I(Y;Z\vert X)=I(Y;Z)+I(X;Z\vert Y).
\end{equation}
\end{thm}
}

For discrete random variables, the mutual information can be expressed in terms of entropy as~$I(X;Y)=H(X)-H(X\vert Y)=H(Y)-H(Y\vert X)$.
For continuous random variables, it can be expressed in terms of the differential entropy as~$I(X;Y)=h(X)-h(X\vert Y)=h(Y)-h(Y\vert X)$.

The relative entropy is a special case of a wider class of information measures called~$f$-divergences.
\begin{dfn}[$f$-divergence]\label{def:f-divergence}
Let~$P$ and~$Q$ be two probability distributions on a common measurable space~$\mathcal X$ such that~$P\ll Q$.
Let~$f:(0,\infty)\rightarrow \reals$ be a convex function with~$f(1)=0$, extended so that~$f(0)=\lim_{x\rightarrow 0^+}f(x)$.
Then, the~$f$-divergence between~$P$ and~$Q$ is defined as
\begin{equation}
\alpharelent{f}{P}{Q} = \Ex{Q}{ f\lefto( \rnPQ \right) }.
\end{equation}
\end{dfn}
By setting~$f(x)=x\log x$, we recover the relative entropy.
Other notable examples include the total variation~$\TV(P,Q)=\Ex{Q}{\abs{\rnPQ-1}}/2$, obtained by setting~$f(x)=\abs{x-1}/2$, and the~$\chi^2$-divergence~$\chisqdiv{P}{Q}=\Ex{Q}{(\rnPQ-1)^2}$, obtained by setting~$f(x)=(x-1)^2$.

For many pairs of~$f$-divergences, one can establish comparison inequalities~\citep[Sec.~7.5]{polyanskiy-22a}.
For our purposes, comparison inequalities involving the relative entropy and total variation will be of particular importance.
We state two such inequalities below~(which are discussed more by, \eg, \citealp{canonne-22a}).
\begin{thm}[Pinsker's inequality and the Bretagnolle-Huber (BH) inequality]\label{thm:pinsker-bh}
Let~$P$ and~$Q$ be two probability distributions such that~$P\ll Q$.
Then, Pinsker's inequality states that (see, \eg,~\citealp[Thm.~7.9]{polyanskiy-22a})
\begin{equation}\label{eq:thm-pinsker}
\TV(P,Q) \leq \sqrt{\frac{\relent{P}{Q} }{2 } }.
\end{equation}
Furthermore, the BH inequality states that~\citep{bretagnolle-78a}
\begin{equation}\label{eq:thm-bh}
\TV(P,Q) \leq \sqrt{1 - \exp(-\relent{P}{Q}) }.
\end{equation}
\end{thm}
We now review some useful properties of~$f$-divergences. For proofs, see~\citet[Thm.~7.4 and 7.5]{polyanskiy-22a}.

\begin{thm}[Properties of~$f$-divergences.]\label{thm:f-div-props}
For every~$f$-divergence, the following properties hold:
\begin{enumerate}
\item \textit{Non-negativity:} $\alpharelent{f}{P}{Q}\geq 0$, and equality holds if and only if~$P=Q$.
\item \textit{Data-processing:} Let~$P_X$ and~$Q_X$ be two distributions on~$\mathcal{X}$, and let~$P_Y$ and~$Q_Y$ be the corresponding distributions on~$\mathcal{Y}$ induced by a kernel~$P_{Y\vert X}$, that is,~$P_Y(\setE)=\int_\mathcal{X}\dv P_X(x) P_{Y\vert X}(\setE\vert x)$ and~$Q_Y(\setE)=\int_\mathcal{X}\dv Q_X(x) P_{Y\vert X}(\setE\vert x)$ for every measurable set~$\setE\subset \mathcal Y$.
Then,
\begin{equation}
\alpharelent{f}{P_X}{Q_X} \geq \alpharelent{f}{P_Y}{Q_Y}.
\end{equation}
\item \textit{Conditioning increases divergence:} Let~$P_X$ be a distribution on~$\mathcal{X}$, and let~$P_Y$ and~$Q_Y$ be the distributions induced on~$\mathcal{Y}$ by two kernels~$P_{Y\vert X}$ and~$Q_{Y\vert X}$ respectively, \ie~$P_Y(\setE)\!=\!\int_\mathcal{X}\dv P_X(x) P_{Y\vert X}(\setE\vert x)$ and~$Q_Y(\setE)=\int_\mathcal{X}\dv P_X(x) Q_{Y\vert X}(\setE\vert x)$ for every measurable set~$\setE\subset\mathcal Y$.
The \emph{conditional}~$f$-divergence is defined as
\begin{equation}
\alphaconrelent{f}{P_{Y\vert X}}{Q_{Y\vert X}}{P_X} = \Ex{P_X}{\alpharelent{f}{P_{Y\vert X}}{Q_{Y\vert X}}}
\end{equation}
and it satisfies the inequality
\begin{equation}
\alpharelent{f}{P_{Y}}{Q_{Y}} \leq \alphaconrelent{f}{P_{Y\vert X}}{Q_{Y\vert X}}{P_X} .
\end{equation}
\end{enumerate}
\end{thm}

Notably, unlike the relative entropy, general~$f$-divergences do \emph{not} satisfy the chain rule (cf. \cref{thm:relent-chain-rule}).

Another special instance of~$f$-divergences is the R\'enyi divergence, also known as the~$\alpha$-divergence \citep{vanerven-14a}.
\begin{dfn}[R\'enyi divergence]\label{def:renyi-divergence}
Let~$\alpha\in (0,1)\cup (1,\infty)$.
The R\'enyi divergence of order~$\alpha$ between~$P$ and~$Q$ is defined as
\begin{equation}
\alpharelent{\alpha}{P}{Q} = \frac{1}{\alpha-1} \log \Ex{Q}{\lefto(\rnPQ\right)^\alpha}.
\end{equation}
For~$\alpha=1$, motivated by continuity, the R\'enyi divergence of order~$1$ coincides with the relative entropy:
\begin{equation}
\alpharelent{1}{P}{Q}=\relent{P}{Q}.
\end{equation}
The conditional R\'enyi divergence of order~$\alpha$ between~$P_{Y\vert X}$ and~$Q_{Y\vert X}$ given~$P_X$ is
\begin{equation}
\alphaconrelent{\alpha}{P_{Y\vert X}}{Q_{Y\vert X}}{P_X} = \alpharelent{\alpha}{P_{Y\vert X}P_X}{Q_{Y\vert X}P_X}.
\end{equation}
\end{dfn}
When~$P=P_{XY}$ and~$Q=P_XP_Y$ are the joint distribution of two random variables and the product of their marginals respectively and~$\alpha\rightarrow \infty$, the R\'enyi divergence reduces to the maximal leakage, defined below~\citep{issa-16a}.
\begin{dfn}[Maximal leakage]\label{def:maximal-leakage}
The maximal leakage from~$X$ to~$Y$ is defined as
\begin{equation}
\maxleakage{X}{Y} = \log \Ex{P_Y}{ \esssup_{P_X} \rnPQxy }.
\end{equation}
Here, the essential supremum of a measurable function $f(\cdot)$ of a random variable $X$ distributed as~$P_{X}$ is defined as
\begin{equation}
\esssup_{P_{X}} f(X) = \infimum_{a\in \reals}\biggo[P_{X}\big(\{X: f(X)>a\}\big)=0 \bigg] .
\end{equation}
The conditional maximal leakage from~$X$ to~$Y$ given~$Z$ is defined as
\begin{equation}\label{eq:def_cond_maximal_leakage}
 \conmaxleakage{X}{Y}{Z}=
 \log\esssup_{P_Z} \Exop_{P_{X\vert Z}}\lefto[ \esssup_{P_{Y\vert Z} }\rnPQxyz \right].
\end{equation}
\end{dfn}
While the maximal leakage is obtained as the infinite limit of the R\'enyi divergence, the same does not hold for the conditional maximal leakage.
Instead, the conditional maximal leakage is the infinite limit of the conditional~$\alpha$-mutual information, defined below~\citep{verdu-15a}.

\begin{dfn}[$\alpha$-mutual information]\label{def:alpha_information}
For~$\alpha\in (0,1)\cup(1,\infty)$, the~$\alpha$-mutual information between~$X$ and~$Y$ is given by
\begin{equation}\label{eq:def_alpha_MI}
I_\alpha(X;Y) =\frac{1}{\alpha-1} \log \Exop_{P_X}^\alpha\lefto[ \Exop_{P_Y }^{1/\alpha} \lefto[\exp\lefto(\rnPQxy\right)^\alpha  \right] \right].
\end{equation}
The conditional~$\alpha$-mutual information between~$X$ and~$Y$ given~$Z$ is
\begin{equation}
\label{eq:def_cond_alpha_MI}
I_\alpha(X; Y\vert Z) \!=\!
\frac{1}{\alpha\!-\!1}\log \Ex{P_Z\!}{ \ExPow{P_{X\vert Z\!} }{\alpha}{  \ExPow{P_{Y\vert Z\!}}{1/\alpha}{\!  \lefto(\rnPQxyz \!\right)^{\!\!\alpha}   }  }}\!.
\end{equation}
\end{dfn}
It should be noted that the definition of the conditional~$\alpha$-mutual information given here is not the only possible one, and other definitions have been proposed~\citep{tomamichel-18a,esposito-21b}.
Our main reason for focusing on this particular definition is its role in generalization bounds and its connection to the conditional maximal leakage.

When~$\alpha>1$, the function~$x^\alpha$ is convex.
Jensen's inequality then implies that, for~$\alpha>1$, the (conditional)~$\alpha$-mutual information is a lower bound to the corresponding (conditional) R\'enyi divergence.
Thus, we have
\begin{align}\label{eq:alpha-cmi-alpha-mi-relation}
I_\alpha(X;Y)  &\leq \alphaconrelent{\alpha}{P_{Y\vert X}}{P_{Y}}{P_X}  \\
I_\alpha(X; Y\vert Z) &\leq \alpharelent{\alpha}{P_{X\!Y}}{P_XP_Y}.
\end{align}
For~$\alpha<1$, the inequalities are reversed, and the two information measures coincide with the (conditional) mutual information for~$\alpha\rightarrow 1$.

All of the aforementioned information measures rely on the Radon-Nikodym derivative in one way or another.
The Wasserstein distance (sometimes called the Kantorovich metric), introduced in the context of optimal transport, is an example of an information measure that does not~\citep{villani-08a}.
One appealing consequence of this is that, while most information measures require absolute continuity, the Wasserstein distance does not.

\begin{dfn}[Wasserstein distance]\label{def:wasserstein}
Let~$\mathcal{X}$ be a set and~$\rho$ be a metric such that $(\mathcal{X},\rho)$ is a Polish metric space. Let~$P$ and~$Q$ be two probability distributions on~$\mathcal{X}$.
Then, for every~$p\in[1,\infty]$, the~$p$-Wasserstein distance is
\begin{equation}
\wasserstein{p}{P}{Q}= \lefto(\inf_{R\in \Pi(P,Q)} \Ex{(x,x')\distas R}{\rho(x,x')^p} \right)^{1/p}
\end{equation}
where~$\Pi(P,Q)$ denotes the set of joint probability distributions on $\mathcal{X}^2$ with marginal distributions~$P$ and~$Q$.
We refer to the~$1$-Wasserstein distance simply as the Wasserstein distance for brevity.
\end{dfn}

The Wasserstein distance can be understood intuitively as follows.
Imagine that the two distributions~$P$ and~$Q$ describe two different ways of distributing a unit of dirt over~$\mathcal X$.
Then, each coupling between~$P$ and~$Q$ can be seen as a scheme of moving the dirt to turn one distribution into the other.
The Wasserstein distance measures the lowest possible cost (in terms of~$\rho$) at which one distribution of dirt can be turned into the other.
This interpretation motivates the alternative name of ‘‘Earth Mover's Distance''~\citep{rubner-98a}.
Note that, due to Jensen's inequality,~$\wasserstein{p}{P}{Q}$ is non-decreasing with~$p$.

\section{Change of Measure}
\label{sec:change}
When studying the generalization gap, as aforementioned, the quantity of interest is the error event under the joint distribution of the hypothesis and the data.
However, this can be difficult to control directly.
Instead, there may be other, auxiliary distributions that allow for direct control of the error event.
For instance, when one considers the hypothesis and the data to be drawn independently from each other, there are many situations where the concentration inequalities that we will introduce in \cref{sec:concentration} readily apply.
The technique of relating the probability of an event under one distribution to its corresponding value under another auxiliary distribution is referred to as \emph{change of measure}.
The penalty incurred by replacing the original distribution with the auxiliary one can be expressed through information measures, such as those introduced in the previous section.

In this section, we introduce several change-of-measure results.
We begin by introducing the Radon-Nikodym theorem, which is the backbone of many change of measure techniques.
After this, we present methods based on variational representations of divergences, starting with the celebrated \donskervtext.
This variational representation can be used to relate averages under different distributions, with a penalty given by the relative entropy between the distributions.
We then show how the notion of Fenchel conjugates can be used to extend the core idea of the Donsker-Varadhan variational representation to the broad family of $f$-divergences.
Finally, we explain how the framework of optimal transport gives rise to a change of measure, in which the Wasserstein distance appears as the information measure.

\subsection{The Radon-Nikodym Theorem}

For any change of measure technique to be sensible, we need some conditions on the measures (or functions) involved.
As an example, consider a random variable~$X$ that follows a standard Gaussian distribution, and assume that we are interested in the expectation of a function~$f(X)$.
Now, assume we want to compute this expectation by drawing samples from a Bernoulli distribution.
Of course, this is doomed to fail from the beginning for almost any~$f$.
While the true distribution is supported on the real line, our auxiliary Bernoulli distribution is limited to~$\{0,1\}$.
Since we have no chance of drawing samples on parts of the space where the Gaussian distribution has a non-zero density, we can only get a good indication from our samples if~$f$ is trivial everywhere except~$\{0,1\}$.
If we instead were to use another distribution supported on all real numbers as our auxiliary distribution---say, another Gaussian or the t-distribution---we could draw samples from our auxiliary distribution and compute the expectation of~$f$ on this basis.
For this procedure to give an accurate result, we would need to scale the samples by the probability ratio between the true distribution and our auxiliary one.
This is related to importance sampling in statistics, and gives some intuition about the information measures that appear in the results of this section.
The intuition described here is formally captured by the concept of absolute continuity, given in \cref{def:absolute-continuity}.

Throughout this section, this property will be crucial for virtually every result.
The importance of the absolute continuity property is that it guarantees the existence of the Radon-Nikodym derivative that appears in the Radon-Nikodym theorem, sometimes simply referred to as ``change of measure.''
Provided that an absolute continuity requirement holds, the change of measure exactly relates the measure of an event under two distributions \citep[Thm.~6.10(b)]{rudin-87a}.

\begin{thm}[Radon-Nikodym theorem]\label{thm:radon-nikodym}
Let~$P$ and~$Q$ be probability distributions on a common space such that~$P\ll Q$. Then, there exists a function~$f$ such that, for any measurable event~$\setE$,
\begin{equation}
P(\setE) = \int_\setE { f \dv Q}.
\end{equation}
The function~$f$ is referred to as the Radon-Nikodym derivative of~$P$ with respect to~$Q$, and is denoted by~$\dv P/\dv Q$.
\end{thm}
For discrete random variables,~$\dv P/\dv Q$ is simply the ratio between the probability mass functions of the two distributions.
For continuous random variables, it is the ratio between the probability density functions.

Recall that when the distributions~$P$ and~$Q$ are chosen as the joint distribution~$P_{XY}$ and the product of marginals~$P_XP_Y$, the logarithm of the Radon-Nikodym derivative is the information density:
\begin{equation}
\imath(X,Y) = \log \frac{\dv P_{XY}}{\dv P_XP_Y}.
\end{equation}
This can be used for the following change of measure: assume that we have~$f(x,y)=0$ whenever~$\imath(x,y)=-\infty$.
Note that, if we assume that~$P_XP_Y\ll P_{XY}$, we always have~$\imath(x,y)>-\infty$ so that the condition is satisfied for any function~$f$.
Then, \newtext{by \cref{thm:radon-nikodym}, we have \citep[Prop.~18.3]{polyanskiy-22a}
\begin{align}
\Ex{P_XP_Y}{f(X,Y )} &= \Ex{P_{XY}}{ \left(\frac{\dv P_{XY}}{\dv P_XP_Y}\right)^{-1} f(X,Y )} \\
&= \Ex{P_{XY}}{e^{-\infdensxy} f(X,Y)}.
\end{align}
}
Of course, the same type of result holds if we replace the product of marginals~$P_XP_Y$ with an auxiliary distribution~$Q_{XY}$, provided that a suitable absolute continuity assumption holds, and that the information density is replaced with the corresponding logarithm of the Radon-Nikodym derivative.

\subsection{The Donsker-Varadhan Variational Representation of the Relative Entropy}
The celebrated \donskervtext\ has its origins in the work of~\citet{donsker-75a}.
It has a rich history and is a core tool in both information theory and machine learning.
Some alternative names include the shift of measure lemma \citep{mcallester-03a} and the compression lemma \citep{banerjee-06a}.
We state this important result below, the proof of which is adapted from~\citet[Lemma~2.2]{alquier-21a}.

\begin{thm}[Donsker-Varadhan variational representation]\label{thm:donskervaradhan}
Let~$Q$ be a probability distribution on a measurable space~$\mathcal X$, and let~$\Pi$ denote the set of probability measures such that, for all~$P\in\Pi$, we have~$P\ll Q$.
For every measurable function~$f:\mathcal{X}\rightarrow \reals$ such that~$\Ex{Q}{e^{ f(X)}}<\infty$, we have
\begin{equation}\label{eq:thm-donskervaradhan}
 \log \Ex{Q}{e^{ f(X)}} =  \sup_{P\in \Pi} \left\{\Ex{P}{f(X)} - \relent{P}{Q} \right\}.
\end{equation}
The supremum is attained by the \emph{Gibbs distribution}~$G$, defined as
\begin{equation}
\frac{\dv G}{\dv Q}(X) = \frac{e^{f(X)}}{\Ex{Q}{e^{f(X)}}}.
\end{equation}
\end{thm}

\begin{proof}
By straight-forward calculation, we find that for every~$P\in\Pi$,
\begin{align}
\relent{P}{G} &= \Ex{P}{\log \frac{\dv P}{\dv Q }} + \Ex{P}{\log \frac{\dv Q}{\dv G }} \\
&= \relent{P}{Q} + \log\Ex{Q}{e^{f(X)}} - \Ex{P}{f(X)}.\label{eq:gibbs-deriv-last}
\end{align}
By \cref{thm:f-div-props}, we have that~$\relent{P}{G}\geq 0$, with equality if and only if~$P=G$.
Thus, the result follows.
\end{proof}

In the above, we view the function~$f$ as fixed, and optimize over the distribution~$P$.
Since the final result holds for the supremum over~$P$, this change of measure will later lead to bounds that hold uniformly over learning algorithms---a celebrated key feature of the PAC-Bayesian approach.
However, we can also consider an alternative view, where the distribution~$P$ is fixed and we allow~$f$ to be any function in~$\mathcal F=\{f: \Ex{Q}{e^{ f(X)}}<\infty\}$.
To see this, note that if we let~$f=\log \dv P/\dv Q$, we automatically get~$G=P$.
Thus, by rearranging~\eqref{eq:gibbs-deriv-last}, we find that
\begin{equation}\label{eq:donsker-varadhan-functional}
\relent{P}{Q} =  \sup_{f\in\mathcal F}  \left\{ \Ex{P}{f(X)} - \log\Ex{Q}{e^{f(X)}} \right\}.
\end{equation}
This provides a dual perspective on Donsker-Varadhan's variational formula.
The version in~\eqref{eq:thm-donskervaradhan}, with the supremum taken over~$P$, is sometimes referred to as ‘‘inverse'' Donsker-Varadhan~\citep[Exercise~III.6]{polyanskiy-22a}.

\cref{thm:donskervaradhan} relates the expectation of~$f(X)$ under~$P$ to the moment-generating function of~$f(X)$ under~$Q$, via the relative entropy between the two distributions.
To see how we can use the theorem in practice for changing measure, consider its weaker form, given in \cref{thm:firstbound-donskervaradhan}, where we consider fixed~$f$ and~$P$ without performing supremization.
As illustrated in the derivation of our first information-theoretic bound in \cref{thm:first-it-bound}, we can use \cref{thm:donskervaradhan} to transition from the joint distribution of~$W$ and~$\traindata$ to an auxiliary distribution where they are independent, at the cost of a relative entropy term.
This allowed us to obtain an explicit generalization bound by direct application of a concentration inequality.

\subsection{Variational Representation of $f$-divergences}
As it turns out, the Donsker-Varadhan variational formula can be seen as a special case of a more general family of variational representations, where the relative entropy is replaced by $f$-divergences (\cref{def:f-divergence}).
This characterization relies on the concept of convex conjugates, sometimes referred to as Fenchel conjugate or Legendre–Fenchel transform \citep{fenchel-49a}.
The convex conjugate is defined as follows.

\begin{dfn}[Convex conjugate]
The convex conjugate~$f^*$ of a convex function~$f$ is defined as
\begin{equation}
f^*(y)=\sup_{x\in\reals} \left\{xy-f(x) \right\}.
\end{equation}
\end{dfn}
\newtext{A useful property of the convex conjugate is \emph{biconjugation}.
This means that, if $f$ is convex and lower semi-continuous,~$(f^*)^*=f$. For more on convex duality, see, for example,~\citet[Part III]{rockafellar-70a}.}

Recall that the~$f$-divergence between two distributions~$P$ and~$Q$ is given by~$\alpharelent{f}{P}{Q} = \Ex{Q}{ f\lefto( \frac{\dv P}{\dv Q} \right) }$, and that the family of~$f$-divergences includes a number of familiar quantities. %
We are now ready to state the variational representation of~$f$-divergences~\citep{nguyen-10a}.
\begin{thm}[Variational representation of~$f$-divergences]\label{thm:f-div_variational_weak}
Let~$P$ and~$Q$ be two probability distributions on a common measurable space~$\mathcal X$ such that~$P\ll Q$. Let~$f:[0,\infty)\rightarrow \reals$ be a convex and lower semi-continuous function with~$f(1)=0$.
Then,

\begin{equation}\label{eq:var_rep_f-divergence_weak}
\alpharelent{f}{P}{Q} = \sup_{\phi\in \Phi} \left\{    \Ex{P}{\phi} - \Ex{Q}{f^*(\phi)}   \right\}.
\end{equation}
Here $\Phi$ denotes the set of all functions~$\mathcal{X}\rightarrow \reals$ such that the expectations in~\eqref{eq:var_rep_f-divergence_weak} are defined.
\end{thm}

While setting $f(x)=x\log x$ allows us to recover the relative entropy as a special case of the~$f$-divergence, the variational representation given in \cref{thm:f-div_variational_weak} does not exactly recover the functional form of the \donskervtext\ in~\eqref{eq:donsker-varadhan-functional}.
Instead, we get
\begin{equation}
\relent{P}{Q} = \sup_{\phi\in \Phi} \left\{ \Ex{P}{\phi} - \frac1e\Ex{Q}{e^\phi} \right\}.
\end{equation}
If we consider a fixed~$\phi$ and use the resulting inequality to bound~$\Ex{P}{\phi}$, we obtain a bound that is strictly weaker than what we get from \cref{thm:donskervaradhan}.
It is possible to derive stronger variational representations of~$f$-divergences which do reduce to the \donskervtext, as is done by~\citet{ruderman-12a} and~\citet[Thm.~7.25]{polyanskiy-22a}.
However, the resulting identities are more involved, and not amenable to analysis for general~$f$-divergences.
For instance, when applied to general~$\alpha$-divergences,~\citet[Thm.~7.25]{polyanskiy-22a} does not yield a closed-form solution.
However, for some cases, the stronger, \emph{constrained} variational representation due to~\citet{ruderman-12a} can be used, as we shall see in \cref{sec:pacb-beyond-relent}.
We state this result below.
\begin{thm}[Constrained variational representation of~$f$-divergences]\label{thm:f-div_variational_strong}
Let~$P$,~$Q$ and~$f$ be defined as in \cref{def:f-divergence}. Then,
\begin{equation}\label{eq:var_rep_f-divergence_strong}
\alpharelent{f}{P}{Q}\geq \sup_{\phi\in \Phi} \bigg\{ \Ex{P}{\phi} - \sup_{p\in \Delta (Q)} \big\{\Ex{Q}{\phi p} + \Ex{Q}{f(p)}\big\} \bigg\}.
\end{equation}
Here $\Phi$ denotes the set of all functions~$\mathcal{X}\rightarrow \reals$ such that the expectations in~\eqref{eq:var_rep_f-divergence_strong} are defined, while~$\Delta(Q)$ denotes the set of probability densities with respect to~$Q$.
\end{thm}

\subsection{Optimal Transport}

Recall the Wasserstein distance from \cref{def:wasserstein}.
The Wasserstein distance can be used to change measure by using the following result \citep[Remark~6.5]{villani-08a}.
\begin{thm}[Kantorovich-Rubinstein duality]\label{thm:KR-duality}
Assume that the first moments of~$P$ and~$Q$ are finite. Then,
\begin{equation}\label{eq:KR-duality}
\wasserstein{1}{P}{Q}= \sup_{f\in 1\textnormal{--Lip}(\rho)} \Ex{P}{f} - \Ex{Q}{f},
\end{equation}
where $1\textnormal{--Lip}(\rho)$ denotes the set of functions~$f:\mathcal{X}\rightarrow \reals$ that are $1$-Lipschitz under the metric~$\rho$ used to define the Wasserstein distance, that is,~$\abs{ f(x)-f(y) }\leq \rho(x,y)$ for all~$x,y\in\mathcal{X}$.
\end{thm}
The usefulness of \cref{thm:KR-duality} for changing measure is apparent: if we fix any $f\in 1\textnormal{--Lip}(\rho)$,~\eqref{eq:KR-duality} implies that
\begin{equation}
\Ex{P}{f} \leq \Ex{Q}{f} + \wasserstein{1}{P}{Q}.
\end{equation}
Thus, knowing the expectation of~$f$ under~$Q$ immediately yields a bound on its expectation under~$P$, provided that we can characterize the Wasserstein distance~$\wasserstein{1}{P}{Q}$.
\cref{thm:KR-duality} allows us to replace the assumption of absolute continuity with a Lipschitz assumption on the function~$f$.

Note that, by Jensen's inequality, the~$p$-Wasserstein distance is an increasing function in~$p$.
Thus, the upper bound above still holds if we use~$\wasserstein{1}{P}{Q}\leq\wasserstein{p}{P}{Q}$ for any~$p\geq 1$.
However, as the~$1$-Wasserstein distance leads to the tightest bound, this would only be useful if~$\wasserstein{p}{P}{Q}$ is easier to control than~$\wasserstein{1}{P}{Q}$ when~$p>1$.

\subsection{H\"older's Inequality}

Finally, we present H\"older's inequality, which relates the expectation of a product to the product of the separate moments.

\begin{thm}[H\"older's inequality]\label{thm:holder}
Let~$p,q\in [1,\infty)$ be constants such that~$1/p+1/q = 1$. For two random variables~$X$ and~$Y$, we have
\begin{equation}
\Ex{}{\abs{XY}} \leq \ExPow{}{1/p}{\abs{X}^p}  \ExPow{}{1/q}{\abs{Y}^q}.
\end{equation}
Here, we use the shorthand~$\ExPow{}{a}{X}=(\Ex{}{X})^a$.
\end{thm}

The utility of H\"older's inequality for the purpose of changing measure is that it relates an expectation under a joint distribution to the product of expectations under the corresponding marginal distributions.

\section{Concentration of Measure}
\label{sec:concentration}

We now turn to \emph{concentration of measure} techniques, which allow us to control the deviation of a random variable from its mean.
Specifically, let~$X$ be a random variable with mean~$\mu$, and let~$S=\frac1n\sum_{i=1}^n X_i$ denote the average of~$n$ independent samples distributed as~$X$.\footnote{\newtext{While concentration inequalities can be derived for dependent random variables (see, for instance, the work of \citealp{marton-96a}, \citealp{samson-00a}, \citealp{kontorovich-08a}, and \citealp{kontorovich-17a} as well as recent results using information measures from~\citealp{esposito-23a}), we focus here on independent random variables. In \cref{sec:conc-martingales}, we consider dependent random variables in the form of martingales. }}
Then, a concentration bound controls the probability that~$S$ deviates from~$\mu$ by a certain amount.
We will use the term ‘‘concentration result'' liberally to include bounds on the moment-generating function of~$X$, since they imply a concentration bound in the sense mentioned above (as we will discuss in \cref{thm:subgauss-tail-bound}).
While the change of measure techniques discussed in \cref{sec:change} are useful for replacing expectations under a hard-to-handle probability distribution with the corresponding one under an easier auxiliary distribution, concentration of measure results are needed to control the expectation under the auxiliary distribution.

For a more detailed review of this vast topic, we refer the reader to, for example, the works of~\citet{massart-07a,raginsky-13a,boucheron-13a,wainwright-19a}, where the proofs of the results presented here can be found.

\subsection{Sub-Gaussian Random Variables}\label{sec:subgauss}

A commonly studied category of random variables is the one of sub-Gaussian random variables.
A random variable is said to be sub-Gaussian with parameter~$\sigma$, or~$\sigma$-sub-Gaussian, if its moment-generating function is dominated by that of a Gaussian random variable with variance~$\sigma^2$.
This ensures that the random variable inherits many of the desirable properties of Gaussian random variables, and in particular, concentration results.
Below, we state the definition of a sub-Gaussian random variable~\citep[Def.~2.2]{wainwright-19a}.
\begin{dfn}[Sub-Gaussian random variable]\label{def:subgauss-rv}
A random variable~$X$ is called $\sigma$-sub-Gaussian if, for all~$\lambda\in \reals$,
\begin{equation}\label{eq:def-subgauss}
\Ex{}{e^{\lambda (X-\Ex{}{X})}} 	\leq e^{\frac{\lambda^2\sigma^2}{2}}.
\end{equation}
\end{dfn}

A useful property of sub-Gaussian random variables is that the sub-Gaussianity parameter~$\sigma$ behaves like a standard deviation under averaging: if we let~$S$ denote the average of~$n$ independent~$\sigma$-sub-Gaussian random variables, then~$S$ is $\sigma/\sqrt{n}$-sub-Gaussian.
We formalize this property below.

\begin{propo}[Averaging sub-Gaussian random variables]
Let~$X$ be a~$\sigma$-sub-Gaussian random variable and let~$S=\frac{1}{n}\sum_{i=1}^n X_i$ be the average of~$n$ independent instances of~$X$. Then,~$S$ is~$\sigma/\sqrt{n}$-sub-Gaussian.
\end{propo}

As indicated earlier in this chapter, a bound on the moment-generating function implies a concentration inequality.
This can be shown through the Chernoff method~\citep[Example~2.1]{wainwright-19a}.
Specifically, for the average of sub-Gaussian random variables, we obtain the following~\citep[Prop.~2.5]{wainwright-19a}.

\begin{thm}[Sub-Gaussian concentration]\label{thm:subgauss-tail-bound}
Let~$X$ be a~$\sigma$-sub-Gaussian random variable and let~$S=\frac{1}{n}\sum_{i=1}^n X_i$ be the average of~$n$ independent instances of~$X$. Then
\begin{equation}\label{eq:subgauss-tail-bound}
P(S-\Ex{}{X}>t) \leq e^{-\frac{nt^2}{2\sigma^2}}.
\end{equation}
\end{thm}
Thus, the average of independent samples of a sub-Gaussian random variable concentrates around its mean exponentially fast.
In later chapters, when deriving generalization bounds, we will typically set~$S$ to be the training loss~$\trainloss$ and~$\mu$ to be the population loss~$\poploss$ (see \cref{sec:notation} for definitions), so that~$\poploss$ can be controlled in terms of~$\trainloss$ and the information measure that arises from the change of measure.
As we will see in \cref{chap:average,chap:probability}, it is often sufficient in the derivations of generalization bounds to have a bound on the moment-generating function, and we do not need to convert it into a concentration inequality as the one above.
Hence, we will focus on bounds on the moment-generating function.

Sub-Gaussian random variables can also be characterized in terms of a bound on the moment-generating function of their square, as we formalized below~\citep[Thm.~2.6]{wainwright-19a}.
\begin{propo}[Squared sub-Gaussian random variables]\label{propo:subgauss-square}
Let~$X$ be a~$\sigma$-sub-Gaussian random variable and let~$S=\frac{1}{n}\sum_{i=1}^n X_i$ be the average of~$n$ independent instances of~$X$.
Then, for all~$\lambda\in[0,1)$,
\begin{equation}\label{eq:subgauss-square-concentration}
\Ex{}{e^{\frac{n\lambda (S-\Ex{}{X})^2}{2\sigma^2}}} 	\leq \frac{1}{\sqrt{1-\lambda}}.
\end{equation}
\end{propo}
While we will not cover it explicitly, we note that sub-Gaussianity can be relaxed to sub-exponentiality and the related Bernstein condition~\citep[Sec.~2.1.3]{wainwright-19a}.

\subsection{Bounded Random Variables}\label{sec:bounded_rv}
We now turn to the special case of bounded random variables. Throughout this section, we will, without loss of generality, assume that the range of the random variable is~$[0,1]$---results for generic bounded intervals can be obtained by shifting and scaling as appropriate.

As stated in the following proposition, bounded random variables are sub-Gaussian.

\begin{propo}[Bounded random variables are sub-Gaussian]
Let~$X$ be a random variable whose range is restricted to~$[0,1]$. Then,~$X$ is~$1/2$-sub-Gaussian.
\end{propo}

By directly exploiting the boundedness of a random variable, tighter characterizations of its concentration can be obtained. In the following, we will use the relative entropy between two Bernoulli random variables to obtain a concentration inequality that leads to significantly tighter bounds on the average of~$X$ when the observed sample mean is small.

\begin{dfn}[Binary relative entropy]
Let~$p,q\in[0,1]$. Then~$\binrelent{q}{p}$ denotes the relative entropy between two Bernoulli random variables with parameters~$q$ and~$p$ respectively, \ie,
\begin{align}\label{eq:def-binrelent}
\binrelent{q}{p}&=\relent{\mathrm{Bern}(q)}{\mathrm{Bern}(p)}\\
&=q\log\frac{q}{p} + (1-q)\log\frac{1-q}{1-p}.
\end{align}
\end{dfn}

Let~$\gamma\in\reals$.
A ``relaxed'' parametric version of the binary relative entropy can be expressed as
\begin{equation}\label{eq:def-gamma-binrelent}
\gammabinrelent{q}{p} = \gamma q-\log(1-p+pe^\gamma).
\end{equation}
Specifically, one can show that~$\binrelent{q}{p}=\sup_\gamma \gammabinrelent{q}{p}$.

The binary relative entropy between a sample mean and its expectation can be shown to display a useful concentration behavior.
The following result is due to \citet{maurer-04a}.
\begin{thm}[Concentration for binary relative entropy]\label{thm:kl_concentration}
Let~$X$ be a random variable with range~$[0,1]$ and mean~$\mu$. Let~$S=\frac{1}{n}\sum_{i=1}^n X_i$ be the average of~$n$ independent instances of~$X$.
Then,
\begin{equation}\label{eq:kl_concentration}
\Ex{}{e^{n\binrelent{S}{\mu} }}\leq 2\sqrt{n}.
\end{equation}
\end{thm}

By using this result, we can obtain upper bounds on the binary relative entropy between the sample average~$S$ and the mean~$\mu$.
Specifically, since~$S$ is known,~\eqref{eq:kl_concentration} leads to a bound on~$\mu$, which can be obtained by numerically evaluating the function
\begin{equation}\label{eq:binrelent-numerical-inverse}
\binrelentinv{S}{c} = \sup\{ \mu\in[0,1]: \binrelent{S}{\mu}\leq c \}.
\end{equation}

While~\eqref{eq:binrelent-numerical-inverse} does not admit an analytical solution, it can be relaxed to obtain the following, more easily interpretable, expression~\citep{mcallester-03b,tolstikhin-13a}.

\begin{propo}[Relaxed inverse of the binary relative entropy]\label{propo:inv_kl_relaxation}
For all $S,c\in[0,1]$, we have
\begin{equation}
\binrelentinv{S}{c} \leq S + \sqrt{2Sc}+2c.
\end{equation}
\end{propo}

In \cref{chap:average}, we will see that this result is useful to derive accurate generalization bounds for small training losses.

An alternative concentration result can be derived by considering the relaxed binary relative entropy in \eqref{eq:def-gamma-binrelent}.
This turns out to be particularly useful in the derivation of average generalization bounds in \cref{chap:average}.
The following result is due to \citet{mcallester-13a}.
\begin{thm}[Concentration for parametric binary relative entropy]\label{thm:parametric_kl_concentration}
Let~$X$ be a random variable with range~$[0,1]$ and mean~$\mu$. Let~$S=\frac{1}{n}\sum_{i=1}^n X_i$ be the average of~$n$ independent instances of~$X$.
Then, for every~$\gamma\in\reals$,
\begin{equation}
\Ex{}{e^{n\gammabinrelent{S}{\mu} }}\leq 1.
\end{equation}
\end{thm}
While the upper bound in \cref{thm:kl_concentration} scales as~$\sqrt n$, this is constant in \cref{thm:parametric_kl_concentration}.
\newtext{This concentration result can be applied for a set of values for $\gamma$, for free in the case of bounds in expectation and at the cost of a union bound for bounds in probability. We will discuss this further in \cref{chap:average} and \cref{chap:probability}.}

\subsection{Binary Random Variables}

While we previously considered bounded random variables within~$[0,1]$, we now restrict our attention to binary random variables within this range.
For such random variables, a concentration result on the weighted difference between the random variable and its complement can be derived, which will turn out useful in \cref{chap:cmi}.
The following is due to \citet{steinke-20a}.

\begin{thm}[Concentration of complementary random variables]\label{thm:binary-rv-concentrate-steinke}
Let~$X$ be a random variable satisfying~$P(X=a)=P(X=b)=1/2$ where~$a,b\in [0,1]$. Let~$\bar X =a+b-X$ denote its complement in the set~$\{a,b\}$. Finally, let~$\lambda,\gamma>0$ be constants satisfying $\lambda(1-\gamma)+(e^{\lambda}-1-\lambda)(1+\gamma^2)\leq 0$.  Then,
\begin{equation}
\Ex{}{e^{\lambda\left(X-\gamma\bar X\right)}} \leq 1.
\end{equation}
\end{thm}

\subsection{Martingales}\label{sec:conc-martingales}

For all of the concentration results we have discussed so far, we have focused exclusively on \emph{independent} samples.
While this assumption often greatly simplifies calculations, it is often not satisfied in practice.
One way to allow dependence between the samples, which still enables us to recover essentially the same type of concentration as with sub-Gaussianity, is the martingale property.

\begin{dfn}[Martingale sequences]
A sequence of random variables~$X_i$, with~$i=1,\dots,n$, is a submartingale if
\begin{equation}
\Ex{}{X_{n+1}\mid X_1,\dots, X_n} \geq X_n.
\end{equation}
The sequence is a supermartingale if
\begin{equation}
\Ex{}{X_{n+1}\mid X_1,\dots, X_n} \leq X_n.
\end{equation}
A sequence that is both a submartingale and a supermartingale is called a martingale.
\end{dfn}

A prototypical example of a martingale is a simple one-dimensional random walk, where~$X_i=X_{i-1}+B_{i-1}$, where~$B_{i-1}$ is independent and uniformly distributed on~$\{-1,+1\}$.
By introducing a bias to the walk, it becomes a sub- or super-martingale.

The martingale property allows us to extend essentially sub-Gaussian concentration results to a much broader class of random variables, as shown in the following~\citep[Cor.~2.20]{wainwright-19a}.

\begin{thm}[Azuma-Hoeffding inequality]\label{thm:azuma-hoeffding}
Let~$\{X_t\}_{t=1}^n$ be a sequence of random variables such that~$\abs{X_t-X_{t-1}}\leq c_t$ almost surely for all~$t\in [n]$ and some constants~$\{c_t\}_{t=1}^n$.
Consider the following bound on the moment-generating function for~$\lambda \in \reals$:
\begin{equation}\label{eq:mgf-bound-martingale}
\Ex{}{e^{\lambda(X_n-X_0)} }\leq  \exp\lefto(\frac{-\lambda^2\sum_{t=1}^n c_t^2}{2} \right).
\end{equation}
If~$\{X_t\}_{t=1}^n$ is a supermartingale,~\eqref{eq:mgf-bound-martingale} holds for every~$\lambda\geq 0$.
If~$\{X_t\}_{t=1}^n$ is a submartingale,~\eqref{eq:mgf-bound-martingale} holds for every~$\lambda\leq 0$.
Finally, if ~$\{X_t\}_{t=1}^n$ is a martingale,~\eqref{eq:mgf-bound-martingale} holds for every~$\lambda\in \reals$.
\end{thm}

Thus, the sum of a bounded martingale sequence satisfies the same kind of sub-Gaussian bound on the moment-generating function (and hence, concentration inequality) as if the sequence were independent.
Notably, \cref{thm:kl_concentration} can similarly be extended to bounded martingale sequences, as shown by~\citet[Lemma~2]{seldin-12a}.\footnote{While the bound that is explicitly stated in \citet[Lemma~2]{seldin-12a} is weaker than \cref{thm:kl_concentration}, this is only for simplicity, as discussed in~\citet[Lemma~13]{seldin-12a}.}

\subsection{Heavy-Tailed Random Variables}

The concentration inequalities that we have discussed so far in this section have all relied on bounds on the moment-generating function.
While this does cover many classes of random variables---and in particular encompasses the bounded random variables that appear in classification---there are many scenarios where such bounds are unrealistic.
Moreover, even in settings where the moment-generating function is bounded \newtext{by some parametric function}, actually confirming this can be untenable, especially if we want to specify the parameters of the bound (such as~$\sigma$ for the sub-Gaussian random variables in \cref{def:subgauss-rv}).
Hence, it is of interest to obtain similar results for \emph{heavy-tailed} random variables.
While there is no definite consensus regarding the exact definition of this term,
it typically refers to random variables for which the moment-generating function does not exist (away from~$0$).
While this precludes the use of the techniques that we have covered so far in this section, it can still be possible to obtain generalization bounds in terms of, for instance, the variance of the involved random variables.
We will see this in more detail in, for instance, \cref{sec:pacb-beyond-relent}.
Finally, an approach to generalization bounds that avoid concentration arguments was taken by~\citet{mendelson-14a} and subsequent works by \citet{lecue-17a,lecue-17b,mendelson-18a}.

\chapter{Generalization Bounds in Expectation}\label{chap:average}

Equipped with the change of measure techniques from \cref{sec:change} and the concentration inequalities from \cref{sec:concentration}, we are now ready to derive bounds on the generalization error of learning algorithms.
In \cref{sec:flavors}, we reviewed bounds of different flavors for randomized learning algorithms.
Specifically, this included average, PAC-Bayesian, and single-draw bounds.
As it turns out, there exists a unified approach to derive bounds of all these flavors simultaneously, sometimes referred to as the exponential stochastic inequality (ESI, see \citealp{grunwald-20a}, \citealp{mhammedi-19a} and \citealp{grunwald-23a}), or simply as the exponential inequality approach.
We will briefly discuss this framework in \cref{sec:exponential-ineq}.
However, the details of the derivations that allow us to obtain the tightest possible bounds with the least restrictive assumptions differ somewhat depending on the type of bound under consideration.
Hence, we will provide separate treatments for each type of bound.
In this chapter, we focus on average generalization bounds---that is, generalization bounds in expectation.
In \cref{chap:probability}, we discuss generalization bounds in probability---that is, PAC-Bayesian and single-draw bounds.

In order to keep the notation compact, we will use the following shorthands: we denote the average population loss as~$\avgpoploss=\Ex{\jointdistro}{\poploss}$, the average training loss as~$\avgtrainloss=\Ex{\jointdistro}{\trainloss}$, and their difference---that is, the average generalization error---as~$\avggengap=\avgpoploss-\avgtrainloss$.

\section{Bounds via Variational Representations of Divergences}\label{sec:average-bounds}

As previously mentioned, most information-theoretic generalization bounds are based on a \textit{change of measure} and a \textit{concentration of measure} step.
In this section, we will first present a generic result where the change of measure is performed using the \donskervtext, stated in \cref{thm:donskervaradhan}.
Under different assumptions on the loss function, this can then be instantiated to obtain particular generalization bounds by applying a suitable concentration of measure step.
This will allow us to recover the first information-theoretic generalization bound that we derived in~\cref{sec:basicitbound}, and generalize and improve it in several ways.
\cref{propo:generic-avg-theorem} below is simply a restatement of the Donsker-Varadhan variational representation for the setup of interest in this chapter.
\newtext{A similar result, for convex functions with bounded inputs, was provided by \citet{goyal-17a}.}

\begin{propo}\label{propo:generic-avg-theorem}
Assume that~$\jointdistro\ll\auxproductdistro$.
Let $f:\mathcal W \times \mathcal Z^n\rightarrow \reals$ be a function satisfying~$\Ex{\jointdistro}{f(W,\traindata)}<\infty$.
Then, the Donsker-Varadhan variational formula for the relative entropy (\cref{thm:donskervaradhan}) implies that
\begin{align}\label{eq:generic-avg-theorem}
\Ex{\jointdistro}{f(W,\traindata)} &\leq \log \Ex{\auxproductdistro}{e^{f(W,\traindata)}} + \relent{\jointdistro}{\auxproductdistro}.
\end{align}
In particular, when~$\auxproductdistro=\productdistro$, we get
\begin{align}\label{eq:generic-avg-theorem-mi}
\Ex{\jointdistro}{f(W,\traindata)} &\leq \log \Ex{\productdistro}{e^{f(W,\traindata)}} + I(W;\traindata).
\end{align}
\end{propo}
The second term in the right-hand side of~\eqref{eq:generic-avg-theorem} is minimized by the choice~$\auxproductdistro=\productdistro$, as a consequence of the golden formula~\citep[Eq.~(8.7)]{csiszar-11a}.
However, the resulting relative entropy may not always be possible to compute, while alternative choices of~$\auxproductdistro$ enable this.
We will discuss this in more detail when turning to PAC-Bayesian bounds, where this is an important aspect.
Throughout this chapter, we will assume that~$\auxproductdistro=\productdistro$, as this allows us to express the information measures as simpler, familiar quantities, but we note that most results hold for arbitrary~$Q_W$.

By suitably choosing the function~$f$, the generic result in~\eqref{eq:generic-avg-theorem-mi} can be instantiated to obtain different generalization bounds.
First, we present the average bound from~\citet{xu-17a}.

\begin{cor}\label{cor:xu_raginsky}
Assume that the loss function~$\ell(W,Z)$ is~$\sigma$-sub-Gaussian under~$\productdistro$ and that~$\jointdistro\ll\productdistro$. Then,
\begin{equation}\label{eq:xu_raginsky}
\avggengap \leq \sqrt{ \frac{2\sigma^2I(W;\traindata)}{n} }.
\end{equation}
\end{cor}
\begin{proof}
We begin by applying~\eqref{eq:generic-avg-theorem-mi} with
\begin{equation}
f(W,\traindata)=\lambda\left(\trainloss - \Ex{P_W}{\poploss}\right) .
\end{equation}
Then, by the sub-Gaussianity assumption~\eqref{eq:def-subgauss}, we have
\begin{equation}
\log \Ex{\productdistro}{e^{\lambda\left(\trainloss - \Ex{P_W}{\poploss}\right)}}  \leq \frac{\lambda^2\sigma^2}{2n}.
\end{equation}
Finally, we observe that
\begin{equation}
\inf_{\lambda > 0} \left(\frac{\lambda\sigma^2}{2n} + \frac{I(W;\traindata)}{\lambda}\right) = \sqrt{ \frac{2\sigma^2I(W;\traindata)}{n} },
\end{equation}
from which the result follows.
\end{proof}

This result subsumes \cref{thm:first-it-bound}, which is a special case for bounded random variables.
Also, we note that the same result also holds under a slightly different assumption.
Instead of the stated sub-Gaussianity and absolute continuity assumptions, one can instead assume that for all~$w\in\mathcal W$, $\ell(w,Z)$ is~$\sigma$-sub-Gaussian under~$P_Z$ and~$\conddistrorevw \ll P_{\traindata}$.
Then, instead of the approach used in \cref{propo:generic-avg-theorem}, we consider a random~$W$ and use Donsker-Varadhan's variational representation to change measure between~$\conddistrorev$ and~$P_{\traindata}$.
The rest of the proof is essentially identical, and we average over~$P_W$ in the end.
In \cref{sec:disintegration}, we will discuss how one can also derive a slightly tighter bound under this different sub-Gaussianity assumption.

At first glance, \cref{cor:xu_raginsky} seems to suggest that the generalization error decays as~$1/\sqrt n$.
However, when discussing the dependence on~$n$ of the right-hand side of~\eqref{eq:xu_raginsky}, one needs to remember that~$I(W;\traindata)$ is also an implicit function of~$n$ via~$\jointdistro$.
In order for this result to be valuable when discussing generalization, we want the upper bound to approach zero as the number of training points goes to infinity, which implies that~$I(W;\traindata)=o(n)$.
Therefore, for most settings of interest, we require that~$I(W;\traindata)$ is sublinear in~$n$.

For bounded losses, an alternative generalization bound can be derived using the concentration result for the binary relative entropy.
We present this below.

\begin{cor}\label{cor:binary-kl-average}
\boundedlosstext\ and that~$\jointdistro\ll\productdistro$.
Then,
\begin{equation}
\binrelent{\avgtrainloss}{\avgpoploss} \leq \frac{I(W;\traindata)}{n}.
\end{equation}
\end{cor}
\begin{proof}
We begin by noting that, due to Jensen's inequality and the convexity of~$\gammabinrelent{\cdot}{\cdot}$,
\begin{align}
\binrelent{\avgtrainloss}{\avgpoploss} &= \sup_\gamma\gammabinrelent{\avgtrainloss}{\avgpoploss} \leq\sup_\gamma \Ex{\jointdistro}{\gammabinrelent{\trainloss}{\poploss} }.
\end{align}
We proceed by applying~\eqref{eq:generic-avg-theorem-mi} with $f(W,\traindata)=n\gammabinrelent{\trainloss}{\poploss}$.
Since~$\ell(\cdot,\cdot)\in[0,1]$, we can apply \cref{thm:parametric_kl_concentration} and obtain
\begin{equation}
\log \Ex{\productdistro}{e^{n\gammabinrelent{\trainloss}{\poploss}}} \leq 0.
\end{equation}
From this, the result immediately follows.
\end{proof}

We pause here to discuss the fact that we used the~$\gammabinrelent{\cdot}{\cdot}$ function as the starting point of our proof, while the end result is expressed in terms of the regular binary relative entropy~$\binrelent{\cdot}{\cdot}$.
The reason for this is that this approach allowed us to use \cref{thm:parametric_kl_concentration} for the concentration of measure step, instead of \cref{thm:kl_concentration}.
Had we not done this, we would end up with an additional~$\log(2\sqrt n)/n$ term in our final result.
Crucially, this was possible due to the fact that we derived a generalization bound in expectation, instead of a bound in probability.
Had we concerned ourselves with tail bounds, the supremization over~$\gamma$ would have been problematic, and would have necessitated using a union bound (or something to that effect).

The tightest explicit bounds on the population loss that can be obtained based on \cref{cor:binary-kl-average} rely on a numerical inversion of the binary relative entropy.
However, by using the upper bound on~$\binrelentinv{\cdot}{\cdot}$ provided in \cref{propo:inv_kl_relaxation}, we can obtain a closed-form relaxation that gives some insight into the~$n$-dependence of the bound.
In particular, for the case of zero training loss, this relaxation reduces to
\begin{equation}
\avgpoploss \leq \frac{2I(W;\traindata)}{n}.\label{eq:binrelent-interp-bound-avg-mi}
\end{equation}
Thus, for the case where~$I(W;\traindata)$ is sublinear in~$n$,~\eqref{eq:binrelent-interp-bound-avg-mi} gives a faster decay rate with respect to~$n$ than the sub-Gaussian bound~\eqref{eq:xu_raginsky}, at the cost of a multiplicative constant (the factor 2 in~\eqref{eq:binrelent-interp-bound-avg-mi}).

The bounds that we have discussed so far in this section are all based on the~\donskervtext.
As previously indicated, alternative changes of measure can be used---for instance, those based on~$f$-divergences in \cref{thm:f-div_variational_weak}.
We will now present a generalization bound derived using \cref{thm:f-div_variational_weak}, following~\citet{jiao-17a}.\footnote{\citet{jiao-17a} state their result in terms of adaptive data analysis, in the same vein as~\citet{russo-16a}. In \cref{thm:f-div-avg-bound-no-n}, we provide a simple adaptation that yields a generalization bound.}

\begin{thm}\label{thm:f-div-avg-bound-no-n}
Let~$\norm{\ell(w,\cdot)}_\beta = \ExPow{P_Z}{1/\beta}{\abs{\ell(w,Z)-\poplossw}^\beta}$ and assume that~$\norm{\ell(w,\cdot)}_\beta \leq \sigma_\beta$ for some~$\beta> 1$.
Also, let~$f_\alpha(x) = \abs{x-1}^\alpha$ for some~$\alpha\geq 1$ satisfying~$\frac1\alpha+\frac1\beta=1$.
Then,
\begin{equation}
\avggengap \leq \sigma_\beta \alpharelent{f_\alpha}{\jointdistro}{\productdistro}^{1/\alpha}.
\end{equation}
In particular, if~$\alpha=1$ (so that~$\beta\rightarrow \infty$),
\begin{equation}
\avggengap \leq \sigma_\infty \TV(\jointdistro,\productdistro).
\end{equation}
\end{thm}
\begin{proof}
We apply \cref{thm:f-div_variational_weak} with~$P=\jointdistro$,~$Q=\productdistro$, and~$\phi=\lambda\genwz$, for some~$\lambda>0$, and obtain
\begin{equation}
\alpharelent{f_\alpha}{\jointdistro}{\productdistro} \geq  \Ex{\jointdistro}{\gen} - \Ex{\productdistro}{f_\alpha^*(\genwz)}.
\end{equation}
Explicitly computing the convex conjugate~$f_\alpha^*$ and optimizing over~$\lambda$, we obtain the desired result.
\end{proof}

As noted by~\citet[Sec.~4]{jiao-17a}, an alternative way to derive this bound is through H\"older's inequality.
The benefit of this bound is clear:
we only needed a bound on the central~$\beta$-moment of the loss function, which allows for a broader range of loss functions than (\eg) sub-Gaussian ones.
Furthermore, the~$f$-divergences that appear in \cref{thm:f-div-avg-bound-no-n} can be bounded even when~$\jointdistro\ll\productdistro$ does not hold.
Therefore, \cref{thm:f-div-avg-bound-no-n} enables us to consider more general distributions and learning algorithms than \cref{cor:xu_raginsky}.
One drawback, however, is that the dependence of the bound on the number of samples~$n$ is less explicit.

\section{The Randomized-Subset and Individual-Sample Technique}\label{sec:individual_sample}
One issue with the bounds from the previous section is that the mutual information that appears in them is infinite if the required absolute continuity criterion does not hold.
For instance, consider the bound in \cref{cor:xu_raginsky}.
If~$W$ and~$\traindata$ are separately continuous random variables and~$W$ is a deterministic function of~$\traindata$, the absolute continuity criterion~$\jointdistro\ll \productdistro$ fails to hold.
Hence, the mutual information~$I(W;\traindata)$ is unbounded.
A similar issue arises for many other information measures that appear in information-theoretic and PAC-Bayesian bounds.
A separate problem, but which can be solved in the same way, is the lack of an explicit decay with~$n$ in \cref{thm:f-div-avg-bound-no-n}.

A possible remedy for this is to use the \emph{randomized-subset} technique, wherein the linearity of the expectation operator is used to obtain an average bound for the loss on randomly chosen subsets of the training set, rather than the loss averaged over the full training set.
A special case of this is the \emph{individual-sample} technique, where the random subsets are single samples chosen uniformly at random.
In the case where~$W$ is a deterministic function of~$\traindata$ but not of any individual sample~$Z_i$, this avoids the unboundedness issue.
This technique was introduced by~\citet{bu-20a}.

\begin{propo}[The randomized-subset technique]\label{propo:randomized-subset}
Consider a population loss bound~$\text{ub}_n$ such that $\avgpoploss\leq \text{ub}_n(\Ex{P_{W\!\traindata}}{\trainloss},P_{W\!\traindata})$ for any~$n\in\naturals^+$.
For any subset~$\subsetm \subseteq [n]$, let~$\trainlosssub{\subsetm}=\sum_{i\in \subsetm} \ell(W,Z_i)/\abs{\subsetm}$ denote the training loss on the samples~$\traindata_{\subsetm}=\{Z_i\}_{i\in \subsetm}$,
Let~$P_{\subsetm}$ be an arbitrary probability mass function on subsets of~$[n]$, and assume that~$\subsetm\distas P_{\subsetm}$. Then,
\begin{equation}
\avgpoploss \leq  \Ex{P_{\subsetm}}{\text{ub}_{\abs{\subsetm}}(\Ex{P_{W\!\traindata_{\subsetm}}}{\trainlosssub{\subsetm}}, P_{W\!\traindata_{\subsetm}}) }.
\end{equation}
In particular, if~$P_{\subsetm}$ is the uniform distribution on~$[n]$,
\begin{equation}
\avgpoploss \leq \frac1n\sum_{i=1}^n\left[ \text{ub}_{1}(\Ex{P_{W\!\traindata_i}}{\ell(W,Z_i)} ,P_{W\!Z_i}) \right].
\end{equation}
\end{propo}
\begin{proof}
For any fixed subset~$\subsetm$, the fact that the samples are~\iid implies that
\begin{equation}
\Ex{\jointdistro}{\trainlosssub{\subsetm}} = \Ex{P_{W\!\traindata_{\subsetm}}}{\trainlosssub{\subsetm}} = \Ex{P_{W\!\traindata_{[\abs{\subsetm}]}}}{\trainlosssub{[\abs{\subsetm}]}},
\end{equation}
where~$[\abs{\subsetm}]=\{1,\dots,\abs{\subsetm}\}$.
The result now follows by applying the generalization bound~$\text{ub}_{\abs{m}}$ and averaging over~$M$.
\end{proof}

Below, we illustrate the technique by deriving the individual-sample mutual information bound of~\citet{bu-20a}, who introduced the technique.

\begin{cor}\label{cor:bu_veeravalli}
Assume that the loss function~$\ell(W,Z_i)$ is~$\sigma$-sub-Gaussian under~$\productdistroi$ and that~$\jointdistroi\ll\productdistroi$. Then,
\begin{equation}
\avggengap \leq \frac{1}{n} \sum_{i=1}^n \sqrt{2\sigma^2 I(W;Z_i)}.
\end{equation}
\end{cor}
\begin{proof}
While the result follows immediately by combining \cref{cor:xu_raginsky} and \cref{propo:randomized-subset}, we give a self-contained proof below.
By exploiting the linearity of the expectation operator and marginalizing out the data points that do not appear in a given term, we see that
\begin{align}
\avggengap&=\frac{1}{n}\sum_{i=1}^n \Ex{\jointdistro}{\poploss - \ell(W,Z_i) } \\
&=  \frac{1}{n}\sum_{i=1}^n \Ex{P_{W\!Z_i}}{\geni }.
\end{align}
The result now follows by applying \cref{cor:xu_raginsky} to each term in the sum.
\end{proof}

The individual-sample mutual-information bound in~\cref{cor:bu_veeravalli} is tighter than its full-sample counterpart in \cref{cor:xu_raginsky}.
Specifically, with~$\traindata_{<i}=(Z_1,\dots,Z_{i-1})$ (where~$\traindata_{<1}=\emptyset$),
\begin{align}
\frac1n\sum_{i=1}^n \sqrt{2\sigma^2 I(W;Z_i)} &\leq  \sqrt{\frac{2\sigma^2}n\sum_{i=1}^n I(W;Z_i)} \label{eq:subsampletighter}\\
&\leq \sqrt{\frac{2\sigma^2}{n}\sum_{i=1}^n I(W;Z_i\vert \traindata_{<i})}\\
&=  \sqrt{\frac{2\sigma^2}{n} I(W;\traindata)}.
\end{align}
Here, the first step follows from Jensen's inequality, the second step uses the fact that conditioning on independent random variables does not decrease mutual information (here,~$Z_i$ and~$\traindata_{<i}$ are independent), and the third follows from the chain rule of mutual information.
In fact, as demonstrated by~\citet{harutyunyan-21a}, a wide family of randomized-subset generalization bounds are non-decreasing functions of the subset size.

The benefit of the individual-sample technique is two-fold. First, as mentioned above, it leads to bounds that depend on the sum of functions of the mutual information between the hypothesis and the individual samples, which can sometimes be shown to be tighter than bounds that depend on the mutual information between~$W$ and the full training set~$\traindata$.
Second, we can obtain an explicit decay with~$n$ even from bounds that are constant with respect to~$n$ (ignoring the implicit dependence on~$n$ of the information measure).
For instance, consider the~$f$-divergence bound in \cref{thm:f-div-avg-bound-no-n}.
By applying the individual-sample technique, we instead obtain
\begin{equation}\label{eq:f-div-bound-for-discuss}
\avggengap \leq \frac{\sigma_\beta}{n} \sum_{i=1}^n \alpharelent{f_\alpha}{\jointdistroi}{\productdistroi}^{1/\alpha}.
\end{equation}
While the chain rule of the relative entropy allowed us to recover the full-sample counterparts of generalization bounds from the individual-sample versions, the same does not hold for the~$f$-divergence bound in~\eqref{eq:f-div-bound-for-discuss}, as the~$f$-divergences do not satisfy the chain rule in general.
Hence, it is less clear to what extent the individual-sample technique actually does yield an improvement, and a more careful characterization of the information measures is needed for each case.

The technique just described may similarly be applied to obtain a samplewise version of \cref{cor:binary-kl-average}.
Now, assume that the learning algorithm is designed so that the loss is zero for all training samples, and assume also that the loss is bounded to~$[0,1]$.
Recall that for the sub-Gaussianity-based bound in \cref{cor:xu_raginsky}, this implies that~$\sigma=1/2$.
Now, while we used the same generalization bound for each subset in \cref{propo:randomized-subset}, it is clear that, instead, we could apply a different generalization bound for each subset.
For instance, when using an individual-sample decomposition, we can apply the minimum of the bounds from \cref{cor:xu_raginsky,cor:binary-kl-average} to each term to obtain
\begin{equation}
\avgpoploss \leq \frac1n\sum_{i=1}^n \min\!\bigg\{\sqrt{I(W;Z_i)/2}, 2 I(W;Z_i)\bigg\}.
\end{equation}
Notice that there is not a general ordering between these two bounds: their ordering depends on the specific value of the mutual information.
Specifically, if the mutual information is lower than~$1/8$, the minimum is achieved by the second term.
Otherwise, the first term achieves the minimum.

While samplewise bounds are powerful tools to obtain average generalization bounds, it can be shown that there are certain formal limitations to how they can be used to obtain bounds on the averaged squared generalization error, as well as bounds in probability, as shown by \citet{harutyunyan-22a}.
We will discuss this further in \cref{chap:probability}.

An alternative approach to obtain samplewise bounds, based on the convexity of probability measures, was taken by~\citet{aminian-22a}.
Their approach is based on the following observation: by the linearity of expectation, we can rewrite the average generalization error as
\begin{align}
\avggengap &= \frac1n \sum_{i=1}^n \Ex{P_{W}\!P_{Z_i}}{\ell(W,Z_i)} -  \frac1n \sum_{i=1}^n \Ex{P_{W\! Z_i}}{\ell(W,Z_i)} \\
&= \Ex{P_W\!P_{\bar Z}}{\ell(W,Z)} - \Ex{P_{W\!\bar Z}}{\ell(W,Z)} \\
&= \Ex{P_{\bar Z}}{ \Ex{P_W}{\ell(W,Z)} - \Ex{P_{W\vert\bar Z}}{\ell(W,Z)} }.\label{eq:conv-of-prob-measure}
\end{align}
Here, $P_{\bar Z}=\frac1n\sum_{i=1}^n P_{Z_i}$,~$P_{W\!\bar Z}=\frac1n\sum_{i=1}^n P_{W\! Z_i}$, and~$P_{W\vert\bar Z}=\frac1n\sum_{i=1}^n P_{W\vert Z_i}$.
Using this characterization, generalization bounds can be derived by similar techniques as already discussed by changing measure from~$P_{W\!\bar Z}$ to~$P_W\!P_{\bar Z}$.
We will give an explicit example of such bounds in \cref{thm:wasserstein-convex-measure} in \cref{sec:average-wasserstein}.

\section{Disintegration}\label{sec:disintegration}

If the generalization bound is a concave function---as is the case for the square-root bound in~\eqref{eq:xu_raginsky}---moving expectations outside of the generalization bound leads to a tighter characterization, by Jensen's inequality.
We used this when showing that the individual-sample technique led to a tighter bound than its full-sample counterpart in~\eqref{eq:subsampletighter}.
This insight, referred to as \emph{disintegration}~\citep{negrea-19a}, can be exploited further to derive bounds where additional expectations are moved outside a concave bound.
Consider the derivation of \cref{cor:bu_veeravalli}: essentially, a sample index~$i$ is fixed, the bound is derived, and only \emph{then} do we average over~$i$.
In fact, the same can be done for~$W$, under the slightly altered sub-Gaussianity assumption discussed after \cref{cor:xu_raginsky}.
\begin{cor}\label{cor:disintegrated_xu_rag}
Assume that, for all~$w\in\mathcal W$, the loss function~$\ell(w,Z)$ is~$\sigma$-sub-Gaussian under~$P_{Z}$ and~$\conddistrorevw\ll P_{\traindata}$. Then,
\begin{equation}\label{eq:cor-disintegrated_xu_rag}
\avggengap\leq \Ex{P_W}{\sqrt{ \frac{2\sigma^2 \relent{\conddistrorev}{P_{\traindata}}}{n} }}.
\end{equation}
\end{cor}
\begin{proof}
By the \donskervtext\ (\cref{thm:donskervaradhan}), we have for all~$w\in\mathcal W$ that
\begin{align}
\Ex{\conddistrorevw}{\lambda\gen} &\leq \log \Ex{P_{\traindata}}{e^{\lambda\gen}} + \relent{\conddistrorevw}{P_{\traindata}}.
\end{align}
By applying our sub-Gaussianity assumption and optimizing over~$\lambda$, we get
\begin{equation}
\Ex{\conddistrorevw}{\gen}\leq \sqrt{ \frac{2\sigma^2 \relent{\conddistrorevw}{P_{\traindata}}}{n} }.
\end{equation}
The result follows after averaging over~$P_W$.
\end{proof}

Naturally, the disintegration approach can be combined with the randomized-subset technique.
We highlight that \cref{cor:disintegrated_xu_rag} relies on the alternative sub-Gaussianity assumption, and~\eqref{eq:cor-disintegrated_xu_rag} does not hold under the sub-Gaussianity assumption used in \cref{cor:xu_raginsky}.
For some cases, such as the important special case of bounded losses, both assumptions hold.

\section{Chaining}\label{sec:chaining}

As previously discussed, one of the main draws of the information-theoretic approach to studying generalization is that it allows us to capture the dependence between the training data and the hypothesis that is induced by the learning algorithm.
However, as pointed out by \citet{asadi-18a}, one relevant aspect that is missing from the bounds that we have seen so far in this chapter is the dependence \emph{between hypotheses}.
If one learning algorithm, for different~$\traindata$, selects hypotheses that are distinct but similar to each other in some sense---for instance, as measured by a metric on the hypothesis space---we may expect it to behave very differently from a learning algorithm that selects distinct, highly dissimilar hypotheses.
This, however, can go unnoticed by quantities such as the mutual information, since it depends only on the probability measures that are involved, but not the underlying hypothesis space.
Hence, if there is a bijection between the output sets for the aforementioned learning algorithms that preserves probabilities, they would be equivalent in terms of mutual information.

One approach to incorporate dependencies between hypotheses is to use the \emph{chaining} technique.
Intuitively, this approach consists of looking at the hypothesis space at a coarse level, and approximating the mutual information with increasingly fine granularity.
To introduce chaining formally, we need the following definition.

\begin{dfn}[$\varepsilon$-partitions and increasing sequences]\label{def:epsilon-partition}
Let~$\mathcal W$ be a set endowed with the metric~$d$.
A partition~$\mathcal P=\{ A_1, \dots, A_m\}$, \newtext{comprising disjoint sets $A_1, \dots, A_m$ such that $\mathcal W = \cup_{i=1^m} A_i$}, is an~$\varepsilon$-partition of~$\mathcal W$ if for each~$A_i$ with~$i\in [m]$, there exists a~$w_i\in \mathcal W$ such that~$A_i \subseteq \mathcal B_d(w_i,\varepsilon)$, where~$\mathcal B_d(w_i,\varepsilon) = \{w\in\mathcal W: d(w,w_i)\leq \varepsilon\}$ is the ball of radius~$\varepsilon$ centered around~$w_i$.
A sequence of partitions~$\{\mathcal P_k\}_{k=k'}^\infty$ is called \emph{increasing} if, for all~$k\geq k'$ and each~$A\in \mathcal P_{k+1}$, there exists~$B\in \mathcal P_k$ such that~$A\subseteq B$.
\end{dfn}

In order to state the generalization bound,
we also need to define sub-Gaussian processes.
The sub-Gaussianity here is essentially the same as we have seen before, with the metric~$d(\cdot,\cdot)$ from \cref{def:epsilon-partition} incorporated into the sub-Gaussianity parameter.

\begin{dfn}[Sub-Gaussian process]
The random process $\{X_w\}_{w\in \mathcal W}$ is \emph{sub-Gaussian} for the metric space~$(\mathcal W,d)$ if~$\Ex{}{X_w}=0$ for all~$w\in\mathcal W$ and, for all~$w,w'\in \mathcal W$ and~$\lambda\in\reals$,
\begin{equation}
\log \Ex{}{e^{\lambda(X_{w}-X_{w'})}} \leq \frac{\lambda^2 d^2(w,w')}{2}.
\end{equation}
\end{dfn}

For the result, we also need to assume that the process is separable~\citep[Definition~2]{asadi-18a}, which is a technical assumption that we refrain from stating explicitly for brevity.
We are now ready to state a generalization bound in terms of the chained mutual information for a bounded hypothesis space, due to~\citet{asadi-18a}.

\begin{thm}\label{thm:chaining-mi-bound}
Assume that~$\{\genw\}_{w\in\mathcal W}$ is a separable sub-Gaussian process on~$\mathcal W$ with metric~$d(\cdot,\cdot)$.
Furthermore, assume that the diameter of~$\mathcal W$, defined as~$\text{diam}(\mathcal W)\!=\!\max_{w,w'\in\mathcal W}d(w,w')$, is finite.
Let~$\{\mathcal P_k\}_{k=k_1}^\infty$ be an increasing sequence of partitions such that, for each~$k\geq k_1$,~$\mathcal P_k$ is a~$2^{-k}$-partition of~$\mathcal W$.
For each~$w\in\mathcal W$ and~$k\geq k_1$, let~$[w]_k$ denote the unique~$A\in\mathcal P_k$ such that~$w\in A$.
Then,
\begin{equation}
\avggengap \leq 3\sqrt 2 \sum_{k=k_1}^\infty 2^{-k} \sqrt{I( [W]_k; \traindata )}.
\end{equation}
\end{thm}
As~$k$ increases, the mutual information~$I( [W]_k; \traindata )$ is evaluated on a finer partition of~$\mathcal W$, which yields an increasingly accurate estimate of the mutual information~$ I(W;\traindata)$.
In fact, the sequence is increasing, and~$I( [W]_k; \traindata )\rightarrow I(W;\traindata)$ as~$k\rightarrow \infty$~\citep[Eq.~(8.54)]{cover-06a}.
Thus, as~$k$ increases, so does our estimate of the mutual information, but the higher-$k$ contributions are exponentially discounted with~$2^{-k}$.
Relatively speaking, the lower-$k$ contributions are therefore more influential, and due to the coarse partitions, these incorporate dependence between hypotheses, leading to lower mutual information.
Indeed, for two distinct~$w,w'\in\mathcal W$, we have~$[w]_k=[w']_k$ if they are sufficiently close as measured by~$d$.

As pointed out by \citet{zhou-22a}, the bound above has some limitations.
Firstly, the hypothesis space is required to be bounded, which precludes many simple settings.
The deterministic and hierarchical partitions also impose certain geometric constraints, and can render the bound challenging to compute.
This can be addressed by using a \emph{stochastic} chaining procedure.
Drawing inspiration from multilevel quantization in data compression, \citet{zhou-22a} derive a similar result as \cref{thm:chaining-mi-bound} in terms of a stochastic partition, as formalized in the following.

\begin{thm}\label{thm:stochastic-chaining}
Assume that~$\{\genw\}_{w\in\mathcal W}$ is a separable sub-Gaussian process on~$\mathcal W$ with metric~$d(\cdot,\cdot)$.
Let~$\{W_k\}_{k=k_0}^\infty$ be a sequence of random variables on~$\mathcal W$ such that:
\begin{enumerate}
\item $\lim_{k\rightarrow\infty}\Ex{P_{W_k\traindata}}{\mathrm{gen}(W_k,\traindata)}=\Ex{\jointdistro}{\gen}$,
\item $\Ex{P_{W_{k_0}\traindata}}{\mathrm{gen}(W_{k_0},\traindata)}=0$, and
\item $\{\genw\}_{w\in\mathcal W} - W - W_k - W_{k-1}$ is a Markov chain for every~$k>k_0$.
\end{enumerate}

Then,
\begin{equation}
\avggengap \leq \sum_{k=k_0+1}^\infty \sqrt{\Ex{}{d^2(W_k,W_{k-1})}} \sqrt{2I( [W]_k; \traindata )}.
\end{equation}
\end{thm}

From the result in \cref{thm:stochastic-chaining}, we can recover the one in \cref{thm:chaining-mi-bound} by setting~$\{W_k\}_{k=k_0}^\infty$ as the deterministic sequence that appears therein.
Naturally, this bound can be combined with the disintegration and samplewise techniques, as detailed by~\citet{zhou-22a}.

\section{Bounds via the Kantorovich-Rubinstein Duality}\label{sec:average-wasserstein}

An alternative approach to obtain bounds that incorporate the dependence between hypotheses and the geometry of the hypothesis class is to use tools from optimal transport.
Recall the Wasserstein distance, introduced in \cref{def:wasserstein}.
This information measure is defined in terms of a metric, and suitable choices for this metric enable the possibility of incorporating dependencies between hypotheses.
This approach obviates the need for absolute continuity, since the Wasserstein distance is still defined and finite in its absence.

The key tool in deriving these bounds is the Kantorovich-Rubinstein (KR) duality, stated in \cref{thm:KR-duality}, which relates the difference in expectation under two different distributions to the Wasserstein distance between them.
Below, we state a first result based on \cref{thm:KR-duality}, given by~\citet{wang-19a}.

\begin{thm}\label{thm:wasserstein-no-n}
\newtext{Recall that $P_{\bar Z}=\frac1n\sum_{i=1}^n P_{Z_i}$,~$P_{W\!\bar Z}=\frac1n\sum_{i=1}^n P_{W\! Z_i}$, and~$P_{W\vert\bar Z}=\frac1n\sum_{i=1}^n P_{W\vert Z_i}$.}
Assume that the loss function~$\ell(\cdot,z)$ is~$L$-Lipschitz on~$\mathcal W$ for all~$z\in \mathcal Z$. Then,
\begin{equation}
\abs{\avggengap} \leq L \Ex{P_{\traindata} }{\wasserstein{1}{\conddistro}{P_W} }.
\end{equation}
\end{thm}
\begin{proof}
Since the loss is~$L$-Lipschitz, the loss normalized by~$L$ is~$1$-Lipschitz.
The result follows immediately by applying \cref{thm:KR-duality} with~$f=\trainloss$,~$P=\jointdistro$, and~$Q=\productdistro$.
\end{proof}

For this bound to decay to zero as~$n$ approaches infinity, the average Wasserstein distance~$\Ex{P_{\traindata} }{\wasserstein{1}{\conddistro}{P_W} }$ would need to be a \emph{decreasing} function of~$n$.
We would prefer that the bound explicitly decays with~$n$ (when ignoring the Wasserstein distance).
This can, for instance, be achieved by observing that if~$\ell(w,z)$ is~$L$-Lipschitz under the~$p$-norm,~$\trainlosswz$ is~$L/n^{1/p}$-Lipschitz under the~$p$-norm.
Under this assumption, the following bound can be derived~\citep[Thm.~1]{lopez-18a}.
\begin{thm}\label{thm:wasserstein-jog-lopez}
Assume that, for some~$p\geq 1$, the loss function~$\ell(w,\cdot)$ is~$L$-Lipschitz on~$\dataspace$ under the~$p$-norm for all~$w\in\mathcal W$. Then,
\begin{equation}
\abs{\avggengap} \leq \frac{L}{n^{1/p}} \ExPow{P_{\traindata} }{1/p}{(\wasserstein{p}{\conddistrorev}{P_{\traindata}})^p }.
\end{equation}
\end{thm}
Note that this bound is in terms of the ‘‘backwards channel''~$\conddistrorev$, which is a result of the Lipschitz assumption being with respect to~$\dataspace$.
While the proof of~\citet{lopez-18a} is a bit more involved, the result for~$p=1$ follows immediately from KR duality.

For generalization bounds in expectation, we have an alternative tool at our disposal: the individual-sample technique.
By applying this, as stated in \cref{propo:randomized-subset}, to the bound in \cref{thm:wasserstein-no-n}, we obtain the following~\citep{rodriguezgalvez-21a}.
\begin{thm}\label{thm:wasserstein-individual-sample}
Assume that the loss function~$\ell(\cdot,z)$ is~$L$-Lipschitz on~$\mathcal W$ for all $z\in \mathcal Z$. Then,
\begin{equation}
\abs{\avggengap} \leq \frac{L}{n} \sum_{i=1}^n \Ex{P_{Z_i} }{ \wasserstein{1}{\conddistroi}{P_W}  }.
\end{equation}
\end{thm}
\begin{proof}
We first use the samplewise decomposition from the proof of \cref{cor:bu_veeravalli} to obtain
\begin{align}
\avggengap&=\frac{1}{n}\sum_{i=1}^n \Ex{\jointdistro}{\poploss - \ell(W,Z_i) } \\
&=  \frac{1}{n}\sum_{i=1}^n \Ex{P_{W}\!P_{Z_i}}{\ell(W,Z_i) } - \Ex{P_{W\!Z_i}}{\ell(W,Z_i) }.
\end{align}
Since each term in the sum is a difference between the average of a random variable taken under two different distributions, the result immediately follows by applying the KR duality given in \cref{thm:KR-duality}.
\end{proof}

As for the mutual information-based bounds, the samplewise Wasserstein bound can be shown to be tighter than its full-sample counterpart~\citep{rodriguezgalvez-21a}.

As mentioned earlier, an alternative approach to obtain samplewise bounds is through the convexity of probability measures~\citep{aminian-22a}.
Using the decomposition in~\eqref{eq:conv-of-prob-measure} and applying the KR duality, we obtain the following result.
\begin{thm}\label{thm:wasserstein-convex-measure}
Assume that the loss function~$\ell(\cdot,z)$ is~$L$-Lipschitz on~$\mathcal W$ for all $z\in \mathcal Z$. Then,
\begin{equation}
\avggengap \leq L \Ex{P_{\traindata}}{ \wasserstein{1}{P_{W\vert\bar Z}}{P_W} }.
\end{equation}
\end{thm}
This bound is always tighter than the one in \cref{thm:wasserstein-individual-sample}, due to Jensen's inequality and the convexity of the supremum.
For symmetric learning algorithms, where~$P_{W\vert Z_i}$ is the same for all~$i$, the bounds in \cref{thm:wasserstein-individual-sample} and \cref{thm:wasserstein-convex-measure} coincide.

For bounded losses, the Wasserstein-based bound in \cref{thm:wasserstein-individual-sample} can be shown to be tighter than the corresponding slow-rate bound in \cref{cor:bu_veeravalli} based on the mutual information.
This improvement also has a clear interpretation, as discussed by~ \citet{rodriguezgalvez-21a}: the Wasserstein distance can account for structure within the hypothesis class by an appropriate choice of metric.
If we use the discrete metric,~$\rho_D(x,y)=1\{x\neq y\}$, we discard this geometric information, but we are able to recover bounds based on the relative entropy.
Specifically, for the discrete metric,~$\wasserstein{1}{P}{Q} = \mathrm{TV}(P,Q)$~\citep[Thm.~6.15]{villani-08a}.
Therefore, we can use either Pinsker's inequality or the BH inequality~(\cref{thm:pinsker-bh}) to upper-bound the Wasserstein distance in \cref{thm:wasserstein-individual-sample}.
Specifically, consider a bounded loss function, with range restricted to~$[0,1]$.
Then, it is~$1$-Lipschitz for any~$z\in\mathcal Z$ under the discrete metric on~$\mathcal W$, \ie,~$\rho_D(w,w')=1\{w\neq w'\}$.
By applying the upper bound from~\eqref{eq:thm-pinsker} to the Wasserstein distance in \cref{thm:wasserstein-individual-sample}, we get
\begin{equation}
\abs{\avggengap} \leq \frac{1}{n} \sum_{i=1}^n \Ex{P_{Z_i} }{ \sqrt{ \frac{\relent{P_{W\vert Z_i}}{P_W}}{2} } } \leq \frac{1}{n} \sum_{i=1}^n \sqrt{ \frac{I(W;Z_i)}{2} } ,
\end{equation}
where the second step is due to Jensen's inequality.
This illustrates that the Wasserstein-based bound is tighter for any bounded loss, through the use of Pinsker's inequality.
Furthermore, the Wasserstein-based bound is never vacuous for this setting.
When the relative entropy is high, the BH inequality in~\eqref{eq:thm-bh} gives a tighter upper-bound than Pinsker's inequality in~\eqref{eq:thm-pinsker} does, and in particular, it is never greater than 1.

Similar arguments can also be made in more general settings. For instance, if~$\rho$ is an arbitrary metric, we still have
\begin{equation}
\wasserstein{1}{P}{Q} \leq  d_\rho(\mathcal X)\mathrm{TV}(P,Q)
\end{equation}
where~$d_\rho(\mathcal X)$ denotes the diameter of~$\mathcal X$.
The relaxation in terms of the BH and Pinsker's inequality can thus be relevant for other metrics, provided that the diameter of the hypothesis space is bounded.

Furthermore, a Pinsker-type relaxation can also be obtained under a sub-Gaussianity assumption.
Specifically, consider a probability distribution~$P_X$ on~$\mathcal X$.
If every~$1$-Lipschitz function~$f:\mathcal X \rightarrow \reals$ is~$\sigma$-sub-Gaussian under~$P_X$, we have~\citep{vanhandel-16a}
 \begin{equation}
 \wasserstein{1}{P}{Q} \leq \sqrt{2\sigma^2 \relent{P}{Q}}.
 \end{equation}
If the loss function~$\ell(W,z)$ is~$1$-Lipschitz and sub-Gaussian under~$P_W$ for all~$z\in\dataspace$, the Wasserstein-based bounds can thus also be shown to be tighter than the corresponding ones based on the relative entropy.
Note that this is a different sub-Gaussianity assumption than the one used in \cref{sec:average-bounds}, where the loss was instead assumed to be sub-Gaussian under~$P_Z$.
Further discussion of this, including the relation to the backward channel, can be found in~\citep[Sec.~B.1]{rodriguezgalvez-21a}.

\section{Bibliographic Remarks and Additional Perspectives}\label{sec:bib-remarks-average-bounds}

In \cref{sec:bib-remarks-intro-to-it}, we discussed the history of information-theoretic generalization bounds and the connection to PAC-Bayesian theory.
In this section, we will specify how the results that we covered in this chapter are related to existing literature, and give a brief overview of some results that we did not cover in detail.
We note that we will not discuss results for the conditional mutual information (CMI) framework, as these will be covered in \cref{chap:cmi}.

As previously pointed out, \cref{propo:generic-avg-theorem} is simply a restatement of the \donskervtext.
This generic formulation of a generalization bound is similar to many PAC-Bayesian bounds which are stated for generic (convex) functions, like those of~\citet{germain-09a,begin-16a,alquier-17a,rivasplata-20a}.
The bound in \cref{cor:xu_raginsky}, under the alternative sub-Gaussianity assumption, is due to~\citet{xu-17a}, but, as discussed in \cref{sec:bib-remarks-intro-to-it}, appeared around the same time in similar forms under slightly different assumptions.
The use of the two different assumptions is explicitly discussed by, \eg,~\citet{rodriguezgalvez-20a}.
The bound with the binary KL divergence in \cref{cor:binary-kl-average} is essentially implicit in the work of~\citet{mcallester-13a}, but only explicitly stated in a looser form.
The exact statement was given explicitly in~\citet{hellstrom-22a}.
\cref{thm:f-div-avg-bound-no-n} was stated by~\citet{jiao-17a} for adaptive data analysis, but obtaining an analogous generalization bound is straightforward by following the same procedure as~\citet{xu-17a} used when adapting the result of~\citet{russo-16a}.

The randomized-subset and individual-sample technique of \cref{propo:randomized-subset} was introduced by~\citet{bu-19a,bu-20a}, and subsequently applied in more settings by, \eg,~\citet{haghifam-20a,rodriguezgalvez-20a,zhou-21a,hellstrom-21b,negrea-19a}, who also introduced the notion of disintegration (\cref{sec:disintegration}).
Several key properties of the randomized-subset approach were studied by~\citet{harutyunyan-21a,harutyunyan-22a}, establishing the general dependence on the size of the subsets and the impossibility of obtaining bounds on the average squared generalization gap in terms of the individual-sample mutual information.
\citet{aminian-22b} provided an alternative perspective, wherein the randomized subsets were instead considered for the probability measures themselves, leading to improved bounds for some settings.

A detailed discussion of chaining and its use in learning theory, where it is used to derive the tightest generalization bounds in terms of the VC dimension, can be found in the book by~\citet[Chapter~8]{vershynin-18a}.
The chaining approach was combined with PAC-Bayesian theory by~\citet{audibert-07a}.
\cref{thm:chaining-mi-bound} is due to~\citet{asadi-18a}, while \cref{thm:stochastic-chaining} is due to~\citet{zhou-22a}.

Bounds based on the Wasserstein distance, discussed in \cref{sec:average-wasserstein}, were obtained by~\citet{raginsky-16a}  for learning algorithms that are stable in terms of Wasserstein distance, in the sense that the distribution of~$W$ does not change much (in terms of Wasserstein distance) if one sample of the training set is replaced.
\newtext{\citet{wintenberger-15a} considered weak transport inequalities, and used this to obtain oracle inequalities with fast convergence rates.}
Results in terms of the Wasserstein distance between the conditional~$\conddistro$ and its marginal~$P_W$ were derived independently by~\citet{lopez-18a} and~\citet{wang-19a}.
The bounds in \cref{thm:wasserstein-no-n,thm:wasserstein-jog-lopez} are based on these works.
Tighter variants of these bounds were obtained by~\citet{rodriguezgalvez-21a}, also through the use of the individual-sample technique.
\citet{aminian-22b} demonstrated that through the convexity of probability measures, tighter bounds can be obtained for non-symmetric learning algorithms.
Finally, the chaining technique was generalized to information measures beyond the mutual information by~\citet{clerico-22a}, who obtained bounds in terms of, for instance, the Wasserstein distance and various~$f$-divergences.

We conclude this chapter by mentioning works on average generalization bounds that we did not cover in detail.
\citet{alabdulmohsin-20a} considered a notion of uniform generalization over all possible parametric loss functions, and showed that this is equivalent to~$\TV(\jointdistro,\productdistro)$, termed the ‘‘variational information.''
\citet{hafezkolahi-20a} discussed methods of tightening information-theoretic generalization bounds through the techniques of conditioning and processing, based on a graphical model perspective.
Many approaches, such as samplewise bounds and chaining, can be expressed through this framework.
\citet{aminian-20a} obtained bounds in terms of the Jensen-Shannon divergence, which can be seen as a symmetrized version of the relative entropy.
\citet{modak-21a} derived variants of the results of~\citet{xu-17a} in terms of the R\'enyi divergence of orders~$\alpha\in(0,1)$, which can potentially be tighter for some settings.
\citet{aminian-21a} considered bounds on higher moments of the generalization error, providing bounds in terms of mutual information and other information measures, based on for instance the~$\chi^2$ divergence.
\citet{raginsky-21a} provided a comprehensive discussion of bounds in terms of information-theoretic stability, while~\citet{sefidgaran-22a} used tools from rate-distortion theory.
\citet{esposito-22a} derived a result that allows them to derive both generalization bounds and transportation-cost inequalities, and used this framework to obtain new bounds in terms of arbitrary divergence measure and recover known bounds in terms of, \eg, the mutual information.
\newtext{\citet{wongso-22a,wongso-23a} considered the \emph{sliced} mutual information, based on one-dimensional random projections, and established connections to generalization both theoretically and empirically.
\citet{chu-23a} provide a unified approach to deriving information-theoretic bounds via a change of measure and Young's inequality. By incorporating other techniques, such as symmetrization and chaining, they obtain new bounds and recover several existing ones.}
Finally, \citet{xu-22a,hafezkolahi-21a} derived bounds for the minimum excess risk in Bayesian learning, while~\citet{hafezkolahi-23a} derived information-theoretic bounds for the minimax excess risk and~\citet{dogan-21a} derived lower bounds on the expected excess risk.

\chapter{Generalization Bounds in Probability} \label{chap:probability}

In the previous chapter, we considered generalization bounds in expectation.
While this allowed for compact derivations, and enabled us to use techniques such as disintegration and randomized subsets that are effective only for average bounds, it is not sufficient for answering the most pertinent question that a practitioner may ask regarding generalization.
Generalization bounds in expectation give us information about the generalization gap that we incur averaged over all possible datasets and all possible instantiations of our learning algorithm.
While this is often sufficient to gain insight, in practice, we usually only have access to a specific, given dataset, and we do not know the distribution that generated it.
In this case, we are interested in whether this specific dataset will allow generalization, and not in the performance averaged over other hypothetical datasets.
Furthermore, the learning algorithm is often used only once for this given data set, and we only concern ourselves with the performance of the specific hypothesis that this yields, rather than the average performance when running the learning algorithm several times.

Motivated by these considerations, we now turn to generalization bounds in probability.
We will focus on two flavors: first, \emph{PAC-Bayesian bounds}, which hold with high probability over the draw of the training set, but are averaged over the learning algorithm.
These bounds apply when we are concerned with \emph{distributions} over the hypothesis class.
Then, we look at \emph{single-draw} bounds, which hold with high probability over the draw of both the dataset and a single hypothesis.
This captures the situation that probably occurs most often in practice.
Finally, we briefly discuss \emph{mean-hypothesis} bounds, which are a sort of hybrid: high-probability bounds on the generalization error of the average hypothesis output from the learning algorithm, given the dataset.

As mentioned in the beginning of \cref{chap:average}, there is a unified way to derive average, PAC-Bayesian, and single-draw bounds through exponential stochastic inequalities.
Hence, many of the bounds that we present in this chapter imply corresponding average bounds.
Below, we give a brief exposition of exponential stochastic inequalities, and then proceed with the PAC-Bayesian and single-draw bounds that are the main focus of this chapter.

\section{Exponential Stochastic Inequality}\label{sec:exponential-ineq}

In this section, we state a basic version of an exponential stochastic inequality, and demonstrate how it can be used to establish bounds of all three flavors (in expectation, PAC-Bayes, and single draw).

\begin{thm}\label{thm:exponential-inequality-general}
Consider two random variables~$X$ and~$Y$ and two functions~$f$ and~$g$ such that, for all~$\eta>0$,
\begin{equation}\label{eq:exp-ineq-base}
\Ex{P_{XY}}{e^{\eta(f(X,Y)-g(X,Y))}}\leq 1.
\end{equation}
Then, we have the ‘‘average'' bound
\begin{equation}\label{eq:exp-ineq-average}
\Ex{P_{XY}}{f(X,Y)} \leq \Ex{P_{XY}}{g(X,Y)}.
\end{equation}
Furthermore,
with probability at least~$1-\delta$ over~$P_{XY}$, we have the ‘‘single-draw'' bound
\begin{equation}\label{eq:exp-ineq-single-draw}
f(X,Y) \leq g(X,Y) + \frac{\log\frac1\delta}{\eta}.
\end{equation}
Finally, with probability at least~$1-\delta$ over~$P_Y$, we have the ``PAC-Bayesian'' bound
\begin{equation}\label{eq:exp-ineq-PAC-bayes}
\Ex{P_{X\vert Y}}{f(X,Y)} \leq \Ex{P_{X\vert Y}}{g(X,Y)} + \frac{\log\frac1\delta}{\eta}.
\end{equation}
\end{thm}
\begin{proof}
To obtain~\eqref{eq:exp-ineq-average}, we use Jensen's inequality to move the expectation inside the exponential.
After re-arranging terms, the result follows.
Next, to obtain~\eqref{eq:exp-ineq-single-draw}, we note that~\eqref{eq:exp-ineq-base} and Markov's inequality imply that
\begin{align}\label{eq:markov-application-thing}
P_{XY}\lefto[ e^{\eta(f(X,Y)-g(X,Y))} \leq \frac1\delta \right] &\geq 1-\Ex{P_{XY}}{e^{\eta(f(X,Y)-g(X,Y))}}\delta \\
&\geq 1-\delta.
\end{align}
From this,~\eqref{eq:exp-ineq-single-draw} follows after re-arranging terms.
Finally, to obtain~\eqref{eq:exp-ineq-PAC-bayes}, we first apply Jensen's inequality only with respect to~$P_{X\vert Y}$.
After using Markov's inequality in the same way as above, the stated result follows.
\end{proof}

When applying \cref{thm:exponential-inequality-general} to obtain generalization bounds, we will typically set~$X=W$,~$Y=\traindata$, let~$f$ be a function of the generalization gap, and let~$g$ be a function of an information measure.
The use of exponential inequalities to derive generalization bounds of different flavors can be traced back at least to the work of~\citet{zhang-06a} and~\citet{catoni-07a}.
For a more thorough discussion of this approach, see the recent work of~\citet{grunwald-23a}.

\section{PAC-Bayesian Generalization Bounds}\label{sec:pac-bayesian-bounds}

The PAC-Bayesian framework, originating in the seminal works of~\citet{shawetaylor-97a} and~\citet{mcallester-98a}, is concerned with high-probability bounds, under the draw of the data, on the averaged loss of the learning algorithm.
The learner, rather than selecting a specific hypothesis given the training data, selects a \emph{distribution} over the hypothesis class.
Then, when we want to use the hypothesis for whichever downstream task we are interested in, we draw a hypothesis according to the distribution.
This is sometimes referred to as a ‘‘Gibbs classifier.''
This stochasticity enables us to capture uncertainty in our choice of hypothesis.

In this section, we overview some PAC-Bayesian generalization bounds.
Now, it should be noted that the PAC-Bayes literature is rich and varied, and we will only cover some of the main developments herein.
In particular, we will highlight how information-theoretic and PAC-Bayesian generalization bounds are closely related via similarities in the derivations and interpretation of the results.
For a complementary overview, with additional bounds, details, and historical comments, the reader is encouraged to consult the excellent introduction to PAC-Bayes by~\citet{alquier-21a}, along with the shorter primer by~\citet{guedj-19a}.

Throughout, to make the notation more compact, we will use the following shorthands: the PAC-Bayesian population loss, averaged over the randomness of the learning algorithm when trained on the training set~$\traindata$, is denoted by~$\pacbpoploss = \Ex{\conddistro}{\poploss}$.
Similarly, the PAC-Bayesian training loss is denoted by~$\pacbtrainloss=\Ex{\conddistro}{\trainloss}$.
The difference between these, that is, the PAC-Bayesian generalization gap, is denoted as~$\pacbgengap=\pacbpoploss-\pacbtrainloss$.

\subsection{Bounds via the Donsker-Varadhan Variational Representation}\label{sec:pacb-donskerv}

To begin, we derive a generic PAC-Bayesian bound, analogous to \cref{propo:generic-avg-theorem}, given in terms of a function~$f(\cdot,\cdot)$ to be specified later.
Similar results are discussed by, \eg,~\citet{germain-09a,begin-14a,alquier-17a,rivasplata-20a}.

\begin{propo}\label{propo:generic-pac-bayes-thm}
Assume that almost surely under~$\traindistro$,~$f:\mathcal W \times \mathcal Z^n\rightarrow \reals$ is a function satisfying~$\Ex{\conddistro}{f(W,\traindata)}<\infty$ and~$\conddistro \ll \auxconddistro$.
Then, we have that with probability at least~$1-\delta$ under~$P_{\traindata}$,
\begin{equation}\label{eq:propo-generic-pac-bayes-thm}
\Ex{\conddistro}{f(W,\traindata)} \leq \log \Ex{\auxjointdistro}{\frac{e^{f(W,\traindata)}}{\delta} } + \relent{\conddistro}{\auxconddistro}.
\end{equation}
\end{propo}
\begin{proof}
By applying the \donskervtext, we can conclude that almost surely,
\begin{equation}
\Ex{\conddistro}{f(W,\traindata)} \leq \log \Ex{\auxconddistro}{e^{f(W,\traindata)} } + \relent{\conddistro}{\auxconddistro}.
\end{equation}
The result follows by noting that Markov's inequality implies that, with probability at least~$1-\delta$ under~$P_{\traindata}$,
\begin{equation}\label{eq:generic-pacb-dv-result}
 \Ex{\auxconddistro}{e^{f(W,\traindata)} } \leq  \Ex{\auxjointdistro}{\frac{e^{f(W,\traindata)}}{\delta}  } .
\end{equation}
\end{proof}
In the PAC-Bayesian vernacular, the distribution~$\conddistro$ is referred to as a \emph{posterior} while the distribution~$\auxconddistro$ is called a \emph{prior}, in line with the historical connection with Bayesian inference.
However, we once again emphasize that these distributions are not required to have any relation to actual Bayesian priors and posteriors.
We only require that the prior is selected so that the first term in the right-hand side of~\eqref{eq:propo-generic-pac-bayes-thm} can be controlled and that the posterior and prior satisfy the absolute continuity criterion.
Typically, the prior~$\auxconddistro$ is selected to be independent of the training data~$\traindata$.
However, as highlighted by, for instance,~\citet{rivasplata-20a}, this is not technically required, although often convenient.

Clearly, the result in \cref{propo:generic-pac-bayes-thm} is very similar to the one in \cref{propo:generic-avg-theorem}, and in fact, the two results are connected through an exponential stochastic inequality.
Predictably, we can therefore derive a result very similar to \cref{cor:xu_raginsky} for sub-Gaussian losses.
However, some care has to be taken.
In the derivation of \cref{cor:xu_raginsky}, we set~$f(W,\traindata)=\lambda\gen$ and applied the concentration result from~\eqref{eq:def-subgauss}, after which we optimized the parameter~$\lambda$.
Such an optimization cannot be performed for \cref{propo:generic-pac-bayes-thm}, since the bound therein holds with probability~$1-\delta$ for a \emph{fixed} function~$f(W,\traindata)$, and hence, a fixed~$\lambda$ if we set~$f(W,\traindata)=\lambda\gen$.
Thus, to use a similar approach here, we would need to use some sort of union bound over the set of candidate values for~$\lambda$.
Now, while this can be done---as can be seen, for instance, in \cref{sec:pacb-martingales} and the work of~\citet{catoni-07a,seldin-12a,rodriguezgalvez-23a}---we will take an alternative approach here, and use the bound on the moment-generating function of the square of sub-Gaussian random variables from \cref{propo:subgauss-square}.

\begin{cor}\label{cor:first_pac-bayes_subgaussian}
Assume that~$\ell(w,Z)$ is~$\sigma$-sub-Gaussian under~$P_{\traindata}$ for all~$w\in\mathcal W$.
Then, with probability at least~$1-\delta$ under~$P_{\traindata}$, we have
\begin{equation}\label{eq:first-pacbayes-bound-subgaussian}
\pacbpoploss \leq \pacbtrainloss  +  \sqrt{2\sigma^2\lefto( \frac{\relent{\conddistro}{Q_W}+\log\frac{\sqrt{n}}{\delta} }{n-1} \right)  }.
\end{equation}
\end{cor}
\begin{proof}
First, we use \cref{propo:generic-pac-bayes-thm} with~$\auxconddistro=Q_W$ and
\begin{equation}
f=\frac{(n-1)\gen^2}{2\sigma^2}.
\end{equation}
The result then follows by applying~\eqref{eq:subgauss-square-concentration} with $\lambda=(n-1)/n$.
\end{proof}

Thanks to the flexibility of the theoretical framework, which allows the prior and posterior to be freely chosen, the relative entropy term can be used in a number of different ways which may be practical for certain applications. For instance, one can choose the prior to have a desirable property---for instance, sparsity \citep{alquier-13a,guedj-13a}---and select the posterior by minimizing~\eqref{eq:first-pacbayes-bound-subgaussian} directly. This encourages the same desirable property in the posterior, subject to fitting the training data.

For bounded losses, we can derive results with better dependence on the sample size~$n$, which are useful in the regime of small training losses.
To do this, we use the techniques from \cref{sec:bounded_rv}.
For instance, by setting~$f(W,\traindata)=n\binrelent{\trainloss}{\poploss}$, we can use \cref{thm:kl_concentration} to bound the second term in the right-hand side of~\eqref{eq:generic-pacb-dv-result}. This result, first obtained by~\citet{maurer-04a}, improves on a previous bound due to~\citet{langford-01a}. It is sometimes referred to as the Maurer-Langford-Seeger (MLS) bound.
\begin{cor}\label{cor:MLS_bound}
\boundedlosstext.
Then, with probability at least~$1-\delta$ under~$P_{\traindata}$,
\begin{equation}
\binrelent{\pacbtrainloss}{\pacbpoploss} \leq \frac{\relent{\conddistro }{Q_W } + \log\frac{2\sqrt n}{\delta} }{2n}.
\end{equation}
\end{cor}
As noted by~\citet{langford-02b} (who attributes this observation to Patrick Haffner), this PAC-Bayesian bound has an appealing dimensional consistency as compared to, say, \cref{thm:first-it-bound}.
Both sides are given in terms of logarithms of probabilities, \ie, nats.

In order to use \cref{cor:MLS_bound} to obtain explicit bounds on the population loss, we somehow need to invert the function~$\binrelent{\trainloss}{\cdot}$.
As discussed after \cref{cor:binary-kl-average}, this can be done via the numerical inverse~$\binrelentinv{p}{\varepsilon}$, defined in~\eqref{eq:binrelent-numerical-inverse}.
In words, given a training loss~$\pacbtrainloss$ and an upper-bound on~$\binrelent{\pacbtrainloss}{\pacbpoploss}$, this ‘‘inverse'' of the binary relative entropy gives the highest possible value of~$\pacbpoploss$ that is consistent with the upper bound and training loss.
While it does not admit an analytical expression, it can be found efficiently via numerical search.
Analytical relaxations can be obtained either by using Pinsker's inequality (\cref{thm:pinsker-bh}) or the more refined bound in \cref{propo:inv_kl_relaxation}.

In the PAC-Bayesian literature, a distinction is sometimes made between parametric and non-parametric bounds.
The MLS bound in \cref{cor:MLS_bound}, for instance, is an example of a non-parametric bound.
It admits a parametric counter-part due to~\citet{catoni-07a} and~\citet{mcallester-13a}.
Unsurprisingly, this is obtained by using the parametric version of the concentration result for binary relative entropy from \cref{thm:parametric_kl_concentration}.

\begin{cor}\label{cor:gamma_kl_div_pacbayes}
\boundedlosstext.
Then, with probability at least~$1-\delta$ under~$P_{\traindata}$, for any constant~$\gamma$,
\begin{equation}\label{eq:cor_gamma_kl_div_pacbayes}
\gammabinrelent{\pacbtrainloss}{\pacbpoploss} \leq \frac{\relent{\conddistro }{Q_W } + \log\frac{1}{\delta} }{n}.
\end{equation}
\end{cor}
We see that, as compared to the MLS bound, this parametric version saves a~$\log(2\sqrt n)/n$-term.
However, this comes at the cost of having to choose the constant~$\gamma$ appropriately (and in a data-independent way).
As discussed following \cref{propo:generic-pac-bayes-thm}, we cannot simply optimize over~$\gamma$, since the bound is probabilistic.
\citet{catoni-07a} discusses how to select~$\gamma$ by constructing a dyadic grid of candidate values and optimizing over it, while \citet{mcallester-13a} advises a heuristic set of candidates values over which one can optimize.

A relaxation of \cref{cor:gamma_kl_div_pacbayes}, which more clearly reveals how the parameter~$\gamma$ can be used to control a trade-off between the training loss and the relative entropy, can be obtained as follows~\citep{mcallester-13a}.

\begin{cor}\label{cor:fast-mcallester-relaxation}
\boundedlosstext.
For any fixed~$\lambda>1$, with probability~$1-\delta$ under~$P_{\traindata}$, we have
\begin{equation}\label{eq:cor_mcallester_fast}
\pacbpoploss  \leq  \lambda \pacbtrainloss + \frac{ \lambda\left(\relent{\conddistro}{Q_W} + \log\frac{1}{\delta}\right)    }{2n(1-1/\lambda)} .
\end{equation}
\end{cor}
\begin{proof}

Starting from~\eqref{eq:cor_gamma_kl_div_pacbayes} and using~\eqref{eq:def-gamma-binrelent}, we obtain
\begin{equation}\label{eq:proof-of-fast-pacb-relax-1}
\gamma \pacbtrainloss - \log(1+(e^\gamma-1)\pacbpoploss) \leq \underbrace{\frac{\relent{\conddistro }{Q_W } + \log\frac{1}{\delta} }{n}}_B.
\end{equation}
Now, assume that~$\gamma\in(-2,0)$.
Then,~\eqref{eq:proof-of-fast-pacb-relax-1} implies that
\begin{equation}
\pacbpoploss \leq \frac{1-\exp(\gamma\pacbtrainloss-B)}{1-e^\gamma}.
\end{equation}
When~$\gamma\in(-2,0)$, we have~$e^\gamma\geq 1+\gamma$ and~$e^\gamma\leq 1+\gamma+\gamma^2/2$, so that
\begin{equation}
\pacbpoploss \leq \frac{1-\exp(\gamma\pacbtrainloss-B)}{1-e^\gamma}\leq \frac{\pacbtrainloss-B/\gamma}{1+\gamma/2}.
\end{equation}
Finally, let~$\lambda=1/(1+\gamma/2)$, and note that~$\gamma\in(-2,0)$ implies~$\lambda>1$, from which the result follows.
\end{proof}

In \cref{sec:individual_sample}, we introduced the individual-sample technique for generalization bounds in expectation.
Given any bound on the average population loss depending on the joint distribution of the hypothesis and the training set~$\jointdistro$, this technique allowed us to obtain a bound depending on the joint distribution of the hypothesis and each individual sample,~$P_{W\!Z_i}$.
In most cases, this allowed us to obtain tighter bounds in expectation, sometimes enabling us to turn a vacuous bound into a nonvacuous one.
A natural question is then the following: can we similarly derive PAC-Bayesian individual-sample bounds?
Unfortunately, the answer is, in general, negative.
As shown by~\citet{harutyunyan-22a}, there exists a counter-example for which~$W$ is independent of each~$Z_i$, but where the PAC-Bayesian generalization gap is high with non-negligible probability.
It is, however, possible to derive such bounds based on subsets of size~$m\geq 2$.

\subsection{PAC-Bayesian Bounds Beyond the Relative Entropy}\label{sec:pacb-beyond-relent}

The bounds that we discussed so far are all based on the \donskervtext.
Similar to the average bound case, PAC-Bayesian bounds in terms of other information measures have also been considered.
Notably, these bounds often allow for heavy-tailed losses.
We will give a brief exposition of two approaches, one based on H\"older's inequality and the other based on the variational representation of~$f$-divergences.
Additional works on this topic are mentioned in the bibliographic remarks of \cref{sec:bib-remarks-probability}, and a more detailed discussion can be found in~\citet[Chapter~5]{alquier-21a}.

The basic idea behind using H\"older's inequality to obtain generalization bounds is as follows.
First, we consider an expectation of a quantity of interest related to generalization under the true, joint distribution.
By using the Radon-Nikodym theorem (\cref{thm:radon-nikodym}), this can be turned into an expectation under an auxiliary distribution, at the cost of a Radon-Nikodym derivative appearing.
Finally, H\"older's inequality can be used to disentangle the Radon-Nikodym derivative and the generalization quantity, which can then be handled separately.
This was done for bounded losses by~\citet{begin-16a}, and extended to unbounded losses by~\citet{alquier-17a}.
We present the result of~\citet[Thm.~1]{alquier-17a} below.
\begin{thm}\label{thm:hostile-data}
Let~$\traindata$ denote a training set, the samples of which are allowed to be dependent and drawn from different distributions.
Furthermore, let~$\poplossb=\Ex{P_{\traindata}}{\trainloss}$.\footnote{While this reduces to the previously defined population loss~$\poploss$ for~\iid data, this does not hold in general.}
For some~$p>1$, we let~$q=p/(p-1)$, and set~$f_\alpha(x)=x^\alpha$.
Assume that~$\conddistro\ll Q_W$ almost surely.
Then, with probability at least~$1-\delta$ under~$P_{\traindata}$,
\begin{multline}
\abs{ \pacbgengap } \leq \left( \frac{ \Ex{\auxproductdistro}{ \abs{\poploss-\trainloss}^q } }{\delta} \right)^{1/q} \\
\times( \alpharelent{f_p-1}{\conddistro}{Q_W}+1 )^{1/p}.
\end{multline}
\end{thm}
\begin{proof}
Let~$\Delta(W,\traindata)=\abs{\poploss-\trainloss}$.
Then, by Jensen's inequality and the Radon-Nikodym theorem (\cref{thm:radon-nikodym}),
\begin{equation}
\abs{\pacbgengap}\leq \Ex{\conddistro}{\Delta(W,\traindata)} = \Ex{Q_W}{\Delta(W,\traindata)  \frac{\dv \conddistro}{\dv Q_W}}.
\end{equation}
Then, by H\"older's inequality (\cref{thm:holder}),
\begin{equation}
\Ex{Q_W}{\Delta(W,\traindata)  \frac{\dv \conddistro}{\dv Q_W}} \leq \ExPow{Q_W}{1/q}{\Delta(W,\traindata)^q } \ExPow{Q_W}{1/p}{  \left(\frac{\dv \conddistro}{\dv Q_W}\right)^p   }.
\end{equation}
Finally, it follows from Markov's inequality that with probability at least~$1-\delta$,
\begin{align}
\ExPow{Q_W}{1/q}{\Delta(W,\traindata)^q } \leq \ExPow{Q_W}{1/q}{ \Ex{P_{\traindata}}{\frac{\Delta(W,\traindata)^q}{\delta} } } .
\end{align}
Note that~$\Ex{Q_W}{  \left(\frac{\dv \conddistro}{\dv Q_W}\right)^p   } = \alpharelent{f_p-1}{\conddistro}{Q_W}+1$.
Thus, the desired result follows after combining the steps above.
\end{proof}
As shown by~\citet{begin-16a,alquier-17a}, this bound can be specialized to settings such as \iid data with bounded variance, and even auto-regressive data with finite moments.
One drawback is the linear dependence on the inverse confidence parameter~$1/\delta$, in contrast to the more benign logarithmic dependence of the previous bounds in this chapter.

An alternative route can be taken based on the unconstrained (\cref{thm:f-div_variational_weak}) or constrained (\cref{thm:f-div_variational_strong}) variational representations for~$f$-divergences.
This was done by~\citet{ohnishi-21a}, who derived explicit bounds in terms of a whole host of divergences under various assumptions.
For instance, they obtained tighter versions of some bounds from~\citet{alquier-17a} for heavy-tailed losses.
To illustrate the benefit of the constrained representation of \cref{thm:f-div_variational_strong}, one can compare the two results in Lemma~2 and~3 of \citet{ohnishi-21a}.
Using the unconstrained representation in \cref{thm:f-div_variational_weak}, \citet{ohnishi-21a} obtain the change of measure result
\begin{equation}
 \Ex{P}{\phi} \leq \chisqdiv{P}{Q} +  \Ex{Q}{\phi} + \frac14 \Ex{Q}{\phi^2},
\end{equation}
where~$\chisqdiv{P}{Q}=\Ex{Q}{(\rnPQ-1)^2}$ is the~$\chi^2$ divergence, which can be expressed as an~$f$-divergence (see \cref{def:f-divergence}) by setting~$f(x)=(\sqrt{x}-1)^2$.
In contrast, using the constrained representation in \cref{thm:f-div_variational_strong}, this can be improved to
\begin{equation}
 \Ex{P}{\phi} \leq \chisqdiv{P}{Q} +  \Ex{Q}{\phi} + \frac14 \sqrt{\Ex{Q}{(\phi - \Ex{Q}{\phi})^2}} .
\end{equation}
Since the variance is upper-bounded by the second moment, this is always tighter.

Below, we state a bound in terms of the R\'enyi divergence for sub-Gaussian losses from~\citet[Prop.~6]{ohnishi-21a}, to illustrate that the variational representation for~$f$-divergences allows for the derivation of bounds with a more benign logarithmic dependence on~$1/\delta$.
\begin{thm}\label{thm:ohnishi}
Assume that the loss function~$\ell(w,Z)$ is~$\sigma$-sub-Gaussian under~$\datadistro$ for all~$w\in\mathcal W$.
Fix~$\alpha>1$.
Then, with probability at least~$1-\delta$,
\begin{equation}
\pacbgengap \leq \sqrt{\frac{2\sigma^2}{m}  \log\lefto(\frac2\delta\right) (\alpha(\alpha-1)\alpharelent{\alpha}{\conddistro}{Q_W})^{1/\alpha}}.
\end{equation}
\end{thm}
This result is obtained by specializing \cref{thm:f-div_variational_weak} to the R\'enyi divergence, from which one obtains~\citep[Lemma~5]{ohnishi-21a}
\begin{multline}
 \Ex{P}{\phi} \leq \alpharelent{\alpha}{P}{Q} +  \frac{(\alpha-1)^{\frac{\alpha}{\alpha-1}}}{\alpha}\Ex{Q}{\phi^{\frac{\alpha}{\alpha-1}}} + \frac{1}{\alpha(\alpha-1)}.
\end{multline}%
The remaining steps follow after setting~$P=\conddistro$,~$Q=Q_W$,~$\phi=\lambda\genwz$, and using a sub-Gaussian concentration argument (\cref{def:subgauss-rv}).

\subsection{Data-Dependent Priors}\label{sec:data-dep-prior}

So far, we have only covered generalization bounds with data-independent priors.
However, as we have indicated when discussing \cref{propo:generic-pac-bayes-thm}, data-dependent priors can also be considered.
There are several ways of obtaining generalization bounds with data-dependent priors.
The approach that is perhaps simplest, but which can lead to very tight bounds in practice, is the data-splitting technique~\citep{ambroladze-06a,dziugaite-21a}.
It does not actually involve any new tool---we simply need to apply the tools we have already introduced in a slightly different way.
The idea is to split the training set into two parts as~$\traindata=(\traindata_B,\traindata_P)$, where~$\abs{\traindata_B}=m$ and~$\abs{\traindata_P}=n-m$.
Then, the training loss in the bound is evaluated only on~$\traindata_B$, which means that we are free to use~$\traindata_P$ to inform our prior.
To be clear: the full training data~$\traindata$ is still used as input to the \emph{posterior} (\ie, the learning algorithm).
The only difference is in how we evaluate the generalization bound itself; the learning procedure remains the same.
Since the prior, which can now be written~$Q_{W\vert \traindata_P}$, is independent of the data~$\traindata_B$ used to compute the training loss in the bound, we can apply the same concentration arguments as in the case of a data-free prior.

As an example, we apply this approach to the bound in \cref{cor:MLS_bound}.
Note that it can be applied to all other PAC-Bayesian bounds reviewed in this chapter (and in fact, also to the average generalization bounds discussed in \cref{chap:average}).

\begin{cor}\label{cor:data-split-MLS-bound}
With probability at least~$1-\delta$ under~$P_{\traindata}$,
\begin{equation}
\binrelent{\Ex{\conddistro}{\trainlossB}}{\pacbpoploss} \leq \frac{\relent{\conddistro }{\auxconddistroP } + \log\frac{2\sqrt m}{\delta} }{2m}.
\end{equation}
\end{cor}

Here, a trade-off emerges between two factors that affect the tightness of the bound.
Evaluating the training loss based only on the~$m$ samples in~$\traindata_B$ means that we divide the right-hand side by a smaller overall factor.
However, this is compensated by the fact that the relative entropy term can be significantly lower, since there may exist a posterior with low training loss that is close (in terms of relative entropy) to a suitably chosen data-dependent prior.

Data-dependent priors with the data-splitting technique can be connected to a class of learning algorithms called \emph{compression schemes}.
We discuss this more in \cref{sec:cmi-pac-bayes}.
Furthermore, data-dependent priors have been used to obtain numerically accurate generalization bounds for neural networks.
We cover this in more detail in \cref{sec:nn-numeric}.

We conclude by noting that there are other ways to obtain data-dependent priors---for instance, through differential privacy~\citep{dziugaite-18b} or algorithmic stability~\citep{rivasplata-18a}.

\section{Single-Draw Generalization Bounds}\label{sec:single-draw-bounds}

The PAC-Bayesian bounds in \cref{sec:pac-bayesian-bounds} apply to losses that are averaged over the posterior,~$\conddistro$.
In practice, it is common to instead use a randomized learning algorithm to select a single hypothesis, and then use this specific instance of~$W$ for future inference.
In the PAC-Bayesian literature, bounds for this scenario are often termed \emph{de-randomized} or \emph{pointwise} PAC-Bayes bounds.
We will refer to this scenario, and bounds that apply for it, as \emph{single-draw}, following the terminology of~\citet{catoni-07a}: the bounds apply to a single draw of the training data and a single draw from the stochastic learning algorithm.
In this section, we present several such single-draw bounds.

\subsection{Bounds via Variational Representations of Divergences}

As we did for both the average case in \cref{chap:average} and the PAC-Bayesian setting in \cref{sec:pac-bayesian-bounds}, we begin by deriving a generic single-draw inequality for a function~$f(\cdot,\cdot)$ to be specified later.
However, this will require slightly stronger absolute continuity assumptions than we had before.

\begin{propo}\label{propo:generic-single-draw-thm}
Assume that~$\jointdistro\ll \auxjointdistro$ and~$\auxjointdistro \ll \jointdistro$.
For any function~$f(\cdot,\cdot)$, with probability at least~$1-\delta$ under~$\jointdistro$,
\begin{equation}
f(W,\traindata) \leq \log \Ex{\auxjointdistro}{\frac{e^{f(W,\traindata)}}{\delta}} + \lRNderiv.
\end{equation}
\end{propo}
\begin{proof}
From~\citet[Proposition~18.3]{polyanskiy-22a}, we have
\begin{equation}
\Ex{\auxjointdistro}{e^{ f(W,\traindata)}} = \Ex{\jointdistro}{\exp\lefto( f(W,\traindata)-\lRNderiv \right)}.
\end{equation}
Rewriting this, we obtain
\begin{equation}\label{eq:gen-single-draw-deriv-before-markov}
\Ex{\jointdistro}{\exp\lefto( f(W,\traindata)-\log\Ex{\auxjointdistro}{e^{ f(W,\traindata)}} -\lRNderiv \right)} = 1.
\end{equation}
Applying Markov's inequality (in the same way as in~\eqref{eq:markov-application-thing}) to~\eqref{eq:gen-single-draw-deriv-before-markov}, we conclude that with probability at least~$1-\delta$ under~$\jointdistro$,
\begin{equation}
\exp\lefto(f(W,\traindata) -\log\Ex{\auxjointdistro}{e^{ f(W,\traindata)}} - \lRNderiv \right) \leq \frac{1}{\delta}.
\end{equation}
The result follows by taking the logarithm and rearranging terms.
\end{proof}
As previously mentioned, for the specific choice of $\auxjointdistro=\jointdistro$, the logarithm of the Radon-Nikodym derivative reduces to the information density~$\infdens$.

By making the same specific choices for~$f(W,\traindata)$ and using the same sub-Gaussian concentration inequality as in \cref{cor:first_pac-bayes_subgaussian}, we can obtain the following analogous single-draw generalization bound.

\begin{cor}\label{cor:single-draw_subgaussian}
With probability at least~$1-\delta$ under~$\jointdistro$, we have
\begin{equation}\label{eq:subgauss-single-draw-bound}
\poploss\leq \trainloss +  \sqrt{2\sigma^2\lefto( \frac{\lRNderivprod+\log\frac{\sqrt{n}}{\delta} }{n-1} \right)  }.
\end{equation}
\end{cor}
In this bound, the population loss, training loss, and information metric~$\lRNderivprod$ all depend on the specific instances of~$W$ and~$\traindata$.
One benefit of this is that the bound is fully empirical: all quantities that appear in the bound can be computed given the training data and hypothesis.
Another benefit of the pointwise information measure~$\lRNderivprod$ is that it can be evaluated in closed form for a wider class of distributions than, say, the relative entropy.
For example, as long as the distributions~$\jointdistro$ and~$\auxproductdistro$ have densities, the Radon-Nikodym derivative can easily be evaluated in closed form.
In contrast, the relative entropy has a closed form only for a limited number of probability distributions.
All PAC-Bayesian bounds that are derived through an exponential stochastic inequality approach admit a single-draw counterpart---provided that the more stringent absolute continuity criterion in \cref{propo:generic-single-draw-thm} is satisfied.
Hence, we can obtain single-draw variants of \crefrange{cor:MLS_bound}{cor:data-split-MLS-bound}, with the PAC-Bayesian losses replaced by their single-draw counterparts and with the relative entropy replaced by the logarithm of the corresponding Radon-Nikodym derivative.

\subsection{Using Hölder's Inequality}
\label{sec:holder}

An alternative way to obtain single-draw generalization bounds is through the use of H\"older's inequality (\cref{thm:holder}), via an approach introduced by~\citet{esposito-21a}.
We start by providing the following general theorem.
\begin{thm}\label{thm:holder_two_variables}
Assume that~$\jointdistro\ll \productdistro$.
For any constants~$\alpha,\alpha',\gamma,\gamma'$ such that~$1/\alpha+1/\gamma=1/\alpha'+1/\gamma'=1$ and all measurable sets~$\setE\in \mathcal{W}\times \mathcal Z^n$, we have
\begin{equation}
\jointdistro[\setE]\leq \Exop_{P_W}^{1/\gamma'} \lefto[\traindistro(\setE_W)^{\gamma'/\gamma}\right]   \Exop_{P_W}^{1/\alpha'}\lefto[ \Exop_{\traindistro}^{\alpha'/\alpha}\lefto[  \left(\frac{\dv \jointdistro }{\dv \productdistro } \right)^{\!\alpha\,} \right] \right].
\end{equation}
Here, $\setE_W=\left\{\traindata: (\traindata,W)\in \setE\right\}$.
\end{thm}
\begin{proof}
Let~$1_\setE$ denote the indicator function of~$\setE$. By the Radon-Nikodym theorem we have
\begin{align}
\jointdistro[\setE] &= \Ex{\jointdistro }{1_\setE} \\
&= \Ex{\productdistro }{ 1_\setE \frac{\dv \jointdistro }{\dv \productdistro } } \\
&=\Ex{P_W}{ \Ex{\traindistro }{ 1_{\setE_W} \frac{\dv \jointdistro }{\dv \productdistro }  } } .
\end{align}
By applying H\"older's inequality twice, we get
\begin{align}
\jointdistro[\setE]&\leq  \Ex{P_W}{ \ExPow{\traindistro }{1/\gamma}{ 1_{\setE_W}^\gamma } \ExPow{\traindistro }{1/\alpha}{ \left(\frac{\dv \jointdistro }{\dv \productdistro } \right)^{\!\alpha\,}  }  } \\
&\leq  \ExPow{P_W}{1/\gamma'}{ \ExPow{\traindistro }{\gamma'/\gamma}{ 1_{\setE_W} }} \ExPow{P_W}{1/\alpha'}{\ExPow{\traindistro }{\alpha'/\alpha}{ \left(\frac{\dv \jointdistro }{\dv \productdistro } \right)^{\!\alpha\,}  }  }
\end{align}
from which the result follows.
\end{proof}

By choosing~$\setE$ and the parameters in \cref{thm:holder_two_variables} appropriately, we can derive many generalization bounds.
We will focus on a bound for sub-Gaussian losses which is expressed in terms of the~$\alpha$-mutual information (see \cref{def:alpha_information}).

\begin{cor}\label{cor:alpha-mi-and-max-leakage}
Assume that the loss function~$\ell(w,Z)$ is~$\sigma$-sub-Gaussian under~$P_Z$ for all~$w\in\mathcal W$. Furthermore, assume that~$\jointdistro\ll \productdistro$. Then, with probability at least~$1-\delta$ under~$\jointdistro$, for any~$\alpha>1$,
\begin{equation}
\abs{\gen} \leq \sqrt{ \frac{2\sigma^2}{n} \lefto(I_\alpha(W;\traindata) + \log 2 + \frac{\alpha}{\alpha-1}\log \frac{1}{\delta} \right) }.
\end{equation}
In particular, when~$\alpha\rightarrow \infty$,
\begin{equation}\label{eq:maxleakage-gen-bound}
\abs{\gen} \leq \sqrt{ \frac{2\sigma^2}{n} \lefto(\maxleakage{\traindata}{W} + \log \frac{2}{\delta} \right) }.
\end{equation}
\end{cor}
\begin{proof}
Let~$\alpha'\rightarrow 1$, implying that~$\gamma'\rightarrow \infty$.
Consider the error event~$\setE=\{W,\traindata: \abs{\poploss-\trainloss} > \varepsilon\}$.
By sub-Gaussianity (see \cref{thm:subgauss-tail-bound}), for each~$w\in\mathcal W$ we have
\begin{equation}\label{eq:holder-arg-concentration-step}
\traindistro[\setE_w] \leq 2\exp\lefto(-\frac{\varepsilon^2}{2\sigma^2}\right).
\end{equation}
Furthermore, note that
\begin{equation}
\Exop_{P_W}\lefto[ \Exop_{\traindistro}^{1/\alpha}\lefto[ \frac{\dv \jointdistro }{\dv \productdistro } \right] \right] = \exp\lefto( \frac{\alpha-1}{\alpha} I_\alpha(W;\traindata) \right).
\end{equation}
By setting~$\delta=\jointdistro[\setE]$ and solving for~$\varepsilon$, we obtain the desired result.
\end{proof}

Notice that, unlike the single-draw bounds we presented previously, the right-hand side of this bound is a constant: it does not depend on the specific instances of~$W$ and~$\traindata$ in the left-hand side.
A drawback of this is that the bound is no longer empirical, and requires knowledge of~$P_Z$ to be computed exactly.
An advantage is that the same bound holds regardless of the specific data or output of the algorithm, and the bound can thus be stated \textit{a priori}.

\section{Mean-Hypothesis Generalization Bounds}\label{sec:mean-hypothesis}

Before concluding this chapter, we will briefly mention a third flavor of generalization bounds in probability: bounds for the mean hypothesis.
To this end, we consider a stochastic learning algorithm~$\conddistro$ with a fixed, randomly drawn~$\traindata$.
There are many ways to define deterministic classifiers based on this, by averaging over the randomness of the learning algorithm in various ways.
For instance, in \citet{banerjee-21a}, the goal is to bound, with high probability over~$\traindistro$, the population loss of the average hypothesis output by the learning algorithm:~$w^*=\Ex{\conddistro}{W}$.
The motivation for this is that, when evaluating PAC-Bayesian generalization bounds, it is common to start from a deterministic learning algorithm and make it stochastic by adding zero-mean Gaussian noise to the parameters.
We will see this in more detail in \cref{chap:iterative-methods}.
While this can allow the bounds to be computed, the new classifier with added noise often has degraded performance relative to the underlying deterministic one.
However, since the mean of this randomized classifier is the underlying, original hypothesis, bounds on~$w^*$ apply to this deterministic classifier.

For binary classifiers, we can obtain various type of majority voting algorithms on the basis of the posterior.
Following the discussion from~\citet{seeger-02a}, we assume that~$W$ denotes the parameters of a map~$f_W: \mathcal X\rightarrow \reals$.
The goal is to predict the binary label of~$X$, and we use the sign of~$f_W(X)$ to achieve this.
For the stochastic predictors we considered previously, the output is given by~$\sign(f_W(X))$, where~$W\distas \conddistro$.
However, we can also consider the majority vote classifier given by~$f_{\text{mv}(X)}=\sign \left(\Ex{\conddistro}{\sign (f_W(X))} \right)$, as well as the averaging classifier, given by~$f_{\text{BPM}(X)}=\sign \left(\Ex{\conddistro}{ f_W(X)}\right)$.
These are referred to as the Bayes voting classifier and the Bayes classifier, respectively, by~\citet{seeger-02a}.
As noted by~\citet{langford-02a}, a bound on the population loss of the stochastic predictor based on~$\conddistro$ leads to a bound on the population loss of the BPM classifier, at the price of a factor 2.
This is based on the observation that, for any sample for which~$f_{\text{BPM}}$ incurs a loss, the underlying stochastic classifier must incur a loss with probability at least~$1/2$.

\section{Bibliographic Remarks and Additional Perspectives}\label{sec:bib-remarks-probability}

In this section, we discuss how the presented results relate to the literature, and briefly mention some results that we did not cover in detail.
As the literature is vast, in particular for PAC-Bayesian bounds, this brief overview will not be exhaustive.
See \cref{sec:bib-remarks-intro-to-it} for further discussion of the early history of PAC-Bayesian bounds, and the monograph of \citet{alquier-21a}.

The underlying concepts of the exponential stochastic inequality framework of \cref{sec:exponential-ineq} can be traced to the works of~\citet{zhang-06a} and~\citet{catoni-07a}.
This was formalized using ESI notation by~\citet{koolen-16a,mhammedi-19a,grunwald-20a}, and was recently given an exhaustive treatment by~\citet{grunwald-23a}.

The generic PAC-Bayesian bound in \cref{propo:generic-pac-bayes-thm} is similar to statements given by~\citet{germain-09a,begin-14a}, while this exact form is due to~\citet{rivasplata-20a}.
While \cref{cor:first_pac-bayes_subgaussian} is very similar to earlier results, such as the one from~\citet{mcallester-03a}, this exact form is from~\citet{hellstrom-20b}.
The bound in \cref{cor:MLS_bound} is due to~\citet{maurer-04a}, where the logarithmic factor is improved compared to the result of~\citet{langford-01a}.
See the work of~\citet{foong-21a} for an in-depth discussion of the tightness of this bound and whether this logarithmic dependence can be improved further.
\cref{cor:gamma_kl_div_pacbayes} is implicit in the work of~\citet{mcallester-13a} (who in turn describes the result as a corollary of statements from~\citet{catoni-07a}), while the loosened version in \cref{cor:fast-mcallester-relaxation} is stated explicitly.
The role of the generic convex function in the bound is studied in \citep{hellstrom-24a}, where optimal choices are established.
\newtext{\cite{jang-23a} used the coin-betting framework from online learning to improve PAC-Bayesian bounds for bounded losses.}

\cref{thm:hostile-data} is due to~\citet{alquier-17a}, and is an extension of a result from~\citet{begin-16a} for bounded losses.
\citet{ohnishi-21a} provided a comprehensive treatment of change of measure inequalities with $f$-divergences and their application in PAC-Bayesian bounds, including \cref{thm:ohnishi}, as well as a whole host of additional results.

Data-dependent priors based on data splitting were introduced by~\citet{ambroladze-06a}, and have since been extended and used in various ways by, for instance,~\citet{parrado-12a,rivasplata-18a,dziugaite-18b,mhammedi-19a,rivasplata-20a,dziugaite-21a}.
\citet{seeger-02a} used a similar technique, whereby an independent set of ‘‘model selection'' samples is used to learn the prior and the model class.
However, unlike in the works mentioned above, this set is disjoint from the training set used to find the posterior.
Data-dependent priors through differential privacy were studied by~\citet{dziugaite-18b}, while~\citet{rivasplata-20a} used algorithmic stability.
Distribution-dependent priors are discussed by, \eg,~\citet{catoni-07a,lever-10a,lever-13a}.
We discuss data-dependent priors further in \cref{sec:cmi-pac-bayes,sec:nn-numeric}.

While mainly focusing on PAC-Bayesian bounds,~\citet[Thm.~1.2.7]{catoni-07a} mentioned in passing that similar techniques can be used to obtain bounds for single draws from the posterior, which is the basis for our terminology of ‘‘single-draw.''
The generic inequality in \cref{propo:generic-single-draw-thm} is due to~\citet{rivasplata-20a}, while \cref{cor:single-draw_subgaussian} can be found in~\citet{hellstrom-20b}.
Explicit derivations of more single-draw generalization bounds can be found in~\citet{hellstrom-21b,hellstrom-21a}.

\cref{thm:holder_two_variables} and \cref{cor:alpha-mi-and-max-leakage} are due to~\citet{esposito-21a}, who also presented several additional bounds and results beyond learning theory.
In \citep{hellstrom-20b}, the ‘‘strong converse'' lemma from binary hypothesis testing is used to obtain single-draw bounds in terms of the tail of the information density.
\citet{xu-17a} adapted the monitor technique from \citet{bassily-16a} to convert their average generalization bound to a single-draw one, albeit with a detrimental linear dependence on the inverse confidence parameter~$1/\delta$.

\citet{langford-02a} pointed out that certain mean-hypothesis generalization bounds follow immediately from standard PAC-Bayesian bounds, stating that this was essentially folklore, with further discussion in the work of~\citet{seeger-02a}.
PAC-Bayesian bounds for aggregated predictors have been studied by, \eg, \citet{leung-06a,dalalyan-07a,dalalyan-08a,salmon-11a,salmon-12a,dalalyan-12a,guedj-13a,alquier-13a}.
Further discussion of this can be found in \citet[Sec.~2.2]{alquier-21a}.
\newtext{\citet{germain-15a} introduced the celebrated $\mathcal{C}$-bound, which studies the behavior of majority votes in binary classification.}
Bounds for voting classifiers are also discussed by~\citet{lacasse-06a}, while \citet{zantedeschi-21a,biggs-22b} consider stochastic majority votes.

Finally, we provide some pointers to results that we did not explicitly cover.
As mentioned, a complementary overview of PAC-Bayesian bounds can be found in the introduction by~\citet{alquier-21a}, as well as the primer by~\citet{guedj-19a}.
Two particularly notable topics that we did not cover are oracle bounds and the localization technique of \citet{catoni-07a}.
Oracle bounds, also called excess risk bounds, bound the difference between the population loss of the posterior (or hypothesis, depending on the flavor under consideration) and the minimal achievable loss for the given hypothesis class.
Such bounds are covered in \citet[Chapter~4]{alquier-21a}.
Some notable works proving oracle bounds are \citet{dalalyan-08a,alquier-11a,salmon-11a,salmon-12a,dalalyan-12a,rigollet-12a,alquier2017oracle}. 
The localization technique of \citet{catoni-07a} is a method for selecting a suitable distribution-dependent prior, and is discussed in \citet[Sec.~4.5]{alquier-21a}.

\citet{tolstikhin-13a} used~\eqref{propo:inv_kl_relaxation} to obtain a relaxation of \cref{cor:MLS_bound} to obtain a bound that interpolates between a fast and slow rate, depending on the value of the training loss, while~\citet{thiemann-17a} considered a relaxation of \cref{cor:MLS_bound}, and provided a procedure for minimizing it.
The connection between PAC-Bayesian bounds and Bayesian inference is discussed by~\citet{germain-16b}, while the connection to KL-regularized objective functions is covered by~\citet{germain-09b}.
PAC-Bayesian bounds for sub-exponential random variables are discussed by, \eg,~\citet{catoni-04b}.
\citet{alquier-06a,alquier-08a} used truncated losses in order to handle unbounded loss functions, while~\citet{catoni-18a} used a robust loss function to handle heavy-tailed distributions.
\citet{holland-19a} derived PAC-Bayesian bounds for heavy-tailed losses, obtaining a novel Gibbs posterior on this basis.
\citet{biggs-23a} obtained tighter bounds based on the excess risk by using the underlying difficulty of the problem.
\citet{langford-02a,herbrich-02a} derived bounds in terms of the margins of the learned predictor---an approach recently used by \citet{biggs2021margins} to establish derandomized generalization bounds.
\citet{audibert-07a} combined the chaining technique (discussed in \cref{sec:chaining}) with PAC-Bayesian bounds.
Similarly,~\citet{asadi-20a} derived bounds based on a multilevel relative entropy, while~\citet{clerico-22a} derived an alternative chained bound.
\citet{yang-19a} derived fast-rate PAC-Bayesian bounds through the use of Rademacher processes.
\citet{arora-19a} derived generalization bounds involving Rademacher complexities for contrastive unsupervised representation learning (CURL), the state-of-the-art technique to learn representations (as a set of features) from unlabelled data.
Their results were generalized by \citet{nozawa-19a} to obtain PAC-Bayesian generalization bounds for CURL, holding for non-\iid data and allowing for new representation learning algorithms.
\citet{mhammedi-20a} noted that while many works study bounds for the expected risk, \ie, the mean performance of an algorithm, this might not be the relevant metrics in many problems (\eg, medical, environmental or sensitive engineering tasks).
Motivated by this, they presented a PAC-Bayesian generalization bound for the Conditional Value at Risk (CVaR).\footnote{For any $\alpha\in (0,1)$ and any random variable $Z$, $\mathrm{CVaR}_\alpha (Z)$ measures the expectation of $Z$ conditioned on the event that $Z$ is greater than its $(1-\alpha)$-th quantile. See, for instance, \citet{Pflug2000}.}
\citet{cherief-21a} analyzed Variational Auto-Encoders (VAE)~\citep{kingma-19a}, a popular generative model, through PAC-Bayesian generalization bounds on the reconstruction error of the VAE, and used it to study the regularization effect of classical VAE objectives.
\newtext{\citet{mbacke-23a} provided further PAC-Bayesian bounds for VAEs, while \citet{mbacke-23b} studied adversarial generative models.}
\citet{haddouche-21a} considered losses with a hypothesis-dependent range, and obtained bounds for these through the use of self-bounding functions. \citet{haddouche2022supermartingales} developed bounds for heavy-tailed loss functions through the use of supermartingales.
\citet{amit-22a} derived bounds in terms of \emph{integral probability metrics} (IPM), which includes the total variation and the Wasserstein distance.
This is achieved by essentially using the definition of IPMs as a change of measure (which is similar to the Kantorovich-Rubinstein duality).
Notably, this can be used to convert uniform convergence bounds, as those discussed in \cref{sec:uniform-convergence}, into algorithm-dependent bounds where the uniform convergence bound is multiplied by a total variation between the posterior and a prior. 
Recently, \citet{haddouche2023wasserstein,viallard2023wasserstein} proposed PAC-Bayesian generalization bounds given in terms of a Wasserstein distance. These bounds hold for unbounded (possibly heavy-tailed) losses, and are used as training objectives.

\chapter{The CMI Framework}\label{chap:cmi}

In previous chapters, the majority of the results that we presented required an absolute continuity assumption to be satisfied.
The reason for this requirement is that without it, quantities such as the mutual information in \cref{chap:average} and the relative entropy in \cref{sec:pac-bayesian-bounds} would be infinite.
This absolute continuity requirement is not satisfied when both the training data and the hypothesis are continuous random variables and the hypothesis is a deterministic function of the training data.
For average bounds, this issue can be alleviated by the individual-sample technique of~\citet{bu-20a}, as discussed in \cref{sec:individual_sample}.
However, this approach still yields a vacuous generalization bound when the hypothesis is a deterministic function of a single training sample.
So, while the individual-sample technique mitigates the problem, the fundamental issue still remains: the information carried by a single training sample can be infinite.

These considerations motivate the conditional mutual information (CMI) approach, introduced to the literature of information-theoretic generalization bounds by~\citet{steinke-20a}.
An essentially equivalent approach was introduced in the PAC-Bayesian context much earlier by~\citet{audibert-04a,catoni-07a}, under the name of ‘‘almost exchangeable priors'' and ‘‘transductive learning,'' the motivation of which was to reduce the variance of PAC-Bayesian generalization bounds.
The terminology used to describe the CMI framework is not uniform---it has also been referred to as the random-subset setting~\citep{hellstrom-20a}, randomized-subsample setting~\citep{rodriguezgalvez-20a}, and the supersample setting~\citep{wang-23a}.
Here, we will stick with the terms ‘‘CMI approach'' or ‘‘CMI framework.''
These names are motivated by one of the main end-products of the approach: generalization bounds in expectation given in terms of a conditional mutual information.
As we will show, many of the techniques covered in the preceding chapters are readily extended to this new setting.

An intuitive view of the CMI framework is that, rather than asking whether one can identify a given training sample based on the chosen hypothesis, we instead ask if, given two candidate samples, we can figure out which one was used for training.
Whereas the first question can reveal infinite information, the second one is a \emph{binary} question, so the answer can carry at most~1 bit.
From a technical standpoint, this will guarantee that the desired absolute continuity criterion is always satisfied.
We now introduce the CMI framework more formally.

\section{Definition of the CMI Framework}
\label{sec:cmi-def}

The CMI framework consists of the following elements.
First, we assume that we generate a \emph{supersample}~$\supersample=(\tilde Z_1,\dots,\tilde Z_{2n}) \in \mathcal{Z}^{2 n}$ consisting of~$2n$ samples drawn \iid from $P_Z$.
Only half of these samples are actually used for training, as determined by a \emph{membership vector}~$\subsetchoice\in\{0,1\}^n$, consisting of~$n$ Bernoulli-$1/2$ random variables that are independent of each other and~$\supersample$.
Specifically, the $i$th training sample~$Z_i(S_i)$ is given by~$\tilde Z_{i+S_in}$, \ie, the Bernoulli-$1/2$ random variable~$S_i$ determines whether~$\tilde Z_{i}$ or~$\tilde Z_{i+n}$ is used for training.
Through this procedure, the training set~$\traindatacmi=(Z_1(S_1),\dots,Z_{n}(S_n))$ is built, and the hypothesis~$W$ is chosen based on this training set.
This leads to the Markov chain~$(\supersample,\subsetchoice)$---$\traindatacmi$---$W$ (\ie, $W$ and $(\supersample,\subsetchoice)$ are conditionally independent given $\traindatacmi$).
We denote the entry-wise modulo-2 complement of~$\subsetchoice$ as~$\bar\subsetchoice$, \ie, the~$i$th element of~$\bar\subsetchoice$ is given by~$\bar S_i=1-S_i$.
Note that~$\testdatacmi$ is conditionally independent from~$W$ given~$\traindatacmi$, and can hence be considered a test set.

We will use the notation~$\conddistrocmi=P_{W\vert \traindatacmi}$ to refer to the conditional distribution on~$\mathcal{W}$ that characterizes the learning algorithm and~$\jointdistrocmi=\conddistrocmi\traindistrocmi$ for the induced joint distribution on $W$,~$\supersample$ and~$\subsetchoice$.

It is important to note that this framework is actually just a reformulation of the standard learning setting from before.
Indeed, the training set~$\traindatacmi$ still consists of~$n$~\iid samples from~$P_Z$, on the basis of which we select~$W$ according to our learning algorithm.
Here, the supersample~$\supersample$ can be viewed as a ‘‘ghost sample,'' which is used purely for the purpose of analysis.

The remainder of this chapter is structured as follows.
First, we present generalization bounds in expectation using the CMI framework.
Then, we review PAC-Bayesian bounds in the CMI framework, with a particular focus on the connection to data-dependent priors, before briefly discussing single-draw bounds.
We end the chapter with some extensions of the CMI framework.
Specifically, we present bounds in terms of the so-called evaluated and functional CMI, which improve upon the standard CMI bounds due to the data-processing inequality.
Finally, we present the leave-one-out setting, where the supersample has size~$n+1$ instead of~$2n$.
This turns out to be closely related to the concept of leave-one-out validation.

\section{Generalization Bounds in Expectation}\label{sec:cmi-avg}

We now derive generalization bounds in expectation using the structure of the CMI framework.
In order to keep the notation more compact, we will use the following shorthands: the average population loss is~$\avgpoploss=\Ex{\jointdistrocmi}{\poploss}$, the average training loss is~$\avgtrainloss=\Ex{\jointdistrocmi}{\trainlosscmi}$, and the average generalization gap is~$\avggengap=\avgpoploss-\avgtrainloss$.
When originally introducing the CMI framework, \citet{steinke-20a} derived the following bound.

\begin{thm}\label{thm:steinke_slow}
\boundedlosstext. Then,
\begin{equation}\label{eq:cmi-sqrt-bound}
\abs{\avggengap} \leq \sqrt{\frac{2\cmi}{n} }.
\end{equation}
\end{thm}

\begin{proof}
The proof is very similar to that of \cref{cor:xu_raginsky}, but with some minor modifications.
We begin by noting that, in expectation, the test loss, \ie, the loss evaluated on the test set~$\testdatacmi$, equals the population loss:
\begin{equation}
\Ex{\jointdistrocmi}{\testlosscmi} = \Ex{\jointdistrocmi}{\poploss} = \avgpoploss.
\end{equation}
Hence, a bound on the average difference between the training and test loss is also a bound on the average generalization gap.
To this end, let~$\genS=\testlosscmi-\trainlosscmi$.
Note that this quantity satisfies the symmetry property~$\genS=-\genSbar$.
Hence, for any~$W$ and~$\traindata$, we have
\begin{equation}
\Ex{P_{\subsetchoice}}{\genS} = \Ex{P_{\subsetchoice}}{\testlosscmi-\trainlosscmi} = 0.
\end{equation}
Furthermore, since~$\ell(\cdot,\cdot)$ is bounded to~$[0,1]$, it follows that for each~$i$, the loss difference~$\ell(W,Z_i(S_i))-\ell(W,Z_i(\bar S_i))$ is bounded to~$[-1,1]$.
Hence, the loss difference is a~$1$-sub-Gaussian random variable under~$P_{\subsetchoice}$ (as well as under~$Q_{W\!\supersample}\!P_{\subsetchoice}$ for every distribution~$Q_{W\!\supersample}$ of~$(W,\supersample)$).
Since~$\genS$ is an average of~$n$ such terms, it is~$1/\sqrt n$-sub-Gaussian.

Next, by the \donskervtext, we have
\begin{align}
\lambda\avggengap &= \Ex{\jointdistrocmi}{\lambda\genS} \\
&\leq \log \Ex{\productdistrocmi}{e^{\lambda\genS}} + \relent{\jointdistrocmi}{\productdistrocmi}.
\end{align}
Note that~$\relent{\jointdistrocmi}{\productdistrocmi}=\cmi$.
The rest of the argument follows the same lines as the proof of \cref{cor:xu_raginsky}: specifically, we apply the sub-Gaussian concentration inequality and optimize over~$\lambda$, from which the result follows.
\end{proof}

The benefit of the CMI framework can now be clearly seen.
Notice that we did not need to impose any absolute continuity assumption.
Since~$\cmi=\conrelent{\jointdistrocmi}{\productdistrocmi}{\traindistrocmi}$, we need~$\jointdistrocmi$ to be absolutely continuous with respect to~$\productdistrocmi$.
But since~$P_{W\vert\supersample}$ is obtained by marginalising~$\conddistrocmi P_{\subsetchoice}$ over the discrete random variable~$\subsetchoice$, this is automatically guaranteed.
More specifically, we actually have~$\cmi\leq H(\subsetchoice)=n\log 2$, where~$H(\subsetchoice)$ is the entropy of~$\subsetchoice$ (see \cref{def:entropy}).
This confirms the motivation for introducing the framework: the training set, consisting of~$n$ samples, cannot carry more information than~$n$ bits.
However, in the worst case scenario where the trivial upper bound holds with equality---which can occur when~$\conddistrocmi$ represents a deterministic learning algorithm and gives distinct outputs for each value of~$\subsetchoice$---the resulting generalization bound is vacuous, since~$\sqrt{2\log 2}>1$.

The following interesting observation regarding the connection between CMI and mutual information was noted by~\citet{haghifam-20a}.
The motivation for having a supersample consisting of~$2n$ data samples was to normalize the information carried by each training sample to~1~bit.
However, we could have a different scheme where~$\supersample$ consisted of~$kn$ samples, for an integer~$k>2$, and instead have~$S_i$ uniformly distributed on an index set of size~$k$.
While such a construction leads to looser bounds than using~$k=2$, it can be shown that, when the hypothesis space~$\mathcal W$ is finite, the resulting conditional mutual information~$\cmi$ equals the mutual information~$I(W;\traindatacmi)$ in the limit~$k\rightarrow \infty$.

We note that the assumption of bounded loss can be somewhat relaxed, as shown in~\citet[Thm.~5.1]{steinke-20a}.
Specifically, assume that there exists a function~$\Delta: \dataspace^2\rightarrow\reals$ such that, for all~$z_1,z_2\in\dataspace$ and~$w\in\mathcal W$, we have~$\abs{\ell(w,z_1)-\ell(w,z_2)}\leq \Delta(z_1,z_2)$.
Furthermore, define~$\bar \Delta=\sqrt{\Ex{Z_1,Z_2\distas\datadistro^2}{ \Delta(Z_1,Z_2)^2 } }$.
Due to boundedness, it is clear that~$\ell(w,z_i(S_i))-\ell(w,z_i(\bar S_i))$ is~$\Delta(z_i(1),z_i(0))$-sub-Gaussian under~$P_{S_i}$ for all~$w\in\mathcal W$,~$i\in[n]$, and~$\supersamplesmall\in\dataspace^{2 n}$.
By following the same argument as above, this therefore leads to the bound
\begin{equation}\label{eq:unbounded-cmi-steinke}
\avggengap \leq \sqrt{\frac{2\bar \Delta^2 \cmi }{n} }.
\end{equation}
For simplicity, we will assume a bounded loss throughout this chapter, but we note that all bounds that are derived through a sub-Gaussianity argument can be generalized in this way.

While the bound in \cref{thm:steinke_slow} achieves a slow~$1/\sqrt n$-rate with respect to the training set size, this can, just as before, be improved at the cost of worse multiplicative constants.
In this vein, \citet{steinke-20a} also presented the following average bound.
The result essentially follows along the same lines as \cref{thm:steinke_slow}, but using \cref{thm:binary-rv-concentrate-steinke} for the concentration step.

\begin{thm}\label{thm:steinke_fast}
\boundedlosstext.
For all constants~$\gamma,\lambda>0$ satisfying~$\lambda(1-\gamma)+(e^{\lambda}-1-\lambda)(1+\gamma^2)\leq 0$, we have
\begin{equation}
\avgpoploss \leq \gamma \avgtrainloss + \frac{\cmi}{\lambda n}.
\end{equation}
\end{thm}

Under the assumption that the learning algorithm interpolates the training data almost surely, meaning that it achieves zero training loss, the constants in the bound can be improved.

\begin{thm}\label{thm:steinke_fast_interpolating}
\boundedlosstext.
Furthermore, assume that~$\avgtrainloss=0$, meaning that the algorithm interpolates the data almost surely. Then, we have
\begin{equation}\label{eq:steinke_fast_interpolating}
\avgpoploss \leq  \frac{\cmi}{n \log 2 }.
\end{equation}
\end{thm}

We will postpone the proof of this result, and instead prove it in \cref{sec:ecmi}, when we introduce the \emph{evaluated} CMI.
While it is possible to prove it without reference to evaluated CMI, as was done by~\citet{steinke-20a}, the proof becomes somewhat shorter once we introduce it.

The constant~$\log 2$ in the bound can be shown to be sharp.
Indeed, as mentioned before, the conditional mutual information is trivially upper-bounded as~$n \log 2$.
Inserting this bound into~\eqref{eq:steinke_fast_interpolating} yields a population loss bound of~$1$. Thus, if the constant could be improved, we would have a non-trivial generalization bound that holds for any algorithm, which is not possible.

We now prove a generalization bound with the binary relative entropy on the left-hand side, as before.
The functional form may appear surprising at first glance, but the reason for this quickly becomes apparent in the proof.

\begin{thm}\label{thm:cmi-binary-kl}
\boundedlosstext.
Then, for every~$\gamma\in\reals$,
\begin{equation}\label{eq:thm:cmi-binary-kl}
\gammabinrelentbig{\avgtrainloss}{\frac{\avgtrainloss+\avgpoploss}{2}} \leq \binrelentbig{\avgtrainloss}{\frac{\avgtrainloss+\avgpoploss}{2}} \leq \frac{\cmi}{n}.
\end{equation}
\end{thm}
\begin{proof}
First, by Jensen's inequality and the definition of the parametrized binary relative entropy, we have
\begin{align}
\binrelentbig{\avgtrainloss}{\frac{\avgtrainloss+\avgpoploss}{2}} &= \sup_\gamma \gammabinrelentbig{\avgtrainloss}{\frac{\avgtrainloss+\avgpoploss}{2}} \nonumber \\
 &\leq \sup_\gamma \Ex{\jointdistrocmi}{\gammabinrelentbig{\trainlosscmi}{\frac{\trainlosscmi+\testlosscmi}{2}}  }   \nonumber \\
  &\leq \sup_\gamma \Ex{\jointdistrocmi}{\gammabinrelent{\trainlosscmi}{\supersamplelosscmi  }   } .\label{eq:cmi-bin-kl-deriv-step}
\end{align}
In the last step, we used that for any~$W$ and~$\traindata$, the value of~$(\trainlosscmi+\testlosscmi)/2=\supersamplelosscmi$ is actually independent of~$\subsetchoice$---it is just the average loss on the entire supersample.
In fact, for fixed~$W,\supersample$, we have
\begin{equation}
\Ex{P_{\subsetchoice}}{\trainlosscmi} = \supersamplelosscmi.
\end{equation}
Therefore, under~$P_{\subsetchoice}$, the second argument of the binary relative entropy in~\eqref{eq:cmi-bin-kl-deriv-step} is the mean of the first argument---this motivates the form of the bound.
As per usual, we use the \donskervtext\ to change measure from~$\jointdistrocmi$ to~$\productdistrocmi$.
Then, we use \cref{thm:parametric_kl_concentration} to find that
\begin{equation}
\log \Ex{\productdistrocmi}{e^{n\gammabinrelent{\trainlosscmi}{\supersamplelosscmi  } }} \leq 0.
\end{equation}
The final result follows after reorganizing terms.
\end{proof}

Without the CMI approach, the corresponding bound in \cref{cor:binary-kl-average} had the training loss as the first argument and the population loss as the second.
Here, we instead have the arithmetic mean of the training and population loss as the second argument.
The reason for this, as seen in the proof, is that the averaging is done over~$\subsetchoice$ rather than~$\traindata$, necessitating a different form in order to use the concentration result for the binary relative entropy.
This gives rise to an additional factor of 2---similar to how this extra factor arose in the sub-Gaussian argument in the proof of \cref{thm:steinke_slow} where we had to apply sub-Gaussianity to a bounded random variable with range~$[-1,1]$ instead of~$[0,1]$.

Naturally, the bound in~\eqref{eq:thm:cmi-binary-kl} can be relaxed as before to obtain a result that more clearly illustrates the scaling of the bound, by following the same recipe used to derive \cref{cor:fast-mcallester-relaxation}.
This yields a result very similar to \cref{thm:steinke_fast}, albeit with slightly different constants.

We conclude this section by discussing the application of the individual-sample technique and disintegration to the CMI framework.
When introducing the individual-sample technique in \cref{sec:individual_sample}, one of the main motivations was to avoid infiniteness of the mutual information.
Now, as mentioned before, this problem has been solved with the CMI, which is always finite.
However, if the CMI reaches its maximum value, our bounds are still vacuous---although finite.
Hence, it is still of interest to apply these techniques in the CMI framework.
This was done by, for instance,~\citet{haghifam-20a}.
We present a bound incorporating these techniques below without proof---as expected, the bound is derived by suitably adapting the proof methods from \cref{sec:individual_sample,sec:disintegration} to the CMI bound in \cref{thm:steinke_slow}.
\begin{thm}\label{thm:haghifam_cmi}
\boundedlosstext. Then,
\begin{equation}
\avggengap \leq \frac{1}{n}\sum_{i=1}^n\Ex{P_{\supersample}}{\sqrt{\frac{I^{\supersample}(W;S_i) } {n} }} ,
\end{equation}
\end{thm}
\noindent where~$I^{\supersample}(W;S_i)=\conrelent{\conddistrocmii}{\margdistrocmi}{P_{S_i}}$.
Similar extensions can be obtained for the other CMI bounds in this section.

The application of the individual-sample technique in \cref{thm:haghifam_cmi} can be naturally extended as follows.
While we have~$\cmi\leq I(W;\traindatacmi)$, meaning that the CMI-based generalization bounds improve on their mutual information-based counterparts (up to constants), the same does not hold true when comparing \cref{thm:haghifam_cmi} to its individual-sample counterpart in \cref{cor:bu_veeravalli}.
The issue is that conditioning on the entire supersample can reveal an unnecessarily large amount of information, so that there are scenarios where we are better off not using any conditioning.
The technical reason behind this is that, in the derivation of \cref{thm:haghifam_cmi}, parts of~$\supersample$ that can be marginalized out in the samplewise decomposition of the generalization error are not marginalized.
This issue was noted by both~\citet{rodriguezgalvez-20a} and~\citet{zhou-21a}, who rectified it to obtain the following result.

\begin{thm}\label{thm:icimi}
\boundedlosstext. Then,
\begin{equation}
\avggengap
\leq \frac{1}{n}\sum_{i=1}^n\Ex{P_{\tilde Z_i \tilde Z_{i+n}}}{\sqrt{\frac{2I^{\tilde Z_i\tilde Z_{i+n}}(W;S_i) } {n} }}.
\end{equation}
\end{thm}

The proof of this is essentially the same as for \cref{thm:haghifam_cmi}, but with more care taken with regards to marginalization.
Due to the data-processing inequality, this always improves on~\cref{cor:bu_veeravalli}, up to a constant factor.

\section{PAC-Bayesian Generalization Bounds}\label{sec:cmi-pac-bayes}

So far, we have used the CMI framework to obtain generalization bounds in expectation.
Indeed, this has been the main focus in the recent CMI literature.
However, as in \cref{chap:probability}, we can also derive bounds in probability, that is, bounds on the PAC-Bayesian or single-draw loss.
PAC-Bayesian bounds were the focus of~\citet{audibert-04a,catoni-07a} in their use of almost exchangeable priors.
In this section, we will discuss such PAC-Bayesian generalization bounds in the CMI framework.
These results---which can be seen as PAC-Bayesian analogues of the results in \cref{sec:cmi-avg} or CMI analogues of the results in \cref{sec:pac-bayesian-bounds}---can be derived for all manner of bounds discussed previously.
Here, we will present one such extension of a previous bound, as well as a simplified version of an excess risk bound due to~\citet{grunwald-21a} based on the Bernstein condition.
We will also discuss the relation between the ‘‘CMI prior'' and the data-dependent prior mentioned in \cref{sec:pac-bayesian-bounds}, and touch upon some connections to other topics in learning theory.
These latter points will be fleshed out further in \cref{chap:info-complexity}.

We begin by presenting a PAC-Bayesian version of \cref{thm:steinke_slow}, which can also be seen as a CMI version of \cref{cor:first_pac-bayes_subgaussian}.
\begin{thm}\label{thm:cmi-pac-bayes}
\boundedlosstext\ and assume that~$\conddistrocmi\ll\auxconddistrocmi$.
Then, with probability at least~$1-\delta$ under~$\traindistrocmi$,
\begin{equation}\label{eq:hellstrom_slow_pacb}
 \Exop_{\conddistrocmi} \lefto[\genS\right]
 \!\leq\! \sqrt{\frac{2}{n\!-\!1}\!\left(\relent{\conddistrocmi}{\auxconddistrocmi } +\log \frac{\sqrt{n}}{\delta}\right)}.
\end{equation}
\end{thm}

The role of the unused data points,~$\traindata(\bar \subsetchoice)$, and the auxiliary conditional distribution~$\auxconddistrocmi$, which acts as the prior in the PAC-Bayesian bounds for the CMI framework, merit some discussion.
In the average bounds presented earlier in this section, these samples are purely hypothetical ‘‘ghost samples,'' and the data-generation process can be seen as a thought experiment that is just used for the proofs.
In the left-hand side, averaging~$\genS$ over~$\jointdistrocmi$ transforms the test loss~$\testlosscmi$ into the ordinary population loss~$\poploss$.
On the right-hand side, the information measure is the conditional mutual information~$\cmi$, where these ghost samples are averaged out.

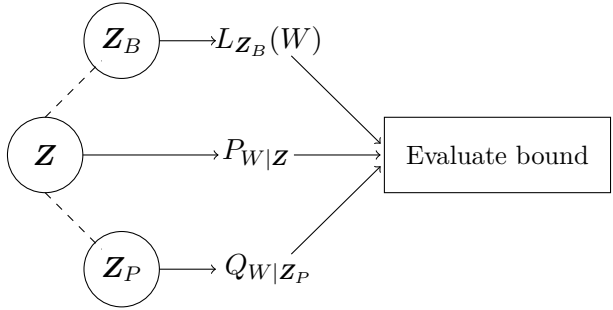
\begin{figure}[t]
\centering
\begin{tikzpicture}
\draw (0,0) circle (0.5) node[pos=.5]{ $\traindata$};
\draw [->] (0.5,0) -- (2.3,0);
\node at (2.8,0) {\normalsize $P_{W\vert\traindata}$ }  ;
\draw [->] (3.3,0) -- (4.45,0);
\draw [dashed] (0,0.5) -- (0.75,1.25);
\draw (1.015165, 1.515165) circle (0.5) node{$\traindata_B$};
\draw [dashed] (0,-0.5) -- (0.75,-1.25);
\draw (1.015165, -1.515165) circle (0.5) node{$\traindata_P$};
\draw [->] (1.515165, -1.515165) -- (2.265165, -1.515165);
\node at (2.965165, -1.515165) {\normalsize $Q_{W\vert \traindata_P}$ }  ;
\draw [->] (3.265165, -1.315165) -- (4.45, -0.15);
\draw [->] (1.515165, 1.515165) -- (2.265165, 1.515165);
\node at (2.965165, 1.515165) {\normalsize $L_{\traindata_B}(W)$ }  ;
\draw [->] (3.265165, 1.315165) -- (4.45, 0.15);
\draw (4.5,-0.5) rectangle (7.5,0.5) node[pos=.5]{\small Evaluate bound };
\end{tikzpicture}
\caption{The data-splitting approach to data-dependent priors, discussed in \cref{sec:data-dep-prior}.}
\label{fig:ambroladze_prior}
\end{figure}

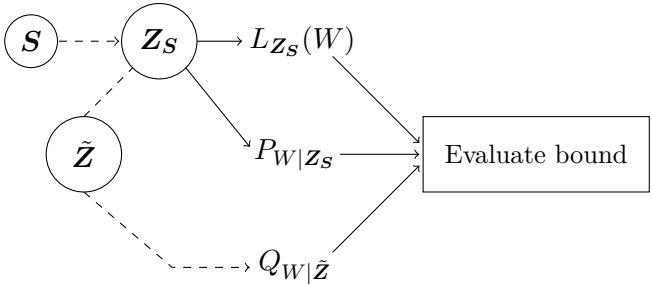
\begin{figure}[t]
\centering
\begin{tikzpicture}
\draw (0,0) circle (0.5) node[pos=.5]{ $\supersample$};
\draw [->] (1.35,1.15) -- (2.2,0.1);
\node at (2.8,0) {\normalsize $P_{W\vert \traindatacmi}$ }  ;
\draw [->] (3.4,0) -- (4.45,0);
\draw [dashed] (0,0.5) -- (0.65,1.15);
\draw (1, 1.5) circle (0.5) node{$\traindatacmi$};
\node at (2.8, -1.5) {\normalsize $Q_{W\vert\supersample}$ }  ;
\draw [->] (3.3, -1.3) -- (4.45, -0.15);
\draw [->] (1.5, 1.5) -- (2.1, 1.5);
\node at (2.9, 1.5) {\normalsize $L_{\traindatacmi}(W)$};
\draw [->] (3.3, 1.3) -- (4.45, 0.15);
\draw (4.5,-0.5) rectangle (7.5,0.5) node[pos=.5]{ \small Evaluate bound };

\draw (-0.7,1.5) circle (0.35) node{ $\subsetchoice$};
\draw [dashed, ->] (-0.33, 1.5) -- (0.47, 1.5);
\draw [dashed, ->] (0, -0.5) -- (1.5-0.33, -1.5) -- (2.2, -1.5);
\end{tikzpicture}
\caption{The CMI approach to data-dependent priors.}
\label{fig:cmi_prior}
\end{figure}

In contrast, for the data-dependent bound in \cref{thm:cmi-pac-bayes}, the left-hand side depends on the PAC-Bayesian test loss~$\Ex{\conddistrocmi}{\testlosscmi}$.
Since this is an unbiased estimate of the population loss, one can convert this into a bound on the PAC-Bayesian population loss through the triangle inequality~\citep[Thm.~3]{hellstrom-20a}.
Furthermore, the right-hand side actually explicitly depends on these unused training samples.
An upside of this is that this leads to bounds that are actually manageable to compute.
Indeed, given a set of~$2n$ training samples, one can just implement the subset-selection procedure in practice, use the obtained~$\traindatacmi$ to select a hypothesis, and select the prior freely based on~$\supersample$---provided that one does not use any knowledge of~$\subsetchoice$.
This shares many similarities with the data-splitting approach for data-dependent priors in PAC-Bayes, discussed in \cref{sec:data-dep-prior}, wherein one splits the training data into two parts: one part~$\traindata_P$ for selecting the prior, and one part~$\traindata_B$ for evaluating the training loss in the generalization bound.
Crucially, in the data-splitting approach, the selected hypothesis is still allowed to depend on all of the training data.
The two approaches are explained pictorially in \cref{fig:ambroladze_prior} and \cref{fig:cmi_prior} respectively.

From a practical standpoint, there are many reasons to prefer the data-splitting approach.
The most obvious difference is that with the data-splitting approach, all available samples can be used as input to the learning algorithm, which typically leads to better performance.
Furthermore, the PAC-Bayesian bound can be directly optimized, meaning that the information measure between posterior and prior is used as a regularizer.
For the CMI approach, such regularization would introduce illegal dependencies between~$W$ and~$\supersample$, violating the Markov property upon which the proof is based.
Thus, the resulting generalization bounds would no longer hold.

However, if the motivation is to theoretically understand learning algorithms, rather than to derive risk certificates for practical hypotheses or to devise the best algorithm, the difference is more conceptual.
The data-splitting approach can be seen as a generalization of \emph{compression schemes} \citep{shalev-shwartz-14a}.
Roughly speaking, a learning algorithm is a compression scheme of size~$k$ if its output on any training set~$\traindata$ with $n>k$ samples is the same as its output on~$\traindata_C$, consisting of~$k$ samples from~$\traindata$.
Indeed, a version of the standard generalization bounds for stable compressors can be derived from the data-splitting PAC-Bayesian bound by setting~$\traindata_P=\traindata_C$, paying a union bound cost for the~$\binom{n}{k}$ possible choices of~$\traindata_C$.
Since the output of the learning algorithm based on~$\traindata$ can be obtained entirely on the basis of~$\traindata_C$, the relative entropy will vanish, as the data-dependent prior exactly matches the posterior.
So, in this compression-related approach, the hypothesis is allowed to depend strongly on a few samples, as long as the dependence on the remaining samples is weak.

The CMI approach, on the other hand, can be seen as drawing on the notion of algorithmic stability, discussed in \cref{sec:alg-stability-bounds}.
Intuitively, algorithmic stability measures how sensitive the output hypothesis is to the inclusion of any one sample in the training set.
In a sense, this is closely related to the CMI.
Indeed, the individual-sample CMI~$\cmii$ measures how strong the dependence of the hypothesis is on the specific~$i$th sample.
This connection is explored in more detail by~\citet{harutyunyan-21a}.
We will discuss all this further in \cref{chap:info-complexity}, where the information complexity of specific algorithms is evaluated.
Of course, by repeating the arguments from \cref{sec:data-dep-prior}, we can combine the data-splitting technique and the CMI approach.
This ensures that the prior and posterior have a fixed set of samples in common, which are absent from the training loss in the bound, while the remaining samples are randomly selected through the CMI procedure.

We conclude this section by stating a variant of a result of~\citet[Cor.~1]{grunwald-21a}, which provides a PAC-Bayesian excess risk bound with potentially fast rates.
Since the statement of the result and its proof are quite involved, we will only provide a simplified version without proof.
The full details, including extensions to average bounds through an exponential stochastic inequality and other variants, can be found in the work of~\citet{grunwald-21a}.

\begin{thm}\label{thm:grunwald-cmi-bound}
\boundedlosstext.
Furthermore, assume that the~$\beta$-Bernstein condition is satisfied, \ie, for some~$\beta\in[0,1]$, there exists a~$w^*\in\mathcal W$ such that, for all~$w\in\mathcal W$,
\begin{equation}
\Ex{\datadistro}{ \left(\ell(w,Z) \!-\! \ell(w^*,Z) \right)^2 } \leq 4\left( \Ex{\datadistro}{ \ell(w,Z) \!-\! \ell(w^*,Z)  } \right)^\beta.
\end{equation}
Then, for some~$C_1,C_2\in\reals^+$ with probability at least~$1-\delta$ under~$P_{\traindatacmi}$,
\begin{multline}\label{eq:grunwald-cmi-bound}
 \Exop_{\conddistrocmi} \lefto[\genS\right]  \leq \min(1,2\beta)\left(\Ex{\conddistrocmi}{\trainlosscmi} - L_{\traindatacmi}(w^*) \right) \\
 + C_1 \left( \frac{ \Ex{P_{\testdatacmi}}{ \relent{\conddistrocmi}{\auxconddistrocmi} } +\log (\sqrt n)}{ n } \right)^{\frac{1}{2-\beta}} + \frac{C_2\log\frac1\delta}{\sqrt n} .
\end{multline}
\end{thm}
The first term on the right-hand side of~\eqref{eq:grunwald-cmi-bound} is the (scaled) \emph{empirical excess risk}, \ie, the degree to which the training loss of the learning algorithm exceeds the training loss of the optimal hypothesis with respect to the population loss.
For many algorithms this is negligible, and for empirical risk minimizers, it is guaranteed to be non-positive.
There are several notable aspects of this result.
Since the loss is bounded, the~$\beta$-Bernstein condition always holds with~$\beta=0$, which means that the slow rate of~$1/\sqrt n$ can be obtained (in which case the empirical excess risk does not enter the bound).
However, for smooth losses such as the squared or logistic loss, it also holds with~$\beta=1$, enabling the relative entropy-dependent term to decay faster.
Furthermore, the relative entropy enters only averaged over the test data.
While this leads to the relative entropy being non-empirical, in the sense that it cannot be computed based on the training set and learning algorithm, it can still be shown to be bounded in several cases, such as for hypothesis classes with bounded VC dimension.
Indeed, this is the main focus of~\citet{grunwald-21a}: deriving a fast-rate bound that is well-behaved for VC classes.
We will discuss this further in \cref{sec:info-comp-vc}.
We emphasize again that the result in \cref{thm:grunwald-cmi-bound} is not stated in its full generality nor tightness, but has been significantly simplified in terms of assumptions, constants, and logarithmic dependencies in the interest of brevity.

\section{Single-Draw Generalization Bounds}

Before moving on to extensions of the CMI framework, we turn to single-draw bounds.
As for the average and PAC-Bayesian bounds, we can also derive CMI versions of the single-draw bounds from \cref{sec:single-draw-bounds}.

We begin by stating a basic single-draw bound, now given in terms of the conditional information density~$\condinfdens$.
When averaged over the joint distribution~$\jointdistrocmi$, the conditional information density gives the CMI~$\cmi$---hence the name.
As the proof is a straightforward adaptation of previous derivations, we do not give it explicitly.

\begin{thm}\label{thm:cmi-single-draw}
\boundedlosstext.
Then, with probability at least~$1-\delta$ under~$\jointdistrocmi$, %
\begin{equation}\label{eq:hellstrom_slow_sd}
\abs{\genS}   \leq  \sqrt{{\frac{2}{n-1}\left( \condinfdens+\log \frac{\sqrt{n}}{\delta}\right)} } .
\end{equation}
\end{thm}

The techniques from \cref{sec:holder}, where repeated uses of Hölder's inequality were used to obtain a generic bound on the probability of an event under one distribution in terms of another, can also be extended to the CMI framework.
In terms of the proof, we need to consider three random variables, perform the change of measure conditioned on one of them, and make use of Hölder's inequality an additional time.
We present the generic result below.

\begin{thm}\label{thm:holder_three_variables}
For all constants $\alpha, \gamma,\alpha',\gamma',\tilde \alpha, \tilde \gamma> 1$ such that $1/\alpha + 1/\gamma =1/\alpha'+1/\gamma'=1/\tilde \alpha + 1/\tilde \gamma =1$ and all measurable sets $\setE\in \mathcal{W}\times \mathcal{Z}^{2n}\times \{0,1\}^n$,
\begin{align}\label{eq:holder_three_variables}
P_{W\! \supersample\! \subsetchoice}[\setE]
\leq &\Exop_{P_{\supersample} }^{1/\tilde \gamma}\lefto[ \Exop_{P_{W\vert \supersample}}^{\tilde \gamma/\gamma '}\lefto[P_{\subsetchoice}^{\gamma'/\gamma}\lefto[\setE_{W\! \supersample}\right] \right] \right]   \times \\ &\Exop_{P_{\supersample}}^{1/\tilde \alpha}\lefto[ \Exop_{P_{W\vert \supersample}}^{\tilde \alpha/\alpha '}\lefto[\Exop_{P_S}^{\alpha ' /\alpha}\lefto[e^{\alpha \condinfdens} \right] \right] \right]. \nonumber
\end{align}
Here,~$\setE_{W\!\supersample}=\{\subsetchoice: (W,\supersample,\subsetchoice)\in \setE \}$.
\end{thm}

\begin{proof}
First, we rewrite~$P_{W\! \supersample\! \subsetchoice}[\setE] $ in terms of the expectation of the indicator function~$1_\setE$ and perform a change of measure:
\begin{align}
P_{W\!\supersample\! \subsetchoice}[\setE]&= \Exop_{P_{W\vert \supersample }P_{\supersample\! \subsetchoice}} \lefto[1_\setE \cdot \frac{\dv P_{W\! \supersample\! \subsetchoice} }{\dv P_{W\vert \supersample} P_{\supersample\! \subsetchoice} } \right] \\
&= \Exop_{ P_{W\vert \supersample}  P_{\supersample } P_{\subsetchoice}} \lefto[ 1_{\setE} \cdot  e^{\condinfdens} \right].
\end{align}
To obtain the desired result, we apply H\"older's inequality thrice. Let~$\alpha$, $\gamma$, $\alpha'$, $\gamma'$, $\tilde \alpha$,  $\tilde \gamma> 1$ be constants such that~$1/\alpha + 1/\gamma =1/\alpha'+1/\gamma'=1/\tilde \alpha + 1/\tilde \gamma =1$. Then,
\begin{align}
&{P_{W\! \supersample\! \subsetchoice}}[\setE]
 \leq  \Exop_{P_{W\vert \supersample}P_{\supersample}\!}\biggo[ \Exop_{P_{\subsetchoice}}^{1/\gamma}\lefto[1_{\setE_{W\! \supersample}}\right]\!\cdot\! \Exop_{P_{\subsetchoice}}^{1/\alpha}\lefto[ e^{\alpha \condinfdens} \right] \bigg]\\
& \leq  \Exop_{P_{\supersample}} \biggo[ \Exop_{P_{W\vert \supersample}}^{1/\gamma'}\lefto[P_{\subsetchoice}^{\gamma'/\gamma}\lefto[\setE_{W\! \supersample}\right] \right] \cdot \Exop_{P_{W\vert \supersample}}^{1/\alpha'}\lefto[\Exop_{P_{\subsetchoice}}^{\alpha'/\alpha}\lefto[ e^{\alpha \condinfdens} \right] \right]  \bigg] \nonumber \\
& \leq  \Exop_{P_{\supersample}}^{1/\tilde \gamma}\biggo[ \Exop_{P_{W\vert \supersample}}^{\tilde \gamma/\gamma'}\lefto[P_{\subsetchoice}^{\gamma'/\gamma}\lefto[\setE_{W\! \supersample}\right] \right] \bigg] \cdot \Exop_{P_{\supersample}}^{1/\tilde \alpha}\biggo[ \Exop_{P_{W\vert {\supersample}}}^{\tilde \alpha/\alpha'}\lefto[\Exop_{P_{\subsetchoice}}^{\alpha'/\alpha}\lefto[e^{\alpha \condinfdens} \right] \right] \bigg]. \nonumber
\end{align}
\end{proof}

Many different types of bounds can be obtained by making different choices for the three free parameters in \cref{thm:holder_three_variables}.
We will focus on a choice that leads to bounds in terms of a version of the conditional~$\alpha$-mutual information (\cref{def:alpha_information}).

We emphasize two properties of the conditional~$\alpha$-mutual information. First, in the limit~$\alpha\rightarrow \infty$, it reduces to the conditional maximal leakage \citep[Thm.~6]{issa-16a}:
\begin{equation}%
 \mathcal{L}(\subsetchoice \rightarrow W\vert \supersample)=
 \log\esssup_{P_{\supersample}} \Exop_{P_{W\vert \supersample}}\lefto[ \esssup_{P_{\subsetchoice\vert \supersample} }e^{\condinfdens} \right].
\end{equation}
Second, for~$\alpha>1$, one can see that the conditional~$\alpha$-mutual information is upper-bounded by the conditional R\'enyi divergence of order~$\alpha$, as shown in~\eqref{eq:alpha-cmi-alpha-mi-relation}.

After this aside, we return to presenting the generalization bound in terms of the conditional~$\alpha$-mutual information.

\begin{cor}\label{cor:singledraw_esposito_cond_alphaMI}
\boundedlosstext.
Then, for any fixed~$\alpha > 1$, the following holds with probability at least~$1-\delta$ under~$P_{W\! \supersample\! \subsetchoice}$:
\begin{equation}\label{eq:cor_singledraw_esposito_cond_alphaMI}
\abs{ \genS } \leq \sqrt{\frac{2}{n}\bigg(\alphaconMI{\alpha}{W}{\subsetchoice}{\supersample }
+\log 2 + \frac{\alpha}{\alpha-1}\log \frac{1}{\delta}\bigg)}.
\end{equation}
\end{cor}
\begin{proof}
In~\eqref{eq:holder_three_variables}, set~$\tilde \alpha = \alpha$ and let~$\alpha' \rightarrow 1$, which implies that~$\tilde \gamma = \gamma$ and~$\gamma' \rightarrow\infty$. Also, let~$\setE$ be the error event
\begin{equation}\label{eq:high_error_event_cond}
\setE=\{W,\supersample, \subsetchoice: \abs{\genS}>\varepsilon \}.
\end{equation}
For this choice of parameters, the second factor in~\eqref{eq:holder_three_variables} reduces to
\begin{multline}\label{eq:esposito_alphami_proof_info_dens_factor}
\Exop_{P_{\supersample}}^{1/\alpha}\lefto[ \Exop^{\alpha}_{P_{W\vert \supersample}}\lefto[\Exop^{1/\alpha}_{P_S}\lefto[\exp\lefto(\alpha\imath(W,\subsetchoice\vert\supersample)\right) \right] \right] \right] \\=
\exp\lefto(\frac{\alpha-1}{\alpha}\alphaconMI{\alpha}{W }{\subsetchoice}{\supersample }\right).
\end{multline}
Furthermore, we can bound~$P_{\subsetchoice}\lefto[\setE_{W\! \supersample}\right]$ in the first factor in~\eqref{eq:holder_three_variables} by using sub-Gaussianity to find that, for all~$W$ and~$\supersample$,
\begin{equation}
P_{\subsetchoice}\lefto[\setE_{W\! \supersample}\right]\leq 2 \exp\lefto(-\frac{n\varepsilon^2}{2} \right) .
\end{equation}
Using this in the first factor of~\eqref{eq:holder_three_variables}, we conclude that
\begin{align}
\lim_{\gamma' \rightarrow\infty} \Exop_{P_{\supersample}}^{1/ \gamma}\lefto[ \Exop_{P_{W\vert \supersample}}^{ \gamma/\gamma'}\lefto[P^{\gamma'/\gamma}_{\subsetchoice}\lefto[\setE_{W\! \supersample}\right] \right] \right]
& = \Exop^{1/\gamma}_{P_{\supersample}}\lefto[ \lefto(\esssup_{P_{W\vert \supersample}}P^{1/\gamma}_{\subsetchoice}\lefto[\setE_{W\! \supersample}\right]\right)^{\gamma} \right] \nonumber \\
 & \leq  \left(  2 \exp\lefto(-\frac{n\varepsilon^2}{2} \right)\right)^{1/\gamma}. \label{eq:esposito_alphami_proof_E_dep_factor}
\end{align}
By substituting~\eqref{eq:esposito_alphami_proof_info_dens_factor} and~\eqref{eq:esposito_alphami_proof_E_dep_factor}  into~\eqref{eq:holder_three_variables}, noting that~$1/\gamma = (\alpha-1)/\alpha$, we conclude that
\begin{equation}\label{eq:deriv_cond_alpha_mi_last}
P_{W\! \supersample \! \subsetchoice}[\setE]
\leq \left( 2 \exp\lefto(-\frac{n\varepsilon^2}{2} \right) \right)^{\frac{\alpha-1}{\alpha} }  \cdot
\exp\lefto(\frac{\alpha-1}{\alpha}\alphaconMI{\alpha}{W }{\subsetchoice}{\supersample }\right).
\end{equation}
We obtain the desired result by requiring the right-hand side of~\eqref{eq:deriv_cond_alpha_mi_last} to equal~$\delta$ and solving for~$\varepsilon$.
\end{proof}

By letting~$\alpha\rightarrow \infty$, we obtain a generalization bound in terms of the conditional maximal leakage:
\begin{equation}\label{eq:cond-maxleakage-gen-bound}
\abs{ \genS } \leq \sqrt{\frac{2}{n}\bigg( \conmaxleakage{\subsetchoice}{W}{\supersample}
+\log \frac{2}{\delta}\bigg)}.
\end{equation}

It can be shown that~$\conmaxleakage{\subsetchoice}{W}{\supersample}\leq \maxleakage{\traindatacmi}{W}$~\citep[Thm.~5]{hellstrom-20a}.
Thus, up to constants and the penalty term incurred to obtain a bound on the population loss (as per \citealp[Thm.~3]{hellstrom-20a}), \eqref{eq:cond-maxleakage-gen-bound} improves on \eqref{eq:maxleakage-gen-bound}.

\section{Evaluated CMI and $f$-CMI}\label{sec:ecmi}

As noted by~\citet{steinke-20a}, there is a potential deficiency that comes with measuring information as captured by the hypothesis~$W$ itself.
For instance, if~$W$ is a real number, we can take the output of an algorithm with low CMI, and change~$W$ so that it encodes the training set in its insignificant digits.
For most settings, this change should have a negligible effect on the generalization of the algorithm, but the CMI will be maximized, leading to vacuous bounds.
Another way to see the issue is to consider the case where~$W$ consists of the weights of a neural network.
Neural networks (which will be discussed in \cref{chap:iterative-methods}) typically possess many symmetries, such as permutation and scaling invariance, so that different values of~$W$ can represent the exact same function.
Ideally, we would want to obtain generalization bounds in terms of a measure that is more directly related to the predictions our hypothesis produces and the losses that they incur.

As it turns out, this can be accomplished in a straightforward way.
In fact, we barely need to change the derivations we have used so far.
Consider, for instance, the derivation of \cref{thm:steinke_slow}.
In the proof,~$W$ only appears in a ‘‘processed'' version, either through the loss on a training sample or the loss on a test sample.
Hence, the derivation can be adapted so that no explicit mention is made of~$W$, but instead, only the losses that it incurs on the supersample appear in both the derivation and the final result.
Motivated by this, we introduce the notation~$\supersampleloss\in [0,1]^{2n}$ to denote the random vector that contains the losses that the hypothesis incurs on the entire supersample.
Specifically, the~$i$th element of~$\supersampleloss$ is~$\Lambda_i=\ell(W,\supersamplez_i)$.
Now, we proceed as in the proof of \cref{thm:steinke_slow}, with~$\supersampleloss$ replacing~$W$.
As observed by~\citet{steinke-20a}, this leads to a bound in terms of~$\ecmi$, referred to as the \emph{evaluated} CMI, or e-CMI for short.
In fact, as pointed out by~\citet{haghifam-22a}, we can even avoid any explicit reference to the supersample, leading to a bound in terms of the evaluated mutual information~$\emi$, abbreviated as e-MI.
For Theorem~\ref{thm:steinke_slow}, this can be taken even further, in fact, by noting that only the \emph{difference} between training and test losses actually enters the derivation, as done by~\citet{wang-23a}.
To this end, we define the vector of loss differences~$\supersamplelossdiff$, with elements given by~$\Delta_i=\ell(W,\supersamplez_{i+ n})-\ell(W,\supersamplez_{i})$, and the resulting loss-difference mutual information~$\edmi$, or ld-MI for short.
This gives rise to the following three upper bounds on the average generalization error.
While we only present the bounds for the full-sample square-root bound, analogous results can be derived for other comparator functions and using individual samples.

\begin{thm}\label{thm:slow_ecmi}
\boundedlosstext.
Then,
\begin{equation}
\avggengap \leq \sqrt{\frac{2\edmi}{n} } \leq \sqrt{\frac{2\emi}{n} } \leq \sqrt{\frac{2\ecmi}{n} }.
\end{equation}
\end{thm}
\begin{proof}
We start off by rewriting the generalization gap in terms of~$\supersamplelossdiff$:
\begin{align}
\Ex{\jointdistrocmi}{\testlosscmi\!-\!\trainlosscmi} \!&=\! \Ex{\jointdistroecmi}{\frac1n \sum_{i=1}^n \left(\Lambda_{i+\bar S_in} \!-\!   \Lambda_{i+S_in} \right)} \\
&= \Ex{\jointdistroemi}{\frac1n \sum_{i=1}^n \left(\Lambda_{i+\bar S_in} \!-\!   \Lambda_{i+S_in} \right)}\\
&= \Ex{\jointdistroedmi}{\frac1n \sum_{i=1}^n  (-1)^{S_i} \Delta_i } .
\end{align}
The remainder of the proof proceeds by changing measure to~$\productdistroedmi$ through the use of the \donskervtext.
The remaining steps of the proof are identical to those used in deriving \cref{thm:steinke_slow}.
The relaxation in terms of the~e-MI follows due to the data-processing inequality.
Finally, since~$\subsetchoice$ and~$\supersample$ are independent, the relaxation in terms of the e-CMI follows since conditioning on independent random variables does not decrease mutual information.
\end{proof}

Expressing generalization bounds in terms of ld-MI, e-MI, and e-CMI can have drastic consequences for the tightness of the resulting bound.
This new approach guarantees that any two hypotheses that lead to the same losses on the supersample will be considered equivalent by our information measure.
While we will not present it explicitly, it is of course possible to apply this approach to the other bounds discussed in the previous sections, including the PAC-Bayesian and single-draw bounds.

By the data-processing inequality, bounds in terms of the CMI from earlier in this chapter can be re-obtained from the ld-MI.
For supervised learning, where the learning algorithm implements a function~$f_W:\mathcal X \rightarrow \mathcal Y$, we can also consider the \emph{functional} CMI ($f$-CMI), studied by~\citet{harutyunyan-21a}.
Specifically, if we assume that each sample consists of label---example pairs~$\supersamplez_i=(\tilde X_i, \tilde Y_i)$, and let~$\mathbf F=(F_1,\dots,F_{2n})$ denote the vector of predictions induced by~$W$, \ie,~$F_i=f_W(\tilde X_i)$, we get the following chain of inequalities:
\begin{equation}
\edmi \leq \emi \leq \ecmi \leq \fcmi \leq \cmi.
\end{equation}
Here, each step is a consequence of the data-processing inequality (or conditioning on independent random variables).
Thus, the tightest of these bounds is the one in terms of the ld-MI, and all the others can be obtained in a straightforward way from this.
It should be noted that the bounds in terms of evaluated mutual informations are, in a sense, more restrictive: they only apply to a specific loss function.
In contrast, bounds in terms of the $f$-CMI and CMI bounds apply to any bounded loss function.
Finally, bounds in terms of the CMI only require knowledge of the hypothesis itself.
In \cref{chap:info-complexity,chap:iterative-methods}, we will provide interpretations of each of these information-theoretic quantities as components of generalization bounds.

We end this section by providing the promised proof of the sharp generalization bound for interpolating learning algorithms, as mentioned after \cref{thm:steinke_fast_interpolating}.
We will do this through a communication-inspired proof, due to \citet{wang-23a}.

\begin{thm}\label{thm:interp-edmi}
Assume that the loss function is binary, meaning that for all~$w\in\mathcal W$ and~$z\in\dataspace$,~$\ell(w,z)\in\{0,1\}$.
Consider an interpolating learning algorithm, so that the training loss~$\avgtrainloss=0$. Then,
\begin{equation}
\avgpoploss = \frac1n \sum_{i=1}^n \frac{\edmii}{\log 2} \leq \frac{\edmi}{n\log 2}.
\end{equation}
\end{thm}
\begin{proof}

Consider the weighted directed graph in \cref{fig:channel-interp-alg}, depicting the communication channel between~$S_i$ and~$\Delta_i=\ell(W,\supersamplez_{i+ n})-\ell(W,\supersamplez_{i})$ that is induced by the learning problem.

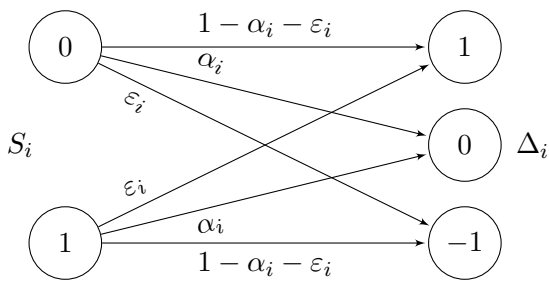
\begin{figure}[t]
\centering\begin{tikzpicture}[shorten >=1pt, node distance=1.3cm, auto, on grid, >=latex']
    \node[state] (A) {$0$};
    \node (mid) [below=of A] {};
    \node (S) [left=of mid, xshift = 0.7cm] {$S_i$};
    \node[state] (B) [below=of mid] {$1$};
    \node[state] (1) [right=of A, xshift = 4cm] {$1$};
    \node[state] (0) [below=of 1] {$0$};
    \node (S) [right=of 0, xshift = -0.4cm] {$\Delta_i$};
    \node[state] (-1) [below=of 0] {$-1$};

    \draw[->] (A) edge node [above,sloped] {$1-\alpha_i-\varepsilon_i$} (1);
    \draw[->] (A) edge node [above,sloped, xshift=-0.8cm] {$\alpha_i$} (0);
    \draw[->] (A) edge node [below,sloped, xshift=-1.8cm] {$\varepsilon_i$} (-1);

    \draw[->] (B) edge node [below,sloped, xshift=-0.8cm] {$\alpha_i$} (0);
    \draw[->] (B) edge node [above,sloped, xshift=-1.8cm] {$\varepsilon_i$} (1);
    \draw[->] (B) edge node [below,sloped] {$1-\alpha_i-\varepsilon_i$} (-1);
\end{tikzpicture}
\caption{Communication channel from~$S_i$ to~$\Delta_i$ induced by the learning algorithm.}
\label{fig:channel-interp-alg}
\end{figure}

We can now interpret the meaning of these transitions, and hence their probabilities.
First, notice that both~$0\rightarrow -1$ and~$1\rightarrow 1$ imply that the learning algorithm incurs a loss on the training sample, which contradicts the interpolating assumption.
Hence, we have~$\varepsilon_i=0$.
Next, note that any transition to~$\Delta_i=0$ means that the test loss is zero, so that we do not incur a loss.
Only the transitions~$0\rightarrow 1$ and~$1\rightarrow -1$ represent situations where a loss is incurred.
Hence, for each input, the probability of incurring a loss on the test sample is~$1-\alpha_i$.
Furthermore, for the specified communication channel, it can be shown that the Shannon capacity (with~$\varepsilon_i=0$) is~$(1-\alpha_i)\log 2$ \citep[Problem~7.13]{cover-06a}.
It is well-known that this equals the mutual information between input and output for a uniform input distribution, \ie,~$I(\Delta_i;S_i)$.
Hence,
\begin{equation}
\avgpoploss = \frac1n\sum_{i=1}^n \Ex{ \jointdistroedmi }{ (-1)^{S_i}\Delta_i } = \frac1n\sum_{i=1}^n (1-\alpha_i) = \frac1n\sum_{i=1}^n \frac{\edmii}{\log 2}.
\end{equation}
The full-sample relaxation follows by the chain rule and conditioning on independent random variables.
\end{proof}

Thus, remarkably, for the case of a binary loss and interpolating learning algorithm, the average population loss can be exactly characterized in terms of the samplewise, loss-difference mutual information.
By progressively upper-bounding this information measure through the data-processing inequality and the chain rule of mutual information, we can go step by step all the way to bounds in terms of the mutual information between the hypothesis and the training data.
More on this result, as well as several interesting extensions, can be found in the work of~\citet{wang-23a}.

Now, \cref{thm:interp-edmi} and data-processing do not directly imply \cref{thm:steinke_fast_interpolating}, since the latter assumes a generic bounded loss.
In order to relate these two results, we need the following observations.
Let~$\ell(\cdot,\cdot)$ be a generic loss function bounded to~$[0,1]$, and let~$\tilde \ell(\cdot,\cdot)$ denote a binarized version of the underlying loss function, given by~$\tilde\ell(w,z)=1\{\ell(w,z) > 0\}$.
Let~$\tilde\avgpoploss$ denote the population loss with respect to~$\tilde \ell(\cdot,\cdot)$, and let~$\tilde \supersamplelossdiff$ denote the loss-difference vector with respect to~$\tilde \ell(\cdot,\cdot)$.
Since~$\tilde \ell(\cdot,\cdot)$ is an upper bound to~$\ell(\cdot,\cdot)$, we have~$\avgpoploss\leq\tilde\avgpoploss$.
Furthermore, since~$\tilde \supersamplelossdiff$ is a processed version of~$\supersamplelossdiff$, we have~$I(\tilde \Delta_i;S_i) \leq I(\Delta_i ;S_i)$.
Thus, \cref{thm:interp-edmi} can be relaxed in order to obtain \cref{thm:steinke_fast_interpolating}.

\section{Leave-One-Out CMI}\label{sec:loo-cmi}

When we introduced the CMI framework, the size of the supersample being~$2n$ was natural:
the aim was to normalize the information carried by each sample to~$1$ bit.
Then, the bounds were derived by comparing the loss on the randomly selected samples, which gives a training loss, and the loss on the parts of the supersample that were \emph{not} selected, which gives a test loss.
In a sense, however, the~$n$ unused samples of~$\supersample$ seem quite wasteful.
Indeed, if we instead were to use all but one sample for the training set, the single remaining sample would suffice for a test loss that provides an unbiased estimate of the population loss.

As it turns out, a variant of the CMI framework where the supersample is of size~$n+1$ is possible, as demonstrated independently by~\citet{haghifam-22a} and~\citet{rammal-22a}.
In order to avoid confusion with the~$2n$-supersample setup, we will use a slightly different notation.
Specifically, we let~$\supersampleloo=(\supersamplelooarg{1},\dots,\supersamplelooarg{n+1})$ denote a vector of~$n+1$ samples drawn independently from~$P_Z$, and let~$U$ be drawn uniformly at random from~$[n+1]=\{1,\dots,n+1\}$.
Based on this, the training set~$\traindatacmiloo$ is formed by removing the~$U$th element from~$\supersampleloo$, while the~$U$th element~$\testdatacmiloo$ is a test sample.
We denote the vector of losses on the supersample by~$\dot\supersampleloss$, with elements given by~$\dot\Lambda_i=\ell(W,\supersamplelooarg{i})$.
Throughout, we assume that the range of~$\ell(\cdot,\cdot)$ is~$[0,1]$.
Using this setup, we can derive bounds in terms of the leave-one-out CMI~$\loocmi$, or loo-CMI for short, or analogous variants as we have seen before, such as the evaluated loo-MI~$\looemi$.
The name is due to its connections to the leave-one-out loss cross validation error, defined as
\begin{equation}
\loocv(\dot\supersampleloss,U) = \frac{1}{n}\sum_{i\neq U} \dot\Lambda_i-\dot\Lambda_U .  %
\end{equation}
Note that, when averaged over the joint distribution of the random variables involved, the leave-one-out cross validation error equals the generalization gap~$\avggengap$.

Compared to the CMI quantities from before, the loo-CMI is significantly less complex to compute.
When computing the CMI (and evaluated versions of it), one needs to average over the~$2^n$ possible values of~$\subsetchoice$.
In contrast, for the loo-CMI,~$U$ can only take~$n+1$ possible values---an exponential reduction in the number of cases that need to be considered.

With this notation in place, let us derive generalization bounds.
For the change of measure step, there are no surprises: we can simply use, for instance, the \donskervtext\ to replace the true joint distribution with one where~$U$ is independent from the other random variables.
For the concentration of measure step, we need the following result.
\begin{lem}
\boundedlosstext.
For all~$\dot\supersampleloss$ and~$\lambda\in\reals$,
\begin{equation}\label{eq:loocv-concentration}
\Ex{P_{U}}{ \exp\lefto( \lambda \loocv(\dot\supersampleloss,U) - \frac{\lambda^2(n+1)^2}{8n^2} \right) } \leq 1.
\end{equation}
\end{lem}
This looks very similar to the exponential inequality for sub-Gaussian random variables from \cref{def:subgauss-rv}, and can be used in an analogous way to derive generalization bounds.
The proof of this result is given by~\citet{rammal-22a}.
A simpler bound (of the same order), as used by~\citet{haghifam-22a}, can be obtained by simply noting that~$\loocv(\dot\supersampleloss,U)$ is bounded to~$[-1,1]$ and using the fact that bounded random variables are sub-Gaussian.
This gives
\begin{equation}
\Ex{P_{U}}{ \exp\lefto( \lambda \loocv(\dot\supersampleloss,U) - \frac{\lambda^2}{2} \right) } \leq 1,
\end{equation}
which can also be directly obtained from~\eqref{eq:loocv-concentration} by using the fact that~$(n+1)/n\leq 2$.

As before, by following the recipe from the proof of \cref{cor:xu_raginsky}, we can obtain the following generalization bound.
\begin{thm}\label{thm:loocmi}
\boundedlosstext.
Then,
\begin{equation}
\abs{\avggengap} \leq \frac{n+1}{n}\sqrt{\frac{\looemi}{2} } .
\end{equation}
\end{thm}
\begin{proof}
We begin from \eqref{eq:loocv-concentration}.
By averaging over~$P_{\dot\supersampleloss}$, changing measure to~$P_{\dot\supersampleloss U}$, and using Jensen's inequality, we get, for~$\lambda>0$,
\begin{equation}
\Ex{P_{\dot\supersampleloss U}}{\loocv(\dot\supersampleloss,U) }   \leq    \frac{\lambda( n+1)^2}{8^2} + \frac{\looemi}{\lambda}.
\end{equation}
As previously mentioned, the average of the leave-one-out cross-validation loss is the generalization gap.
The result follows by optimizing over~$\lambda$ and repeating the argument for~$\lambda<0$.
\end{proof}

Unlike most generalization bounds that we reviewed so far, the result in \cref{thm:loocmi} does not decay with~$n$ (ignoring the~$n$-dependence of the information measure).
Since~$U$ can take at most~$n+1$ values, a trivial upper bound on the evaluated loo-MI is~$\looemi\leq \log(n+1)$, so the bound could grow logarithmically with~$n$ in the worst case.

Finally, we present a bound for interpolating learning algorithms in terms of the evaluated loo-MI due to \citet{haghifam-22a}.
As for \cref{thm:interp-edmi}, this result can be derived through an argument that, essentially, just uses the Shannon capacity formula of a suitably chosen discrete memoryless communication channel.
The proof below, which follows the one of~\citet{haghifam-22a}, proceeds without explicit reference to such a channel, and instead simply relies on the manipulation of information-theoretic quantities.
\begin{thm}\label{thm:interp-looemi}
Assume that the loss function is binary, meaning that for all~$w\in\mathcal W$ and~$z\in\dataspace$,~$\ell(w,z)\in\{0,1\}$.
Consider an interpolating learning algorithm, so that the training loss~$\avgtrainloss=0$. Then,
\begin{equation}
\avgpoploss = \frac{\looemi}{\log(n+1)} .
\end{equation}
\end{thm}
\begin{proof}
To prove this result, we will simply compute the entropies in the decomposition~$\looemi=H(\dot\supersampleloss)-H(\dot\supersampleloss\vert U)$.
For~$i\in[n+1]$, let~$0^{(i)}$ denote the~$n+1$ vector with $0^{(i)}_j=0$ for~$j\neq i$ and~$0^{(i)}_i=1$, and let~$0^{(0)}$ denote the all-zeros vector of size~$n+1$.
Since the learning algorithm is interpolating,~$\dot\supersampleloss$ can incur a loss for at most one element of~$\supersampleloo$, and hence, the support of~$\dot\supersampleloss$ is the set~$\{0^{(i)}:i\in\{0,\dots,n+1\}\}$.
For~$i>0$,~$P[\dot\supersampleloss = 0^{(i)}]$ is the probability of not training on the~$i$th sample times the probability of incurring a loss on that sample if it is not used for training---\ie, the test loss. Hence,
\begin{equation}
P[\dot\supersampleloss = 0^{(i)}] = \frac{1}{n+1}P[\dot\supersampleloss = 0^{(i)} \vert U=i] = \frac{\avgpoploss}{n+1}.
\end{equation}
Furthermore,~$P[\dot\supersampleloss = 0^{(0)}]$ is the probability of not incurring a loss on the test sample, \ie,~$1-\avgpoploss$.
Hence, we can calculate the entropy of~$\dot\supersampleloss$ as
\begin{align}
H(\dot\supersampleloss) &= - \sum_{i=0}^{n+1} P[\dot\supersampleloss = 0^{(i)}] \log\lefto( P[\dot\supersampleloss = 0^{(i)}]\right) \\
&= -(1-\avgpoploss)\log(1-\avgpoploss) - \avgpoploss \log\lefto( \frac{\avgpoploss}{n+1} \right) .
\end{align}
Through a similar calculation, we get
\begin{equation}
H(\dot\supersampleloss\vert U) = -(1-\avgpoploss)\log(1-\avgpoploss) - \avgpoploss \log\lefto( \avgpoploss \right).
\end{equation}
Putting it all together, we find that
\begin{equation}
\looemi = H(\dot\supersampleloss) - H(\dot\supersampleloss\vert U) = L\log(n+1),
\end{equation}
from which the result follows.
\end{proof}
As noted after \cref{thm:interp-edmi}, this result can be leveraged to obtain a bound for generic bounded losses.

We thus have two characterizations of the binary loss of interpolating learning algorithms that hold with \emph{equality}, both from \cref{thm:interp-edmi} and \cref{thm:interp-looemi}.
Hence, it follows that the two characterizations must be equivalent, implying that
\begin{equation}
 \frac1n \sum_{i=1}^n \frac{\edmii}{\log 2} = \frac{\looemi}{\log(n+1)}.
\end{equation}
This dual perspective may be beneficial in specific applications, making it possible to choose whichever representation is easier to analyze.

\section{Bibliographic Remarks and Additional Perspectives}\label{sec:bib-remarks-cmi}

The underlying concept of the CMI framework can be traced back to the work of~\citet{audibert-04a} under the name of almost exchangeable priors.
Specifically, a function~$Q$ on~$\dataspace^{2n}$ is almost exchangeable if, for any permutation~$\pi$ such that~$\{\pi(i),\pi(i+n)\}=\{i,i+n\}$ for all~$i\in[n]$, it satisfies~\citep[Definition~1.1]{audibert-04a}
\begin{equation}
Q(\tilde Z'_1,\dots,\tilde Z'_{2n}) = Q(\tilde Z'_{\pi(1)},\dots,\tilde Z'_{\pi(2n)})
\end{equation}
for any~$\supersample'\in\dataspace^{2n}$.
Given a~$\supersample'=(\traindata,\traindata')\in\dataspace^{2n}$, where the first~$n$ samples are the training set and the last~$n$ are~$n$ independent samples (a ‘‘ghost sample''), an almost exchangeable prior is an almost exchangeable function of~$\supersample'$.
Thus, it has to be invariant to permutations that swap~$n$-separated pairs, \ie, the~$i$th training sample with the~$i$th independent ghost sample.
This ensures that the prior does not encode knowledge of which samples are in the training set.
Such priors were also used by~\citet{catoni-07a} in the transductive learning setting.

This is equivalent to the CMI prior~$\auxconddistrocmi$, from the independently constructed CMI framework of~\citet{steinke-20a}.
Specifically, in both cases, the prior is allowed to depend on the training set and a ghost sample, as long as it cannot tell one from the other.
There is a very slight difference in formulation between the two settings: for almost exchangeable priors,~$\supersample'$ is ordered so that the training set comes first, followed by the ghost sample, and the function is required to be invariant to permutations within pairs.
For the CMI framework, the~$n$-separated pairs of~$\supersample$ are instead randomly assigned, which directly eliminates the information of which samples are in the training set.
Hence, the CMI prior is allowed to depend in an arbitrary way on~$\supersample$.
In the end, both formulations lead to equivalent results.
We remark once more, though, that the construction of~\citet{steinke-20a} was formulated independently and with a different motivation.

Many of the results from this chapter were introduced already in the work of~\citet{steinke-20a}, namely \cref{thm:steinke_slow,thm:steinke_fast,thm:steinke_fast_interpolating}; the extension to unbounded losses in~\eqref{eq:unbounded-cmi-steinke}; and the concept of~e-CMI from \cref{sec:ecmi}.
Bounds in terms of a generic convex function, and the specific bound from \cref{thm:cmi-binary-kl}, can be found in~\citep{hellstrom-22a}.
The use of disintegration and the random-subset technique for the CMI framework, as in \cref{thm:haghifam_cmi}, was introduced by~\citet{haghifam-20a}, later extended in the form of \cref{thm:icimi} independently by~\citet{rodriguezgalvez-20a} and~\citet{zhou-21a}.

The tail bounds in \cref{thm:cmi-pac-bayes,thm:cmi-single-draw} can be found in~\citet{hellstrom-20b}.
As mentioned, the PAC-Bayesian bound in \cref{thm:grunwald-cmi-bound} is a heavily simplified version of the result of~\citet[Thm.~1]{grunwald-21a}.
This result, which gives fast-rate CMI-flavored PAC-Bayesian bounds under the Bernstein condition, is significant in how is leads to non-trivial fast-rate bounds in terms of the VC dimension, which was previously shown to be impossible for standard PAC-Bayesian bounds~\citep{livni-17a}.
We will discuss this further in \cref{sec:info-comp-vc}.
The single-draw bounds in \cref{thm:holder_three_variables} and \cref{cor:singledraw_esposito_cond_alphaMI} can be found in~\citet{hellstrom-20b}, and are extensions of bounds from~\citet{esposito-21a} to the CMI setting.

As pointed out, the concept of e-CMI was introduced by~\citet{steinke-20a}.
This was studied further by~\citet{harutyunyan-21a,haghifam-22a,hellstrom-22a,rammal-22a,wang-23a}, who extended the original bounds in various ways.
The form given in \cref{thm:slow_ecmi} is due to~\citet{wang-23a}, as is the result in \cref{thm:interp-edmi}.
The extension to leave-one-out CMI, discussed in \cref{sec:loo-cmi}, was introduced essentially simultaneously by~\citet{haghifam-22a} and~\citet{rammal-22a}.
The result in \cref{thm:loocmi} is due to~\citet{rammal-22a}, while \cref{thm:interp-looemi} is from~\citet{haghifam-22a}.
\newtext{The information-theoretic approach to generalization was combined with techniques from algorithmic stability by \citet{wang-23c}, leading to improved bounds for certain stochastic convex optimization problems.}
Recently,~\citet{sachs-23a} derived bounds in terms of an algorithm-dependent Rademacher complexity, which is conceptually similar to the CMI framework.
\newtext{Finally, \citet{sefidgaran-23a} used related ideas, combined with the information bottleneck and the minimum description length principle, to obtain generalization bounds for representation learning.}

\part{Applications and Additional Topics}

\chapter{The Information Complexity of Learning Algorithms}\label{chap:info-complexity}

As argued in \cref{sec:why-information-theoretic}, one of the benefits of the information-theoretic approach to analyzing generalization is that the resulting bounds depend on both the learning algorithm and the data distribution.
This is in contrast to the uniform convergence-flavored bounds of \cref{sec:uniform-convergence}, \ie, bounds that hold uniformly over all data distributions, or even uniformly over all hypotheses.
Still, this is not very useful if we cannot compute or bound the information measures that appear in the information-theoretic generalization bounds.

In this chapter, we study these information measures for specific learning algorithms.
We begin by looking at the Gibbs posterior, which naturally emerges as the minimizer of some PAC-Bayesian bounds, and whose generalization error can be exactly characterized via a symmetrized relative entropy.
After this, we discuss the Gaussian location model, wherein the learner aims to estimate the mean of a Gaussian distribution.
This simple setting allows us to exactly evaluate the training and population losses, as well as several information measures, and thus allows us to compare various bounds for a concrete setting.
Next, we consider the VC dimension, which plays a fundamental role in uniform convergence-flavored generalization bounds, as well as bounds for compression schemes.
It can be shown that, in many cases, such uniform convergence-flavored bounds can (essentially) be recovered from the information-theoretic bounds from the previous chapters.
We refer to this property as the \emph{expressiveness} of the bounds---\ie, the extent to which the information-theoretic bounds are able to express results from alternative frameworks.
Finally, we discuss connections to algorithmic stability and privacy measures.
We postpone applications to neural networks and gradient-based algorithms, such as stochastic gradient descent and stochastic gradient Langevin dynamics, to \cref{chap:iterative-methods}.

\section{The Gibbs Posterior}\label{sec:gibbs}

Given a generalization bound, it is tempting to design a learning algorithm so as to minimize it.
So far, when presenting information-theoretic bounds, we have considered a specific learning algorithm, characterized in terms of a posterior~$\conddistro$.
Given this posterior, we mostly focused on the prior given by the marginal distribution~$P_W$, as this typically minimizes the bounds in expectation.
However, a slightly different approach is possible, as we exemplified when discussing PAC-Bayesian bounds in \cref{sec:pac-bayesian-bounds}.
There, we discussed bounds that hold for any prior and posterior.
Crucially, the bounds based on the \donskervtext\ in \cref{thm:donskervaradhan} actually hold simultaneously for \emph{all} posteriors.
This is because of the supremum over~$P$ in~\eqref{eq:thm-donskervaradhan}.
This implies that for a fixed prior, we can choose the posterior that minimizes the bound.

Of particular relevance is the Gibbs posterior.
Given a prior~$Q_W$, a training loss~$\trainloss$, a parameter~$\lambda$ referred to as the inverse temperature, the Gibbs posterior for any measurable set~$\setE\subseteq\mathcal W$ is given by
\begin{equation}
\gibbs(\setE \vert \traindata) =\frac {\int_{\setE} \exp(-\lambda \trainlossw) \dv Q_W(w)}{ \int_{\mathcal W}  \exp(-\lambda L_{\traindata}(w)) \dv Q_W(w)}.
\end{equation}
The normalization constant in the denominator, referred to as the \textit{partition function}, is a random variable that depends on $\traindata$. This terminology comes from statistical physics, where the Gibbs posterior also appears under the name of Boltzmann distribution. For later use, it will be convenient to define the \textit{log-partition function}
\begin{equation}
\Psi_\lambda(\traindata) = \log \int_{\mathcal W}  \exp(-\lambda L_{\traindata}(w)) \dv Q_W(w).
\end{equation}
The relevance of the Gibbs posterior is that it is the minimizer of many PAC-Bayesian bounds. Specifically, we have the following result, which is a simple consequence of the \donskervtext\ applied conditionally on $\traindata$.

\begin{lem}\label{lm:Gibbs} Let the prior $Q_W$ be given. Then, for any $\conddistro$,
	\begin{equation}
		\Ex{\conddistro}{\trainloss} + \frac{\relent{\conddistro}{Q_W}}{\lambda} \ge -\frac{1}{\lambda}\Psi_\lambda(\traindata),
	\end{equation}
	and equality is achieved uniquely by the Gibbs posterior $\gibbs$.
\end{lem}

The inverse temperature parameter~$\lambda$ controls the trade-off between the influence of the prior and the influence of the data, and the relative entropy~$\relent{\conddistro}{Q_W}$ acts as a regularizer.
On the one hand, when~$\lambda\rightarrow \infty$, we completely ignore this regularizer and perform unfettered empirical risk minimization.
On the other hand, if~$\lambda\rightarrow 0$, the optimal posterior equals the prior, and we pay no mind to the collected data.
In PAC-Bayesian bounds such as~\eqref{eq:cor_mcallester_fast}, the inverse temperature is typically chosen to be proportional to~$n$.
This leads to a very sensible trade-off: when the amount of data is small, we are not easily convinced to stray far from the prior.
However, when the amount of data grows large, we are inclined to place more importance on it, without relying much on the prior.

\cref{lm:Gibbs} can be used to obtain bounds on the average generalization error of the Gibbs posterior. To that end, we start with a simple observation based on \cref{cor:xu_raginsky} and the identity
\begin{align}
	\inf_{\lambda > 0} \left(a \lambda + \frac{b}{\lambda}\right) = 2\sqrt{ab}.
\end{align}
Suppose that $\ell(w,Z)$ is $\sigma$-subgaussian for all $w \in {\cal W}$. Then, for any $\conddistro$ and any $\lambda > 0$,
\begin{equation}\label{eq:pre_gibbs}
	\Ex{}{\poploss} \le \Ex{}{\trainloss} + \frac{I(W; \traindata)}{\lambda} + \frac{\lambda \sigma^2}{2n}.
\end{equation}
It is tempting to use this inequality to construct a learning algorithm with small expected population loss as follows: fix the inverse temperature $\lambda > 0$ and then choose $\conddistro$ to minimize the right-hand side of \eqref{eq:pre_gibbs}.
However, the mutual information $I(W; \traindata)$ depends on both $\conddistro$ and on the marginal distribution $P_Z$, while the learning algorithm has to be designed without knowledge of $P_Z$.
This can be solved by relaxing the bound using the so-called \textit{golden formula} for the mutual information: for any $Q_W \ll P_W$, we have~\citep[Eq.~(8.7)]{csiszar-11a}
\begin{equation}\label{eq:golden-formula}
	I(W; \traindata) = D(\conddistro \| Q_W | P_{\traindata}) - D(P_W \| Q_W).
\end{equation}
Using this, along with the fact that the relative entropy is nonnegative, we can weaken \eqref{eq:pre_gibbs} to
\begin{align}
		\!\!\!\Ex{\jointdistro}{\poploss} &\le \Ex{\jointdistro}{\trainloss} + \frac{D(\conddistro \| Q_W | P_{\traindata})}{\lambda} + \frac{\lambda \sigma^2}{2n} \\
		&=\Ex{P_{\traindata}}{\Ex{\conddistro}{\trainloss} + \frac{D(\conddistro \| Q_W)}{\lambda}} + \frac{\lambda \sigma^2}{2n}.\!
\end{align}
Thus, applying \cref{lm:Gibbs} conditionally on $\traindata$, we arrive at the following.
\begin{thm} Assume $\ell(w,Z)$ is $\sigma$-subgaussian under $P_Z$ for all $w \in {\cal W}$.
Then, the expected population loss of the Gibbs posterior $\gibbs$ at inverse temperature $\lambda$ satisfies
	\begin{align}
		\Ex{}{\poploss} \le - \frac{1}{\lambda} \Ex{}{\Psi_\lambda(\traindata)} + \frac{\lambda \sigma^2}{2n}.
	\end{align}
\end{thm}
\noindent Bounds of this sort are common in the PAC-Bayes literature \citep{mcallester-98a,mcallester-99a,zhang-06a,catoni-07a}.
To instantiate them in a given setting, we need lower bounds on the log-partition function $\Psi_\lambda(\traindata)$, which are typically derived on a case-by-case basis.
As an example, we give the following result, due to~\citet{pensia-18a}.
\begin{thm}\label{thm:gibbs-bound-max}
Assume the following:
	\begin{enumerate}
		\item The hypothesis space ${\cal W}$ is the $d$-dimensional Euclidean space $\reals^d$.
		\item The loss function $\ell(w,z)$ is differentiable in $w$, and its gradient $\nabla \ell(w,z)$ with respect to $w$ is Lipschitz-continuous uniformly in $z$, that is, there exists a constant $M > 0$, such that for all~$w,w' \in {\cal W}$
		\begin{align}
			\sup_{z \in {\cal Z}} \| \nabla \ell(w,z) - \nabla \ell(w', z) \| \le M \| w - w' \|
		\end{align}
		where $\| \cdot \|$ denotes the Euclidean $(\ell^2)$ norm on $\reals^d$.
		\item For every realization of $\traindata$, all global minimizers of the training loss $\trainloss$ lie in the ball of radius $R$ centered at $0$.
		\item The loss $\ell(w,Z)$ is $\sigma$-subgaussian under $P_Z$ for all $w \in {\cal W}$.
	\end{enumerate}
Let $\gibbs$ be the Gibbs posterior with inverse temperature $\lambda > 0$ associated to the Gaussian prior $Q_W = {\cal N}(0, \rho^2 I_d)$. Then
\begin{align}
	&\Ex{}{\poploss} - \min_{w \in {\cal W}} L_{P_Z}(w) \nonumber\\
	& \le \frac{M\pi \rho^2 d}{\lambda} + \frac{1}{2\lambda \rho^2} \left( R + \sqrt{\frac{2\pi\rho^2d}{\lambda}}\right)^2 + \frac{d}{2\lambda} \log \frac{\lambda}{d} - \frac{1}{\lambda}\log V_d + \frac{\lambda \sigma^2}{2n}, \label{eq:Gibbs_excess_risk}
\end{align}
where $V_d$ is the volume of the unit ball in $(\reals^d, \|\cdot\|)$.
\end{thm}
\begin{proof} Fix $\traindata$ and let $w^*_{\traindata}$ be any global minimizer of $\trainloss$, where $\| w^*_{\traindata} \| \le R$ by hypothesis. Since the gradient $w \mapsto \nabla \ell(w,\traindata)$ is $M$-Lipschitz and $\nabla L_{\traindata}(w^*_{\traindata}) = 0$, we have
	\begin{align}
		L_{\traindata}(w) - L_{\traindata}(w^*_{\traindata}) \le \frac{M}{2}\| w - w^*_{\traindata} \|^2.
	\end{align}
	Therefore,
	\begin{align}
		\Psi_\lambda(\traindata) &= -\lambda L_{\traindata}(w^*_{\traindata}) + \log \Ex{Q_W}{\exp\left(-\lambda\left(\trainloss - L_{\traindata}(w^*_{\traindata})\right)\right)} \\
		&\ge -\lambda L_{\traindata}(w^*_{\traindata}) + \log\Ex{Q_W}{\exp\left(-\frac{\lambda M}{2}\| W - w^*_{\traindata}\|^2\right)},
	\end{align}
	so, in order to lower-bound the log-partition function $\Psi_\lambda(\traindata)$, we need to lower-bound the Gaussian integral
	\begin{align}
		G = \frac{1}{(2\pi \rho^2)^{d/2}} \int_{\reals^d} e^{-\frac{1}{2\rho^2}\|w\|^2}e^{-\frac{\lambda M}{2} \| w - w^*_{\traindata} \|^2} \dv w.
	\end{align}
	Let ${\cal B}$ be the $\ell^2$ ball of radius $\varepsilon > 0$ (to be tuned later) centered at $w^*_{\traindata}$ with volume~$\mathsf{Vol}_d({\cal B})$. Then
	\begin{align*}
		G &\ge  \frac{1}{(2\pi\rho^2)^{d/2}}e^{-\frac{\lambda M \varepsilon^2}{2}} \cdot  \int_{\cal B} e^{-\frac{1}{2\rho^2}\|w\|^2}\dv w \\
		&\ge \frac{1}{(2\pi\rho^2)^{d/2}} e^{-\frac{\lambda M \varepsilon^2}{2}} \cdot  e^{-\frac{1}{2\rho^2}(\|w^*_{\traindata}\|+\varepsilon)^2} \mathsf{Vol}_d({\cal B}) \\
		&= \frac{1}{(2\pi\rho^2)^{d/2}} e^{-\frac{\lambda M \varepsilon^2}{2}} \cdot  e^{-\frac{1}{2\rho^2}(\|w^*_{\traindata}\|+\varepsilon)^2} \varepsilon^d V_d \\
		&\ge \left(\frac{\varepsilon^2}{2\pi\rho^2}\right)^{d/2} \exp\left(-\frac{\lambda M \varepsilon^2}{2}-\frac{1}{2\rho^2}(R+\varepsilon)^2\right) V_d.
	\end{align*}
	For all~$\varepsilon>0$, this leads to the estimate
	\begin{align}
	&	-\frac{1}{\lambda} \Ex{}{\Psi_\lambda(\traindata)}  \le \Ex{}{\min_{w \in {\cal W}}\trainloss}  \\
	& \qquad  + \frac{M\varepsilon^2}{2} + \frac{1}{2\lambda\rho^2} (R+\varepsilon)^2 + \frac{d}{2\lambda}\log \left(\frac{2\pi\rho^2}{\varepsilon^2}\right) - \frac{1}{\lambda}\log V_d,.
	\end{align}
Choosing $\varepsilon = \frac{2\pi \rho^2 d}{\lambda}$ and using that
\begin{align}
	\Ex{}{\min_{w \in {\cal W}}\trainloss} = \Ex{}{L_{\traindata}(w^*_{\traindata})} \le \min_{w \in {\cal W}} L_{P_Z}(w),
\end{align}
we get \eqref{eq:Gibbs_excess_risk}.
\end{proof}

Recently,~\citet{aminian-21b} provided an exact information-theoretic characterization of the average generalization error of the Gibbs posterior.
Let~$\gibbsmarginal=\Ex{\traindistro}{\gibbs}$ denote the marginal distribution on~$W$ induced by the Gibbs posterior.
Then, for the Gibbs posterior, we let the symmetrized KL information between~$W$ and~$\traindata$ be given by
\begin{equation}\label{eq:iskl-def-gibbs}
\Iskl{W}{\traindata} = \relent{\traindistro \gibbs}{\traindistro \gibbsmarginal} + \relent{\traindistro \gibbsmarginal} {\traindistro \gibbs}.
\end{equation}
This symmetrized relative entropy, where we sum two relative entropies with their arguments swapped, is sometimes referred to as Jeffreys' divergence.
Notice that the term~$\relent{\traindistro \gibbs}{\traindistro \gibbsmarginal}$ is the mutual information~$I(W;\traindata)$ while the term~$\relent{\traindistro \gibbsmarginal} {\traindistro \gibbs}$ is sometimes referred to as the lautum information~\citep{palomar-08a}.\footnote{This provides a strong incitement to refer to~$\Iskl{\cdot}{\cdot}$ as the mutualautum information, but we digress.}
With this, \citet{aminian-21b} derived the following exact characterization of the average generalization error of the Gibbs posterior.
\begin{thm}\label{thm:gibbs-aminian}
Given an inverse temperature~$\lambda$ and a prior distribution~$Q_W$, the average generalization error of the Gibbs posterior is given by
\begin{equation}
\Ex{\traindistro \gibbs}{\poploss-\trainloss} = \frac{\Iskl{W}{\traindata}}{\lambda}.
\end{equation}
\end{thm}
\begin{proof}
Note that~$\Ex{\traindistro\gibbs}{\log\gibbsmarginal}=\Ex{\traindistro \gibbsmarginal}{\log \gibbsmarginal}$. 
Hence, using \eqref{eq:iskl-def-gibbs}, we can write
\begin{align}
\Iskl{W}{\traindata} &= \Ex{\traindistro \gibbs}{\log  \frac{\gibbs}{\gibbsmarginal}} + \Ex{\traindistro \gibbsmarginal}{\log \frac{\gibbsmarginal}{\gibbs}} \\
&= \Ex{\traindistro \gibbs}{\log \gibbs} - \Ex{\traindistro \gibbsmarginal}{\log \gibbs}.
\end{align}
From the definition of the Gibbs posterior, we see that
\begin{equation}
\log \gibbs(W\vert\traindata) = \log Q_W(W) - \Psi_\lambda(\traindata) - \lambda\trainloss.
\end{equation}
Since the marginal distributions of~$W$ and~$\traindata$ are the same under~$\traindistro \gibbs$ and~$\traindistro \gibbsmarginal$ we have
\begin{equation}
\Ex{\traindistro \gibbs}{\log Q_W(W) \!-\! \Psi_\lambda(\traindata)} \!=\! \Ex{\traindistro \gibbsmarginal}{\log Q_W(W) \!-\! \Psi_\lambda(\traindata)}.
\end{equation}
From this, it follows that
\begin{align}
&\Ex{\traindistro \gibbs}{\log \gibbs} - \Ex{\traindistro \gibbsmarginal}{\log \gibbs} \nonumber \\
=&\Ex{\traindistro \gibbs}{-\lambda\trainloss} - \Ex{\traindistro \gibbsmarginal}{-\lambda\trainloss}\\
=& \lambda\Ex{\traindistro \gibbs}{\poploss-\trainloss}.
\end{align}
From this, the result follows.
\end{proof}

In order to interpret this result, we need to discuss the extreme cases.
First, if~$\lambda\rightarrow \infty$, it may seem as if the generalization error vanishes.
This is the case if~$\Iskl{W}{\traindata}$ remains finite when we perform exact empirical risk minimization.
For this to occur, we need not only that~$\gibbs\ll\gibbsmarginal$, but also that~$\gibbsmarginal\ll\gibbs$.
Since the Gibbs posterior with infinite temperature is supported only on empirical risk minimizers, the second criterion can only be fulfilled if the prior is also supported only on empirical risk minimizers.
For any non-trivial case, we expect the prior to assign some probability mass to non-minimizers as well, meaning that~$\Iskl{W}{\traindata}$ would diverge as~$\lambda\rightarrow \infty$.
In a similar vein, when~$\lambda\rightarrow 0$, the posterior does not change relative to the prior, so~$\Iskl{W}{\traindata}\rightarrow 0$ as well.

While the Gibbs posterior has many attractive properties theoretically, it is not always straightforward to implement in practice.
This is discussed further by, for instance,~\citet{alquier-16a,perlaza-23a}.

\section{The Gaussian Location Model}

We now turn to a simple learning problem in which many of the quantities in the generalization bounds that we discussed can be evaluated explicitly, allowing us to perform a direct comparison between different bounds for a concrete setting.
Specifically, we assume that the data distribution~$\datadistro=\normal(\mu, \sigma^2)$ is a Gaussian distribution with mean~$\mu$ and variance~$\sigma^2$, and the training set~$\traindata=(Z_1,\dots,Z_n)\in\reals^n$ consists of~$n$ independent samples from~$\datadistro$.
Based on this, the goal is to learn the mean of the Gaussian distribution.
Thus, the hypothesis space consists of the real numbers~$\mathcal W=\reals$.
A natural choice for the loss function, which we will consider throughout, is the squared loss~$\ell(w,z)=(w-z)^2$.
We will focus on the empirical risk minimizer obtained by taking the sample average,~$W=\frac1n\sum_{i=1}^n Z_i$.

For this setting, the average generalization error can in fact be computed explicitly as~\citep{bu-20a}
\begin{align}
\avggengap &\!=\! \Ex{\jointdistro}{ \Ex{Z'\distas\datadistro}{(Z'\!-\!W)^2}  \! - \! \frac{1}{n}\sum_{i=1}^n (Z_i\!-\!W)^2 }\\
&= \frac{2\sigma^2}{n}.
\end{align}
We thus have a known baseline with which to compare the generalization bounds that we derived in \cref{chap:average,chap:cmi}, and for this setting, many of them can be computed exactly.
It should be noted here that if a bound gives a loose characterization of the generalization error for this specific problem, this is not an indictment of the bound as a whole.
Since all of the bounds that we will discuss have been derived for a very general class of learning problems and learning algorithms, it is not unexpected that they will be loose for many specific problems and algorithms.
Nevertheless, due to its analytical tractability, this setting serves as an instructive case study.
Also note that, as mentioned in \cref{sec:gibbs}, the average generalization error of the Gibbs posterior is exactly characterized by the symmetrized KL information.
By evaluating this information-theoretic quantity, one can show that the Gibbs posterior also has a generalization error of order~$\sigma^2/n$.
For more details, see the work of~\citet{aminian-22a}.

First, we note that the mutual information~$I(W;\traindata)$ gives a vacuous bound on the generalization gap.
Indeed, since the training data and hypothesis are continuous and we use a deterministic learning algorithm, the mutual information is infinite.
However, as noted by~\citet{bu-20a}, this can be rectified by using the individual-sample technique: since the hypothesis is not a deterministic function of any single sample, the individual-sample mutual information is finite.
Indeed, it can be computed in closed form as~\citep{bu-20a}
\begin{align}
I(W;Z_i) = \frac{1}{2}\log\frac{n}{n-1} .
\end{align}
Inserting this into the generalization bound in \cref{cor:bu_veeravalli}, we find that
\begin{align}
\avggengap &\leq \frac1n\sum_{i=1}^n \sqrt{2\sigma^2 I(W;Z_i)} \\
&= \sigma\sqrt{\log\lefto(\frac{n}{n-1} \right) } \\
&\leq \sigma\sqrt{\frac{1}{n-1} }.
\end{align}
Thus, this gives a bound of order~$1/\sqrt{n}$, which is quadratically worse than the true generalization gap.

Next, let us consider the CMI framework.
To do this, one needs to go beyond the assumption of a bounded loss that was considered throughout most of \cref{chap:cmi}.
As indicated in~\eqref{eq:unbounded-cmi-steinke}, the main results extend to certain unbounded losses.
This includes the squared loss under a Lipschitz condition, provided that the fourth moment of the data is finite~\citep[Sec.~5.4]{steinke-20a}.
This is satisfied for the Gaussian location problem---see the work of~\citet{zhou-21a} for details.
While the CMI yields a finite result, unlike the mutual information, it is significantly looser than the individual-sample mutual information bound.
Indeed, we have~\citep{zhou-21a}
\begin{equation}
\cmi =  \frac{n}{\log_2(e)} .
\end{equation}
The reason for this is that conditioning on the supersample reveals too much information, due to the continuous nature of the output.
In fact, if we consider a n\aive\ individual-sample version of the CMI, where we still condition on the full supersample, that is,~$\cmiiz$, we still get a constant---leading to a generalization bound that does not decay with~$n$.
Motivated by this,~\citet{zhou-21a} argue for the individually conditioned CMI, where the conditioning is also on individual pairs of the supersample---as discussed in \cref{thm:icimi}.
With this, it can be shown that~\citep[Lemma.~4]{zhou-21a}
\begin{equation}
\cmiidis = \frac{(z_i-z_{i+n})^2}{8\sigma^2(n-1)} + o\lefto(\frac1n\right) .
\end{equation}
Inserting this into the corresponding generalization bound of~\citet{zhou-21a}, we again get a bound that decays as~$1/\sqrt n$, but with a slightly improved constant factor.

This raises the question: is it possible to obtain the correct~$1/n$-dependence from information-theoretic generalization bounds?
The answer turns out to be yes.
Through the use of stochastic chaining, as mentioned in \cref{sec:chaining},~\citet[Sec.~4.1]{zhou-22a} obtained a generalization bound of~$\avggengap\leq 13\sigma^2/n$, thus matching the dependence of the true generalization error but with a larger constant.
An alternative approach was taken by~\citet{wu-22b}, who derived a bound that, on its face, is identical to the individual-sample bound of~\citet{bu-20a}, but with a key modification---instead of assuming the loss to be sub-Gaussian, the \emph{excess risk},~$r(w,Z)=\ell(w,Z)-\ell(w^*,Z)$, is assumed to be sub-Gaussian under~$P_Z$ for all~$w\in\mathcal W$, where~$w^*$ is a minimizer of the population loss.
For sufficiently large~$n$, the excess risk of the Gaussian location problem with the sample-averaging algorithm actually turns out to be~$\sqrt{4\sigma^4/ n}$-sub-Gaussian---the sub-Gaussianity parameter decays with~$n$.
Evaluating the generalization bound with this yields an~$O(1/n)$ rate.

However, it is possible to demonstrate that this fast rate is achievable with arguably simpler techniques.
In fact, it turns out that it is possible to derive an information-theoretic generalization bound that is exactly tight for this problem, even up to constants, which was done by~\citet{zhou-23a}.
This is achieved through a variant of the individual-sample approach of~\citet{bu-20a}, with some key modifications: the change of measure is applied to the generalization gap rather than the training loss; disintegration is used; a different prior than the true marginal is used; and the straight-forward sample-averaging algorithm is replaced with a weighted one where Gaussian noise is added (which has the same performance as the sample-averaging algorithm in expectation).
This includes many of the techniques that we covered in \cref{chap:average}, applied in a very careful way.
If we are satisfied with a bound that is optimal only in an asymptotic sense, the alternative prior and weighted sample-averaging are not needed.
The interested reader is referred to the work of~\citet{zhou-23a} for the full details.

\section{The VC Dimension}\label{sec:info-comp-vc}

As discussed in \cref{sec:vc-dimension}, a fundamental quantity that characterizes distribution- and algorithm-independent learnability for binary classification is the VC dimension.
While our original motivation for pursuing information-theoretic generalization bounds was to go beyond this style of uniform convergence analysis, an interesting question is whether or not the information-theoretic approach is still expressive enough to capture complexity measures such as the VC dimension.
More precisely, we seek to answer the following question: consider a hypothesis class~$\mathcal W$ with bounded VC dimension~$\dVC$.
Can we provide a bound on the information measures that appear in our generalization bounds in terms of~$\dVC$, and if so, do the resulting bounds coincide with the best available generalization bounds?

To partially answer this question, we will focus on the case of generalization bounds in expectation and consider binary classification with the~$0-1$ loss.
Throughout, we assume that the instance space~$\dataspace$ factors into a feature space~$\featurespace$ and label space~$\labelspace=\{0,1\}$, and we associate each hypothesis~$w\in\mathcal W$ with a function~$f_w:\featurespace\rightarrow\labelspace$.

\subsection{Mutual Information}\label{sec:vc-dimension-mi}

We begin by considering the mutual information between the training data~$\traindata$ and hypothesis~$W$,~$I(W;\traindata)$, that appears in, \eg, \cref{cor:xu_raginsky}.
As an illustrative example of a class with finite VC dimension, we consider threshold classifiers: that is, the set of classifiers is given by~$\{ f_w(x) = 1\{x\geq w\} \given w\in\reals \}$.
As this hypothesis class can induce arbitrary labels for a set with a single element, but not a set with two elements (as achieving~$f_w(x_1)=1$ and~$f_w(x_2)=0$ for~$x_1<x_2$ is not possible), its VC dimension is one.
Throughout, we shall refer to data distributions for which an element of the hypothesis class achieves zero population loss as \emph{realizable}.

Immediately, we can establish one negative result: the mutual information~$I(W;\traindata)$ can be unbounded, even for very reasonable empirical risk minimizers.
Consider, for instance, the case of threshold classifiers for a realizable distribution.
Let us denote each training sample as~$Z_i=(X_i,Y_i)$, which consists of a real number feature~$X_i$ and a label~$Y_i\in\{0,1\}$.
A reasonable empirical risk minimizer is an algorithm that outputs~$f_{\hat W}$, where~$\hat W = \min \{x: (x,1)\in\traindata\} $, \ie, the smallest feature labelled~$1$.
Due to the realizability assumption, this must achieve zero training loss.
However, since the learning algorithm is a deterministic function of the training set with a continuous output,~$I(W;\traindata)=\infty$.

In order to circumvent this,~\citet{xu-17a} considered the following two-stage algorithm.
First, split the training set into two halves, so that~$\traindata_a=(Z_1,\dots,Z_{n/2})$ and $\traindata_b=(Z_{n/2+1},\dots,Z_{n})$, where we assume~$n$ to be even for simplicity.
In the first stage of the algorithm, one constructs an empirical cover of~$\mathcal W$ on the basis of~$\traindatax_a=(X_1,\dots,X_{n/2})$, \ie, a subset~$\mathcal W_a\subset\mathcal W$ such that~$\abs{ \{ (f_w(X_1),\dots,f_w(X_{n/2})) : w\in\mathcal W_a \} } = \abs{\mathcal W_a }$, meaning that each element of~$\mathcal W_a$ induces a distinct classification, and $\abs{ \{ (f_w(X_1),\dots,f_w(X_{n/2})) : w\in\mathcal W \} }= \abs{\mathcal W_a }$, meaning that each possible classification using~$\mathcal W$ is induced by an element of~$\mathcal W_a$.
In the second stage of the algorithm, one selects an empirical risk minimizer for~$\traindata_b$ from the finite~$\mathcal W_a$.
By applying \cref{cor:xu_raginsky} conditional on~$\traindata_a$, evaluating the training loss with respect to~$\traindata_b$, we thus find that
\begin{align}
\Ex{\jointdistro}{\poploss\!-\!L_{\traindata_b}(W)} &\!=\! \Ex{P_{\traindata_a}}{  \Ex{P_{W\!\traindata_b\vert\traindata_a}}{  \poploss\!-\!L_{\traindata_b}(W) }    } \\
&\leq \sqrt{ \frac{I(W;\traindata_b\vert\traindata_a)}{n} } ,
\end{align}
where we used the fact that the~$0-1$ loss is~$1/2$-sub-Gaussian.
Now, given~$\traindata_a$,~$W$ can only take values in the finite set~$\mathcal W_a$.
Furthermore, the cardinality of~$\mathcal W_a$ can be bounded using the Sauer-Shelah lemma (\cref{lem:sauer-shelah}).
We thus conclude that
\begin{align}
I(W;\traindata_b\vert\traindata_a) \leq H(W\vert \traindata_a) \leq \log (\abs{\mathcal W_a}) \leq \dVC \log\lefto(\frac{en}{2\dVC}\right),
\end{align}
where the first step follows from the non-negativity of entropy, the second step from the fact that entropy is maximized by a uniform distribution, and the final step from the Sauer-Shelah lemma.
Note that, through these arguments, we have obtained an average version of the standard generalization guarantee in terms of the VC dimension from \cref{thm:vc-dim-classical}, up to constants and logarithmic dependencies.
Still, this applies only to a very particular algorithm, and not the standard empirical risk minimizer.
Indeed,~\citet{bassily-18a,nachum-18a} showed that for any empirical risk minimizer over a finite input space, there exists a realizable data distribution for which the mutual information~$I(W;\traindata)$ scales with the cardinality of the input space.
Furthermore,~\citet{livni-17a} demonstrated that for any learning algorithm for threshold classifiers, there exists a realizable distribution for which either the population loss or the mutual information is large (in fact, their result applies more generally to the relative entropy that appears in PAC-Bayesian bounds).
On the positive side,~\citet{nachum-19a} showed that there does exist learning algorithms with bounded mutual information for ‘‘most'' hypotheses in VC classes.

\subsection{Conditional Mutual Information}

We now turn to the CMI framework of \cref{chap:cmi}.
Specifically, we consider the conditional mutual information between the hypothesis~$W$ and the membership vector~$\subsetchoice$ given the supersample~$\supersample$, that is,~$\cmi$.
As discussed in \cref{chap:cmi}, bounds in terms of the CMI are tighter (up to constants) than the ones based on the mutual information~$I(W;\traindata)$.
In contrast to the mutual information, there is a wide class of natural empirical risk minimizers for which the CMI can be shown to be bounded by (approximately) the VC dimension.
In particular, this applies to any algorithm satisfying the following consistency property.
For simplicity, following~\citet{steinke-20a}, we restrict ourselves to deterministic learning algorithms.

\begin{dfn}[Global consistency property]\label{def:global-consistency-prop}
Let~$W(\traindatasmall)$ denote the point mass on which~$\conddistroz$ concentrates when trained on~$\traindatasmall=(\traindataxsmall,\traindataysmall)\in\dataspace^n$.
Let~$\traindatasmall'=(\traindataxsmall',\traindataysmall')\in\mathcal Z^m$ with~$m\geq n$ be constructed so that~$(i)$: for all~$i\in[n]$, there is a~$j\in[m]$ such that~$x_i=x_j'$, and,~$(ii)$: for all~$i\in[m]$,~$f_{W(\traindatasmall)}(x_i) = y_i'$.
Then, the learning algorithm characterized by~$\conddistro$ has the global consistency property if, for any~$\traindatasmall\in \mathcal Z^n$,~$\conddistrozp$ concentrates on~$W(\traindatasmall)$.
\end{dfn}

\noindent This property requires that if a training set~$\traindatasmall$ is re-labelled to obtain~$\traindatasmall'$, which is fully consistent with the output hypothesis~$W(\traindata)$ obtained from training on~$\traindatasmall$ and possibly expanded with more consistent samples, the output hypothesis obtained from training on~$\traindatasmall'$ should still be~$W(\traindata)$.
Clearly, this property is satisfied for many reasonable empirical risk minimizers.

With this, we can show the following.
\begin{thm}\label{thm:cmi-vc-dim}
Consider the~$0-1$ loss and assume that the VC dimension~$\dVC$ of~$\mathcal W$ is finite.
Assume that the learning algorithm satisfies the global consistency property.
Then, if~$n > \dVC$,
\begin{equation}
\cmi\leq  \dVC \log \lefto(  \frac{2en}{\dVC} \right).
\end{equation}
\end{thm}
\begin{proof}
Let~$\supersamplesmall_*=\argmax_{\supersamplesmall} I(W; \subsetchoice\vert \supersample=\supersamplesmall)$.
Also, let~$\hat {\mathcal W}\subseteq \mathcal W$ denote the set of possible output hypotheses obtainable by varying~$\subsetchoice$ given the fixed supersample~$\supersamplesmall_*=(\supersamplesmallx_*,\supersamplesmally_*)$.
Then, we have
\begin{align}
\cmi \leq I(W; \subsetchoice\vert \supersample=\supersamplesmall_*) \leq \log\abs{\hat {\mathcal W}}.
\end{align}
Now, by the global consistency property, the output hypothesis~$w(\supersamplesmall_*(\subsetchoicesmall))$ obtained by running the learning algorithm on the training set~$\supersamplesmall_*(\subsetchoicesmall)$ can also be obtained by running the learning algorithm on the training set~${\supersamplesmall_*}'=({\supersamplesmallx_*}',{\supersamplesmally_*}')$, which is constructed so that~$\supersamplesmallx_*=\supersamplesmallx_*'$ and, for all~$i\in[2n]$,~$f_{w(\supersamplesmall_*(\subsetchoicesmall))}((\tilde{x}_*)_{i})=({\tilde{y}_*}')_{i}$.
In words: the output hypothesis~$w(\supersamplesmall_*(\subsetchoicesmall))$ from the training set~$\supersamplesmall_*(\subsetchoicesmall)$ can be obtained by running the learning algorithm on~${\supersamplesmall_*}'$, which only contains samples that are consistent with~$w(\supersamplesmall_*(\subsetchoicesmall))$.
Hence, the number of distinct possible output hypotheses~$\abs{\hat {\mathcal W}}$ is upper-bounded by the number of possible labellings of~$\supersamplesmallx_*$ using hypotheses from~$\mathcal W$.
This, in turn, can be bounded using the Sauer-Shelah lemma (\cref{lem:sauer-shelah}).
Specifically,
\begin{equation}
\cmi \leq \log\abs{\hat {\mathcal W}} \leq \dVC \log \lefto(\frac{2en}{\dVC} \right).
\end{equation}
\end{proof}
To complete this argument, it remains to show that there exist deterministic empirical risk minimizers with the global consistency property.
Since the argument is quite technical, we will not reproduce it here.
The proof can be found in~\citet[Lemma~4.15]{steinke-20a}.

Note that this result does not imply that \emph{every} empirical risk minimizer over a hypothesis class with finite VC dimension has bounded CMI.
For this, we need to consider further processed versions of the CMI.

\subsection{Evaluated and Functional CMI}

We now turn to the evaluated and functional versions of the CMI, or e-CMI and $f$-CMI for short.
Specifically, recall that the $f$-CMI is given by the mutual information between the predictions~$\mathbf F$ (for the supersample~$\supersample$ induced by the hypothesis~$W$) and the membership vector~$\subsetchoice$ given~$\supersample$, that is,~$\fcmi$.
The e-CMI is obtained by replacing the predictions with the losses~$\supersampleloss$ that they induce, that is,~$\ecmi$.
For binary classification with the~$0-1$ loss, there is a bijection between~$\mathbf F$ and~$\supersampleloss$ given~$\supersample$: the loss of a prediction is~$0$ if and only if it matches the corresponding label, otherwise the loss is~$1$.
Thus, for this particular case,~$\ecmi=\fcmi$, although the latter more generally only gives an upper bound.
We will thus consider only the~$f$-CMI.
In contrast to the CMI, it is possible bound the $f$-CMI for \emph{every} learning algorithm over a hypothesis class with finite VC dimension.
We establish this result in the following theorem.

\begin{thm}\label{thm:fcmi-vc-dim}
Consider the~$0-1$ loss and assume that the VC dimension~$\dVC$ of~$\mathcal W$ is finite.
Then, if~$n > \dVC$,
\begin{equation}
\fcmi \leq \dVC \log \lefto(\frac{2en}{\dVC} \right).
\end{equation}
\end{thm}
\begin{proof}
Let~$\supersamplesmall_*=\argmax_{\supersamplesmall} I(F; \subsetchoice\vert \supersample=\supersamplesmall)$.
Also, let~$\hat{\mathcal{F}} \subseteq\mathcal Y^{2\times n}$ denote the set of possible predictions obtainable by varying~$\subsetchoice$ given the fixed supersample~$\supersamplesmall_*=(\supersamplesmallx_*,\supersamplesmally_*)$.
Then, we have
\begin{align}
\fcmi \leq I(F; \subsetchoice\vert \supersample=\supersamplesmall_*) \leq \log\abs{\hat{\mathcal{F}}}.
\end{align}
The number of distinct possible output predictions~$\hat{\mathcal{F}}$ is upper-bounded by the number of possible labellings of~$\supersamplesmallx_*$ using hypotheses from~$\mathcal W$.
This can be bounded using the Sauer-Shelah lemma (\cref{lem:sauer-shelah}), from which the final result follows.
\end{proof}

Again, we emphasize that this result holds for \emph{every} learning algorithm, even beyond empirical risk minimizers.
Furthermore, by using the $f$-CMI, the proof of this result just involves an application of the Sauer-Shelah lemma.
In a sense, this provides an information-theoretic re-interpretation of this classic uniform convergence argument (discussed in \cref{sec:vc-dimension}).
Specifically, when the hypothesis class has low complexity as measured by the VC dimension, any learning algorithm for the hypothesis class has low information complexity, as measured by the $f$-CMI.

While this demonstrates that one can obtain bounds for the f-CMI of any learning algorithm, this does not generally lead to optimal generalization bounds, as they are off by a~$\log$-factor~\citep[Thm.~4.4]{haghifam-21a}.

\subsection{Leave-One-Out CMI}\label{sec:info-comp-loo-cmi}

We conclude the discussion of the VC dimension by describing a bound for learning of VC classes over realizable distributions obtained through the leave-one-out evaluated CMI (loo-e-CMI), due to~\citet{haghifam-22a}.
Since the proof of this result is somewhat more involved, we will not give it in full detail, but instead just sketch the arguments.

For the purposes of this discussion, we consider the leave-one-out CMI setting introduced in \cref{sec:loo-cmi} with the~$0-1$ loss, and assume the data distribution to be realizable.
First, we connect the binary loss loo-e-CMI of interpolating learning algorithms and the leave-one-out-error, defined as
\begin{equation}
\Rloo =  \Ex{P_U}{  \Ex{P_{\dot \supersampleloss\vert U\!\supersampleloo}}{ \dot\Lambda_U  } } .
\end{equation}
In words, given a supersample~$\supersampleloo$, $\Rloo$ is the test loss when leaving out the~$U$th sample, averaged over~$U$ and the randomness of the learning algorithm.
Notice that~$\Rloo\in[0,1]$.
It can be shown that the loo-e-CMI~$\looecmi$ can be bounded by~$H_b(\Rloo)+\Rloo\log(n+1)$, where~$H_b(\Rloo)$ denotes the binary entropy (\ie, the entropy of a Bernoulli random variable with parameter~$\Rloo$)~\citep[Thm.~3.1]{haghifam-22a}.

Next, we briefly describe the one-inclusion graph algorithm introduced by~\citet{haussler-88a}.
Given~$\supersampleloo=(\supersampleloox,\supersamplelooy)\in\dataspace^{n+1}$, let~$\mathcal V$ denote the set of possible labellings of~$\supersampleloox=(\dot X_1,\dots,\dot X_{n+1})$ with hypotheses from~$\mathcal W$.
We refer to elements of $\mathcal V$ as adjacent if they differ in only one element.
We define a probability assignment~$P:\mathcal V\times \mathcal V\rightarrow [0,1]$ so that~$P(\mathbf v,\mathbf w)=0$ if~$\mathbf v,\mathbf w\in \mathcal V$ are not adjacent, and~$P(\mathbf v,\mathbf w)+P(\mathbf w,\mathbf v)=1$ if they are, where~$P$ is chosen solely on the basis of~$\supersampleloox$.
Recall that~$\traindatacmiloo$ denotes the training set, formed by removing the~$U$th entry of~$\supersampleloo$, while~$\testdatacmiloo$ is a test sample.
Due to the realizability assumption, either one or two elements of~$\mathcal V$ are consistent with~$\supersamplelooargb{\bar u}$ for~$u\in[n]$.
The one-inclusion graph algorithm, given the training set~$\supersamplelooargb{\bar u}$, predicts the label of~$\dot y_u$ as follows: if only one element~$\mathbf v\in \mathcal V$ is consistent with~$\supersamplelooargb{\bar u}$, it predicts~$v_u$.
If two elements~$\mathbf v,\mathbf w\in \mathcal V$ are consistent with~$\supersamplelooargb{\bar u}$, it predicts~$v_u$ with probability~$P(\mathbf v,\mathbf w)$ and~$w_u$ otherwise.
Let~$\mathbf v^*$ denote the vector of correct labels for~$\supersampleloox$.
When using~$\supersamplelooargb{\bar u}$ as training set, the probability of incurring an error on~$\supersamplelooargb{u}$ is given by~$P(\mathbf v',\mathbf v^*)$ for~$\mathbf v'$ such that~$v'_u\neq v^*_u$ but all other entries of~$\mathbf v'$ and~$\mathbf v^*$ are equal, provided that such a~$\mathbf v'$ exists in~$\mathcal V$.
Otherwise, it is zero.
Therefore, the leave-one-out error is given by
\begin{equation}
\Rloo = \sum_{\mathbf v'\in \mathcal V} \frac{P(\mathbf v',\mathbf v^*)}{n+1}.
\end{equation}
\citet[Lemma~5.2]{haussler-88a} established that there exists a probability assignment such that~$\sum_{\mathbf v'\in \mathcal V} P(\mathbf v',\mathbf w)\leq \dVC$ uniformly for~$\mathbf w\in \mathcal V$.
By combining this with the bound on~$\looecmi$ in terms of~$\Rloo$ provided in the first step, a bound for learning realizable VC classes can be established.

\newtext{Notably, in the works of \citet{haghifam-21a,haghifam-22a}, the CMI of a learning algorithm is demonstrated to provide a \emph{universal} characterization of realizable generalization in a certain sense: specifically, for every interpolating learning algorithm and data distribution, the population loss vanishes as $n$ goes to infinity \emph{if and only if} the CMI of the learning algorithm grows sub-linearly in $n$.
For the loo-e-CMI, an even stronger characterization can be established, in the sense that the loo-e-CMI also captures the decay rate when the population loss decays polynomially or converges to a positive value.
For more details, the reader is referred to \citet{haghifam-21a,haghifam-22a}.}

\section{Compression Schemes}\label{sec:info-comp-compression-schemes}

We now consider a class of learning algorithms known as \emph{compression schemes}~\citep{littlestone-86a}.
A compression scheme of size~$k$ consists of two components: a sequence of maps~$\kappa:\mathcal Z^n\rightarrow \mathcal Z^k$ for~$n\geq k$, which given an input vector~$\traindata$ of size~$n$ outputs a vector~$\kappa(\traindata)$ consisting of~$k$ elements of~$\traindata$; and a map~$\rho:\mathcal Z^k\rightarrow \mathcal W$ that selects a hypothesis based on this compressed training set.
By composing these maps, we obtain a learning algorithm for training sets of size~$n\geq k$.

As an example, consider threshold classifiers, as introduced in \cref{sec:vc-dimension-mi}, and a learning algorithm that simply sets the threshold~$W$ to be the smallest training feature with the label~$1$, \ie,~$W=\min \{x: (x,1)\in\traindata\} $ (and~$W=\infty$ if there is no sample with the label~$1$).
Clearly, this can be written as the composition of a map~$\kappa$ that outputs~$\kappa(\traindata)=(x_{i^*}, y_{i^*})$, where $i^*=\argmin_i \{x_i: (x_i,1)\in\traindata\} $, and a map
\begin{equation}
\rho(x,y)=
\begin{cases}
x \text{ if } y= 1\\
\infty \text{ otherwise}.
\end{cases}
\end{equation}
Therefore, it is a compression scheme of size~$1$.

The mutual information~$I(W;\traindata)$ of such algorithms will generally be unbounded, since we are dealing with deterministic algorithms with continuous inputs and outputs.
However, for the CMI, the following can be established, as per~\citet[Thm~4.2]{steinke-20a}.
\begin{thm}\label{thm:cmi-comp-scheme}
Assume that~$\conddistro$ is a compression scheme of size~$k$.
Then, we have~$\cmi\leq k\log(2n)$.
\end{thm}
\begin{proof}
Since~$W$ is a function of~$\kappa( \traindatacmi )$,
\begin{align}
\cmi &\leq I( \kappa( \traindatacmi ) ; \subsetchoice\vert \supersample) \leq H(\kappa( \traindatacmi ) \vert \supersample) \leq k\log(2n).
\end{align}
Here, the last step follows since, given~$\supersample$, there are at most~$\tbinom{2n}{k}\leq (2n)^k$ possible values of~$\kappa( \traindatacmi )$.
This establishes the result.
\end{proof}

Up to constants, this bound cannot be improved for general compression schemes.
However, for the important subclass of \emph{stable} compression schemes, the logarithmic dependence on~$n$ can be removed.
A compression scheme is said to be stable if it is invariant to permutations of its input, and~$\kappa(\traindata)=\kappa(\traindata')$ if~$\kappa(\traindata)\subseteq \traindata'\subseteq\traindata$---that is, if only elements that are not in the compressed set are removed from the training set, this does not change the output.
For stable compression schemes,~\citet[Thm.~3.4]{haghifam-21a} showed that~$\cmi\leq 2k\log(2)$.
This result demonstrates that the CMI suffices to obtain generalization bounds for stable compression schemes without a logarithmic dependence on~$n$, which is optimal up to constants~\citep[Thm.~3.1]{haghifam-21a}.

\section{Algorithmic Stability}\label{sec:info-comp-stability}

We now turn to algorithmic stability, as discussed in \cref{sec:alg-stability-bounds}.
As mentioned therein, several notions of stability have been discussed in the literature.
In this section, following~\citet[Thm.~4.2]{harutyunyan-21a}, we will focus on average prediction stability with respect to sample replacement and bound the~$f$-CMI.
This notion of stability is comparable to the pointwise hypothesis stability in \citet[Def.~4]{bousquet-02a}.
Note that~\citet{harutyunyan-21a} also consider other notions of stability, which we do not cover for brevity.
We will discuss further connections between algorithmic stability and information-theoretic and PAC-Bayesian generalization bounds in \cref{sec:bib-remarks-info-comp}.

\begin{thm}\label{thm:harutyunyan-stability}
Assume that~$\dataspace=\featurespace\times\reals^d$ and~$\ell(w,z)=\ell_f(f_w(x),y)$, where each~$w\in\mathcal W$ induces a function~$f_w:\featurespace \rightarrow \reals^d$.
Let~$\traindatacmi^{(i)}$ equal~$\traindatacmi$ for all entries except the~$i$th, which we denote by~$Z'=(X',Y')$, and assume to be independently drawn from~$P_Z$.
Consider a deterministic learning algorithm, and let~$f_{W\vert \traindatacmi}:\featurespace\rightarrow \reals^d$ denote the function that the learning algorithm induces given the training set~$\traindatacmi$.
Assume that the learning algorithm is~$\beta$-stable, meaning that for all~$i\in[n]$,
\begin{equation}
\Ex{\jointdistrocmi P_{Z'}}{ \norm{ f_{W\vert \traindatacmi}(\tilde X_{i+S_in}) - f_{W\vert \traindatacmi^{(i)}}(\tilde X_{i+S_in}) }^2  } \leq \beta^2.
\end{equation}
Roughly speaking, this means that the prediction that the hypothesis issues for~$\tilde X_{i+S_in}$ does not depend too strongly on whether or not this specific sample is included in the training set.
Furthermore, suppose that the loss function~$\ell_f(\cdot,\cdot)$ is~$\gamma$-Lipschitz in its first argument.
Then, we have that
\begin{equation}
\abs{\avggengap} \leq d^{1/4}\sqrt{ 8\gamma\beta } .
\end{equation}
\end{thm}
\begin{proof}

In order to establish this result, we will relate the deterministic algorithm to a stochastic one.
Specifically, let
\begin{equation}
f^\sigma_{W\vert \traindatacmi,N}(x) = f_{W\vert \traindatacmi}(x) + N_\sigma.
\end{equation}
Here, the Gaussian noise~$N_\sigma\distas \normal(0,\sigma^2 I_d)$, where~$I_d$ denotes the~$d$-dimensional identity matrix, is independent for all training sets and inputs.
With this, we find that the average generalization gap of the learning algorithm with added noise is
\begin{align}
\avggengap_\sigma = &\bigg| \Exop_{\jointdistrocmi P_{Z'} }\bigg[ \Ex{P_{N_\sigma}}{\ell_f( f^\sigma_{W\vert \traindatacmi,N}(X'), Y' )}   \nonumber\\
&\qquad\qquad\qquad- \frac1n\sum_{i\in[n]} \Ex{P_N}{\ell_f( f^\sigma_{W\vert \traindatacmi,N}(X_{i+S_in}), Y_{i+S_in} ) }   \bigg]    \bigg| \nonumber\\
= &\bigg| \Exop_{\jointdistrocmi P_{Z'} } \bigg[ \ell_f( f_{W\vert \traindatacmi}(X'), Y' ) + \Ex{P_{N_\sigma}}{\Delta'}  \\
&\qquad\qquad- \frac1n\sum_{i\in[n]} \bigg(\ell_f( f_{W\vert \traindatacmi}(X_{i+S_in}), Y_{i+S_in} ) +\Ex{P_{N_\sigma}}{\Delta_i} \bigg)  \bigg]     \bigg|,\nonumber
\end{align}
where
\begin{align}
\Delta' &\!=\! \ell_f( f^\sigma_{W\vert \traindatacmi,N}(X'), Y' )  \!-\! \ell_f( f_{W\vert \traindatacmi}(X'), Y' ), \\
\Delta_i &\!=\!  \ell_f( f^\sigma_{W\vert \traindatacmi,N}(X_{i+S_in}), Y_{i+S_in} ) \!-\! \ell_f( f_{W\vert \traindatacmi}(X_{i+S_in}), Y_{i+S_in} ).
\end{align}
Due to the Lipschitz assumption, we have~$\abs{\Delta'}\leq \gamma\norm{{N_\sigma}'}$, where~${N_\sigma}'\distas \normal(0,\sigma^2 I_d)$.
Similarly,~$\abs{\Delta_i}\leq \gamma\norm{{N_\sigma}'}$.
Since~$\Ex{}{\norm{{N_\sigma}'}}\leq 2\sigma\sqrt{d}$, we find that
\begin{align}
\avggengap_\sigma \geq \avggengap - 2\gamma\sigma\sqrt{d}.\label{eq:alg-stab-proof-bound-stoch-gap}
\end{align}
We now need to bound~$\avggengap_\sigma$.
Let~$\mathbf F^\sigma$ denote the vector of predictions on~$\supersamplex$ induced by~$f^\sigma_{W\vert\traindatacmi, N}$.
By the individual-sample $f$-CMI version of \cref{thm:slow_ecmi}, we have
\begin{align}
\avggengap_\sigma &\leq \frac1n\sum_{i\in[n]} \sqrt{2I(F^\sigma_i,F^\sigma_{i+n} ; S_i\vert\supersample )} \\
&\leq \frac1n\sum_{i\in[n]} \sqrt{2I(F^\sigma_i,F^\sigma_{i+n} ; S_i \vert \subsetchoice_{-i}, \supersample )},\label{eq:alg-stab-proof-pre-stability}
\end{align}
where~$\subsetchoice_{-i}$ is~$\subsetchoice$ with the~$i$th entry removed.
Here, the last step follows since~$\subsetchoice_{-i}$ is independent from~$S_i$.
To establish the result, it remains to bound the conditional mutual information in~\eqref{eq:alg-stab-proof-pre-stability}.
Intuitively, computing this quantity involves comparing the conditional joint distribution of~$(F^\sigma_i,F^\sigma_{i+n})$ and~$S_i$, given~$\subsetchoice_{-i}$ and~$\supersample$, with the products of their conditional marginals.
When~$S_i$ is drawn independently from all other random variables, there is a~$50\%$ chance of drawing the ‘‘matching'' instance, in which case the two distributions coincide, and a~$50\%$ chance of drawing the ‘‘opposite'' instance, in which case the~$i$th sample of the training set is replaced.
Hence, we are comparing two Gaussian distributions with covariance~$\sigma^2I_d$ and means given by the predictions based on the training set corresponding to~$\subsetchoice_{-i}$ and either~$S_i=1$ or~$S_i=0$.
By the stability assumption, the difference between the means is on average bounded by~$\beta^2$ (for more details, see the work of~\citealp[Prop.~4.2 and Eq.~(175)-(179)]{harutyunyan-21a}).
Since~$\relent{\normal(x_1,\sigma^2 I_d)}{\normal(x_2,\sigma^2 I_d)}=\norm{x_1-x_2}^2/(2\sigma^2)$, we get
\begin{align}
I(F^\sigma_i,F^\sigma_{i+n} ; S_i \vert \subsetchoice_{-i}, \supersample ) \leq \frac{\beta^2}{2\sigma^2} .\label{eq:alg-stab-proof-post-stability}
\end{align}
By combining~\eqref{eq:alg-stab-proof-bound-stoch-gap},~\eqref{eq:alg-stab-proof-pre-stability}, and~\eqref{eq:alg-stab-proof-post-stability}, setting~$\sigma^2=\beta/(2\gamma \sqrt d)$ to optimize the bound, we obtain the desired result.
\end{proof}

Thus, for Lipschitz losses, certain notions of algorithmic stability imply bounds on certain information measures for the learning algorithm, allowing us to (essentially) recover known generalization bounds (cf.\ \cref{sec:alg-stability-bounds}).
The technique used in this proof, where a learning algorithm is compared to a noisy surrogate in order to more easily evaluate the mutual information, is a fruitful approach that has also been used to establish generalization bounds for stochastic gradient descent~\citep{neu-21a}.

\section{Differential Privacy and Related Measures}\label{sec:info-comp-dp}

We now discuss differential privacy, which can be seen as a type of stability measure.
As the name suggests, this measure was originally constructed as a guarantee on the privacy of the training data used by a learning algorithm.
Specifically, let~$\traindatasmall,\traindatasmall'\in\dataspace^n$ be two training sets that differ in a single element.
Then, the algorithm~$\conddistro$ is~$\varepsilon$-differentially private if, for any measurable set~$\setE \in\mathcal W$~\citep{dwork-15a}
\begin{equation}
\conddistroz(\setE \vert \traindatasmall) \leq e^{\varepsilon}\conddistrozp(\setE \vert \traindatasmall') .
\end{equation}
This is related to so-called $\varepsilon$-MI stability, which requires that for any random~$\traindata\in\dataspace^n$~\citep{feldman-18a}
\begin{equation}
\frac1n \sum_{i=1}^n I(W;Z_i\vert \traindata_{-i}) \leq \varepsilon,
\end{equation}
where~$\traindata_{-i}$ denotes~$\traindata$ with the~$i$th element removed.
As shown by~\citet{feldman-18a}, an algorithm that is~$\sqrt{2\varepsilon}$-differentially private is~$\varepsilon$-MI stable.
If the elements of~$\traindata$ are independent, we have
\begin{equation}
I(W;\traindata) = \sum_{i=1}^n I(W;Z_i\vert\traindata_{<i}) \leq  \sum_{i=1}^n I(W;Z_i\vert \traindata_{-i}) \leq \varepsilon n,
\end{equation}
where $\traindata_{<i}=(Z_1,\dots,Z_{i-1})$ (and~$\traindata_{<1}=\emptyset$).
Thus, any~$\varepsilon$-MI stable (including any~$\sqrt{2\varepsilon}$-differentially private) learning algorithm has mutual information bounded by~$\varepsilon n$.

We conclude with a brief mention of max information, defined as~\citep{dwork-15a}
\begin{equation}
I_{\text{max}}(W;\traindata) = \esssup_{\jointdistro} \infdens.
\end{equation}
As established by~\citet[Lemma~12]{esposito-21a},~$\maxleakage{\traindata}{W}\leq I_{\text{max}}(W;\traindata)$.
Furthermore, since the~$\alpha$-mutual information is non-decreasing with~$\alpha$~\citep{verdu-15a}, and it coincides with the mutual information for~$\alpha=1$ and the maximal leakage for~$\alpha\rightarrow \infty$, we have
\begin{equation}
I(W;\traindata)  \leq \maxleakage{\traindata}{W} \leq I_{\text{max}}(W;\traindata) .
\end{equation}
Thus, bounds in terms of max information, as discussed by~\citet{dwork-15a}, can be recovered from bounds in terms of the mutual information and maximal leakage.

\section{Bibliographic Remarks and Additional Perspectives}\label{sec:bib-remarks-info-comp}

In this section, we discuss the relation of the results we presented to the literature, and give a brief overview of results that we did not cover explicitly.
For the Gibbs posterior, \cref{thm:gibbs-bound-max} is largely based on~\citet[Chapter~10]{raginsky-21a}, while \cref{thm:gibbs-aminian} is due to~\citet{aminian-21a}.

The Gaussian location model has been studied as an example application of information-theoretic generalization bounds since the work of~\citet{bu-19a}, with later improvements by~\citet{zhou-21a,zhou-22a,wu-22a}.
An information-theoretic bound that is tight up to constants was provided by~\citet{zhou-23a}.

For learning with VC classes,~\citet{xu-17a} constructed a two-phase learning algorithm with finite mutual information, but this result does not apply to standard empirical risk minimizers.
As shown by~\citet{livni-17a,bassily-18a,nachum-18a}, there are certain limitations in obtaining finite PAC-Bayesian and information-theoretic generalization bounds using the standard, non-CMI framework.
Recently,~\citet{pradeep-22a} showed that under the stricter requirement of a finite Littlestone dimension, it can be shown that learnability is possible with finite mutual information, demonstrating a gap compared to just having finite VC dimension.
Through the use of the CMI framework,~\citet{steinke-20a} obtained \cref{thm:cmi-vc-dim} for all empirical risk minimizers satisfying the consistency property of \cref{def:global-consistency-prop}.
As shown by~\citet{harutyunyan-21a}, the use of functional CMI enables \cref{thm:fcmi-vc-dim}, which applies to any learning algorithm.
An extension to the Natarajan dimension, which is an analogue of the VC dimension for the multiclass setting, was provided by~\citet{hellstrom-22a}.
Finally, the leave-one-out CMI framework enables optimal bounds for VC classes in certain situations, as shown by~\citet{haghifam-22a} and discussed in \cref{sec:info-comp-loo-cmi}.
Further discussion of the expressiveness of information-theoretic generalization bounds can be found in the work of~\citet{haghifam-21a}.
Notably, generalization bounds in terms of the VC dimension obtained from PAC-Bayesian bounds were originally derived in the work of~\citet[Corollary~2.4]{catoni-04a}.
The derivation is very similar to the CMI case, and based on the formalism of exchangeable priors.
This was extended to almost exchangeable priors by~\citet{audibert-04a,catoni-07a}.
Recently, a further extension that allows for bounds with fast rates under a Bernstein condition was provided by~\citet{grunwald-21a}.
Furthermore,~\citet{grunwald-19a} also explored connections between PAC-Bayesian bounds and the Rademacher complexity.

For compression schemes, \citet{steinke-20a} obtained the result of \cref{thm:cmi-comp-scheme}.
This was improved by a logarithmic factor for stable compression schemes by~\citet[Theorem~3.1]{haghifam-21a}.
\citet[Sec.~3]{catoni-04a} studied the use of exchangeable priors to obtain bounds for compression schemes.

The result in \cref{thm:harutyunyan-stability} is due to~\citet{harutyunyan-21a}, who also established results for other notions of algorithmic stability.
Bounds based on average stability, with connections to information-theoretic generalization bounds, were also established by~\citet{banerjee-22a}.
PAC-Bayesian generalization bounds in terms of stability have been established by, for instance,~\citet{london-14a,london-17a,rivasplata-18a,sun-22a,zhou-23a2}.

The discussion of privacy measures, such as the differential privacy of~\citet{dwork-15a}, in \cref{sec:info-comp-dp} is largely based on results from~\citet{feldman-18a}, with additional results due to~\citet{esposito-21a}.
For further discussion of these and other privacy measures, see for instance the work of~\citet{steinke-20a,oneto-20a,hellstrom-20b,esposito-21a,rodriguezgalvez-21b}.

\chapter{Neural Networks and Iterative Algorithms}\label{chap:iterative-methods}

In this chapter, we apply the bounds from \crefrange{chap:average}{chap:cmi} to learning algorithms that are \emph{iterative} in nature, in the sense that they proceed by updating a hypothesis step-by-step with the aim to converge to a final output hypothesis with good properties.
A key example of such an algorithm is the ubiquitous \emph{gradient descent}, which updates the current hypothesis by adding the negative gradient of the training loss, scaled by a parameter called the learning rate.
Of particular importance in modern machine learning are neural networks, which are typically trained using variants of (stochastic) gradient descent.
However, the framework of iterative learning algorithms applies to a much broader class of learning algorithms.

In \cref{sec:noisy_algos}, we discuss iterative, noisy algorithms in general, before specializing to the case of stochastic gradient Langevin dynamics (SGLD).
SGLD is a variant of stochastic gradient descent (SGD) with added Gaussian noise, which makes it particularly well-suited to analysis via information-theoretic bounds.
In \cref{sec:nn-numeric}, we discuss the application of generalization bounds from \crefrange{chap:average}{chap:cmi} to neural networks.
\newtext{Clearly, some bounds cannot be computed for practical scenarios---for instance, the mutual information depends on the unknown data distribution, and some information metrics can be prohibitively expensive to estimate due to high dimensionality or the lack of closed-form expressions.
For many bounds, however, it is possible to obtain informative values, for instance by using Monte Carlo estimates.
}%
We will mainly focus on methods for numerically evaluating the bounds, and discuss training algorithms inspired by them.
We will also provide pointers to methods for obtaining generalization bounds in closed form.

\section{Noisy Iterative Algorithms and SGLD}\label{sec:noisy_algos}

Here, we consider iterative learning algorithms of the following general form.
The hypothesis space ${\cal W}$ is the $d$-dimensional Euclidean space $\reals^d$. Given the training data $\traindata = (Z_1,\dots,Z_n)$, we generate the hypothesis $W$ as follows:
\begin{align}\label{eq:noisy_algo}
	\begin{split}
	W &= f(V_1,\dots,V_T) \\
	V_t &= g(V_{t-1}) - \eta_t F(V_{t-1},Z_{J_t}) + \xi_t, \qquad t = 1, \dots, T
	\end{split}
\end{align}
where $V_0$ is a random initial condition independent of everything else; $T \in \naturals$ is a fixed number of iterations; $J_1, \dots, J_t$ is a sequence of random elements of $[n] = \{1,\dots,n\}$; $\xi_t \sim {\cal N}(0, \rho_t^2 I_d)$ is a sequence of independent Gaussian random vectors which are also independent of everything else; and finally, $f(\cdot), g(\cdot), F(\cdot,\cdot)$ are deterministic mappings.
We will use the shorthand~$\boldsymbol{V}=(V_0,\dots,V_T)$.

The analysis relies on the following regularity assumptions:

\begin{enumerate}
	\item The following holds for the algorithm's \textit{sampling strategy}, \ie, the conditional probability law of $\boldsymbol{J} = (J_1,\dots,J_T)$ given $(\traindata,\boldsymbol{V})$: for each $t \in [T-1]$,
	\begin{align}\label{eq:sampling_strategy_assumption}
		P_{J_{t+1}|J_1,\dots,J_t,\boldsymbol{V},\traindata} = P_{J_{t+1}|J_1,\dots,J_t,\boldsymbol{Z}}.
	\end{align}
That is, the index of the sample in round~$t+1$ does not depend on the iterates
 $V_1,\ldots,V_t$, given the previous choices $J_1,\ldots,J_t$ and the data $\traindata$.
 \item The update function $F(\cdot,\cdot)$ is bounded:
 \begin{align}
	 \sup_{v \in \reals^d} \sup_{z \in {\cal Z}} \| F(v,z) \| \le L < \infty.
	\end{align}
\end{enumerate}
To control the generalization error, we will upper-bound the mutual information $I(W;\traindata)$. Let $\traindata^{\boldsymbol{J}} = (Z_{J_1},\dots,Z_{J_T})$ denote the random $T$-tuple of the training instances ``visited'' by the algorithm and observe that~$\traindata$ and~$\boldsymbol{V}$ are conditionally independent given~$\traindata^{\boldsymbol{J}}$.
Using this fact together with the data processing inequality and the chain rule, we have the following:
\begin{align}
	I(W; \traindata) &= I(f(\boldsymbol{V}); \traindata) \\
	&\le I(\boldsymbol{V}; \traindata) \\
	&\le I(\boldsymbol{V}; \traindata^{\boldsymbol{J}}) \\
	& = \sum^T_{t=1} I(V_t; \traindata^{\boldsymbol{J}}| V^{t-1}).\label{eq:the-above-sun}
\end{align}
Each term in~\eqref{eq:the-above-sun} admits a simple expression involving only random variables from two successive time steps, as we show in the following lemma.
\begin{lem}\label{lem:cond-cmi-iterative-rewrite}
Under the conditional independence assumption on the sampling strategy in~\eqref{eq:sampling_strategy_assumption},
	\begin{align}%
	I(V_t; \traindata^{\boldsymbol{J}} | V^{t-1}) = I(V_t; Z_{J_t} | V_{t-1}) . \label{eq:cond-indep-assumption}
	\end{align}
\end{lem}
\begin{proof} First, we express $I(V_t; \traindata^{\boldsymbol{J}}| V^{t-1})$ as
	\begin{align}
		I(V_t; \traindata^{\boldsymbol{J}}| V^{t-1}) = h(V_t|V^{t-1}) - h(V_t|V^{t-1},\traindata^{\boldsymbol{J}}),\label{eq:CMI_at_t}
	\end{align}
where $h(\cdot|\cdot)$ is the conditional differential entropy (\cref{def:differential-entropy}).
From the update rule for $V_t$ in \eqref{eq:noisy_algo} and the assumption on $\{\xi_t\}_{t\in[T]}$, it follows that $V_t$ is conditionally independent from~$(V^{t-2},\traindata^{\boldsymbol{J}\setminus\{J_t\}})$ given~$(V_{t-1},Z_{J_t})$.
Using this, we conclude that
	\begin{align*}
		h(V_t|V^{t-1},\traindata^{\boldsymbol{J}}) &= h(V_t|V_{t-1},Z_{J_t},V^{t-2},\traindata^{\boldsymbol{J} \setminus \{J_t\}}) \\
		& = h(V_t|V_{t-1},Z_{J_t}).
	\end{align*}
By the same token, $h(V_t|V^{t-1}) = h(V_t|V_{t-1})$. Using these expressions in \eqref{eq:CMI_at_t}, we obtain the desired result.
\end{proof}
\noindent The following lemma provides an easy-to-compute upper bound on $I(V_t; Z_{J_t}|V_{t-1})$ .
\begin{lem}\label{lem:cond-cmi-iterative-bound}
For every $t \in [T]$,
	\begin{align}
		I(V_t;Z_{J_t}|V_{t-1}) \le \frac{d}{2}\log \left( 1 + \frac{\eta^2_t L^2}{d\rho^2_t} \right) \le \frac{\eta^2_t L^2}{2\rho^2_t}.
	\end{align}
\end{lem}
\begin{proof} Given $V_{t-1} = v_{t-1}$, we have
	\begin{align}
		V_{t} = g(v_{t-1}) - \eta_t F(v_{t-1},Z_{J_t}) + \xi_t,
	\end{align}
	where $Z_{J_t}$ and $\xi_t$ are independent. Consequently, by the shift-invariance property of differential entropy,
	\begin{align}
		h(V_t|V_{t-1}=v_{t-1}) &= h(V_t - g(v_{t-1})|V_{t-1} = v_{t-1}) \\
		&= h(-\eta_t F(v_{t-1},Z_{J_t})+\xi_t|V_{t-1}=v_{t-1}).
	\end{align}
Now, recall that for any $d$-dimensional random vector $U$ with finite second moment, \ie, $\Ex{}{ \|U\|^2 } < \infty$, we have~\citep[Thm.~2.7]{polyanskiy-22a}
\begin{align}
	h(U) \le \frac{d}{2}\log \left(\frac{2\pi e \Ex{}{ \|U\|^2 } }{d}\right).
\end{align}
Since $Z_{J_t}$ and $\xi_t$ are independent and $\xi_t$ has zero mean, we obtain
\begin{align}
	&\Ex{}{\|-\eta_t F(v_{t-1},Z_{J_t})+\xi_t\|^2 \given V_{t-1} = v_{t-1}}\nonumber\\
	&= \eta^2_t \Ex{}{\|F(v_{t-1},Z_{J_t})\|^2 \given V_{t-1} = v_{t-1}} + \Ex{}{\|\xi_t\|^2} \nonumber\\
	&\le \eta^2_t L^2 + \rho^2_t d,
\end{align}
where we have also used the uniform boundedness assumption on $F(\cdot,\cdot)$. Consequently,
\begin{align}
	h(V_t|V_{t-1}) \le \frac{d}{2}\log \left(\frac{2\pi e(\eta^2_t L^2 + \rho^2_t d)}{d}\right).
\end{align}
By the same reasoning,
\begin{align}
	h(V_t|V_{t-1},Z_{J_t}) &= h(g(V_{t-1})-\eta_t F(V_{t-1},Z_{J_t})+\xi_t|V_{t-1},Z_{J_t}) \\
	&= h(\xi_t|V_{t-1},Z_{J_t}) \\
	&= h(\xi_t) \\
	&= \frac{d}{2}\log (2\pi e \rho^2_t),
\end{align}
where we have used the fact that $\xi_t$ is independent of the pair $(V_{t-1},Z_{J_t})$. Hence,
\begin{align}
	I(V_t; Z_{J_t}| V_{t-1}) &= h(V_t|V_{t-1}) - h(V_t|V_{t-1},Z_{J_t}) \\
	&\le \frac{d}{2}\log\left(1 + \frac{\eta^2_t L^2}{\rho^2_t d}\right) \\
	&\le \frac{\eta^2_t L^2}{2\rho^2_t},
\end{align}
where the last step follows from the inequality $\log x \le x - 1$.
\end{proof}
Combining \cref{lem:cond-cmi-iterative-rewrite,lem:cond-cmi-iterative-bound} and the mutual information generalization bound in \cref{cor:xu_raginsky}, we get the following result, due to~\citet{pensia-18a}.
\begin{thm}\label{thm:PJL} Suppose that $\ell(w,Z)$ is $\sigma^2$-subgaussian for every $w \in {\cal W}$ under~$P_Z$.
Then, under the assumptions on the sampling strategy and on $F$ stated in~\eqref{eq:noisy_algo} and~\eqref{eq:sampling_strategy_assumption}, we have
	\begin{align}\label{eq:thm-PJL}
		\Ex{\jointdistro}{\gen} \le \sqrt{\frac{\sigma^2}{n}\sum^T_{t=1}\frac{\eta^2_t L^2}{\rho^2_t}}.
	\end{align}
\end{thm}

We now specialize the result in~\eqref{eq:thm-PJL} to the case of SGLD.
Specifically, we assume that the loss $\ell(w,z)$ is differentiable as a function of $w$ for every $z$, and take
\begin{align}\label{eq:SGLD}
	\begin{split}
		V_0 &= 0 \\
		V_t &= V_{t-1} - \eta_t \nabla \ell(V_{t-1},Z_{J_t}) + \xi_t, \qquad t = 1, \dots, T \\
		W &= V_T
	\end{split}
\end{align}
where $J_1,\dots,J_T$ are i.i.d.\ samples from the uniform distribution on~$[n]$ (in each iteration, we sample with replacement from the $n$-tuple $\traindata$); $\eta_1,\dots,\eta_T$ are positive step sizes; and $\xi_t \sim {\cal N}(0,\rho^2_t I_d)$, with $\rho_t^2 = \frac{\eta_t}{\beta}$ for some $\beta > 0$.
The resulting SGLD algorithm is a special case of \eqref{eq:noisy_algo} with $g(v) = v$, $F(v,z) = \nabla \ell(v,z)$, and $f(v_1,\dots,v_T) = v_T$. Thus, $W$ is the last iterate $V_T$, although other choices are possible, such as $f(v_1,\dots,v_T) = \frac{1}{T}\sum^T_{t=1} v_t$ (trajectory averaging).
There exists a large literature on generalization bounds in expectation for SGLD; here we provide one such result due to~\citet{pensia-18a}, obtained under the (restrictive) assumption of a Lipschitz-continuous loss.

\begin{thm} %
Suppose that the loss function $w\mapsto \ell(w,z)$ is $L$-Lipschitz uniformly in $z$:
	\begin{align}
		\sup_{z \in {\cal Z}} |\ell(w,z) - \ell(w',z)| \le L \| w - w' \|.
	\end{align}
Assume that the SGLD algorithm in~\eqref{eq:SGLD} (with an arbitrary postprocessing step) runs for $T = nk$ steps, where $k$ is a positive integer, and let $\eta_t = \frac{1}{t}$. Then
\begin{align}
	\Ex{\jointdistro}{\gen} &\le \sqrt{\frac{\beta \sigma^2 L^2}{n} \sum^{nk}_{t=1}\frac{1}{t}} \\
	&\le \sqrt{\frac{\beta\sigma^2L^2}{n}(\log n + \log k + 1)}.
\end{align}
\end{thm}
\begin{proof} By the Lipschitz assumption on $\ell$, its gradient $\nabla \ell(\cdot,\cdot)$ is bounded by $L$ in $\ell^2$ norm. The result then follows from \cref{thm:PJL}.
\end{proof}

\section{Numerical Bounds for Neural Networks}\label{sec:nn-numeric}

In recent years, many practical successes in machine learning have relied on neural networks (NNs).
Although a comprehensive discussion of NNs is beyond the scope of this monograph, we will provide a very brief description of NNs and introduce some notation.
Further details can be found in, for instance,~\citet[Chapter~III]{murphy-22a}.
While a whole host of different NN architectures have been developed for specific application areas, we will focus solely on so-called feedforward NNs.
We proceed by defining a single layer, from which NNs can be formed through composition.
Each layer consists of two components: an affine transformation and an activation function.
Denote the input to the~$l$th layer as~$x_{l-1}\in\reals^{d_{l-1}}$.
The weights of the~$l$th layer are denoted by~$A_l\in\reals^{d_l\times d_{l-1}}$, while the bias vector is~$b_l\in\reals^{d_l}$.
We refer to~$d_l$ as the width of the layer.
Then, the pre-activation output is given by~$a_l=A_l x_{l-1} + b_l$, which is simply an affine transformation of the input.
In order to allow the network to express non-linear functions, we also use an activation function~$\phi_l:\reals\rightarrow \reals$.
Then, the final output from the layer is given by~$x_l=\phi_l(a_l)$, where the activation function is applied elementwise to the pre-activation vector~$a_l$.
Since NNs are typically trained through a gradient-based algorithm, this activation function is often required to be (almost everywhere) differentiable.
An NN~$f_W(\cdot)$ of depth~$L$ consists of~$L$ such layers, where we let~$W\in\reals^p$, with~$p=\sum_{l=1}^L (d_{l-1}+1) d_l$, denote the concatenation of all weights and biases expressed in vector form.
We will typically also denote the output as~$\hat y = x_L\in\reals^{d_L}$ and the input as~$x = x_0\in\reals^{d_0}$.
Thus, the final output is~$\hat y=f_W(x)=\phi_L(a_L)$.

For a given sample~$z=(x,y)$, the loss is given by~$\ell(W,z)=\ell_f(\hat y, y)$.
Given the training set~$\traindata$, we assume that the NN is trained as follows: first, the weights and biases of the network are initialized as~$W_0$.
At each time step~$t$, they are then updated as
\begin{align}
W_t &= W_{t-1} - \eta\nabla_W \trainloss \\
&= W_{t-1} - \eta\sum_{i=1}^n \nabla_W \ell_f(\hat y_i, y_i) \\
&= W_{t-1} - \eta\sum_{i=1}^n \nabla_Wf_W(x_i) \frac{\dv \ell_f(\hat y_i, y_i)}{\dv \hat y_i} . \label{eq:form-of-gd}
\end{align}
Here,~$\eta>0$ is the learning rate.
The exact form of this update depends on the specific activation function under consideration, and can be computed for each parameter of the network through the chain rule.
This process may, for instance, continue for a fixed number of steps or until a certain target loss, either evaluated on the training set or on a held-out validation set, is reached.
One common variant of~\eqref{eq:form-of-gd} is SGD, where the training loss gradient is not evaluated with respect to the entire training set at each time step.
Instead, a ‘‘mini batch'' of~$K<n$ samples is selected at each time step, and the weight update is computed with respect to these samples.
This has several benefits, such as speeding up computation and reducing memory requirements.

Typically, NNs operate in the so-called \emph{overparameterized} regime.
This means that~$p$, which is determined by the widths and depth of the network, is greater than what would be needed in order to interpolate the~$n$ training samples in~$\traindata$ after gradient descent training.
In many practically relevant scenarios,~$p$ is many orders of magnitude greater than~$n$.
In fact, NNs often have the capacity to interpolate the training data even with randomly assigned labels.
This indicates that they do not operate in a regime where notions like the VC dimension are relevant~\citep{zhang-21a}.
Still, when trained using data with the correct labels, NNs display impressive generalization performance.
So, in the regime that is relevant in practice, NNs generalize well when trained with true labels, but generalize poorly when trained with random labels.
This appears to indicate that any generalization guarantee that is uniform over all data distributions is doomed to be vacuous, since such a guarantee would need to hold for both scenarios.
This provides a motivation for considering PAC-Bayesian and information-theoretic bounds, as these can incorporate data-distribution dependence.
We now discuss various ways to evaluate information-theoretic and PAC-Bayesian bounds for NNs.

\subsection{Weights with Gaussian Noise}

One issue with applying many standard PAC-Bayesian and information-theoretic generalization bounds, as repeatedly discussed, is that they are often vacuous for deterministic learning algorithms.
For instance, training an NN using gradient descent with a fixed initialization and stopping criterion would yield infinite mutual information between the training data and the parameters of the NN.
Now, typically, there are sources of stochasticity in NN training.
First, the initialization is often not fixed, but instead drawn from some distribution.
Second, training is usually based on SGD, or one of its variants, rather than deterministic gradient descent.
However, characterizing information-theoretic quantities in the presence of these sources of stochasticity is not entirely straightforward.
Furthermore, one would still expect the bulk of generalization performance to be present even for deterministic gradient descent---while the stochasticity of SGD, for instance, may provide a marginal benefit, it is unlikely to make the difference between very poor and very good generalization.
This was empirically demonstrated by~\citet{geiping-22a}.

An alternative approach builds on the popular hypothesis that the generalization capabilities of an NN are related to the \emph{flatness} of the loss function in the vicinity of its global minima.
If the training loss of the NN is not significantly affected when its parameters are perturbed, this indicates some kind of robustness that could lead to good generalization.
This is intimately related to the concept of margins, which has previously been successfully used to analyze the performance of support vector machines~\citep{cristianini-00a}.
It is with this motivation that~\citet{langford-01b} considered stochastic NNs, for which the parameters are randomly drawn from a particular distribution each time the NN is used.
The distribution of each parameter is set as an independent Gaussian distribution, whose mean coincided with the underlying deterministic NN and with variance selected to be as large as possible without degrading the training loss by more than a given threshold.
Exploiting this randomization, they were able to evaluate PAC-Bayesian generalization bounds, which can be related to the performance of the underlying deterministic NN through parameters such as the margin and Lipschitz properties of the NN.
In order to be able to select reasonable parameters for the prior, \citet{langford-01b} considered a suitable dyadic grid of candidate values, applying a union bound over these to obtain bounds that hold simultaneously for all candidates on the entire grid.
This led to bounds that are nonvacuous, and significantly better than known generalization bounds for deterministic networks---although the NNs that were considered by~\citet{langford-01b} were naturally significantly less complex than what has been used in recent years.

This approach was adapted to more modern settings by~\citet{dziugaite-17a}.
While~\citet{langford-01b} performed a sensitivity analysis for each parameter separately, this is not tenable for large NNs.
Instead, given a trained NN, \citet{dziugaite-17a} selected the weight distributions by directly optimizing a PAC-Bayesian bound, using \cref{cor:MLS_bound} as a starting point.
By using the relaxation obtained via Pinsker's inequality, replacing the training loss with a convex surrogate, fixing the prior to be a Gaussian distribution centered on the underlying deterministic network, and restricting the posterior to be an isotropic Gaussian, they obtained a training objective that can be optimized via gradient-based methods.
The underlying motivation for why this procedure is successful is, as already indicated, the hypothesized flatness of the loss landscape around minimizers of the training loss.
While certain measures of flatness have been criticized as insufficient to explain generalization, since they can be arbitrarily altered through reparameterizations that do not affect the neural network itself~\citep{dinh-17a}, measuring flatness through the relative entropy avoids such drawbacks.
Indeed, the relative entropy is invariant under parameter transformations.

This idea was further developed by~\citet{dziugaite-21a}, who pointed out the crucial role that data-dependent priors, discussed in \cref{sec:data-dep-prior}, can play in the tightness of PAC-Bayesian bounds, as  observed earlier by, \eg,~\citet{ambroladze-06a,mhammedi-19a}.
In fact, as demonstrated in~\citet[Lemma~3.3]{dziugaite-21a}, there exist learning settings for which data-dependent priors are necessary in order to obtain a nonvacuous PAC-Bayesian bound.

Motivated by this,~\citet{dziugaite-21a} proceed to evaluate such data-dependent priors for NNs.
Roughly speaking, a fraction~$\alpha$ of the training set,~$\traindata_P$, is used to train an NN upon which the prior is based, while the full training set~$\traindata$ is used to train another NN that corresponds to the posterior.
In order to obtain a tighter characterization, this is done in such a way that both NNs process the same samples in the initial epochs, since these will have the largest impact on the final weights.
Experiments are also performed where the prior is further informed by a ghost sample, which is not used for selecting the posterior, in order to approximate an oracle prior.
The use of data-dependent priors leads to tighter bounds than just the use of a ghost sample.
Crucially, unlike the aforementioned results, this leads to nonvacuous bounds when the posterior is chosen through a standard SGD-based procedure (with added noise).
However, an even tighter bound can be obtained by optimizing the PAC-Bayesian bound via SGD, as shown in~\citet[Fig.~5]{dziugaite-21a}.
Even tighter results, where bounds with data-dependent priors were directly optimized, were obtained by~\citet{perez-ortiz-21a}, who argued that this could potentially be used for self-certified learning, where no separate test set is needed to certify the performance of the learned hypothesis.
Still, the utility of these data-dependent priors is not entirely clear.
As argued by~\citet[Fig.~1(a)]{lotfi-22a}, similar or better bounds can be obtained by simply letting the posterior equal the data-dependent prior, and using the remaining data to obtain an unbiased estimate of the population loss.

\subsection{Using the CMI Framework}

As discussed in \cref{sec:cmi-pac-bayes}, the CMI framework of \cref{chap:cmi} can be viewed as an alternative path to data-dependent priors.
This was exploited in~\citet{hellstrom-21b,hellstrom-21a}, wherein an approach similar to that of~\citet{dziugaite-17a,dziugaite-21a} was used, in that Gaussian distributions centered on the outputs of SGD are set as the posterior and prior.
Specifically, given a supersample~$\supersample$ of training samples, half of them are selected to form the training set~$\traindatacmi$.
The mean of the posterior is then found by running SGD for a fixed set of iterations on~$\traindatacmi$.
Next, the true marginal distribution~$\conddistrocmi$ in \cref{thm:cmi-pac-bayes} is replaced by an auxiliary~$\auxconddistrocmi$, the mean of which is obtained by averaging the output of SGD trained on a number of samples of~$\traindatacmi$ with a fixed~$\supersample$.
For both the posterior and prior, the variance is set to be as large as possible while not degrading the training loss of the randomized NN too much---similar to~\citet{langford-01b}, but with a uniform choice for all parameters.
While this yields similar numerical bounds as~\citet{dziugaite-21a}, there is one notable drawback---the bound cannot be directly optimized, as this would introduce a direct dependence of the posterior on~$\testdatacmi$.
This would violate the required conditional independence between~$\supersample$ and~$W$ given~$\traindatacmi$.

All these bounds apply to stochastic networks, with noise added to the parameters, and not to the underlying, deterministic ones that are typically used in practice.
While the CMI bounds are finite without this added noise, as guaranteed by the CMI framework, they are typically vacuous.
This can be avoided through the use of evaluated or functional CMI (e-CMI or $f$-CMI).
Motivated by the aim of obtaining information-theoretic generalization bounds that depend on the predictions induced by a learning algorithm, instead of the hypothesis itself,~\citet{harutyunyan-21a} derived several bounds in terms of the $f$-CMI.
To illustrate the benefits of this shift, consider the case of binary classification.
Then, the $f$-CMI~$\fcmi$ measures the mutual information between the predictions~$F$ and the membership vector~$\subsetchoice$---two discrete random variables---given the supersample~$\supersample$.
Furthermore, for individual-sample $f$-CMI bounds,~$\fcmiiz$ measures mutual information between binary random variables.
This dramatically expands the set of possible scenarios where the information measure, and thus the bound itself, can be small even for deterministic learning algorithms, while being easy to evaluate numerically.
Specifically,~\citet{harutyunyan-21a} evaluated an average, disintegrated, individual-sample $f$-CMI bound through Monte Carlo estimation, and obtained nearly accurate estimates of the test error for deterministic NNs with relatively small training set sizes.
These numerical evaluations were extended to tighter generalization bounds and e-CMI by~\citet{hellstrom-22a}.
In subsequent work,~\citet{wang-23a} obtained further improvements through the use of ld-MI.

\begin{figure}
 \centering
    \includegraphics[width=0.8\textwidth]{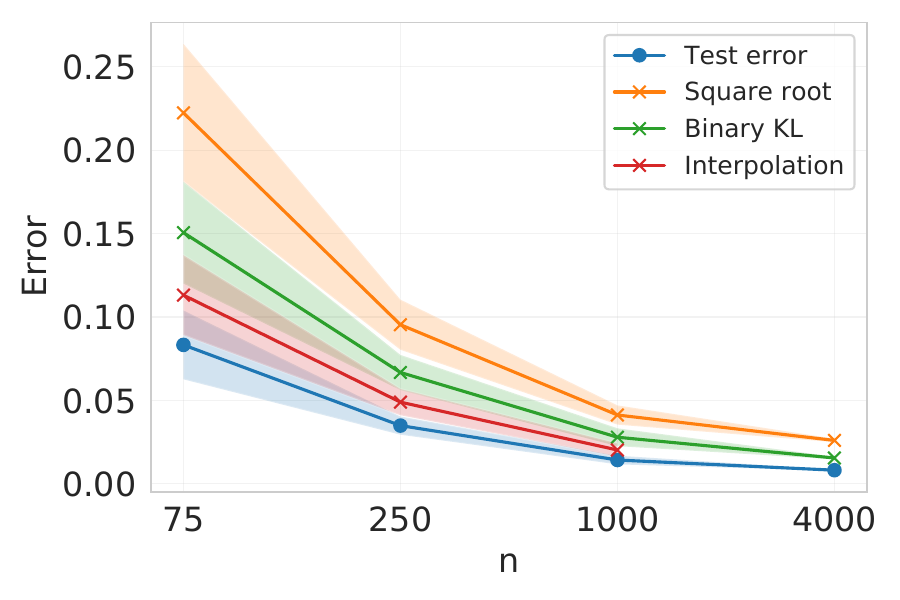}
    \caption{Numerical evaluation for a CNN trained on a binary version of MNIST \citep[Fig.~2(a)]{hellstrom-22a}.}
    \label{fig:mnist}
\end{figure}
\newtext{For a concrete example, consider \cref{fig:mnist} \citep[Fig.~2(a)]{hellstrom-22a}. The setting under consideration is binary classification for a version of the MNIST data set, which consists of $32\times 32$ images of handwritten digits. Specifically, the data set is restricted to the digits 4 and 9, and a CNN trained with Adam (a variant of SGD) is used. The plot shows the test error, \ie, the test loss using the~$0-1$ loss, along with several upper bounds. Specifically, these are samplewise, disintegrated e-CMI versions of the square-root bound in \eqref{eq:cmi-sqrt-bound}, the binary KL bound in \eqref{eq:thm:cmi-binary-kl}, and the interpolation bound in \eqref{eq:steinke_fast_interpolating}. To be explicit, the bounds are, recalling the notation of \cref{sec:ecmi},
\begin{align}
\avgpoploss 
&\leq \avgtrainloss + \frac1n\sum_{i=1}^n\Ex{\supersample }{ \sqrt{2 I^{\supersample\!}( \Lambda_i ;\subsetchoice_i) } } \label{eq:num-sqrt}\\
\avgpoploss &\leq \Exop_{\supersample}\bigg[ d^{-1}_2\bigg(\Ex{ P_{W\! \subsetchoice \vert \supersample\! } }{\trainlosscmi},
\frac1n\sum_{i=1}^n  I^{\supersample\!}(\Lambda_i ;\subsetchoice_i) \bigg)\bigg].\label{eq:num-kl}\\
 \avgpoploss &\leq\sum_{i=1}^n\frac{ I( \Lambda_i ; \subsetchoice_i \vert \supersample ) }{n\log(2)}.\label{eq:num-interp}
\end{align}
where $d^{-1}_2(q,c) = \sup \big\{p \in [0,1]: d\big(q\,||\,\frac{q+p}{2}\big) \leq c\big\}$. The disintegrated samplewise e-CMI $I^{\supersample\!}( \Lambda_i ;\subsetchoice_i)$ is evaluated via sampling: for each $n\in\{75, 250, 1000, 4000\}$, a supersample of $2n$ samples is drawn from the full data set. Half of these are selected to obtain the $n$ training samples, and the network is then trained and evaluated. This is repeated several times to build an empirical distribution of the relevant random variables, which is used to compute the mutual information term via a plug-in estimator. The results show that, whenever it is applicable, the interpolating bound~\eqref{eq:num-interp} is tightest. For $n=4000$, not all training losses were zero, precluding its use. Thus, the binary KL bound of \eqref{eq:num-kl} is tightest of the applicable bounds. For all values, it improves on the square-root bound~\eqref{eq:num-sqrt}. Thus, these results demonstrate that the bounds can be estimated and are numerically fairly accurate. For more details and results for other settings, the reader is referred to, for instance, the work of \citet{harutyunyan-21a,hellstrom-22a,wang-23a}. }

Note that, in contrast to the aforementioned bounds for stochastic NNs, these bounds hold only in expectation.
While corresponding results can be obtained in probability, this would limit the possibility of using the individual-sample technique, potentially degrading the bounds significantly.

\subsection{Compression-Based Bounds}

An alternative approach to obtaining numerically nonvacuous generalization bounds for NNs is through the lens of \emph{compression}~\citep{arora-18a,bu-21a}.
This approach builds on the observation that, often, well-performing NNs can be significantly compressed without noticably affecting their performance.
While generalization bounds for the original NN may be far from accurate, applying the same bound to a compressed NN can yield much better results.
While these bounds still do not explain the generalization capabilities of the original NN, they can provide guarantees for the compressed counterparts.

This approach was used by~\citet{zhou-18a}, who obtained non-vacuous generalization bounds for NNs by combining off-the-shelf compression algorithms and PAC-Bayesian bounds.
The idea is essentially to set the posterior in the PAC-Bayesian bound to be a point mass centered on the output of the combined NN training and compression algorithm, and combine this with a suitably chosen prior on the set of possible hypotheses following the compression step.
The specific compression algorithm considered by~\citet{zhou-18a} is weight pruning, whereby a large number of parameters are set to zero in a way that aims to minimize adversely affecting predictive performance~\citep{han-16a}.
Finally, in order to further exploit the flatness of the loss surface, Gaussian noise is added to the non-zero weights, similar to the approach taken by~\citet{dziugaite-17a}.

This approach was extended in several ways by~\citet{lotfi-22a}, who aimed to leverage these bounds to shed light on various factors behind generalization in NNs.
First, they perform training only in a carefully constructed random linear subspace of the parameters, constraining the space of possible hypotheses and thus enabling smaller compressed sizes.
Instead of pruning,~\citet{lotfi-22a} use trainable quantization, whereby the quantization levels and the weights themselves can be learned simultaneously.
Furthermore, whereas~\citet{zhou-18a} considered a prior based on a uniform distribution,~\citet{lotfi-22a} replaced it with a so-called universal prior, which places greater weight on more compressed hypotheses.
This leads to nonvacuous bounds, which can be further tightened through the use of data-dependent priors in the style of~\citet{ambroladze-06a,dziugaite-21a}.
However,~\citet{lotfi-22a} argue that while this leads to numerically accurate bounds, it does not explain generalization for the full learning procedure: such bounds only compare the posterior to the data-dependent prior, but the question of why the prior is good is left unanswered.
Finally, numerical experiments by~\citet{lotfi-22a} indicate that one possible explanation for why techniques such as transfer learning and the use of symmetries improve generalization is that they improve compressibility.

\section{Bibliographic Remarks and Additional Perspectives}\label{sec:bib-remarks-nn}

The results in \cref{sec:noisy_algos} are based on the work of~\citet{pensia-18a}.
Additionally, information-theoretic bounds for SGLD have also been derived by, for instance,~\citet{mou-18a,li-20a,bu-20a,negrea-19a,haghifam-20a,wang-21a,wang3-21a,wang2-23a,issa-23a,futami-23a}.
By relating the parameter trajectory of SGLD to the corresponding noise-free trajectory of SGD,~\citet{neu-21a,wang2-22a} obtained bounds for SGD.
However, as demonstrated by~\citet{haghifam-23a}, current information-theoretic approaches are not sufficient to obtain minimax optimal rates for stochastic convex optimization problems.
\newtext{This was rectified to some extent by \citet{wang-23c}, who combined the information-theoretic approach with techniques from algorithmic stability.}

In addition to the results for NNs that we have discussed so far, several alternative approaches to obtain generalization bounds for neural networks have been explored in the literature, both within the scope of information-theoretic and PAC-Bayesian bounds and beyond it.
While a comprehensive overview of all such work is beyond the scope of this monograph, we will mention some of the approaches here.
For instance, bounds have been derived based on the norms of the weights of the NN~\citep{neyshabur-15a,bartlett-17a}.
A PAC-Bayesian view on this approach was taken by~\citet{neyshabur-18a}, who used the robustness of NNs to parameter perturbations in order to obtain a de-randomized bound in terms of a relative entropy that can be evaluated explicitly.
\citet{bartlett-02a} derived norm-based bounds for NNs starting from the Rademacher complexity.
The connection between PAC-Bayesian bounds and flatness has also been explored by, \eg,~\citet{tsuzuku-20a,foret-21a}.
Several works have derived generalization bounds for NNs trained via SGLD~\citep{bu-20a,haghifam-21a}, and other noisy versions of SGD~\citep{banerjee-22a}.
\citet{pitas-20a} explored the use of Gaussian posteriors in PAC-Bayesian bounds for NNs, while~\citet{dziugaite-18a} established a connection to entropy-SGD.

In the limit of infinite width, and under certain conditions on their initialization, NNs can be described as a Gaussian process~\citep{neal-94a}, a correspondence referred to as the NN Gaussian process (NNGP---\citealp{lee-18a}).
For certain loss functions and suitably scaled learning rates, the evolution of the infinitely wide NN during training is also tractable, and is described by the neural tangent kernel (NTK)~\citep{jacot-18a}.
\citet{valleperez-18a} combined PAC-Bayesian bounds with the NNGP correspondence to argue that the functions learned by NN tend to be simple in a sense that leads to generalization, and support their arguments by numerically estimating the relevant quantities.
\citet{bernstein-21a} took a similar approach, but derived analytical upper bounds that lead to nonvacuous generalization guarantees.
\citet{shwartz-ziv-20a} used the NTK formalism to analytically study many information metrics for NNs, such as~$I(W;\traindata)$.
\citet{huang-23a,clerico-23a,clerico-24a} extended the NTK formalism to networks trained by optimizing PAC-Bayesian bounds, while~\citet{wang-22a} explored connections to the information bottleneck.

\citet{viallard-19a} used the PAC-Bayesian framework to analyze a particular two-phase procedure to train NNs.
\citet{rivasplata-19a} considered a broad family of methods for training stochastic NNs by minimizing PAC-Bayesian bounds.
\citet{letarte-19a} considered NNs with binary activation functions, and used PAC-Bayesian bounds to both formulate a framework for training and to obtain nonvacuous generalization guarantees.
\citet{biggs-21a} considered ensembling over stochastic NNs, obtaining differentiable PAC-Bayes objectives, while \citet{biggs-22a} derived a de-randomized PAC-Bayesian bound for shallow NNs, using data-dependent priors to get nonvacuous generalization bounds.
\citet{zantedeschi-21a} used PAC-Bayesian bounds to learn stochastic majority votes, while \citet{nagarajan-19a} obtained de-randomized PAC-Bayes bounds via noise-resilience.
\citet{tinsi-22a} obtained tractable bounds for certain aggregated shallow NNs, using a PAC-Bayesian bound with Gaussian priors as the starting point, while~\citet{clerico-22b} derived a training algorithm for stochastic NNs without the need for a surrogate loss.
\citet{jin-22a} discussed how the use of dropout affects PAC-Bayesian generalization bound through the concept of weight expansion.
\citet{liao-21a} used PAC-Bayes to derive generalization bounds for graph NNs, while~\citet{viallard-21b,xiao-23a} derived bounds for adversarial robustness.

Comprehensive surveys of various complexity measures and their connection to generalization can be found in, for instance, the works of~\citet{neyshabur-17a,jiang-20a,dziugaite-20a}.

\chapter{Alternative Learning Models}
\label{chap:alternative-settings}

So far, we have considered a generic learning model in which the learner has access to~$n$ (typically \iid) data points from a fixed data distribution, and the goal is to achieve a small loss on new samples from the same distribution.
While this learning model covers many learning settings of interest, it is not all-encompassing.
In this chapter, we consider learning problems that do not fit neatly into the generic setting we discussed so far.
We will not analyze any of these settings in depth.
Our aim is merely to illustrate the wide applicability of the information-theoretic and PAC-Bayesian approaches to generalization.

First, we discuss the setting of meta learning, wherein the learner observes training data from several related tasks, and the goal is to learn how to perform well on a new task.
Next, we consider transfer learning, wherein the distribution of the training data is not the same as the distribution of the test data.
This is closely related to domain adaptation and out-of-distribution generalization.
Following this, we present an information-theoretic generalization bound for federated learning, where a set of distributed nodes separately observe training samples, on the basis of which a composite hypothesis is formed under certain communication constraints.
Finally, we look at reinforcement learning, wherein the learner collects observations by interacting with an environment.
Specifically, it observes states, takes actions according to a policy, and receives rewards, with the goal of learning a policy that yields high rewards.
We conclude by briefly discussing the application of information-theoretic and PAC-Bayesian generalization bounds to online learning, active learning, and density estimation.

\section{Meta Learning}

In typical supervised learning, each learning task is considered in a vacuum: the learner has access to~$n$ training samples from the task, and this is all it has to go by.
In reality, this is usually not the case: different tasks of interest may have many commonalities.
For instance, any computer vision task is based on processing of visual data, which may be similar across many different tasks.

This idea is captured by the framework of meta learning~\citep{caruana-97a,thrun-98a,baxter-02a}.
In this setting, we assume that there exists a task space~$\taskspace$, paired with a task distribution~$\taskdistro$.
For each task~$\task\in\taskspace$, there is a corresponding in-task data distribution~$\traindistroarg{\task}$.
In order to form the meta-training set~$\metatrainsetb\in\dataspace^{\tasknr \times n}$,~$\newtext{\tasknr}$ tasks are drawn from~$\taskdistro$, and for each of these,~$n$ samples are drawn from the corresponding~$\traindistroarg{\task}$.
Thus, for each~$i\in[\tasknr]$,~$\task_i$ is drawn independently from~$\taskdistro$, and for each~$j\in[n]$,~$\metatrainset_{i,j}$ is drawn independently from~$\traindistroarg{\tau_i}$.
On this basis, the meta learner aims to find a hyperparameter (or meta hypothesis)~$U\in\mathcal U$ on the basis of the meta-learning algorithm~$\metaalg$.
This hyperparameter will serve as an additional input to a base learner, allowing it to use information from the meta-training set for new tasks.
Specifically, for~$\traindata\in\dataspace^n$, the base learner is characterized by the conditional distribution~$\basealg$.
The performance of the meta learner is evaluated through the test loss of the base learner on a test task.
Specifically, let~$\task$ be drawn from~$\taskdistro$, independently from~$\metatrainset$, let the ‘‘test-training set''~$\traindata^\task$ consist of~$n$ \iid samples from~$\traindistroarg{\task}$, and let the ‘‘test-test sample''~$Z^\task\distas\traindistroarg{\task}$.
Then, the average meta-test loss is defined as
\begin{equation}
\avgpoploss = \Ex{ P_{\metatrainsetb} \metaalg  P_{\traindata^\task} \basealgt P_{Z^\task} }{ \ell(W,Z^\task) }  =  \Ex{ P_W P_{Z^\task} }{ \ell(W,Z^\task) }.
\end{equation}
While the meta learner does not have access to~$\avgpoploss$, it can compute the meta-training loss, defined as
\begin{equation}
\avgtrainloss = \Ex{ P_{\metatrainset} \metaalg }{\frac1{\tasknr}\sum_{i=1}^{\tasknr} \Ex{  \basealgi  }{ \frac1n\sum_{j=1}^n \ell(W_i,\metatrainset_{i, j}) }  }.
\end{equation}
Here,~$\metatrainsetb_{i,:}=(\metatrainset_{i,1},\dots,\metatrainset_{i,n})$ denotes the training set for the~$i$th task and~$W_i$ is the corresponding hypothesis of the base algorithm.
For simplicity, we only focus on generalization bounds in expectation.
We can extend all of these results to obtain PAC-Bayesian and single-draw counterparts, by following the approach detailed in \cref{chap:probability}.

In the standard learning setting, a key step was to perform a change of measure to handle the dependence between the training data and the hypothesis.
In the meta-learning setting, there is an additional dependence between the training data and the hyperparameter.
One way to handle this additional dependence is to use a two-step approach, wherein an auxiliary loss is introduced as an intermediate step between the meta-training and meta-population loss.
This allows us to obtain generalization bounds by applying two changes of measure, separately: one to relate the meta-training loss to the auxiliary loss, and one to relate the auxiliary loss to the meta-population loss.
This allows us to apply standard generalization bounds on the intra-task and inter-task levels separately.
However, tighter bounds can be obtained by dealing with them simultaneously.
This joint approach leads to the following generalization bound for meta learning, due to~\citet{chen-21a}.
\begin{thm}\label{thm:meta-learning-it-bound}
Assume that the loss is~$\sigma$-sub-Gaussian.
Let~$\hat W=(W_1,\dots,W_{\tasknr})$ denote the output hypotheses of the base learners for the~$\tasknr$ training tasks.
Then,
\begin{equation}
\abs{ \avgpoploss-\avgtrainloss } \leq \sqrt{ \frac{2\sigma^2  I(U,\hat W; \metatrainsetb)}{n\tasknr} } .
\end{equation}
\end{thm}
\begin{proof}
The proof essentially follows immediately by the same approach as was used in the proof of \cref{cor:xu_raginsky}, once we make the following observation:
the average loss on the meta-training set under the joint distribution of~$U$,~$\hat W$, and~$\metatrainsetb$ equals~$\avgtrainloss$.
If we instead draw~$(U,\hat W)$ independent from~$\metatrainsetb$, it equals~$\avgpoploss$.
We begin by re-writing the training loss as
\begin{align}
\avgtrainloss =& \Ex{ P_{\metatrainsetb} \metaalg }{\frac1{\tasknr}\sum_{i=1}^{\tasknr} \Ex{  \basealgi  }{ \frac1n\sum_{j=1}^n \ell(W_i,\metatrainset_{i, j}) }  } \\
=& \frac1{\tasknr}\sum_{i=1}^{\tasknr} \Ex{  \basealgi P_{\metatrainsetb} \metaalg }{ \frac1n\sum_{j=1}^n \ell(W_i,\metatrainset_{i, j}) }  .
\end{align}
Furthermore, since the tasks and samples are~\iid,
\begin{equation}
 \frac1{\tasknr}\sum_{i=1}^{\tasknr} \Ex{  P_{W_i\vert U} P_{\metatrainsetb} P_U }{ \frac1n\sum_{j=1}^n \ell(W_i,\metatrainset_{i, j}) }  =  \Ex{ P_W P_{Z^\task} }{ \ell(W,Z^\task) } = \avgpoploss.
\end{equation}
We conclude the proof by changing measure from~$P_{U \hat W  \! \metatrainset}$ to~$P_{U\hat W}P_{\metatrainset}$ and using sub-Gaussian concentration.
\end{proof}

\noindent The effects of the environment level and in-task level in \cref{thm:meta-learning-it-bound} can be disentangled through the use of the chain rule:
\begin{align}
\sqrt{ \frac{2\sigma^2  I(U,\hat W; \metatrainsetb)}{n\tasknr} }  &= \sqrt{ \frac{2\sigma^2 (I(U; \metatrainsetb)  + I(\hat W;\metatrainset\vert U))}{n\tasknr} }\\
& \leq \sqrt{ \frac{2\sigma^2 I(U; \metatrainsetb) }{n\tasknr} } + \sqrt{ \frac{2\sigma^2  I(W_1;\metatrainsetb_{1,:}\vert U)}{n} }.
\end{align}
In the second step, we used the fact that~$I(\hat W;\metatrainset\vert U)$ can be separated as~$\tasknr$ mutual information terms, one for each task, with the same underlying distributions.

The bound in \cref{thm:meta-learning-it-bound} can be tightened through the use of alternative changes of measure and concentration methods, disintegration, and the individual-sample technique.
We will not discuss this explicitly, but instead provide pointers for such extensions and to additional results.
PAC-Bayesian bounds for meta learning have been derived, often with a focus on algorithms that minimize these bounds to improve generalization, by, \eg,~\citet{pentina-14a,amit-18a,rothfuss-21a,rezazadeh-22a}.
Information-theoretic bounds were provided by \citet{jose-21a,jose-22a}, who used a two-step derivation, and~\citet{chen-21a} who used the one-step derivation above.
A CMI formulation of meta learning was introduced by~\citet{rezazadeh-21a}, which was later extended to incorporate one-step derivations, disintegration, and alternative comparator functions by~\citet{hellstrom-22b}.
Finally,~\citet{jose-21c} derived generalization bound that explicitly incorporate task similarity, as measured through, for instance, the relative entropy.

\section{Out-of-Distribution Generalization and Domain Adaptation}

In the standard learning setting, the population loss is defined with respect to the same distribution from which the training set was drawn.
While this is a natural assumption to make from a theoretical standpoint, there are many situations in which one expects a distribution shift when deploying a model.
There are also scenarios where there is an abundance of data from a surrogate distribution, while there is a lack of data from the actual distribution of interest.
This motivates theoretical settings where the population loss is defined with respect to a target distribution, which may differ from the source distribution used to generate the training data.

For the purposes of this discussion, we assume that the sample space factors into a feature space and a label space as~$\dataspace=\featurespace\times\labelspace$.
The overarching framework, where the only assumption is that the training data is drawn from a source distribution~$\sourcedistro$ but we evaluate the model on a target distribution~$\targetdistro$, is usually referred to as \emph{out-of-distribution} (OOD) generalization~\citep{liu-23a}.
When the marginal distribution on~$\featurespace$ induced by~$\sourcedistro$ differs from the one induced by~$\targetdistro$, but the conditional distributions of the label given the features are identical, we refer to this as \emph{domain adaptation}~\citep{kouw-19a,redko-22a}.
Finally, whenever the learner has access to (partial) samples from the target distribution, we refer to this as \emph{transfer learning}, categorized as \emph{unsupervised} if the learner only has access to unlabelled target features and \emph{supervised} if it has access to full target samples~\citep{weiss-16a}.
While the definitions of OOD generalization and domain adaptation provided above are fairly established, the term transfer learning is overloaded, and is sometimes used to refer to OOD generalization more broadly, or even to certain variations of meta learning.

For simplicity, we will only consider bounds in expectation.
As usual, we denote the training set as~$\traindata$, drawn from~$\traindistro=\sourcedistro^n$, and the output hypothesis from the stochastic algorithm~$\conddistro$ as~$W$.
Similarly, the average training and population loss with respect to the source distribution are still given by
\begin{equation}
\avgtrainloss = \Ex{\jointdistro}{ \trainloss }, \qquad \avgpoploss = \Ex{\jointdistro}{ \Ex{\sourcedistro}{\ell(W,Z)} }.
\end{equation}
However, the performance metric that we actually wish to minimize is the average \emph{target} population loss, given by
\begin{equation}
\avgpoploss^T = \Ex{\jointdistro}{ \Ex{\targetdistro}{\ell(W,Z^T)} }.
\end{equation}

\subsection{Generic OOD Generalization Bounds}

Our first approach to obtaining OOD generalization bounds is natural.
Since we have already established bounds for the population loss under the source distribution, but are now interested in bounds under the target distribution, we can just apply a change of measure.
By a direct application of the \donskervtext, we obtain the following~\citep{wang-23b}.
\begin{propo}\label{propo:ood-gen-pp}
Assume that the loss function is~$\sigma$-sub-Gaussian under~$\sourcedistro$ almost surely under~$P_W$ and that~$\targetdistro\ll\sourcedistro$.
Then,
\begin{equation}
\abs{\avgpoploss^T-\avgpoploss} \leq \sqrt{2\sigma^2\relent{\targetdistro}{\sourcedistro}}.
\end{equation}
\end{propo}
\begin{proof}
By the \donskervtext\ in \cref{thm:donskervaradhan}, we have for any~$\lambda\in\reals$
\begin{equation}
  \relent{\targetdistro}{\sourcedistro} \! \geq \!  \Ex{\targetdistro}{\lambda\Ex{\jointdistro\!}{\ell(W,Z^T)}} \!-\!  \log \Ex{\sourcedistro}{e^{ \lambda\Ex{\jointdistro}{\ell(W,Z)}  \!}} \! . \!
\end{equation}
Due to the sub-Gaussianity assumption, we have
\begin{align}
 &\log \Ex{\sourcedistro}{e^{ \lambda\Ex{\jointdistro}{\ell(W,Z)}  \!}} \nonumber\\
 =&  \log \Ex{\sourcedistro}{e^{ \lambda(\Ex{\jointdistro}{\ell(W,Z)} - \Ex{\sourcedistro}{\Ex{\jointdistro}{\ell(W,Z)}} + \Ex{\sourcedistro}{\Ex{\jointdistro}{\ell(W,Z)}})  \!}}\\
 \geq & \lambda\Ex{\sourcedistro}{\Ex{\jointdistro}{\ell(W,Z)}} + \frac{\lambda^2\sigma^2}{2}.
\end{align}
By combining these steps and optimizing over~$\lambda$ for the two cases~$\lambda>0$ and~$\lambda<0$, we obtain the final result.
\end{proof}

\cref{propo:ood-gen-pp} allows us to turn any generalization bound for standard learning into an OOD generalization bound via the triangle inequality, at the cost of a term depending on~$\relent{\targetdistro}{\sourcedistro}$.
This result confirms the intuition that OOD generalization works well if the target and source distribution are similar, with the added specificity that similarity in terms of relative entropy is sufficient.
One drawback of the relative entropy is that it requires absolute continuity for finiteness.
This can be alleviated to some extent: the role of the source distribution~$\sourcedistro$ and target distribution~$\targetdistro$ in the derivation above can be swapped, leading to a bound in terms of~$\relent{\sourcedistro}{\targetdistro}$.
For this to work, we instead need to assume that the loss function is~$\sigma$-sub-Gaussian under~$\targetdistro$ almost surely under~$P_W$ and that~$\sourcedistro\ll\targetdistro$.

Unfortunately, there are scenarios where neither of these conditions are satisfied---for instance, if the two distributions have disjoint supports.
This motivates bounds in terms of other information measures, such as the Wasserstein distance.
The following result follows directly from the Kantorovich-Rubinstein duality.

\begin{propo}\label{propo:wasserstein-ood}
Assume that the loss is~$1$-Lipschitz. Then,
\begin{equation}
\abs{\avgpoploss^T-\avgpoploss} \leq \wasserstein{1}{\sourcedistro}{\targetdistro}.
\end{equation}
\end{propo}

The benefit of this result is that, unlike for the relative entropy, it remains finite even for the case where the source and target distributions have disjoint support.

\subsection{Unsupervised Transfer Learning}

In the previous section, we derived generic bounds in which minimal assumptions were made on the distributions and task, and the learning algorithm did not have access to any samples from the target distribution.
While this led to explicit bounds in terms of discrepancy measures between the source and target distribution, the utility is limited---we cannot minimize these discrepancy measures, and in fact, we do not have any access to the source and target distributions.

In order to gain algorithmic insights, we will now consider unsupervised transfer learning.
More precisely, we assume that the sample space factors into a feature space and label space as~$\dataspace=\featurespace\times\labelspace$.
Hence, the target distribution also factors as~$\targetdistro=\targetdistrox\targetdistroyvx$.
Furthermore, we assume that the hypothesis~$W$ implements a function~$f_W:\featurespace\rightarrow\labelspace$, and that its loss depends on the true label and the corresponding prediction as~$\ell(W,Z)=\ell_f(f_W(X), Y)$.
In addition to the training set~$\traindata$ drawn from~$\traindistro$, the learning algorithm now also has access to a set of unlabelled features~$\unsuptrain=(X_1^T,\dots,X_m^T)$, with each element drawn independently from~$\targetdistrox$.
The learning algorithm is now characterized by the conditional distribution~$\conddistrozx$, and the training loss and target population loss are thus given by
\begin{equation}
\avgtrainloss = \Ex{\jointdistrowzx}{ \trainloss } ,\qquad
\avgpoploss^T = \Ex{\jointdistrowzx}{ \Ex{\targetdistro}{\ell(W,Z^T)} } .
\end{equation}
Following~\citet{wang-23b}, we can derive bounds on~$\avgpoploss^T$ directly from~$\avgtrainloss$, \ie, without going through the source-distribution population loss.

\begin{thm}\label{thm:transfer-learning-unsup}
Assume that the loss function is~$\sigma$-sub-Gaussian under~$\sourcedistro$ almost surely under~$P_W$ and that~$\targetdistro\ll\sourcedistro$.
Then,
\begin{equation}
\abs{\avgpoploss^T-\avgtrainloss} \leq \Ex{P_{\unsuptrain}}{ \sqrt{ \frac{2\sigma^2 I^{\unsuptrain}(W;\traindata)}{ n }  + 2\sigma^2\relent{\targetdistro}{\sourcedistro}} } .
\end{equation}
\end{thm}
\begin{proof}
We begin by considering a specific~$\unsuptrain$.
Then, by the same argument as used in \cref{propo:ood-gen-pp}, for all~$\lambda\in\reals$
\begin{align}
&\relent{P_{W\!Z_i\vert X^T_j }}{P_{W\vert X^T_j} \targetdistro} \nonumber \\
&\geq \Ex{P_{W\!Z_i\vert X^T_j }}{\lambda\ell(W,Z_i)} - \Ex{P_{W\vert X^T_j} \targetdistro}{\lambda\ell(W,Z^T)} - \frac{\sigma^2\lambda^2}{2n}.
\end{align}
Now, note that
\begin{equation}
\relent{P_{W\!Z_i\vert X^T_j }}{P_{W\vert X^T_j} \targetdistro}  = I^{X^T_j}(W;Z_i) + \relent{P_Z}{P_Z^T}.
\end{equation}
Hence, by optimizing over~$\lambda$ as before, we get
\begin{align}
 &\abs{\Ex{P_{W\!Z_i\vert X^T_j }}{\lambda\ell(W,Z_i)} - \Ex{P_{W\vert X^T_j} \targetdistro}{\lambda\ell(W,Z^T)} } \nonumber \\
 &\leq\sqrt{ 2\sigma^2 I^{X^T_j}(W;Z_i) + \relent{P_Z}{P_Z^T} }\label{eq:transfer-learning-samplewise-conditioned-bound}
\end{align}
The stated result now follows by decomposing~$\abs{\avgpoploss^T-\avgtrainloss}$, applying~\eqref{eq:transfer-learning-samplewise-conditioned-bound} termwise, and performing a full-sample relaxation.
\end{proof}

The role of~$\unsuptrain$ in the disintegrated mutual information here is not entirely clear.
Indeed, if we use Jensen's inequality to move the expectation inside the square root, we get
\begin{equation}
 \Ex{P_{\unsuptrain}}{ \sqrt{ I^{\unsuptrain}(W;\traindata)  } } \leq  \sqrt{  I(W;\traindata\vert \unsuptrain) }  .
\end{equation}
This conditional mutual information is lower-bounded as~$ I(W;\traindata\vert \unsuptrain) \geq   I(W;\traindata) $.
If we had not fixed~$\unsuptrain$ at the beginning of the derivation, and had instead just averaged it out, we would have obtained a generalization bound in terms of~$I(W;\traindata)$, where the role of~$\unsuptrain$ is ignored, as was done by~\citet{jose-21b}.
However, the relation between~$\Ex{P_{\unsuptrain}}{ \sqrt{ I^{\unsuptrain}(W;\traindata)  } }$ and~$I(W;\traindata)$ is not clear.
Indeed, the unlabelled target features could potentially be used to decrease the information measure that appears in the bound, as discussed by~\citet{wang-23b}.

Still, this does not address the term~$\relent{P_Z}{P_Z^T}$ in \cref{thm:transfer-learning-unsup}.
This term can be controlled to some extent when the function implemented by the learning algorithm can be expressed as a composition~$f_W=g_W\circ h_W$, where~$h_W:\featurespace\rightarrow\representationspace$ is a mapping to a \emph{representation} space~$\representationspace$ and~$g_W:\representationspace\rightarrow\labelspace$ is the final mapping to the prediction.
Here,~$f_W(\cdot)$ can for instance be an~$N$-layer neural network, where~$h_W(\cdot)$ consists of the first~$N-k$ layers and~$g_W(\cdot)$ consists of the remaining~$k$ layers, for some~$k\in[N]$.
For this setting, we can try to align the distributions on the representation induced by the source and target distributions.

For the purposes of this discussion, we will look at the relative entropy~$\relent{\targetdistro}{\sourcedistro}$, but similar techniques can be applied to, \eg, the Wasserstein distance.
First, consider a fixed function~$h: \featurespace\rightarrow\representationspace$, and let~$\targetdistroh$ denote the pushforward of~$\targetdistrox$ with respect to~$h$---that is, the distribution on~$\representationspace$ induced by~$h$ acting on~$\targetdistrox$---and similarly for~$\sourcedistroh$.
Furthermore, let~$\targetdistroyvh$ and~$\sourcedistroyvh$ denote the conditional target and source distributions for the label, given the representation.
Then, for a fixed~$W$, we have
\begin{align}
 \avgpoploss^T(W) = \Ex{\targetdistro}{\ell(W,Z)} &= \Ex{\targetdistroh\targetdistroyvh}{\ell( g_W(h_W(X)) )}, \\
 \avgpoploss(W) = \Ex{\sourcedistro}{\ell(W,Z)} &= \Ex{\sourcedistroh\sourcedistroyvh}{\ell( g_W(h_W(X)) )} .
\end{align}
Therefore, by repeating the argument of \cref{propo:ood-gen-pp} with this re-formulation at the start, we obtain
\begin{equation}
\abs{\avgpoploss^T-\avgpoploss} \leq \Ex{P_W}{ \sqrt{2\sigma^2\relent{\targetdistroh\targetdistroyvh}{\sourcedistroh\sourcedistroyvh}} }.
\end{equation}
The result in \cref{thm:transfer-learning-unsup} can be adapted similarly.
Next, note that the relative entropy can be decomposed as
\begin{equation}
\relent{\targetdistroh\targetdistroyvh\!}{\!\sourcedistroh\sourcedistroyvh}\! =\! \relent{\targetdistroh\!}{\!\sourcedistroh}  \!+\! \relent{\targetdistroyvh\!}{\!\sourcedistroyvh} .
\end{equation}
Consequently, we have two components of the discrepancy measure: the representation discrepancy~$\relent{\targetdistroh\!}{\!\sourcedistroh}$ and the conditional discrepancy~$\relent{\targetdistroyvh\!}{\!\sourcedistroyvh}$.
The representation discrepancy is something that we actually \emph{can} aim to minimize by suitably designing our learning algorithm.
While we do not have access to the underlying feature distribution for neither the source nor the target, we have empirical estimates based on the source features in~$\traindata$ and the unlabelled target features~$\unsuptrain$.
Thus, as part of choosing~$W$, we can aim to minimize the discrepancy between the pushforward of these empirical source and target feature distributions with respect to~$h_W$.

Now, the relative entropy between the two conditional distributions is not under our control in the same sense, but there are situations where its contribution can be minor.
For the setting of domain adaptation, this term will be zero, as we assume that the conditional distribution on the label given the features is identical for the source and target distributions.
This implies that the corresponding pushforward measures are also equal.
Under some additional assumptions, this relative entropy can also be replaced by a term that is small for settings of practical relevance.
Specifically, as shown by~\citet[Thm.~4.2]{wang-23b}, if we assume that the loss is symmetric and satisfies the triangle inequality, we find that, for any fixed~$W$,
\begin{equation}
\avgpoploss^T(W)-\avgpoploss(W) \leq  \sqrt{2\sigma^2\relent{\targetdistrox}{\sourcedistrox}}  + \min_{w^*\in\mathcal W} \{\avgpoploss^T(w^*) + \avgpoploss(w^*) \}.
\end{equation}
Thus, the relative entropy between the conditional distributions can be replaced by the smallest possible sum of source and target population losses.
If transfer learning is to be successful in the sense that we should be able to find a hypothesis that works well for both the source and the target distributions---even given oracle knowledge of the true distributions---this quantity has to be small.

We conclude this section by presenting a generalization bound for \emph{supervised} transfer learning, where the learning algorithm has access to labelled data from the target distribution.
This bound is in terms of the~$f$-mutual information and uses total variation as discrepancy measure, and is due to~\citet{wu-22a}.
We shall assume that, in addition to the source training set~$\traindata$, the learning algorithm also has access to a set of~$m$ labelled examples from the target distribution~$\traindata^T=(Z^T_1,\dots,Z^T_m)$, with all elements drawn independently from~$\targetdistro$.
Thus, the learning algorithm is characterized by a conditional distribution~$P_{W\vert \traindata \traindata^T}$.
We define the weighted training loss as
\begin{align}
\avgtrainloss &=  \Ex{P_{W\!\traindata\!\traindata^T}}{ \frac{\alpha}{m}\sum_{i=1}^m \ell(W,Z^T_i) }  +  \Ex{P_{W\!\traindata\!\traindata^T}}{ \frac{1-\alpha}{n}\sum_{i=1}^n \ell(W,Z_i) } \\
 &=      \frac{\alpha}{m}\sum_{i=1}^m \Ex{P_{W\!Z^T_i}}{ \ell(W,Z^T_i) }  +    \frac{1-\alpha}{n}\sum_{i=1}^n \Ex{P_{W\!Z_i}}{ \ell(W,Z_i)  }  .
\end{align}
Here, the parameter~$\alpha\in[0,1]$ determines the relative emphasis that we place on the data from the target distribution.
When~$\alpha=1$, it reduces to the standard training loss for supervised learning.
When~$\alpha=0$, we are instead back to a generic OOD setting with no target data to learn from.
\begin{thm}\label{thm:trasnfer-supervised}
Assume that, for any~$w\in\mathcal W$, the loss is bounded by~$\sigma$ in~$L_\infty$-norm, \ie,
\begin{equation}
\abs{ \ell(w,Z) }_\infty = \inf\{ s: \targetdistro( \ell(w,Z) > s ) = 0 \} \leq \sigma.
\end{equation}
Then, we have
\begin{multline}
\abs{\avgpoploss^T-\avgtrainloss} \leq  \frac{2\alpha\sigma}{m}\sum_{i\in[m]} \TV( \jointdistroi , \productdistroit  ) \\ +  \frac{2(1-\alpha) \sigma }{n} \sum_{i\in[n]}  \left( \TV( \jointdistroi , \productdistroi  ) + \TV(\sourcedistro,\targetdistro) \right).
\end{multline}
\end{thm}
\begin{proof}
First, we decompose the generalization gap as
\begin{align}\label{eq:proof-step-1}
\abs{\avgpoploss^T\!\!-\!\avgtrainloss} &= \abs{  \avgpoploss^T -    \frac{\alpha}{m}\sum_{i=1}^m \Ex{P_{W\!Z^T_i}}{ \ell(W,Z^T_i) }  -    \frac{1-\alpha}{n}\sum_{i=1}^n \Ex{P_{W\!Z_i}}{ \ell(W,Z_i)  }  } \nonumber \\
&\leq  \frac{\alpha}{m} \sum_{i=1}^m \abs{ \Ex{P_W\!P_{Z^T_i}}{  \ell(W,Z^T) } -  \Ex{P_{W\!Z^T_i}}{ \ell(W,Z^T_i) } }  \\&\;\;+  \frac{1-\alpha}{n}\sum_{i=1}^n\abs{  \Ex{P_W\!P_{Z^T_i}}{  \ell(W,Z^T) } - \Ex{P_{W\!Z_i}}{ \ell(W,Z_i)  } }. \nonumber
\end{align}
Now, the terms in the first sum are individual-sample generalization gaps.
By applying \cref{thm:f-div-avg-bound-no-n} to each term, we can bound them as
\begin{equation}\label{eq:proof-step-2}
\Ex{P_W\!P_{Z^T_i}}{  \ell(W,Z^T) } -  \Ex{P_{W\!Z^T_i}}{ \ell(W,Z^T_i) } \leq \TV(P_{W\!Z^T_i},P_W\!P_{Z^T_i}).
\end{equation}
Proceeding similarly with the second sum, we can bound each term as
\begin{equation}\label{eq:proof-step-3}
\Ex{P_W\!P_{Z^T_i}}{  \ell(W,Z^T) } - \Ex{P_{W\!Z_i}}{ \ell(W,Z_i)  } \leq \TV(P_{W\!Z_i},P_W\!P_{Z^T_i}).
\end{equation}
To isolate the effect of the distribution shift, we can decompose this last upper bound as
\begin{align}
\TV(P_{W\!Z_i},P_W\!P_{Z^T_i}) &= \frac12 \int_{\mathcal W\times\dataspace} \abs{ \dv P_{W\!Z_i} - \dv P_W\!P_{Z^T_i} } \\
&\leq \frac12 \int_{\mathcal W\times\dataspace} \abs{ \dv P_{W\!Z_i} - \dv P_W\!P_{Z_i} }  \\
&\qquad + \frac12 \int_{\mathcal W\times\dataspace} \abs{ \dv P_W\!P_{Z_i}  - \dv P_W\!P_{Z^T_i}  } \nonumber  \\
&=\TV(P_{W\!Z_i},P_W\!P_{Z_i}) +  \TV(\sourcedistro,\targetdistro)  .\label{eq:proof-step-4}
\end{align}
We obtain the desired result by substituting \eqref{eq:proof-step-2},~\eqref{eq:proof-step-3} and~\eqref{eq:proof-step-4} into~\eqref{eq:proof-step-1}.
\end{proof}

While we only covered bounds in expectation, many of these results can be extended to PAC-Bayesian and single-draw variants.
Further discussion regarding many of these topics, as well as practical algorithms based on these bounds, are provided by~\citet{wu-22a,aminian-22b,wang-23b}.

\section{Federated Learning}

Federated learning is a framework for describing distributed learning, for instance in mobile networks~\citep{kairouz-21a}.
Specifically, we assume that there are~$K$ separate nodes, all having access to their own training set~$\traindata_k=(Z_{k,1},\dots,Z_{k,n})$ of size~$n$, for each~$k\in[K]$.
We assume that~$Z_{k,i}\distas\datadistro$ for all~$(k,i)\in [K] \times [n]$, and denote the collection of all training sets as~$\traindata=(\traindata_1,\dots,\traindata_K)$.
Each node uses a learning algorithm~$\conddistrok$ to generate the hypothesis~$W_k$ on the basis of~$\traindata_k$.
These local models are then combined to form the final model~$W$ through an aggregation algorithm~$\conddistroaggrw$.
A common choice is to use averaging, so that~$W=\frac1K\sum_{k=1}^K W_k$.
Composing the local learning algorithms and the aggregation algorithm induces a conditional distribution on~$W$ given the full training set~$\traindata$, denoted as~$\conddistroaggr$.
As usual, our aim is to bound the population loss~$\poploss$.

One way to obtain generalization bounds is simply to consider~$\conddistroaggr$ as a learning algorithm acting on~$nK$ samples, and use a generalization bound for standard supervised learning.
Alternatively, assuming that the aggregation algorithm performs averaging and that the loss is convex, we have
\begin{align}
\poploss &= \Ex{\datadistro}{ \ell\lefto(\frac1K\sum_{k=1}^K W_k,Z\right) } \\
&\leq \frac1K\sum_{k=1}^K \Ex{\datadistro}{ \ell( W_k,Z) } .\label{eq:compared-to-this-approach}
\end{align}
This allows us to apply a standard generalization bound for each node.
Neither of these approaches, as noted by~\citet{barnes-22a}, exploits the specific structure of federated learning, except potentially implicitly through the information measures that appear in the bounds.
We will therefore focus here on the result in~\citet[Thm.~4]{barnes-22a}, in which an explicit improved dependence on the number of nodes~$K$ is achieved.

To this end, we need to assume that the loss can be described as a \emph{Bregman divergence}.
Specifically, for a continuously differentiable and strictly convex function~$f:\reals^m \to \reals$, the Bregman divergence between two points~$p,q\in\reals^m$ is defined as
\begin{equation}
\bregmandiv{f}{p}{q} = f(p)-f(q) - \langle \nabla f(q) , p-q \rangle ,
\end{equation}
where~$\langle\cdot,\cdot\rangle$ is the inner product.
Notably, this includes the squared loss, which is obtained by setting~$f(\cdot)$ to be the squared two-norm.
With this, the following can be established.
\begin{thm}\label{thm:federated-learning}
Assume that the loss function is a Bregman divergence~$\ell(w,z)=\bregmandiv{f}{w}{z}$.
Furthermore, assume that~$\ell(w,Z)$ is~$\sigma$-sub-Gaussian under~$P_Z$ for all~$w\in\mathcal W$.
Then, if~$W=\frac1K\sum_{k=1}^K W_k$,
\begin{equation}
\Ex{\jointdistro}{\poploss - \trainloss} \leq \frac{1}{K^2} \sum_{k\in[K]} \sqrt{ \frac{I(W_k; \traindata_{k})}{n} }.
\end{equation}
\end{thm}
\begin{proof}
Let~$\traindata'=(\traindata'_1,\dots,\traindata'_K)$ be an independent copy of~$\traindata$, and let~$\traindata^{(k,i)}$ equal~$\traindata$ for all elements except~$Z^{(k,i)}_{k,i}=Z'_{k,i}$.
Then, we have~\citep[Lemma~11]{shalev-shwartz-10a}
\begin{align}
\Ex{\jointdistro}{\poploss} \!&=\! \frac{1}{nK}\sum_{k,i} \Ex{\jointdistro P_{\traindata'}}{\ell(W,Z'_{k,i}) } \\
\!&=\! \frac{1}{nK}\sum_{k,i}\Ex{\jointdistro P_{\traindata'}\!}{f(W)\!-\!f(Z'_{k,i})\!-\!\langle \nabla f(Z'_{k,i}), W\!-\!Z'_{k,i}}\! ,\! \nonumber
\end{align}
since~$Z'_{k,i}$ is independent from~$W$.
Here, the summation indices implicitly run over~$k\in[K]$ and~$i\in[n]$.
Let~$W^{k,i}$ be drawn according to~$P_{W^{k,i}\vert \traindata^{(k,i)}}$.
Then,
\begin{align}
\Ex{\jointdistro}{\trainloss} &= \frac{1}{nK}\sum_{k,i}  \Ex{P_{W^{k,i} \traindata \! \traindata'}} {\ell(W^{k,i},Z'_{k,i}) } \\
&=\frac{1}{nK}\sum_{k,i} \Exop_{P_{W^{k,i} \traindata \! \traindata'}} \bigg[ f(W^{k,i})-f(Z'_{k,i}) \\
&\qquad\qquad-\langle \nabla f(Z'_{k,i}), W^{k,i}\!-\!Z'_{k,i} \bigg] , \nonumber
\end{align}
since~$Z'_{k,i}$ is in the training set of~$W^{k,i}$.
It follows that
\begin{multline}
\Ex{\jointdistro}{\poploss-\trainloss} \\
=\!\frac{1}{nK}\sum_{k,i}   \Ex{P_{W W^{k,i} \traindata \! \traindata'}} {  \langle \nabla f(Z'_{k,i}) , W^{k,i}-W \rangle }  .\!
\end{multline}
Here, we used that~$\Ex{P_W}{f(W)} = \Ex{P_{W^{k,i}}}{f(W^{k,i})}$ since~$W$ and~$W^{k,i}$ have the same marginal distributions.
The key observation that leads to the improved dependence on~$K$, compared to an approach using~\eqref{eq:compared-to-this-approach}, is that~$W$ and~$W^{k,i}$ are the average of~$K$ sub-models, but they differ only in the~$k$th sub-model.
Hence,~$W^{k,i}-W=\frac1K(W^i_k-W_k)$, where~$W^i_k$ denotes the~$k$th submodel trained on~$\traindata_k^{(i)}$.
Therefore,
\begin{multline}
\Ex{\jointdistro}{\poploss-\trainloss} \\
=\!\frac{1}{nK^2}\sum_{k,i}   \Ex{P_{W W^{k,i} \traindata \! \traindata'}} {  \langle \nabla f(Z'_{k,i}) , W^i_k-W_k \rangle }  .
\end{multline}
Hence, we can conclude that
\begin{align}
\Ex{\jointdistro}{\poploss \!-\! \trainloss} \!=\! \frac{1}{K^2}\! \sum_{k\in[K]} \! \Ex{\jointdistro}{L_{\datadistro}(W_k) \!-\!L_{\traindata_k}(W_k)} . \!
\end{align}
We obtain the desired result by applying \cref{cor:xu_raginsky}.
\end{proof}
If~$z=(x,y)$, this result also holds if~$\ell(w,(x,y))=\bregmandiv{f}{\langle w,x\rangle}{y}$, with a nearly identical proof.
Intuitively, the improved dependence on~$K$ arises because the dependence of the final hypothesis~$W$ on any individual sample is dampened by~$1/K$ due to the averaging.
Naturally, this result can be extended to incorporate disintegration, the individual-sample technique, or by using other generalization bounds than \cref{cor:xu_raginsky} in the proof.
For further discussion and extensions of these bounds, see for instance the work of~\citet{yagli-20a,barnes-22a}.

\section{Reinforcement Learning}\label{sec:reinforcement-learning}

So far, we have assumed that the training data is independent from the learning algorithm.
In this section, we instead look at reinforcement learning, wherein the learner collects observations by taking observation-dependent actions in an environment.
Specifically, in \cref{sec:pacb-martingales}, we present extensions of PAC-Bayesian bounds from \iid data to martingales, which allows us to capture some of the interactions that occur in reinforcement learning.
Then, in \cref{sec:markovdecisionprocess}, we discuss information-theoretic bounds for Markov decision processes (MDP), which constitute an important class of reinforcement learning problems.

\subsection{PAC-Bayesian Bounds for Martingales}\label{sec:pacb-martingales}

We begin by presenting a PAC-Bayesian bound for martingales (described in \cref{sec:conc-martingales}) due to~\citet{seldin-12a}.
This can be used to apply generalization bounds like those in \cref{sec:pac-bayesian-bounds} developed for \iid training samples to various types of interactive settings.
\begin{thm}\label{thm:martingale-pacb}
Let~$ M_i$ for~$i\in[n]$ be a martingale sequence of random functions~$ M_i:\mathcal W\rightarrow [-1,1]$ such that~$\Ex{}{ M_{i+1}(w)\vert  {\boldsymbol{M}}_{\leq i}(w)}=0$ for all~$w\in\mathcal W$, where~$ {\boldsymbol{M}}_{\leq i}(w)=( M_1(w),\dots, M_i(w))$.
Suppose that the randomness of each~$ M_i$ is captured by a random variable~$Z_i$, and let~$\bar M_t=\sum_{i=1}^t  M_i$ and~$\traindata=(Z_1,\dots,Z_n)$.
Fix a prior distribution~$Q_W$ on~$\mathcal W$ and a~$\delta\in(0,1)$.
Then, for every distribution~$\conddistro$ on~$\mathcal W$, with probability at least~$1-\delta$ over~$P_{\traindata}$,
\begin{equation}\label{eq:thm-martingale-pacb-eq}
\abs{\Ex{\conddistro}{\frac{\bar M_n(W)}{n}}} \leq \sqrt{  \frac{ \relent{\conddistro}{Q_W} + \log\frac{4en}\delta }{ 2n }  } .
\end{equation}
\end{thm}
\begin{proof}
By the \donskervtext, we have, for a fixed~$\lambda>0$,
\begin{align}
 \Ex{\conddistro}{\frac{\lambda\bar M_n(W)}{n}}  &\leq \relent{\conddistro}{Q_W}  + \log \Ex{Q_W}{e^{\frac{\lambda\bar M_n(W)}{n} } } .
\end{align}
By Markov's inequality, we have with probability at least~$1-\delta$
\begin{align}
\log \Ex{Q_W}{e^{\frac{\lambda\bar M_n(W)}{n} } }
 &\leq \log \Ex{Q_W \datadistro}{\frac1\delta e^{\frac{\lambda \bar M_n(W)}{n} } } \\
  &\leq  \log\frac1\delta  + \frac{\lambda^2}{8n},
\end{align}
where the last step is due to \cref{thm:azuma-hoeffding}.
After repeating this argument for~$-\bar M_n$ and using the union bound, we find that with probability at least~$1-\delta$,
\begin{align}\label{eq:unopt-martingale-bound}
 \abs{ \Ex{ \conddistro }{\frac{\bar M_n(W)}{n}} } &\leq \frac{\relent{ \conddistro }{Q_W} + \log\frac2\delta}{\lambda}  + \frac{\lambda}{8n} .
\end{align}
To complete the proof, we need to select~$\lambda$.
We will do this by optimizing the bound over a grid of candidate values, using a union bound to ensure that the result is valid for all possible values.\footnote{In the original proof,~\citet{seldin-12a} use a dyadic grid and a weighted union bound over an infinite range. We restrict ourselves to a finite range, similar to~\citet{rodriguezgalvez-23a}, in order to simplify the proof.}
First, note that if~$\relent{ \conddistro }{Q_W} > 2n$, the right-hand side of~\eqref{eq:unopt-martingale-bound} is lower-bounded by~$1$ for all~$\lambda$, meaning that the resulting bound is vacuous~(since~$\bar M_n(W)\leq n$).
Hence, the result in~\eqref{eq:thm-martingale-pacb-eq} holds trivially in this case.
Thus, we only consider~$\relent{ \conddistro }{Q_W} \leq 2n$.
Specifically, assume that~$\relent{ \conddistro }{Q_W} \in [k-1,k]$ for~$k\in[2n]$.
Then, by~\eqref{eq:unopt-martingale-bound}, we have
\begin{equation}
 \abs{ \Ex{ \conddistro }{\frac{\bar M_n(W)}{n}} } \leq \frac{k + \log\frac2\delta}{\lambda}  + \frac{\lambda}{8n} .
\end{equation}
For a fixed~$k$, this is minimized by~$\lambda=2\sqrt{ 2 n( k + \log\frac2\delta ) }$, which gives
\begin{equation}
 \abs{ \Ex{ \conddistro }{\frac{\bar M_n(W)}{n}} } \leq \sqrt{  \frac{ k + \log\frac2\delta }{ 2n }  } .
\end{equation}
By the union bound, this holds simultaneously for %
$k\in[2n]$ with probability at least~$1-2n\delta$.
Hence, by substituting~$\delta$ with $\delta/(2n)$, noting that~$k\leq \relent{ \conddistro }{Q_W} + 1 $,
\begin{equation}
 \abs{ \Ex{ \conddistro }{\frac{\bar M_n(W)}{n}} } \leq \sqrt{  \frac{ \relent{ \conddistro }{Q_W} + 1 + \log\frac{4n}\delta }{ 2n }  }
\end{equation}
with probability at least~$1-\delta$.
From this, the desired result follows.
\end{proof}
By suitably selecting~$\bar M_i$---for instance, as the difference between the loss for a training instance and its expectation---this bound can be instantiated for various settings with martingale data, extending the applicability of PAC-Bayesian bounds beyond  \iid data.
For instance,~\citet{seldin-11a,seldin-12b} apply these bounds to the case of multiarmed bandits.
It is worth noting that~\citet{seldin-12a} derive additional bounds using martingale versions of the concentration for binary relative entropy in \cref{thm:kl_concentration} as well as Bernstein's inequality.

\subsection{Markov Decision Processes}\label{sec:markovdecisionprocess}

In reinforcement learning, the learner is viewed as an ‘‘agent'' that interacts with an environment and takes actions according to a strategy, also known as policy, obtaining rewards on this basis.
The goal of this is to learn a good policy for how to select actions depending on the state of the environment.
A defining characteristic of reinforcement learning is that the environment is only partially observed through the agent's interaction with it.
A specific example of this is the setting of contextual bandits, where the PAC-Bayesian bounds for martingales can be applied, as demonstrated by~\citet{seldin-11a}.
Here, following~\citet{gouverneur-22a}, we will focus on Bayesian regret in an MDP, presenting a bound that extend the result obtained by~\citet{xu-22a} for supervised learning.

In order to formally describe an MDP, we need the following definitions.
We let~$\statespace$ denote a set of states, let~$\actionspace$ denote a set of actions, and let~$\labelspace$ denote a set of outcomes.
At each time~$t\in[T]$, the learner observes the state~$S_t\in\statespace$ and takes an action~$A_t\in\actionspace$, after which the environment produces an outcome~$Y_t\in\labelspace$.
This leads to the reward~$R_t=r(Y_t,A_t)\in\reals$.
The environment is characterized by a random variable~$\theta\in\Theta$, drawn according to~$P_\theta$.
More specifically, it consists of a transition kernel~$P_{S_{t+1}\vert S_t,A_t,\theta}$, an outcome kernel~$P_{Y_t\vert S_t,\theta}$, an initial state distribution~$P_{S\vert\theta}$, from which~$S_1$ is drawn, and the reward function~$r:\labelspace\times\actionspace\rightarrow\reals$.
The stochastic mapping from the state~$S_t$ and action~$A_t$ to the reward~$R_t$ is characterized by the kernel~$P_{R_t\vert S_t,A_t,\theta}$.
The goal is to learn a policy~$\varphi=\{ \varphi_t: \statespace\times(\statespace,\actionspace,\reals)^t\rightarrow \actionspace \}_{t\in[T] }$, which selects an action~$A_t$ on the basis of~$S_t$ and the observed history~$H_{\leq t}=(H_1,\dots,H_{t-1})$, where~$H_t=(S_t,A_t,R_t)$.
Specifically, the policy should be chosen to obtain a high cumulative expected reward~$r_c (\varphi)$, defined as
\begin{equation}\label{eq:cum-exp-rew}
r_c (\varphi) = \Ex{}{ \sum_{t\in[T] } r(Y_t,\varphi_t(S_t,H_{\leq t})) }.
\end{equation}
We refer to the maximal expected cumulative reward as the Bayesian cumulative reward, and denote it by~$R_c=\sup_{\varphi} r_c(\varphi)$, where the supremum is taken over all policies that lead to a finite expectation in~\eqref{eq:cum-exp-rew}.
We will compare this to the maximal expected cumulative reward that can be obtained by an oracle that has knowledge of~$\theta$.
Specifically, we consider decision rules~$\psi = \{\psi_t: \statespace\times\Theta \rightarrow \actionspace\}_{t\in [T]} $ and define the oracle Bayesian cumulative reward as
\begin{equation}
R^o_B  = \sup_{\psi}  \Ex{}{ \sum_{t\in[T] } r(Y_t,\psi_t(S_t, \theta )) }.\label{eq:suppsietc}
\end{equation}
We let~$\psi^*=\{\psi_t^*\}_{t\in[T]}$ denote the policy that achieves the supremum in~\eqref{eq:suppsietc}, and assume that it exists.
With this, we are ready to define the key quantity that we wish to bound: the minimum Bayesian regret (MBR) given by
\begin{equation}
\MBR = R^o_B - R_c .
\end{equation}
This quantity %
is the difference between the reward that is obtainable based only on observing the system through interactions and the one that is obtainable when the underlying system parameters are known.

In order to bound the MBR, we will consider a specific learning algorithm, related to Thompson sampling~\citep{thompson-33a,russo-16b}.
One approach to selecting~$\phi_t$ is to use~$H_{\leq t}$ to compute an estimate~$\hat\theta_t$ through a kernel~$P_{\hat \theta_t\vert H_{\leq t}}$, and then select an action on the basis of~$(S_t,\hat\theta_t)$.
Since this is a special instance of a learning algorithm, the resulting cumulative expected reward cannot be greater than the Bayesian cumulative reward.
\begin{align}
 R_c &=  \sup_\varphi \Ex{}{ \sum_{t\in[T] } r(Y_t,\varphi_t(S_t,H_{\leq t})) }  \\
  &\geq \sup_\psi\Ex{}{ \sum_{t\in[T] } r(Y_t,\psi_t(S_t,\hat \theta_t)) } \\
  &\geq \Ex{}{ \sum_{t\in[T] } r(Y_t,\psi^*_t(S_t,\hat \theta_t)) }.
\end{align}
We now introduce~$Y^*_t$ and~$S^*_t$ as the outcomes and states that are obtained through~$\psi^*$ acting on the MDP with the true~$\theta$ as input.
Similarly, we let~$\hat Y_t$,~$\hat S_t$, and~$\hat H_t$ denote the outcomes, states, and histories that are obtained through~$\psi^*$ acting on the MDP with the estimated~$\{\hat\theta_t\}_{t\in[T]}$ as input.
Now, by expanding the expression above, we find that the MBR can be bounded as
\begin{align}
&\MBR \leq  R^o_B - \Ex{}{ \sum_{t\in[T] } r(Y_t,\psi^*_t(S_t,\hat \theta_t)) } \\
 = &\!\sum_{t\in[T]} \! \Ex{ P_{\theta \hat\theta_t \hat H_{\leq t} \!\! } }{   \Ex{  P_{Y^*_t S^*_t \hat Y_t  \hat S_t \vert \theta \hat\theta_t \hat H_{\leq t}}  }{   r( Y^*_t, \psi^*_t(S^*_t,\theta))  \!-\!  r(\hat Y_t, \psi^*_t( \hat S_t, \hat \theta _t ))   \!    }   }. \nonumber
\end{align}
Now, observe that the following Markov chain holds:
\begin{equation}
Y^*_t,S^*_t)-\theta-(\hat Y_t,\hat S_t)-\hat H_{\leq t} - \hat \theta_t.
\end{equation}
From this, it follows that for each~$t\in[T]$, the first term of the inner expectation is distributed according to~$P_{Y^*_t,S^*_t\vert \theta}$, while the second is distributed according to~$P_{\hat Y_t,\hat S_t\vert H_{\leq t}}$.
Therefore, we can use change of measure techniques to relate the two terms, by following the same arguments as in \cref{chap:average} (and in particular, \cref{sec:individual_sample}).
This leads to the following result~\citep[Prop.~1]{gouverneur-22a}.
\begin{thm}\label{thm:reinforcement-learning}
Assume that, for all~$t\in[T]$,~$r(\hat Y_t,\psi^*_t(\hat S_t,\theta))$ is~$\sigma^2_t$-sub-Gaussian under~$P_{ \hat Y_t, \hat S_t\vert \hat H_{\leq t} }$ for all~$\theta\in\Theta$. Then,
\begin{equation}
\MBR \leq \sum_{t\in[T] }  \Ex{ P_{\theta \hat H_{\leq t}}  }{   \sqrt{2\sigma^2_t \relent{ P_{ Y^*_t, S^*_t  \vert \theta } }{ P_{ \hat Y_t, \hat S_t\vert \hat H_{\leq t} } }  } } .
\end{equation}
\end{thm}

More discussion of these results, including applications to special cases and results in terms of the Wasserstein distance, can be found in the work of~\citet{gouverneur-22a}.

\section{Bibliographic Remarks and Additional Perspectives}\label{sec:bib-remarks-alternative-learning}

The result in \cref{thm:meta-learning-it-bound} is due to~\citet{chen-21a}.
Information-theoretic generalization bounds for meta learning can also be found in the work of~\citet{jose-21a,jose-22a}, and were extended to the case of e-CMI in~\citet{hellstrom-22b}.
Additional works that provide PAC-Bayesian and information-theoretic generalization bounds for meta learning include, \eg,~\citet{pentina-14a,amit-18a,rothfuss-21a,liu-21a,farid-21a,meunier-21a,flynn-22a,rezazadeh-22a,jose-22b,riou-23a}.
The bounds for OOD generalization in \cref{propo:ood-gen-pp,propo:wasserstein-ood,thm:transfer-learning-unsup} are due to~\citet{wang-23b}, while \cref{thm:trasnfer-supervised} is due to~\citet{wu-22a}.
\citet{jose-22a} considered a combination of transfer learning and meta learning, while~\citet{jose-23a} analyzed transfer learning for quantum classifiers.
Additional results for transfer learning and domain adaptation can be found in the works of~\citet{germain-16a,achille-21a,jose-21c,aminian-22a,bu-22a}.
Relatedly, \citet{he-22a} derived bounds for iterative semi-supervised learning.
\cref{thm:federated-learning} is due to~\citet{barnes-22a}, with earlier work by~\citet{yagli-20a}.
\citet{sefidgaran-22b} derived generalization bounds for distributed learning using rate-distortion techniques.
The extension of PAC-Bayesian bounds to martingales in \cref{thm:martingale-pacb} is due to~\citet{seldin-12a};~\citet{seldin-11a} applied these to contextual bandits.
\cref{thm:reinforcement-learning} is due to~\citet{gouverneur-22a}.
Additional PAC-Bayesian results for reinforcement learning can be found in the work of~\citet{fard-10a,wang2-19a}.

We conclude by mentioning alternative learning models, and their connections to PAC-Bayesian and information-theoretic generalization bounds.
\citet{seeger-02a} applied PAC-Bayesian bounds to Gaussian process classification, while \citet{shawetaylor-09a} considered the problem of maximum entropy classification.
Unsupervised learning models, such as various types of clustering, were studied by, \eg,~\citet{seldin-10a,higgs-10a,li-18a}.
\citet{alquier-11a} considered the sparse regression model in high dimension, while \citet{guedj-18a} derived PAC-Bayesian bounds for the bipartite ranking problem in high dimension.
\citet{ralaivola-10a} derived bounds for non-\iid\ data, with applications to certain ranking statistics, while \citet{li-13a} extended PAC-Bayesian bounds to the nonadditive ranking risk.
\citet{jose-21d} used PAC-Bayesian bounds to analyze machine unlearning, where a learning algorithm has to ‘‘forget'' specific samples.
Online learning, where the learner has to sequentially select hypotheses to minimize losses set by a potentially adversarial environment (a recent introduction is provided by~\citealp{orabona-23a}), is intimately related to PAC-Bayesian and information-theoretic bounds.
In particular, there is a formal relationship between the Gibbs posterior and the exponential weights algorithm.
PAC-Bayesian bounds for a version of online learning were studied by~\citet{haddouche-22a}.
Recently,~\citet{lugosi-22a,lugosi-23a} established a method for converting regret bounds from online learning to PAC-Bayesian and information-theoretic bounds, allowing them to (essentially) recover established results and derive new ones.
\newtext{Finally, \citet{sharma-23a} exploited PAC-Bayesian generalization bounds in the context of inductive conformal prediction, allowing the calibration data set to be used for learning the hypothesis and score function.}

\chapter{Concluding Remarks}\label{chap:perspectives}

In this monograph, we provided a broad overview of information-theoretic and PAC-Bayesian generalization bounds.
We highlighted the connection between these fields; presented a wide array of bounds for different settings in terms of different information measures; detailed analytical applications of the bounds to specific learning algorithms; discussed recent applications to iterative methods and neural networks; and covered extensions to alternative settings.
We hope that this exposition demonstrates the versatility and potential of the information-theoretic approach to generalization results.

Still, there are many unanswered questions and directions to explore.
On the one hand, as shown by~\citet{haghifam-21a,haghifam-23a}, there are certain settings for which the information-theoretic approaches discussed in this monograph yield provably suboptimal bounds.
On the other hand, there are bounds in terms of the evaluated mutual information that equal the population loss for interpolating settings~\citep{haghifam-22a,wang-23a}, as discussed in Section~\ref{sec:ecmi}, and by appropriately adapting standard information-theoretic bounds, optimal characterizations of the generalization gap in the Gaussian location model can be derived~\citep{zhou-23a}.
This raises the question of which settings the information-theoretic approach to generalization is suitable for, and whether or not it can be extended further through new ideas, or whether alternative approaches are necessary.

As discussed in Section~\ref{sec:nn-numeric}, information-theoretic and PAC-Bayesian bounds have been shown to be numerically accurate for certain settings with neural networks.
However, the utility and interpretation of these results is not entirely clear.
\citet{dziugaite-17a} connect their bound to the flatness of the loss landscape;~\citet{harutyunyan-21a} draw parallels to stability; and~\citet{lotfi-22a} point towards compressibility, exploring its relation to, \eg, equivariance and transfer learning.
Pinning down these connections more precisely, and developing the bounds to such an extent that they can guide model selection \textit{a priori}, are intriguing avenues to explore.

Regarding the structure of the bounds themselves,~\citet{foong-21a,hellstrom-24a} explore the question of what the tightest attainable bound is.
For instance, what is the best comparator function to use in \cref{propo:generic-pac-bayes-thm}?
Can the~$\log\sqrt n$ dependence in \cref{cor:MLS_bound} be removed?
Another question is whether the most suitable information measure to use for a given setting can be determined.
As discussed throughout, the specific information measure that arises in a bound is just a consequence of the change of measure technique that is used in its derivation.

Finally, there are several interesting extensions to other settings and connections to other approaches that can be explored.
While we covered some topics in~\cref{chap:alternative-settings}, the relation to, for instance, active learning, wherein the information carried by a sample is a central quantity~\citep{settles-12a}, and online learning, the analysis of which shares many tools with the information-theoretic approach~\citep{orabona-23a}, is a promising direction.
For instance, recently,~\citet{lugosi-23a} showed that any regret bound for online learning implies a corresponding generalization bound for statistical learning.

While this discussion is far from comprehensive, addressing these questions and exploring the aforementioned connections may be a fruitful path forward.
We hope that this monograph will be valuable in pursuing these goals.

\begin{acknowledgements}

\noindent \newtext{The authors thank Yunwen Lei, Alex Olshevsky, Olivier Wintenberger, the anonymous referees, and the associate editor for providing valuable comments on an earlier version of this work.}
\medskip

\noindent F.H.\ and G.D.\ acknowledge support by the Wallenberg AI, Autonomous Systems and Software Program (WASP) funded by the Knut and Alice Wallenberg Foundation.
\medskip

\noindent G.D.\ also acknowledges support by the Swedish Foundation for Strategic Research under grant number FUS21-0004.
\medskip

\noindent B.G.\ acknowledges partial support by the U.S. Army Research Laboratory and the U.S. Army Research Office, and by the U.K. Ministry of Defence and the U.K. Engineering and Physical Sciences Research Council (EPSRC) under grant number EP/R013616/1. B.G. acknowledges partial support from the French National Agency for Research, through grants ANR-18-CE40-0016-01 and ANR-18-CE23-0015-02, and through the programme ``France 2030'' and PEPR IA on grant SHARP ANR-23-PEIA-0008. 
\medskip

\noindent M.R. acknowledges partial support by the U.S. National Science Foundation (NSF) through the Illinois Institute for Data Science and Dynamical Systems (iDS${}^2$),
an NSF HDR TRIPODS Institute, under Award CCF-193498.

\end{acknowledgements}

\backmatter  %
\DeclareNameAlias{sortname}{family-given}

\printbibliography
 
\end{document}